\documentclass[11pt]{report}         %% LaTeX2e document.
\usepackage {theapa}
\usepackage {utthesis-compact}       %% Preamble.

\input{epsf}

  \phdthesis                         %% use either, the default is \phdthesis.

  \singlespace                       %% Uncomment one of these if you want
\renewcommand{\thesisauthor}{Ulf Hermjakob}  %% Your official UT name.

\renewcommand{\thesismonth}{May}     %% Your month of graduation.

\renewcommand{\thesisyear}{1997}     %% Your year of graduation.

\renewcommand{\thesistitle}{Learning Parse and Translation Decisions\\[2mm] From Examples With Rich Context}

\renewcommand{\thesisauthorpreviousdegrees}{Dipl.-Inform.}
\renewcommand{\thesissupervisor}{Raymond J.\ Mooney}
\renewcommand{\thesisauthoraddress}{Moltkestr.\ 40a\\ 32257 B\"{u}nde\\ Germany}
\renewcommand{\thesisdedication}{To my parents Udo and Juliane \\ for their loving support and continuous encouragement for a good education}
\begin{document}

\thesiscopyrightpage                 %% Generate the copyright page.

\thesistitlepage                     %% Generate the title page.

\vspace*{11cm}                       %% Preplace the signatures
\hspace*{3.6cm}                      %% on the signature page
\epsfxsize=7.88cm
\centerline{\epsfbox{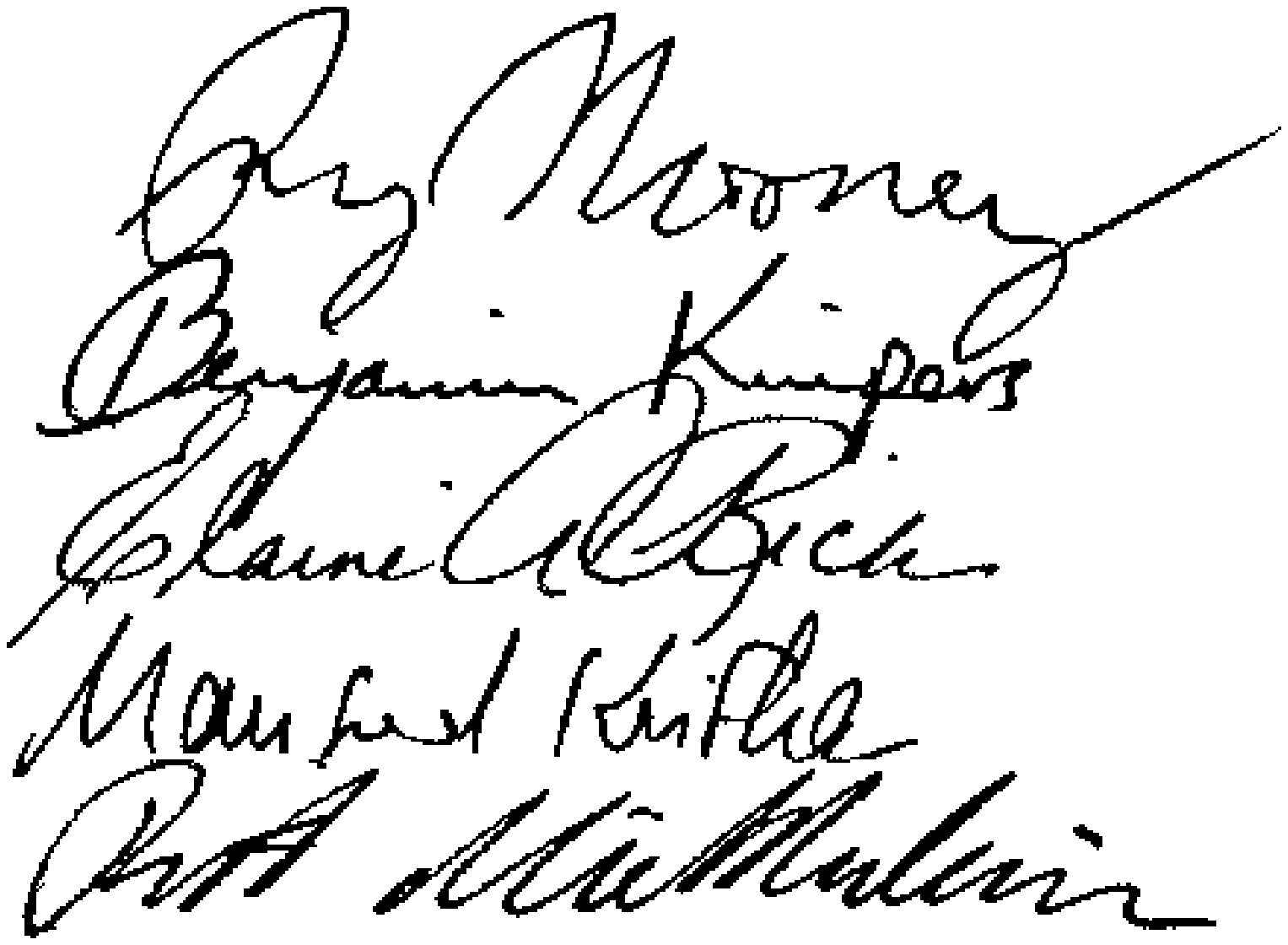}}
\vspace*{-17.9cm}
\thesissignaturepage                 %% Generate the signature page.

\thesisdedicationpage                %% Generate the dedication page.

\begin{thesisacknowledgments}        %% Use this to write your
I am very thankful for the contributions of my first advisor, the late
Robert F.\ Simmons, who helped me develop the basic ideas underlying my
doctoral research. I also deeply appreciate the dedicated support of my 
current advisor, Raymond J.\ Mooney, who guided the completion of my 
dissertation with many helpful insights.
Many thanks for valuable comments also to the other members of my committee,
Risto Miikkulainen, Benjamin J.\ Kuipers, Manfred Krifka and Elaine A.\ Rich.

To Elaine Rich and Jim Barnett, who led the Knowledge Based Natural Language
group at the Microelectronics and Computer Technology Corporation, I owe 
many stimulating ideas and much support during the four years I worked at MCC.

Many thanks also to everybody in the Natural Language Acquisition discussion
group at our computer science department; besides Bob Simmons,
Ray Mooney and Risto Miikkulainen in particular to John Zelle, for the many
interesting discussions we had since the group was started in 1992.

Finally I would like to give thanks to
Brent Adamson, Ulrich Ehrenberger, Ralf Gerlich, Thomas G\"{o}hlert, 
Mike Hoefelein, Martin Kracklauer, Michael Menth, Florian Mertens, 
Christoph P\"{u}ntmann and Joe Sullivan for volunteering as translation 
evaluators.\\
\end{thesisacknowledgments}          %% allowed in LaTeX2e par-mode.

\begin{thesisabstract}               %% Use this to write your thesis
                                     %% abstract; it can be anything

The parsing of unrestricted text, with its enormous lexical and structural ambiguity,
still poses a great challenge in natural language processing.
The difficulties with traditional approaches, which try to master the complexity of parse
grammars with hand-crafted rules, have led to a trend towards more empirical techniques.

We therefore propose a system for parsing and translating natural language that
learns from examples and uses some background knowledge.
As our parsing model we choose a deterministic shift-reduce type parser that
integrates part-of-speech tagging and syntactic and semantic processing, which
not only makes parsing very efficient, but also assures transparency during the
supervised example acquisition.
Applying machine learning techniques, the system uses parse action examples to
generate a parser in the form of a decision structure, a generalization
of decision trees.
To learn good parsing and translation decisions, our system relies heavily on context,
as encoded in currently 205 features describing the morphological, syntactical and 
semantical aspects of a given parse state. Compared with recent probabilistic systems
that were trained on 40,000 sentences,
our system relies on more background knowledge and a deeper analysis, but radically 
fewer examples, currently 256 sentences.

We test our parser on lexically limited sentences from the Wall Street Journal
and achieve accuracy rates of 89.8\% for labeled precision, 98.4\% for part of 
speech tagging and 56.3\% of test sentences without any crossing brackets.
Machine translations of 32 Wall Street Journal sentences to German have been evaluated
by 10 bilingual volunteers and been graded as 2.4 on a 1.0 (best) to 6.0 (worst)
scale for both grammatical correctness and meaning preservation.
The translation quality was only minimally better (2.2) when starting each translation
with the correct parse tree, which indicates that the parser is quite robust and that
its errors have only a moderate impact on final translation quality.
These parsing and translation results already compare well with other systems and,
given the relatively small training set and amount of overall knowledge used so far,
the results suggest that our system {\sc Contex} can break previous accuracy ceilings 
when scaled up further.

\end{thesisabstract}                 %% allowed in LaTeX2e par-mode.

\tableofcontents                     %% Generate table of contents.
% \listoftables                      %% Uncomment this to generate list
                                     %% of tables.
% \listoffigures                     %% Uncomment this to generate list
                                     %% of figures.

\chapter{Introduction}          
\label{ch-intro}

Natural language processing in general and machine translation in particular face many 
difficulties.  Consider for example the following sentences:\\[3mm]
\noindent \begin{tabular}{lp{12.4cm}}
(1) & He had the right to go. (`right' is a noun)\\
(2) & He was right in deciding not to go. (adjective)\\
(3) & The exit door was right in the line of fire. (adverb)\\
(4) & Should he wait until the problem might right itself? (verb)\\[3mm]
(5) & I ate the pasta with a fork. (PP attaches to verb)\\
(6) & I ate the pasta with a delicious tomato sauce. (PP attaches to preced.\ NP)\\[1mm]
(7) & I sent a package to New York. (PP attaches to verb)\\
(8) & I booked a flight to New York. (PP attaches to preceding NP)\\[3mm]
(9) & I saw the Grand Canyon flying to New York. (semantics vs.\ syntactic bias)\\
(10) & I will pick up the car at the airport. (attachment unclear and irrelevant)\\[3mm]
(11) & I know your friend in New York. (``to know'' $\rightarrow$ ``conocer'')\\
(12) & I know your friend is in New York. (``to know'' $\rightarrow$ ``saber'')\\[3mm]
\end{tabular}

A parser\footnote{A {\it parser} is a program that analyses text to find structural
descriptions of the text that encode useful information of some kind about it.
The term {\it parser} is used for the analysis of both source code of computer programs
and natural languages, such as English and Swedish. In natural language understanding,
the structural descriptions built by the parser typically include a tree reflecting the
syntactic structure of the text, but can also contain semantic and other information.
A good introductory textbook on natural language understanding is \cite{allen:book95}.}
has to cope with ambiguous
words like ``right'' (sentences 1-4), proper structural attachment (sentences 5-8), even
if the preposition phrases are identical (sentences 7\&8). Sometimes the existence of ambiguity
is relatively obvious to the human reader, particularly when syntactic and semantic clues point
in opposite directions (sentence 9), while in other cases the ambiguity is more subtle and
possibly irrelevant (sentence 10).
Even when there seems to be little ambiguity for words like ``to know'', other
languages often make distinctions, depending on how exactly a word is used. In the case
of sentences 11\&12, ``to know'' has different translations for most languages,
e.g.\ ``conocer''/``saber'' in Spanish.
\newpage
\noindent \begin{tabular}{lp{12.4cm}}
(13) & The statue of liberty in New York, which was sent to the United States as a gift
       from France, was dedicated in 1886. \\
(14) & The airline said that it would report a loss for the first quarter,
       but that it would be less than \$100 million.\\[3mm]
\end{tabular}

Sentences 13\&14 demonstrate the problem of anaphora resolution\footnote{An {\it anaphor}
is an expression which cannot have independent reference, but refers to another expression,
the so-called {\it antecedent} \cite{radford:tg88}. Examples of anaphora in sentences 13\&14 
are the pronouns ``which'' and ``it'' (both occurrences), which refer to the antecents 
``the statue of liberty'', ``the airline'', and ``a loss''. 
The process of finding the proper antecedent of an anaphor is called {\it anaphora resolution}.}.
The German translation of the second occurrence of the personal pronoun `it' in sentence 14
for example
depends on whether that pronoun refers to ``the airline'', ``a loss'', or ``the first quarter'',
because the grammatical gender of the translated German pronoun ({\it ``sie'', ``er''} or 
{\it ``es''}) has to agree with the grammatical gender of its antecedent 
(``the airline'' $\rightarrow$ {\it ``die Fluggesellschaft''} (feminine),
``a loss'' $\rightarrow$ {\it ``ein Verlust''} (masculine), or 
``the first quarter'' $\rightarrow$ {\it ``das erste Quartal''} (neuter)).
The same holds for the relative pronoun ``which'' in sentence 13. 

These few examples already show that an automatic parsing and machine translation system
has to make many difficult decisions, including part of speech assignment, proper
structural analysis, word sense disambiguation and anaphora resolution.
Natural language processing is a truly formidable task.

\section{Previous Approaches}

\subsection{Traditional Approaches}

The parsing of unrestricted text, with its enormous lexical and structural
ambiguity, still poses a great challenge in natural language processing (NLP).
Traditional approaches try to master the complexity of parse grammars
with hand-crafted rules, as in augmented transition networks \cite{bates:atn78},
definite clause grammars \cite{pereira:dcg80},
lexical functional grammars \cite{kaplan:lfg82}, 
functional unification grammars \cite{kay:fug82},
tree-adjoining grammars\footnote{Tree-adjoining grammars are based on 
hand-coded trees instead of rules.} \cite{joshi:tag85},
or generalized phrase structure grammars \cite{gazdar:gpsg95}.
The manual construction of broad-coverage grammars turned out to be much more 
difficult than expected, if not impossible. 
In his {\it Machine Translation `forum'} lead paper, \citeA{somers:mt93}
reflects on the relatively modest advancement in natural language processing
in the 70's and 80's by asking
``What's wrong with classical second generation\footnote{By {\it second generation
architecture} Somers refers to systems which incorporate most of the typical
design features such as
``linguistic rule-writing formalisms with software implemented independently of the
linguistic procedures, stratificational analysis [in particular morphology - surface
syntax - deep syntax] and generation, and an intermediate linguistically motivated
representation which may or may not involve the direct application of contrastive 
linguistic knowledge.''}
architecture?''. 
\nocite{nirenburg:book87,slocum:book88}
\nocite{hermjakob:acl97}

\subsection{Empirical Approaches}

In recent years, there has been a trend towards more empirical approaches that 
augment or replace the hand-coded rule paradigm with statistical and machine 
learning techniques. These techniques automatically derive parse grammars and 
other classification structures from examples.

A number of researchers have already applied machine learning techniques to
various NLP tasks like
accent restoration by \citeA{yarowsky:acl94}, who achieves 99\% accuracy,
relative pronoun disambiguation \cite{cardie:aaai92,cardie:ml93,cardie:emnlp96},
(Japanese) anaphora resolution \cite{aone:acl95},
part of speech tagging \cite{weischedel:compling93,brill:compling95,daelemans:wvlc96},
where tagging accuracies between 96\% and 97\% are achieved,
cue phrase classification \cite{siegel:aaai94,litman:jair96},
and word sense disambiguation \cite{yarowsky:coling92,yarowsky:acl95,ng:acl96,mooney:emnlp96}.

Many of the statistical approaches have been made possible by the increased availability of
large corpora of natural language data like the Brown corpus \cite{francis:brown82},
which consists of about a million words, all labeled with their parts of speech,
or the 4.5 million word Penn Treebank \cite{marcus:compling93},
in which more than half of the sentences have been annotated for their skeletal
syntactic structure.

Newer treebank-based probabilistic approaches \cite{magerman:acl95,collins:acl96}
have produced encouraging results.
It is however not clear how these systems can still be improved significantly,
because (1) using about 40,000 training sentences, they have already exhausted 
the large Penn Treebank corpus and (2) they still use a fairly limited context 
and sharply restrict the amount of background knowledge.

To cope with the complexity of unrestricted text, parse rules in any kind of formalism
will have to consider a complex context with many different morphological, syntactic,
semantic and possibly other features to make good parsing decisions.
This can present a significant problem, because even linguistically trained natural
language developers have great difficulties writing and even more so extending explicit
parse grammars covering a wide range of natural language.
On the other hand it is much easier for humans to decide how {\it specific} sentences
should be analyzed and what feature might discriminate a pair of {\it specific} parse
states.

\section{A Context-Oriented Machine Learning Approach}

We therefore propose an approach to parsing based on learning from examples with
a very strong emphasis on context, integrating morphological, syntactic, 
semantic and other aspects relevant to making good parse decisions, thereby
also allowing the parsing to be deterministic.
In order to reduce the complexity of the learning task, we break the parsing
process into a sequence of smaller steps, so-called {\it parse actions}. This
reduces parsing to a decision problem, where the system has to learn which
parse action to perform next. As we have shown at the beginning of this introduction,
these decisions can be quite difficult. We therefore provide (1) a rich context, 
encoded by currently 205 morphological, syntactic and semantic features and
(2) some background knowledge, in particular dictionaries, a concept hierarchy,
and a subcategorization table.

As our parsing model we choose a deterministic shift-reduce type parser, which not
only makes parsing very efficient, but also assures transparency during the example
acquisition, when parse action examples are collected interactively: the
partially trained parser proposes the next parse action and a human supervisor
confirms or overwrites the proposal. Compared to the original shift-reduce parser
\cite{marcus:book80}, our parser includes additional types of operations and
allows additional operation parameters, so that the parser can produce a deeper
analysis of the sentence. The parse tree integrates semantic information,
phrase-structure and case-frames. Not only does this lead to a final parse tree
that is powerful enough to be fed into a transfer and a generation module
to complete the full process of machine translation, it also provides much richer
intermediate results, which are used to improve the quality of parsing decisions.

Applying machine learning techniques, the system uses the acquired parse action
examples to generate a parse grammar in the form of an advanced decision structure,
an extension of decision trees \cite{quinlan:ijcai87,quinlan:book93}. Such a parse
grammar can be used to parse previously unseen sentences into parse trees. Based
on such parse trees, our system translates the sentences, re-using the empirical
methods developed for parsing with only minor adaptations.

Following this corpus based approach, we relieve the NL-developer from the hard if 
not impossible task of writing explicit grammar rules and make grammar coverage
increases very manageable.
Compared with recently published probabilistic methods, our system relies on
some background knowledge, a rich context, a deeper analysis and more supervision
per sentence, but radically fewer examples, currently 256\footnote{Note however
that these 256 sentences contain some 11,000 individual parse action examples.}, 
as opposed to 40,000 sentences for the treebank-based approaches cited above.

\section{Summary of Experimental Results}
We tested our parser on lexically limited sentences from the Wall Street Journal 
and achieved accuracy rates of 89.8\% for labeled precision\footnote{See 
chapter~\ref{sec-parse-eval-methodology} for exact definitions of parsing
metrics.}, 98.4\% for part of speech tagging and 56.3\% of test sentences without 
any crossing brackets. Our accuracy results are about as good as for competing 
probabilistic systems trained on 40,000 sentences. Based on our learning curve,
we expect that, when increasing the number of our training sentences from 256, 
but still staying far below the 40,000 training sentences used by probabilistic 
systems, our results will be significantly better than for those treebank-based 
probabilistic system.

Machine translations of 32 randomly selected Wall Street Journal sentences
from English to German have been evaluated
by 10 bilingual volunteers and been graded as 2.4 on a 1.0 (best) to 6.0 (worst)
scale for both grammatical correctness and meaning preservation. Our system has been
trained and tested on the same lexically limited corpus, but despite this advantage
we believe that it is quite an achievement that after only very limited training,
our system already produces better translations than all three competing commercial
systems.
The translation quality was only minimally better (2.2) when starting each translation
with the correct parse tree, indicating that the parser is quite robust and that
its errors have only a moderate impact on final translation quality.

\section{Organization of Dissertation}

Chapter~\ref{ch-overview} presents the task definition and gives an overview
of the architecture of the system and its major components.
Chapter~\ref{ch-background} describes the background knowledge we use,
namely the knowledge base, the monolingual lexicons, the bilingual dictionary,
the subcategorization table, and the so-called morphological pipelines.
Chapter~\ref{ch-preparsing} briefly explains the two pre-parsing modules,
segmentation, and morphological analysis.
The key chapter~\ref{ch-parsing} on parsing explains our parsing paradigm and
presents the details of parse actions, features, the training of the parser and
the decision structures we use for learning.
Chapter~\ref{ch-p_exp} evaluates the parser, including a series of `ablation' tests
to investigate the practical contributions of various system components.
Chapter~\ref{ch-transfer} describes how the techniques used for parsing can
be applied to the parse tree transfer to the target language.
Chapter~\ref{ch-generation} explains how that target language tree is reordered 
and morphologically propagated in order to compute the surface forms of words and 
phrases in the target language.
Chapter~\ref{ch-t_exp} reports the results of our translation experiments, including
comparisons to three commercial systems.
Chapter~\ref{ch-related} describes related work and
chapter~\ref{ch-future} discusses several potential future extensions of our system,
before chapter~\ref{ch-conclusions} concludes.
          %\chapter{Introduction} \label{ch-intro}
\chapter{System Architecture}
\label{ch-overview}
 
\section{Task Definition}
Our goal is to parse and translate natural language. In this dissertation, we 
investigate how machine learning techniques can be used to first parse input
text into a parse tree and subsequently, based on this parse tree, translate 
the text into another language.

To develop a parser for English and a translator from English to German, we use
training sentences from the Wall Street Journal.
We evaluate the parser by testing it on sentences from the Wall Street
Journal that were not used for training. We use standard evaluation criteria for parsing,
including precision, recall and violations of constituent boundaries 
(``crossing brackets'').
The German translations are graded by outside bilingual evaluators.

\subsection{Corpus}
The WSJ corpus used in our work is a subset of sentences from Wall Street
Journal articles from 1987, as provided on the ACL data-disc.
In order to limit the size of the required lexicon, we work on a reduced corpus
that includes all those sentences that are fully covered by the 3000 most
frequently occurring words (ignoring numbers etc.) in the entire corpus.
The lexically reduced corpus contains 105,356 sentences, a tenth of the full corpus.

For our training and testing we use the first 272 sentences from this lexically
restricted corpus. They vary in length from 4 to 45 words, averaging at 17.1 words.
We believe that it is important to use sentences from the ``real world'' instead
of artificially generated sentences in order to capture the full complexity of
natural language.
A complete listing of the 272 training and testing sentences can be found in
appendix~\ref{app-corpus} and a more detailed discussion of the testing and training 
corpus is presented in section~\ref{ch-p_exp_corpus} of the chapter on 
``Parsing Experiments''.

\section{Transfer-Based vs.\ Interlingua Approach}

\begin{figure}[htb]
\epsfxsize=16.4cm
\centerline{\epsfbox{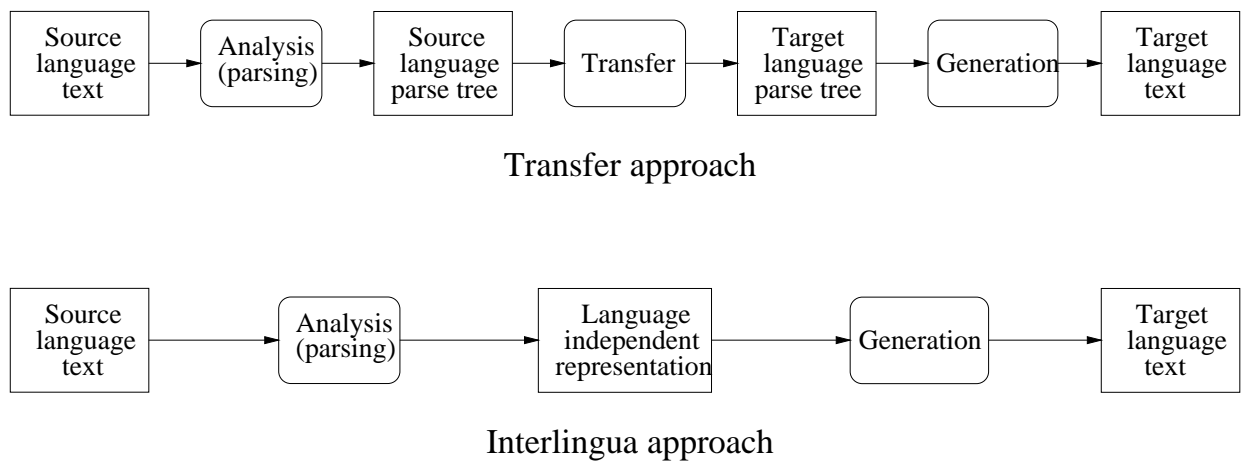}}
\caption{The transfer and interlingua approaches}
\label{fig_transfer_vs_interlingua}
\end{figure}

The overall system architecture follows the classical subdivision of parsing, transfer
and generation, thereby rejecting a pure interlingua approach
(see figure~\ref{fig_transfer_vs_interlingua}).
As the authors and interlingua proponents of \cite{nirenburg:mt_kb92} point out,
the major distinction between the interlingua- and transfer-based systems
is the ``attitude toward comprehensive analysis of meaning''.
While the pure interlingua approach might be quite appealing academically,
we believe that the transfer approach is easier and more practical to handle
and can demonstrate the main paradigm and ideas introduced by our system equally well.

An interlingua is a formal language which has to represent
the meaning of source text fully and unambiguously.
The task to extract all information contained in natural language text and to resolve
all shades of ambiguity is so complex and would require such an amount of world knowledge
that it would overburden any system development.

For specific tasks such as machine translation between particular languages, a lot of this
extracted and disambiguated information will be irrelevant, thus wasting both computing
resources at runtime and human resources at development time.

On the other hand, classical transfer systems, which during parse time analyze the source text
only up to some syntactic level and avoid a semantic analysis, will often not only face
insurmountable difficulties in selecting the correct target concepts during the transfer phase,
but also experience significant control problems during the parse phase itself, e.g.\ when making
decisions concerning preposition phrase attachment.
We therefore include a ``shallow''
semantic level during parsing, thereby substantially incorporating a major characteristic
of the interlingua approach.

Another argument in favor of interlingua is that it
requires only one analysis and one generation module per language involved, thus only $2*n$
modules for $n$ languages, as opposed to an additional $n*(n-1)$ transfer modules.
This argument could be
refuted by finding a way to largely automatically construct a transfer module from
language $L_{A}$ to $L_{C}$ from existing modules for transfer from $L_{A}$ to $L_{B}$ and
from $L_{B}$ to $L_{C}$. It appears that this would be quite feasible to do given the way
our system is set up. This last argument is nevertheless somewhat moot
as long as machine translation is not mature enough to perform well with a single language pair.

\section{Major System Components}

\begin{figure}[htb]
\epsfxsize=16.4cm
\centerline{\epsfbox{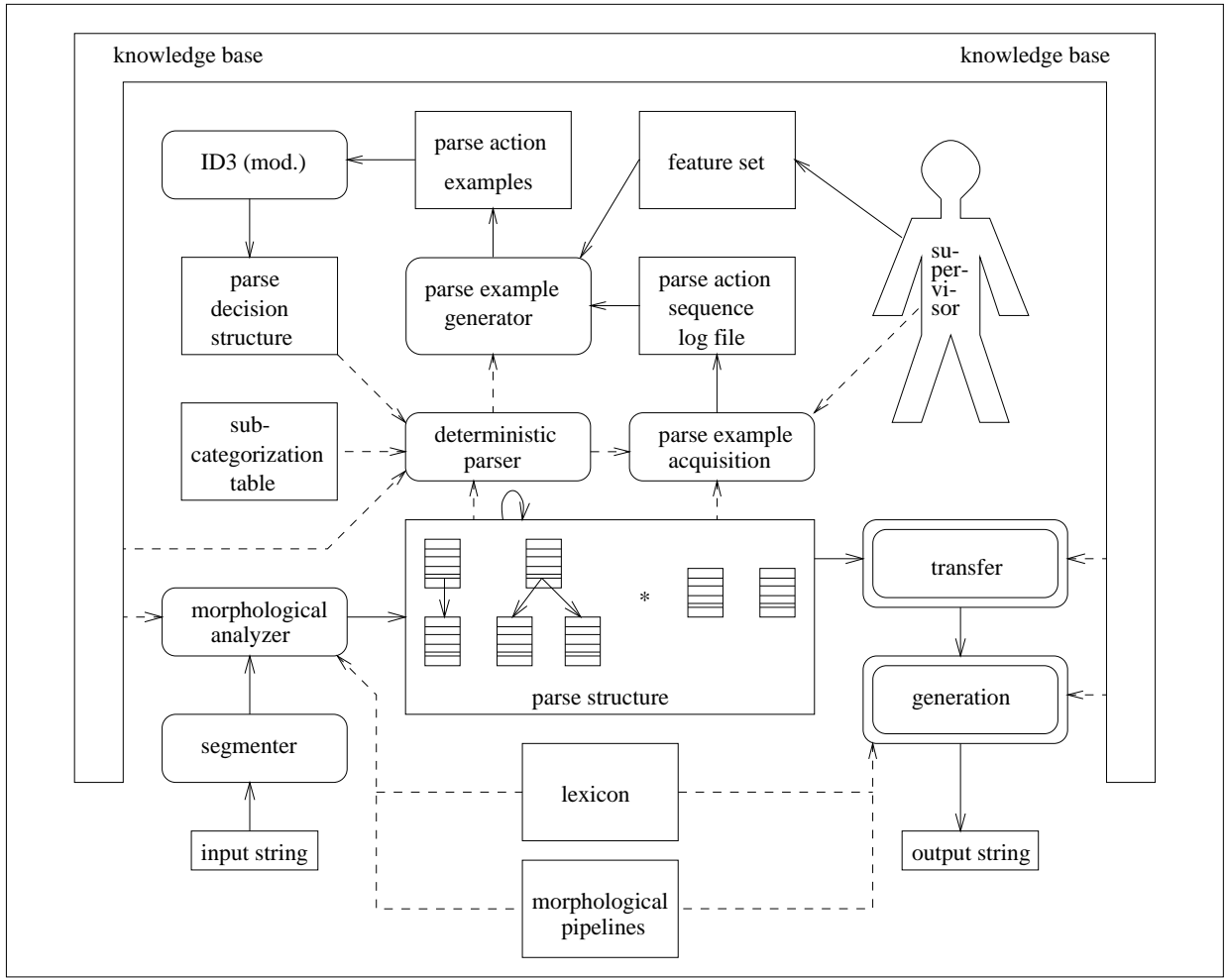}}
\caption{System Architecture Overview with Focus on Parsing}
\label{fig_sys_arch1}
\end{figure}

Figure~\ref{fig_sys_arch1} gives an overview of the system architecture.
Boxes represent data, ovals represent programs and arrows indicate the
flow of information. In the case of solid arrows, the entire data structure is passed on,
whereas in the case of the dashed arrows, only the information requested by
the recipient is passed on.

Before the parsing itself starts, the input string is segmented into a list of words 
including punctuation marks, which then are sent through a morphological analyzer.
The parsing itself is broken into an ordered sequence of small and manageable parse 
actions. Using the output of the morphological analyzer as an initial parse state,
the {\it deterministic parser} applies a sequence of {\it parse actions} to the
parse structure until it decides that it is done. The parse structure is composed
of a {\it parse stack} and an {\it input list}. The asterisk in the parse structure
in figure~\ref{fig_sys_arch1} marks the boundary between the parse stack (left) and
the input list (right) and can be interpreted as the `current position' in a
partially parsed sentence. The two most typical parse actions, {\it shift}, which
moves an element from the input list to the parse stack, and {\it reduce}, which
combines one or more elements on the parse stack into a more complex element, give
the shift-reduce parser its name. The input list is initialized with the output
of the morphological analyzer and the result of the parsing is the typically single
but complex element that is left on the parse stack. The resulting parse tree is then sent
through a transfer module that maps it to an equivalent tree in the target language
and a generation module that reorders the tree and generates the proper surface
words and phrases in the target language.

The key data structure that the deterministic parser uses to decide what the next
parse action should be is the {\it parse decision structure}. This structure, a hybrid
extension of decision trees and lists, is learnt from parse action examples. Parse action examples
are collected by interactively running training sentences through the parser.
Initially, when there are no parse action examples and hence no parse decision structure yet, 
a human supervisor provides the full and correct sequence of parse actions for a 
sentence.  These are collected in a {\it log file}, which, together with the feature set, 
is used to construct a set of full parse action examples. The feature set is 
provided by the supervisor and describes the context of a parse state. The same
feature set is used for all parse action examples.  For each parse step, the parse action
example generator makes the parse engine of the deterministic parser automatically 
determine the values for all
features in the feature set and then complements the feature vector with the correct 
parse action from the log file as the classification of the parse action example. The
resulting parse action examples are then used by a machine learning program, an extension of
the classic decision tree constructor ID3 \cite{quinlan:ijcai87}, to build the parse 
decision structure. Starting
with the second sentence, this parse decision structure can be used by the 
deterministic parser to propose the next parse action to the supervisor. As the
number of parse action examples grows and the parse decision structure becomes more and more
refined, the supervisor has to correct fewer and fewer of the proposed parse actions.

Note that we directly use the proper parse action sequence to train our parser,
whereas in most empirical methods, the example input consists of the final parse
trees only. By providing
examples with actual parse actions to the machine learning unit, our system provides
more information and a more direct and therefore better guidance.
The various types of background knowledge that support the parsing process (lexicons,
morphological pipelines, knowledge base and subcategorization table) are described
in the chapter~\ref{ch-background}.

\subsection{Features}

To make good parse decisions, a rich context in the form of a wide range of features 
at various degrees of abstraction have to be considered.
To express such a wide range of features, we defined a {\it feature language}.
The following examples,
for easier understanding rendered in English and not in feature language syntax,
further illustrate the expressiveness of the feature language:
\begin{itemize}
\item the general syntactic class of $frame_{-1}$ (the top frame of the parse 
      stack\footnote{The parse stack and the input list form the parse structure.
      For more details, see section~\ref{sec-core-parsing-mechanism}.}):
   e.g.\ verb, adj, np,
\item whether or not $frame_{-1}$ could be a nominal degree adverb,
\item the semantic role of $frame_{-1}$ with respect to $frame_{-2}$:
   e.g.\ agent, time; this involves pattern matching with corresponding entries
    in the verb subcategorization table,
\item whether or not $frame_{-2}$ and $frame_{-1}$ agree as np and vp.
\end{itemize}

The feature collection is basically independent from the supervised parse action 
acquisition. Before learning a decision structure for the first time, the supervisor
has to provide an initial set of features that can be considered obviously relevant.
Later, the feature set can be extended as needed.
All concepts and methods introduced in this overview are described in detail in the
following chapters, notably features in section~\ref{sec-features}.

   %\chapter{System Architecture} \label{ch-overview}
\chapter{Background Knowledge}
\label{ch-background}

In our approach to natural language processing, learning is complemented by some prespecified
knowledge. This knowledge can be accessed both indirectly through the use of features
when learning and using decision structures for parsing and transfer, as well as directly,
as happens for example with the morphological knowledge during generation.
The background knowledge is organized in a general
{\it knowledge base (KB)}, {\it monolingual lexicons}, {\it a bilingual dictionary},
{\it subcategorization tables} and what we call {\it morphological pipelines}.

Much of the background knowledge is used at various stages of the translation process,
e.g.\ the general KB at virtually all stages and the morphological pipelines for both pre-parsing
analysis and generation, while other knowledge is used exclusively in a single phase, as e.g.\
the bilingual dictionary in transfer. 

Before describing in further detail {\it what} background knowledge we use, we first want to give
reasons {\it why} we use it.
In chapter~\ref{ch-intro} we have already explained why we use machine learning in
making parse and transfer decisions: essentially because the lexical and structural ambiguity
of unrestricted natural language is so complex that hand-crafted approaches as tried in the past
have failed. Why then not learn everything, and do without any background knowledge?
Statistical approaches such as SPATTER \cite{magerman:acl95} or the Bigram Lexical Dependency
based parser \cite{collins:acl96} produce parsers on very limited linguistic background
information and Inductive Logic Programming systems such as CHILL \cite{zelle:ml94,zelle:thesis95}
have even generated linguistically relevant categories such as {\it animate}.

In short, the background knowledge we use is at least qualitatively easy to provide and mostly
already conceptually available in the form of traditional (paper) dictionaries and grammar books;
using available background knowledge, we can let the learning focus better on core decision
problems in parsing and transfer, on decisions that are known to be extremely hard if not impossible
to make using hand-written rules. With a better focus, training sizes can be smaller, because less
has to be learnt; no wheels need to be reinvented. 

The nature of the background knowledge we use is much simpler than what we try to learn,
because it is what we call {\it micromodular}.
We use the term {\it micromodular} for knowledge that is composed of very small pieces
that are very independent of each other, e.g.\ single entries in a monolingual
lexicon, a bilingual dictionary, or a subcategorization table.
These entries are without any claim of completeness: the nominal lexical entry for {\it high}
for example only asserts that
there is a concept which can manifest itself as a noun with the stem form {\it high},
but does {\it not} preclude any other interpretations for {\it high} or indicate which 
interpretation might be most appropriate in a given context.
Similarly, subcategorization table entries and other elementary assertions make no claim about
the existence or preferences of potentially competing entries. This micromodular nature of the
background knowledge makes it easy for humans to express it.
The acquisition of background knowledge is therefore mostly a quantitative problem, calling for
tools and methods that allow rapid information acquisition.
In contrast to this micromodular background knowledge, the knowledge encompassing parse grammars
is qualitatively very complex; it is this knowledge that has to disambiguate natural language text
with all its regularities, sub-regularities, pockets of exceptions, and idiosyncratic exceptions.

The following sections describe the different types of background knowledge in detail.
Section~\ref{sec-acquisition} shows how additions to the KB and lexicon are supported.

\section{Knowledge Base}
\label{sec-kb}
The {\it knowledge base (KB)} consists of concepts linked by relations. The vast majority of its 
currently 4356 concepts are semantic or syntactic. 

Most semantic concepts represent a specific word in a specific language. 
A separate notion of a concept is necessary, because words are too ambiguous. Not only can
words have different meanings in different languages, e.g.\ {\it gift} can mean {\it poison} in
Swedish; some words, called homonyms, have unrelated meanings, e.g.\ {\it go}, a common verb and
a Japanese board game, or have related meanings, but differ in their syntactic function, in which
case the word is sometimes also referred to as a homomorph, as e.g.\ for {\it increase} which
can be both a verb and a noun. Finally, concepts allow to distinguish between different shades of
meanings of a word.

Even if two words from the same or different language basically share the same meaning, they are
still represented by different concepts, because their meanings typically never match perfectly
due to possibly ever so slight different connotations.

Semantic concepts are typically represented
by identifiers of the form\\ 
I-$<$language-tag$>$$<$part-of-speech-tag$>$-$<$specific-tag$>$, e.g.\
I-EN-PILOT,
I-EN-TANGIBLE-OB\-JECT
or I-GADJ-DEUTSCH,
where the `I' stands for {\it internal concept}, the `E' for {\it English}, the `G' for {\it German},
the `N' for {\it noun}, the `ADJ' for {\it adjective}. Other semantic concepts include semantic roles,
e.g.\ R-AGENT or R-TO-LOCATION.

Syntactic concepts include parts of speech, prefixed by ``S-'', e.g.\ S-VERB or S-TR-VERB, for verbs
and, more specifically, transitive verbs respectively. Other syntactic concepts include forms, e.g.\
F-NUMBER, F-PLURAL, F-TENSE, F-FINITE-TENSE, F-PAST-TENSE and syntactic roles such as R-SUBJ.

\begin{figure}[htb]
\epsfxsize=16.4cm
\centerline{\epsfbox{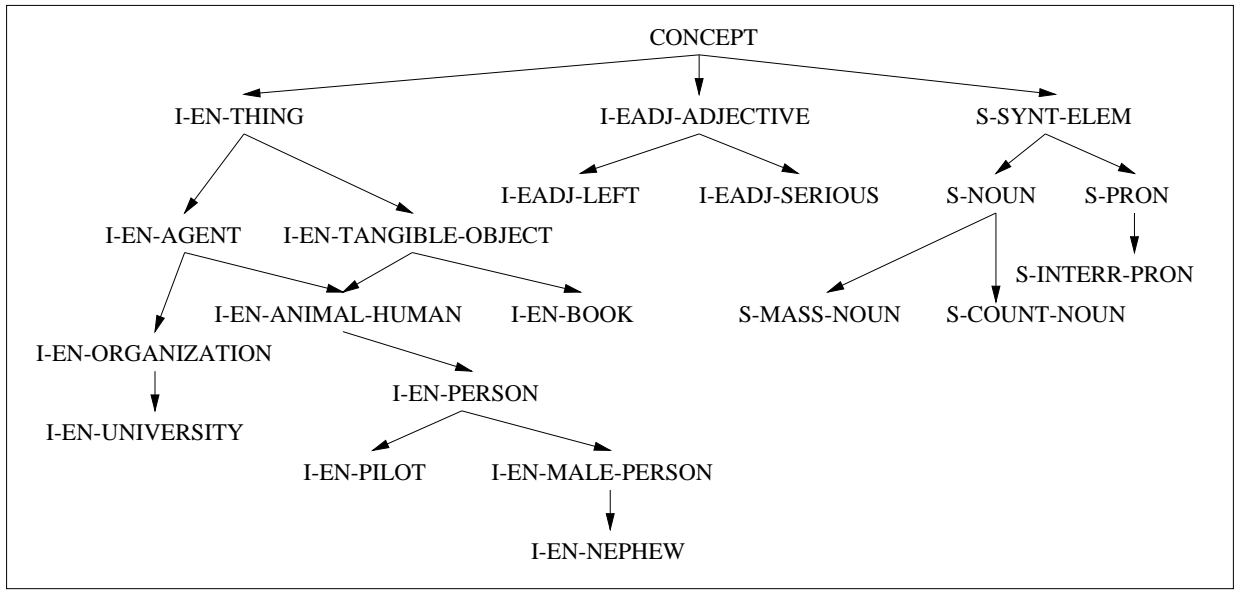}}
\caption{A subsection of the knowledge base}
\label{fig-kb_hierarchy}
\end{figure}

In its current form, there is only one relation in the KB, the binary {\it is-a}.
4518 such {\it is-a} links connect the 4356 concepts to form an acyclic graph.
More than 95\% of the network is semantics, the rest syntax (90 links), forms (to express
concepts for case, tense etc.; 52 links), roles (46 links) and miscellaneous (10 links).

While specific words are represented by one or more concepts specifically tied to the language the word
is from, many generalizations such as 
I-EN-TANGIBLE-OBJECT are shared as super-concepts 
for concepts representing words from different languages.

More than 90\% of the KB concepts are leaves that lexical entries refer to.
The granularity of the KB is basically application driven. Concepts can for example be grammatically
relevant because they can help determine
\begin{itemize}
  \item the need for an article ({\it S-COUNT-NOUN})
  \item the proper antecedent based on natural gender ({\it I-EN-MALE-PERSON})
  \item the semantic roles a constituent can fill ({\it I-EN-AGENT}).
\end{itemize}
As the direct superconcepts of I-EN-BOOK and I-EADJ-LEFT in figure~\ref{fig-kb_hierarchy}
already suggest, a very coarse classification is often sufficient. 
A distinction between mammals and reptiles for example would hardly help in parsing and machine translation.
We don't even have a concept for {\it noninherent} adjectives, a class of adjectives that typically 
can not be used in predicative position (as in *``The president is {\it former}.''). 
While this could be a useful concept for a grammar checker, 
it probably doesn't serve a purpose in machine translation, where the source
sentences are expected to be (more or less) correct and where it is probably best to conserve any
semantic oddity in the translation.

\section{Monolingual Lexicons}
\label{sec-lexicons}

A lexicon is a set of lexical entries, which link
surface words to semantic concepts as used in the KB. Besides the surface word and a corresponding
concept, the lexicon entry will always include the part of speech for the word. 
Optional attributes include:
\begin{itemize}
  \item irregular forms
  \item {\it value}, the numerical value for numerals
  \item the predicates {\it is-short-form} and {\it is-abbreviation}
  \item {\it grade} for superlative and comparative forms of adjectives
  \item {\it prob-rank} a relative rank of a priori likelihood for ambiguous words (see explanation below)
  \item {\it dummy-index, class, restriction} and {\it forms} for dummy words (see explanation below)
  \item {\it components} for contracted words (e.g.\ German ``im'' = ``in dem'')
  \item {\it connective-form} for German noun compounds {\it (as explained later)}
  \item {\it gender, genetive-type, plural-type} for German nouns
  \item {\it separable-prefix, has-inseparable-prefix, is-strong-verb, perfect-auxiliary}
for 
German verbs
\end{itemize}

The attribute {\it prob-rank} affects the order of morphological analyses of words with ambiguous
part of speech at the beginning of parsing, after morphological processing. Based on the
following excerpt from the English lexicon, the English word ``man'' will be recognized
as both a noun and a verb. With a default {\it prob-rank} of 1, the nominal analysis
will be placed at the first position of the list of alternatives, and the verbal analysis 
at place two, since it has a {\it prob-rank} value of 2, indicating that it is a priori less likely.\\

The English lexicon currently contains 3015 entries, fully covering the restricted WSJ vocabulary.
The German stands at 1039 entries, sufficient to cover all words that could occur in our translation
training and test sentences.

\noindent {\bf From the English lexicon}:
\begin{verbatim}
((LEX "man") (CONCEPT I-EN-MAN) (SYNT S-COUNT-NOUN) 
  (IRR-FORMS ("men" (NUMBER F-PLURAL))))
((LEX "man") (CONCEPT I-EV-MAN) (SYNT S-TR-VERB) 
  (PROPS (PROB-RANK 2)))
((LEX "Sen") (CONCEPT I-EN-SENATOR-TITLE) (SYNT S-COUNT-NOUN) 
  (PROPS (IS-ABBREVIATION TRUE)))
((LEX "sixty") (CONCEPT I-ENUM-SIXTY) (SYNT S-CARDINAL) 
  (PROPS (VALUE 60)))
((LEX "be") (CONCEPT I-EV-BE) (SYNT S-AUX) 
  (IRR-FORMS
    ("am" (TENSE F-PRES-TENSE) (PERSON F-FIRST-P) (NUMBER F-SING))
    ("are" (TENSE F-PRES-TENSE) (PERSON F-SECOND-P) (NUMBER F-SING))
    ("is" (TENSE F-PRES-TENSE) (PERSON F-THIRD-P) (NUMBER F-SING))
    ("are" (TENSE F-PRES-TENSE) (NUMBER F-PLURAL))
    ("was" (TENSE F-PAST-TENSE) (PERSON F-FIRST-P) (NUMBER F-SING))
    ("were" (TENSE F-PAST-TENSE) (PERSON F-SECOND-P) (NUMBER F-SING))
    ("was" (TENSE F-PAST-TENSE) (PERSON F-THIRD-P) (NUMBER F-SING))
    ("were" (TENSE F-PAST-TENSE) (NUMBER F-PLURAL))
    ("been" (TENSE F-PAST-PART))
    ("being" (TENSE F-PRES-PART))))

((LEX "'be") (CONCEPT I-EV-BE) (SYNT S-AUX)  ;;; contracted form
  (IRR-FORMS
    ("'m" (TENSE F-PRES-TENSE) (PERSON F-FIRST-P) (NUMBER F-SING))
    ("'re" (TENSE F-PRES-TENSE) (PERSON F-SECOND-P) (NUMBER F-SING))
    ("'s" (TENSE F-PRES-TENSE) (PERSON F-THIRD-P) (NUMBER F-SING))
    ("'re" (TENSE F-PRES-TENSE) (NUMBER F-PLURAL))
    (NO-FORM (TENSE F-PRES-INF))
    (NO-FORM (TENSE F-PAST-TENSE))
    (NO-FORM (TENSE F-PAST-PART))
    (NO-FORM (TENSE F-PRES-PART)))
  (PROPS (IS-SHORT-FORM TRUE)))
((LEX "better") (CONCEPT I-EADJ-GOOD) (SYNT S-ADJ) 
  (PROPS (GRADE COMPARATIVE)))
((LEX "SOMEBODY_1") (SYNT S-NP) 
  (PROPS (DUMMY-INDEX 1) (CLASS I-EN-PERSON)))
\end{verbatim}

The lexicon does not only contain normal word entries, but also {\it dummies}. These words
will not occur in any normal text to be processed, but only in 
the bilingual dictionaries, where they mark the place for words or phrases. 
A typical English component of such a dictionary entry is the verb phrase
{\bf ``to get SOMEBODY\_1 TO\_DO\_SOMETHING\_2''}.
The English lexicon describes SOMEBODY\_1 as a dummy for a noun phrase semantically
restricted to be of class I-EN-PERSON. The possible restriction attributes {\it forms} and
{\it restriction} offer alternative ways to indicate restrictions by respectively providing 
limitations on forms, e.g.\ tense or number, or through the reference to the name of a 
hard-coded predicate.\\

\noindent {\bf From the German lexicon}
\begin{verbatim}
((LEX "Mann") (CONCEPT I-GN-MANN) (SYNT S-COUNT-NOUN)
  (PROPS (PLURAL-TYPE F-INFL+ER) 
         (GENETIVE-TYPE F-INFL-ES) 
         (GENDER F-MASC)))
((LEX "abbestellen") (CONCEPT I-GV-ABBESTELLEN) (SYNT S-TR-VERB)
  (PROPS (IS-STRONG-VERB :FALSE) 
         (SEPARABLE-PREFIX "ab")
         (HAS-INSEPARABLE-PREFIX :TRUE)
         (HAS-SEPARABLE-PREFIX :TRUE)))
((LEX "im") (CONCEPT I-GP-IM) (SYNT S-PREP-PLUS-ART) 
  (PROPS (COMPONENTS ("in" "dem")) 
         (HAS-COMPONENTS TRUE)))
((LEX "Westen") (CONCEPT I-GN-WESTEN) (SYNT S-COUNT-NOUN) 
  (PROPS (GENETIVE-TYPE F-INFL-S)
         (PLURAL-TYPE F-INFL-)
         (GENDER F-MASC)
         (CONNECTIVE-FORM "West")))
\end{verbatim}

The identifiers for the genetive and plural types (``F-INFL...'') already indicate the way in 
which the respective forms are 
constructed. The sign following F-INFL, + or -, indicates whether or not the form requires an
umlaut and the rest of the identifier represents the respective ending. In the case of {\it Mann} 
(which means {\it man}), the
(singular) genetive form is {\it Mannes} and the plural (nominative) form is {\it M\"{a}nner}.
Other forms can be constructed in a regular way from these forms.\\
All German nouns possess a grammatical gender (masculine, feminine or neuter).\\
German nouns can take on a special connective forms in noun compounds. The head noun (at the
last position) remains in the usual form with the appropriate number and case, whereas modifying nouns
appear in their connective forms, typically not separated by any spaces, e.g.\ {\it Westafrika}.

\section{Acquisition of Lexicon and KB}
\label{sec-acquisition}
The entries in the KB and the lexicons are basically entered manually, facilitated
by some useful tools though. So, to get the above entries for ``Mann'' and ``abbestellen'' ({\it to cancel}),
along with some others, one can just enter

\noindent{\bf Aquisition example 1:}
\begin{verbatim}
e "der Mann -es +er" nc spec I-EN-MALE-PERSON "der Junge -n -n" 
  "der Bruder -s +"
\end{verbatim}
{\footnotesize {\it Junge}: ``boy''; {\it Bruder}: ``brother''; {\it nc} stands for {\it count noun}, 
{\it -es} etc. for genetive and plural endings}\\
This entry command produces the following system feedback:
\begin{verbatim}
To lex:
   ((LEX "Mann") (CONCEPT I-GN-MANN) (SYNT S-COUNT-NOUN) 
                 (GENDER F-MASC) (PLURAL-TYPE F-INFL+ER) 
                 (GENETIVE-TYPE F-INFL-ES))
   ((LEX "Junge") (CONCEPT I-GN-JUNGE) (SYNT S-COUNT-NOUN) 
                  (GENDER F-MASC) (PLURAL-TYPE F-INFL-N) 
                  (GENETIVE-TYPE F-INFL-N))
   ((LEX "Bruder") (CONCEPT I-GN-BRUDER) (SYNT S-COUNT-NOUN) 
                   (GENDER F-MASC) (PLURAL-TYPE F-INFL+) 
                   (GENETIVE-TYPE F-INFL-S))
To KB:
   (I-GN-MANN IS-A I-EN-MALE-PERSON)
   (I-GN-JUNGE IS-A I-EN-MALE-PERSON)
   (I-GN-BRUDER IS-A I-EN-MALE-PERSON)
OK [y n l k]?
\end{verbatim}

For German verbs,
the double dash as used in the following example is not part of the word, 
but indicates that the preceding prefix is separable;
similarly, a single dash indicates an inseparable prefix.

\newpage
\noindent{\bf Aquisition example 2:}
\begin{verbatim}
e "ab--be-stellen" vt spec I-EV-PROCESS "ab--stellen" 
  "be-stellen" "stellen"
\end{verbatim}
This entry command produces the following system feedback:
\begin{verbatim}
To lex:
   ((LEX "abbestellen")
       (CONCEPT I-GV-ABBESTELLEN) (SYNT S-TR-VERB)
       (SEPARABLE-PREFIX "ab") (HAS-SEPARABLE-PREFIX TRUE)
       (HAS-INSEPARABLE-PREFIX TRUE) (IS-STRONG-VERB FALSE))
   ((LEX "abstellen")
       (CONCEPT I-GV-ABSTELLEN) (SYNT S-TR-VERB)
       (SEPARABLE-PREFIX "ab") (HAS-SEPARABLE-PREFIX TRUE)
       (HAS-INSEPARABLE-PREFIX FALSE) (IS-STRONG-VERB FALSE))
   ((LEX "bestellen")
       (CONCEPT I-GV-BESTELLEN) (SYNT S-TR-VERB)
       (HAS-SEPARABLE-PREFIX FALSE) (HAS-INSEPARABLE-PREFIX TRUE)
       (IS-STRONG-VERB FALSE))
   ((LEX "stellen")
       (CONCEPT I-GV-STELLEN) (SYNT S-TR-VERB)
       (HAS-SEPARABLE-PREFIX FALSE) (HAS-INSEPARABLE-PREFIX FALSE)
       (IS-STRONG-VERB FALSE))
To KB:
   (I-GV-ABSTELLEN IS-A I-EV-PROCESS)
   (I-GV-ABBESTELLEN IS-A I-EV-PROCESS)
   (I-GV-BESTELLEN IS-A I-EV-PROCESS)
   (I-GV-STELLEN IS-A I-EV-PROCESS)
OK [y n l k]?
\end{verbatim}

Entries for English verbs and nouns are simpler because English nouns don't have
a grammatical gender or case and the English equivalent of a prefix is always
separate (e.g.\ ``take off''). There are special tables for irregular nouns and verbs.
The irregular forms contained in these tables are automatically added to the
lexical entries. 

The part of speech of a word, as well as applicable information about grammatical gender, 
inflectional class, prefixes etc., are commonly listed in traditional (paper) dictionaries,
but will typically be obvious to the educated speaker of a language.

Section~\ref{sec-lexical-expansion} in the chapter on future work sketches how the lexical
acquisition process could be further automated.

\section{Bilingual Dictionary}
\label{sec-dictionary}

A {\it bilingual dictionary} links words and expressions between different languages
and is used for translation.
The {\it surface} dictionary is a listing of words and phrases very similar to as one would find them
in a traditional (paper) dictionary. Its format was designed to be very intuitive and user-friendly
so that additions (or changes) can be made very easily. The following samples from the English-German
transfer lexicon is presented in a table for better legibility and closely follows
the actual format of the surface dictionary file, which, for the first table entry would be
{\it (``be'' S-VERB ``sein'')}.\\

\begin{center}
\begin{tabular}{|ll|} \hline
``be''  &                               S-VERB    \\  ``sein'' & \\ \hline
``to be dissatisfied with SOMETHING\_1'' & S-VP    \\  ``mit ETWAS\_1 unzufrieden sein'' & \\ \hline
``to be lucky''                         & S-VP    \\  ``Gl\"{u}ck haben'' & \\ \hline
``because''                             & S-CONJ  \\  ``weil'' & \\ \hline
``book''                                & S-NOUN  \\  ``Buch'' & \\ \hline
``downtown PLACENAME\_1''                & S-ADV   \\  ``im Stadtzentrum von ORTSNAME\_1'' & S-PP \\ \hline
``know''                                & S-VERB  \\  ``kennen'' & \\ \hline
``know''                                & S-VERB  \\  ``wissen'' & \\ \hline
``to make SOMETHING\_1 equal to SOMETHING\_2'' & S-VP \\ ``ETWAS\_ACC\_1 auf ETWAS\_ACC\_2 bringen'' & \\ \hline
``primarily''                           & S-ADV   \\  ``in erster Linie'' & S-PP \\ \hline
``some''                                & S-ADJ   \\  ``einige'' & \\ \hline
``some''                                & S-ADJ   \\  nil & \\ \hline
``up QUANTITY\_1''            & S-PARTICLE-PHRASE  \\  ``ein Plus von QUANTITAET\_1 '' & S-NP \\ \hline
``it takes SOMEBODY\_3 SOMETHING\_1     & S-CLAUSE \\
                                    TO\_DO\_SOMETHING\_2'' & \\
     ``JEMAND\_3 braucht ETWAS\_ACC\_1,                           & \\
                                        um ETWAS\_ZU\_MACHEN\_2`` & \\ \hline
I-EART-INDEF-ART                        &         \\  I-GART-INDEF-ART & \\ \hline
I-EN-PERSONAL-PRONOUN                   &         \\  I-GN-PERSONAL-PRONOUN & \\ \hline
I-EPRT-'S                               &         \\  I-GP-GEN-CASE-MARKER & \\ \hline
\end{tabular}\\
\end{center}

A typical dictionary entry consists of a pair of words or phrases along with
part of speech restrictions. Note that the parts of speech don't necessarily
have to be the same (e.g. ``primarily''/``in erster Linie'').

Phrases can contain variables, e.g.\ {\it SOMETHING\_1}, {\it PLACENAME\_1}, or {\it ETWAS\_ZU\_MA\-CHEN\_2}.
Variables in the source and target phrase are linked by a common index.
They have special entries in their respective monolingual lexicons, which list any
syntactic, semantic or form restrictions. The German lexicon for example restricts {\it JEMAND\_3}
to be a noun phrase (SYNT S-NP), a person (CLASS I-EN-PERSON), and nominative case (CASE F-NOM).

The dictionary can also contain a pair of concepts, e.g.\ I-EART-INDEF-ART/I-GART-INDEF-ART.
There are only relatively few such direct concept pair entries, but these entries allows the 
pairing of concepts whose level of abstraction can not easily be captured by specific words.

Note that a word or phrase can have multiple translations 
(e.g.\ ``know'' \hspace{-0.5mm}$\leftrightarrow$\hspace{-0.5mm} 
``kennen''\hspace{-1mm}/\hspace{-0.5mm}``wissen'') or an empty one 
(e.g.\ ``some'' $\leftrightarrow$ ``eingige''/{\it nil}).
(Chapter~\ref{ch-transfer} will describe how the system learns to make the proper choice.)

When a {\it surface} dictionary is loaded, it is compiled into an {\it internal} dictionary.
All strings are parsed (in their respective languages) and mapped to a KB concept or a complex parse tree. 
So the internal dictionary does not map from surface word or phrase to surface word or phrase, 
but from parse tree to parse tree. 

The transformation of transfer entries from surface strings to their
parse trees can be accomplished by using the {\bf same} parsers that are used for the parsing
of normal input surface sentences. Of course parsers for both the source and the target
language are necessary, but since the phrases in the lexicon are not anywhere as complex and ambiguous 
as normal input surface sentences, the `parsing power' requirements for the bilingual dictionary
are relatively small, so that fairly few parse examples are necessary for a language that is only used
as a machine translation target language.

The intuitive surface representation, which closely follows the format of good traditional
dictionaries, enormously facilitates the building of the transfer lexicon;
it is very transparent, can be built, extended and checked easily, and is not very susceptible to errors.

\section{Subcategorization Tables}
\label{sec-subcat-tables}
Consider the following two sentences:
\begin{enumerate}
  \item I booked a flight to New York.
  \item I sent a letter to New York.
\end{enumerate}
Superficially, both sentences look very similar, but in sentence 1, the prepositional phrase
{\it to New York} has to attach to the preceding noun phrase {\it a flight}, whereas in sentence 2,
it belongs directly to the predicate {\it sent}. When deciding where to attach the prepositional
phrase, an analysis of the parts of speech will not suffice, because it will not discriminate
between the two sentences.

The fundamental difference lies in the different argument structure of the verbs and nouns. While both take
a direct object, {\it to send} also allows an indirect object or a {\it to-location} argument.
Additionally, the noun {\it flight}, a nominal form of {\it to fly}, allows a {\it to-location} argument.

This argument information can be represented in {\it subcategorization tables}. 
The following is a subset of the currently 242 entries in the {\bf English Verb Patterns}. 
The asterisk (*) indicates that the
following argument is optional.

\begin{verbatim}
("to book SOMETHING")

("to decline *I-EN-QUANTITY *C-TO-QUANT"
    (SUBJ = THEME) (OBJ = DIFF-QUANT))
("to decline by I-EN-QUANTITY *C-TO-QUANT"
    (SUBJ = THEME) (APP = DIFF-QUANT))
("to decline TO-DO-SOMETHING"
    (INF-COMPL = THEME))

("to get SOMETHING"
    (SUBJ = BEN))
("to get SOMEBODY TO-DO-SOMETHING"
    (OBJ = BEN) (INF-COMPL = THEME))
("to get out of SOMETHING"
    (APP = THEME))
("to get through SOMETHING"
    (SUBJ = THEME) (APP = PRED-COMPL))
("to get under way"
    (SUBJ = THEME) (APP = PRED-COMPL))

("to send *SOMEBODY SOMETHING")
("to send SOMETHING C-TO-LOCATION")
\end{verbatim}

Each subcategorization table entry consists of a phrase, e.g.\ a verb phrase in a verb 
subcategorization table,
plus a mapping of syntactic to semantic roles for the phrase pattern. The default 
mappings are
\begin{itemize}
  \item subject (SUBJ) $\rightarrow$ agent
  \item object (OBJ) $\rightarrow$ theme
  \item indirect object (IOBJ) $\rightarrow$ beneficiary (BEN)
  \item infinitival complement (INF-COMPL) $\rightarrow$ purpose
  \item adjective complement (ADJ) $\rightarrow$ predicate complement (PRED-COMPL)
  \item C-TO-LOCATION $\rightarrow$ TO-LOCATION
  \item C-TO-QUANT $\rightarrow$ TO-QUANT
\end{itemize}

With this information, the subcategorization tables can not only be used to decide 
where to attach phrases, but also
help in finding the proper semantic role of the new sub-component in the larger phrase.

Assuming that the two sentences at the beginning of this section have been partially parsed into
a noun phrase, verb, noun phrase and prepositional phrase, i.e.\ 
(I) (sent/booked) (a flight/letter) (to New York),
the parser could now exploit the verb subcategorization table by evaluating the feature 
``(SEMROLE OF -1 OF -3)''.
As explained in more detail in section ~\ref{sec-features}, this feature is interpreted as 
``what is the semantic role of the last item (at position -1; here {\it to New York})
with respect to the item at position -3 (in our examples the verb)?''. 
After matching the partially parsed phrase to the best verb pattern, 
the system would return feature values `UNAVAIL' for {\it (I) (booked) (a flight) 
(to New York)} and `R-TO-LOCATION' for {\it (I) (sent) (a letter) (to New York)}.
Using this discriminating feature, a parser can make the decision that will
properly attach the prepositional phrase {\it to New York}.

The currently 242 entries of the verb subcategorization table have been entered manually.
Some dictionaries and grammar books such as \cite{hornby:dict74,engel:dt88} contain 
verb patterns that could serve as a basis for subcategorization tables, but the entries
were actually made without any such support, because the micromodular nature of the knowledge 
and its intuitive format already made the task easy.
In future extensions it might be useful to have a similar subcategorization table for nouns.

\section{Morphology}
\label{sec-morphology}
At least from a pragmatic computational standpoint, morphological processing has basically
already been solved for English as well as other Germanic and Romance languages, Japanese, 
and, we suspect, all other languages. Morphology is computationally much simpler than 
syntax or semantics, because it operates only very locally.

Some approaches might be more elegant or linguistically motivated than others, but
inflection tables, regardless of how many classes, cases, tenses, numbers, genders, voices, 
and levels of definiteness or politeness etc. have to be considered,
pose no problem to computational processing, nor do irregular forms, and any approach can 
be made fast by caching results.

In our system, the symmetric morpholgy module is used for both analysis and generation.
For analysis, given the surface form of a word, say {\it ((``increases''))}, it finds its
annotated stem-forms, e.g.
\begin{verbatim}
(((lex "increase")
  (surf "increases") 
  (synt s-count-noun)
  (forms (((number f-plural))))
  (concept i-en-increase))

 ((lex "increase")
  (surf "increases")
  (synt s-tr-verb)
  (forms (((tense f-pres-tense) (person f-third-p) (number f-sing))))
  (concept i-ev-increase)))
\end{verbatim}

\newpage \noindent
In the other direction, given the annotated stem-form, e.g.
\begin{verbatim}
 (("increase"
      (synt s-noun)
      (number f-plural)
      (concept i-en-increase)))
\end{verbatim}
it finds the corresponding surface form, here {\it ((``increases''))}.

Such generation is done by sending the annotated forms through a {\it morphological pipeline}, 
a data structure with elements that are interpreted to manipulate these annotated forms.
A morphological pipeline can automatically be inverted. A generation pipeline thus implicitely
also defines an analysis pipeline, which can then be used to gradually manipulate initially
unannotated forms to become fully annotated stem forms.

A morphological pipeline is a sequence of {\it pipeline elements}. Each of these elements
manipulates or filters an annoted form, yielding a {\it set} of annotated output 
forms.
Manipulations include the replacement of a word-ending string by another string,
change of stems, the addition or deletion of umlauts (for German) or the addition
of an annotation; filters include tests whether or not the word has a certain ending 
or whether or not it has a certain prefix. Finally the pipeline element can be a set
of parallel sub-pipelines, a so-called {\it fork} element.

The following figures illustrate this morphological generation and analysis. Morphological
pipelines are represented by polygons, the annotated forms at various stages of processing
by circles.

\begin{figure}[htb]
\epsfxsize=16.4cm
\centerline{\epsfbox{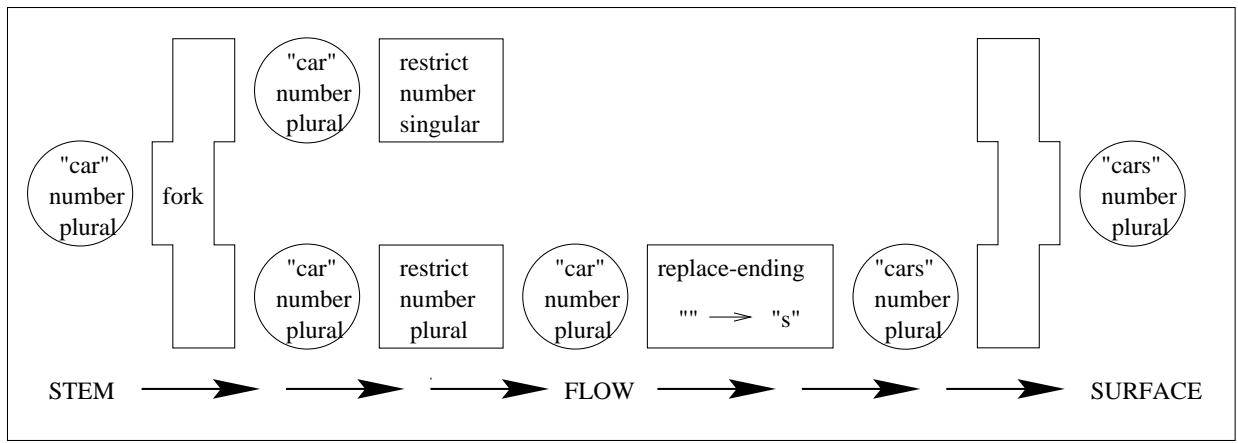}}
\caption{Generating the plural form of ``car''}
\label{fig-morph_pipeline2}
\end{figure}

Figure~\ref{fig-morph_pipeline2} depicts a simplified noun pipeline. With ``car'' annotated
by the restriction {\it number plural} as input, the pipeline forks, passing along the
data to both sub-pipelines. While the {\it restrict number singular} element filters out
the data because of the conflicting number, the plural filter element passes the data on to
the next element, which adds an ``s'' to the word. The closing fork bracket combines the
result of both sub-pipelines, in this case only the ``cars'' from the plural sub-pipeline.

\newpage
The pipeline depicted in figure~\ref{fig-morph_pipeline2} is textually represented as
\begin{verbatim}
   `((fork ((restrict number f-sing))
           ((restrict number f-plural)
            (repl "" "s"))))
\end{verbatim}

\begin{figure}[htb]
\epsfxsize=16.4cm
\centerline{\epsfbox{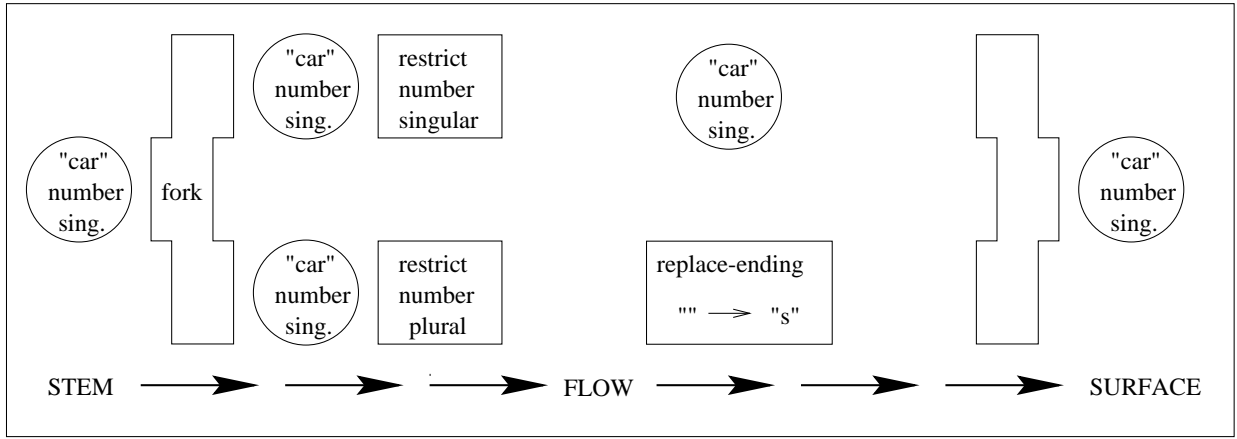}}
\caption{Generating the singular form of ``car''}
\label{fig-morph_pipeline1}
\end{figure}

Figure~\ref{fig-morph_pipeline1} depicts the same pipeline as figure~\ref{fig-morph_pipeline2}, 
but different data. In this case, the plural sub-pipeline filters out the incoming data.

\begin{figure}[htb]
\epsfxsize=16.4cm
\centerline{\epsfbox{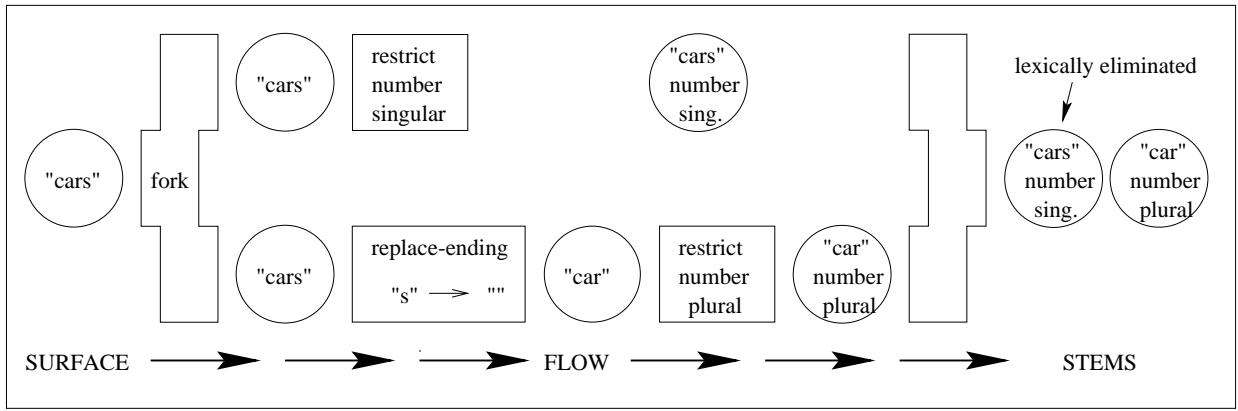}}
\caption{Analyzing the word ``cars''}
\label{fig-morph_pipeline3}
\end{figure}

Figure~\ref{fig-morph_pipeline3} depicts the automatically inverted pipeline, which is used
for analysis. Since the input data does not have any prespecified restrictions, the data
passes through both sub-pipelines and finally yields a result of two units. In the full system,
other pipeline elements add the filter {\it restriction synt noun} and perform a lexical 
check, which eliminates the stem form ``cars'', because no such noun stem will be found in the 
lexicon. Had the input been ``bus'' instead, the result {\it ``bus'' number singular} would have
been kept and {\it ``bu'' number plural} been lexically eliminated.

Once morphological pipelines are defined, they can be used as sub-pipelines thereafter. This can
be particularly useful since there are related morpological phenonoma even across parts of speech.
The standard plural formation, obviously more complex than the simplified version depicted above
{\it replace-ending ``'' $\rightarrow$ ``s''}, since it also has to cover like {\it bush/bushes},
{\it baby/babies}, e.g.\ closely resembles the third person singular present tense formation with
{\it push/pushes} and {\it cry/cries}.

When a pipeline element such as {\it replace-ending ``s'' $\rightarrow$ ``''} encounters a data unit
that it can not operate on, e.g.\ ``car'', where no final ``s'' can be deleted, that pipeline
just filters out that non-applicable data unit.

(Very) irregular forms are handled by forking off a special sub-pipeline that basically performs
an irregular form table lookup; and the ``regular'' sub-pipeline filters out any forms that conflict
with the irregular form table. Special pipeline elements for German handle umlauts and stem changes
for strong verbs, in the latter case with access to irregular verb tables.

\subsection{Specific Pipeline Elements}

The {\bf replacement} of a word-ending string by another string,
e.g.\ the pipeline element {\it (repl ``e'' ``ed'')}
\footnote{Since the system is implemented in Lisp and morphological changes tend to occur at
the end of words, words are actually represented as their reversed character lists, e.g.\
``increases'' is represented as ({\#}$\setminus$s {\#}$\setminus$e {\#}$\setminus$s
{\#}$\setminus$a {\#}$\setminus$e {\#}$\setminus$r {\#}$\setminus$c {\#}$\setminus$n {\#}$\setminus$i).
For easier reading,
strings are left as such in the following descriptions of the morphological module.}
would modify a form with ``increase'' to one with ``increased''; if the incoming word
does not match the first ending, the output is empty.

There are pipeline elements to add or delete umlauts (in German), e.g.\ {\it (add-umlaut)}
changes ``Apfel'' to ``\"{A}pfel'' and {\it (delete-umlaut)} does the opposite.

{\bf {\it (double-last-letter)}} and {\bf {\it (undouble-last-letter)}} do what their names
suggest, e.g.\ ``run'' is changed to ``runn'' and vice versa.

{\bf Restrictions} add annotations, and automatically filter out annotated forms with
conflicting annotations. E.g.\ {\it (restrict number f-plural)} changes an incoming 
(``increase'') to (``increase'' (number f-plural)), leaves an incoming (``increase'' (number f-plural))
as is, and returns an empty output for an incoming (``increase'' (number f-sing)).

Other specialized pipeline elements can perform {\bf stem changes} or take care of
{\bf irregular forms} using lookup-tables.

Filters, which return a set containing the incoming annotated form or an emtpy set,
check word-endings, e.g.\ (ends-in ``ed''), (not-ends-in ``ing'') or (memb (nth-last-letter x {\it n})
{\it $<$set of characters$>$}), which checks whether or not the {\it n}th last letter is a member of
the given character set.

The last important pipeline element of a pipeline is a {\it fork}, which channels incoming
input forms into parallel sub-pipelines.

\newpage
Using a previously defined sub-pipeline {\it needs-doubling} the following sub-pipeline describes
the formation of the regular past tense and past participle for English:
\footnote{The {\it x} is defined to refer to the word without its annotations.}

\begin{verbatim}
   `((fork ((repl "e" "ed"))                    ;;; like, liked
           ((not-memb (nth-last-letter x 2)
                      '(#\a #\e #\i #\o #\u))   ;;; pray, prayed
            (repl "y" "ied"))                   ;;; cry,  cried
           (,needs-doubling              ;; prev. def. sub-pipeline
            (double-last-letter)                ;;; stop, stopped
            (repl "" "ed"))         
           ((not-ends-in x "e")                 ;;; push, pushed
            (or (not-ends-in x "y")    
                (memb (nth-last-letter x 2)
                      '(#\a #\e #\i #\o #\u)))
            (not ,needs-doubling)        ;; prev. def. sub-pipeline 
            (repl "" "ed"))))
\end{verbatim}

A major advantage of these pipelines is that they can easily be inverted automatically,
so that the preceding sub-pipeline could also be used to find stem forms for regular
past tense or particple forms. The inverse for {\it (repl ``e'' ``ed'')} for example is
{\it (repl ``ed'' ``e'')} and inverting other pipeline elements and pipelines is equally
straightforward:\\

\begin{tabular}{|l|l|} \hline
element & inv(element) \\ \hline
replace-ending s1 s2 & replace-ending s2 s1 \\
{\it any filter} & {\it stays the same} \\
double-last-letter & undouble-last-letter \\
undouble-last-letter & double-last-letter \\
{\it sequence} (s1 ... sn) & {\it sequence} (inv(sn) ... inv(s1)) \\
fork (s1 ... sn) & fork (inv(s1) ... inv(sn)) \\
\hline 
\end{tabular}\\

Referring to the just defined pipeline as {\it pv-ed},
and given similarly defined sub-pipelines for present tense ({\it pv-prt}) 
and present participle ({\it pv-prp}) forms, a pipeline for regular English verbs would then
be\footnote{{\it append} concatenates two morphological pipelines}
\begin{verbatim}
   `((fork ((restrict tense f-pres-inf))
           (append '((restrict tense f-past-tense))
                   pv-ed)       ;;; previously defined sub-pipeline
           (append '((restrict tense f-past-part))
                   pv-ed)       ;;; previously defined sub-pipeline
           (append '((restrict tense f-pres-tense))
                   pv-prt)      ;;; previously defined sub-pipeline
           (append '((restrict tense f-pres-part))
                   pv-prp)))    ;;; previously defined sub-pipeline
\end{verbatim}

If we call this sub-pipeline {\it pv}, and with a similarly defined sub-pipelines for nouns,
{\it pn}, the top-level pipeline for regular words looks like:
\begin{verbatim}
   `((fork ,(append '((restrict synt s-verb))
                    pv)
           ,(append '((restrict synt s-noun))
                    '((restrict  person f-third-p))
                    pn)
           ((restrict synt s-prep))
           ((restrict synt s-adv))
           ...))
\end{verbatim}

The monolingual lexicon is used to link stem-forms to actual lexicon entries which then
provide more detailed information like (semantic) concept, a more detailed syntactic 
category and other annotations. 

A morphological pipeline for English was defined in a file with 178 lines, whereas the
German equivalent needed 640 lines.
  %\chapter{Background Knowledge} \label{ch-background}
\chapter{Segmentation and Morphological Analysis}          
\label{ch-preparsing}

Before the actual parsing starts, the input text is preprocessed. The text, already segmented into 
sentences in the original corpus, is first segmented into
words and punctuation marks; the resulting words are then morphologically analyzed. The list of
morphologically analyzed words and punctuation marks serves as input to the parser.

Segmentation and morphological analysis are not particularly challenging, because
both steps need to consider only very local properties of substrings and words
(at least for Germanic and Romance languages, at least
from a pragmatic perspective, and at least qualitatively).
Segmentation is fairly simple and morphology is well described in the literature 
\cite{lederer:dt69,quirk:engl85}.

\section{Segmentation}

Given a string of text, e.g.\ a sentence, the first task is to segment the string into a list of words,
numbers and punctuation marks. E.g.\ the string ``In the 1970s, the St. Louis-based 'company' traded at \$2.45.''
gets segmented into

\begin{flushleft}
((``In'' 0 0 2 2) (``the'' 2 3 6 6) (1970 6 7 11 11) (``s'' 11 11 12 13 ,)\\
 (D-COMMA 12 12 13 13) (``the'' 13 14 17 17) (``St'' 17 18 20 21 .)\\
 (D-PERIOD 20 20 21 21) (``Louis'' 21 22 27 27) (D-DASH 27 27 28 28)\\
 (``based'' 28 28 33 33) (D-APOSTROPHE 33 34 35 35) (``company'' 35 35 42 43 ')\\
 (D-APOSTROPHE 42 42 43 43) (``traded'' 43 44 50 50) (``at'' 50 51 53 53)\\
 (D-DOLLAR-SIGN 53 54 55 55) (2.45 55 55 59 60 .) (D-PERIOD 59 59 60 60))
\end{flushleft}

The four integers following the word, number or punctuation tokens indicate the {\it soft} and {\it hard}
boundary positions of a token. The {\it hard} boundaries include only the token itself, whereas the {\it soft}
boundaries include preceding whitespaces and some following punctuation.

\begin{quotation}
\begin{small}
This spacing information is maintained and used in later stages of parsing to reproduce precisely the 
surface
strings associated with partially parsed sentence segments. Consider for example the two sentences 
``The new Japans - Taiwan and South Korea - were thriving economically.'' and 
``In tests it has worked for many heart-attack victims.'' The character `-' serves as a dash in the
first example and as a hyphen in the second. Since spaces are not part of the segmented words per se,
we attach the spacing information to the segmented words so that the original surface string can
easily be reconstructed, as in the display of partially parsed sentences.

The spacing information also allows the surface string associated with partially parsed trees to be
kept in the proper order, which is not trivial, because of discontinuous elements. E.g.\ when
reducing the sentence elements ``has been'' + ``always'', we want to represent the resulting tree,
even if it is structured [[has been] always], by its portion of the original surface string 
``has always been'' for easier legibility. Finally, the spacing information is also used to properly
represent so-called empty categories (see~\ref{sec-empty-category}).
\end{small}
\end{quotation}

In many languages, including English, segmentation is relatively simple, 
since words are fairly easily identifiable as such.
For languages with significant compounding, such as German or Swedish, where noun compounds
are typically written as one word, an advanced segmenter needs to break noun compounds down into its
components.
Yet more difficult, in languages such as Japanese, where words are generally not separated by spaces,
sophisticated segmentation tools, such as JUMAN \cite{matsumoto:tr94}, have to be used. Since we currently
only translate {\it from} English, and therefore only need to parse general text in English, 
a relatively simple segmenter is sufficient and so that's all we have implemented at this point.

\section{Morphological Analysis and Parse Entries}
\label{morphological_analysis_and_parse_entries}

The words put out by the segmenter are then morphologically analyzed. As already described in
more detail in section~\ref{sec-morphology},
the morphological analyzer, given the surface form of a word, 
computes a list of {\bf {\it parse entries}}, i.e.\ frames containing information about the lexical 
and surface form of a word,
its syntactic and semantic category, form restrictions such as number, person, tense and other attributes
that apply to certain types of words only, such as a value attribute for numerals. 
{\it Complex} parse entries have {\it subentries}, i.e.\ parse entries annotated by one or more roles, that can
be syntactic or
semantic, and describe the function of the subentry with respect to the super-entry. 
{\it Simple} parse entries,
which is what the morphological analyzer returns, do not have such subentries.
Given for example the word {\it ``increases''}, the morphological analyzer returns two simple parse entries,
one for the nominal and one for the verbal interpretation of the word:
\begin{verbatim}
(((lex "increase")
  (surf "increases")
  (synt s-count-noun)
  (forms (((number f-plural))))
  (concept i-en-increase))

 ((lex "increase")
  (surf "increases")
  (synt s-tr-verb)
  (forms (((tense f-pres-tense) (person f-third-p) (number f-sing))))
  (concept i-ev-increase)))
\end{verbatim}

\noindent Note: 
\begin{itemize}
 \item ``{\bf lex}'' is the lexical root form of the head of the entry.
 \item ``{\bf surf}'' is the surface string associated with an entry.
 \item ``{\bf ssurf}'', not displayed here, contains spacing information.
 \item ``{\bf synt}'' is the part of speech.
 \item ``{\bf forms}'' displays any restrictions with respect to person, number, tense, etc.
             The following subsection contains a full list of these forms and their values.
 \item ``{\bf concept}'' is the semantic class.
 \item ``{\bf subs}'' denotes the list of sub-entries.
 \item ``{\bf props}'' includes any other properties.
\end{itemize}

\noindent Morphological ambiguity is captured within a parse entry. E.g.\ the verb "put"
would be analyzed as
\begin{verbatim}
(((lex "put")
  (surf "put")
  (synt s-tr-verb)
  (forms (((person f-first-p) (number f-sing) (tense f-pres-tense))
          ((person f-second-p) (number f-sing) (tense f-pres-tense))
          ((number f-plural) (tense f-pres-tense))
          ((tense f-pres-inf)) 
          ((tense f-past-part))
          ((tense f-past-tense))))
  (concept i-ev-put)))
\end{verbatim}
The forms present different morphological alternatives. Each form consists of a set of
form restrictions. Note that the attribute value pair ``{\bf (forms nil)}'' indicates that 
there are no morphological alternatives, which, in pathological cases, can sometimes result 
when incompatible forms are unified, as for example when determining the number of the 
noun phrase ``a cats'', where the article imposes {\it singular} and the noun {\it plural}. 
On the other hand
``{\bf (forms (nil))}'' indicates that there is one alternative without any restrictions.
Words from syntactic categories without any morphological variation, e.g.\ adverbs,
typically have this form.

\newpage
\subsection{List of Morphological Forms}

In our system, we use the following morphological forms. Possible values, as they might
arise in English or German, are given in parentheses.

\begin{itemize}
 \item  person (first, second, third)
 \item  number (singular, plural)
 \item  case (nominative, genitive, dative, accusative, object)
 \item  voice (active, passive)
 \item  tense (present, past, present perfect, past perfect, future, future perfect, 
               past participle, present participle, present infinitive, 
               present `to'-infinitive, past infinitive, past `to'-infinitive)
 \item  gender (masculine, feminine, neuter)
 \item  mood (indicative, subjunctive)
 \item  mode (declaration, wh-question, yn-question, imperative, exclamation)
 \item  aspect (continuous, non-continuous)
 \item  det-adj-ending (primary, secondary, uninflected) [for German adjectives]
 \item  you-caps (true, false) [for politeness capitalization of German pronouns]
 \item  ref-number (singular, plural) [number of the noun phrase that a pronoun refers to]
 \item  ref-person (first, second, third)
 \item  ref-gender (masculine, feminine, neuter)
\end{itemize}
            %\chapter{Segmentation and Morphological Analysis} \label{ch-preparsing}
\chapter{Parsing}          
\label{ch-parsing}

The parsing of unrestricted text, with its enormous lexical and structural
ambiguity, still poses a great challenge in natural language processing.
The traditional approach of trying to master the complexity of parse grammars
with hand-crafted rules turned out to be much more difficult than expected,
if not impossible. 
Many parsers leave ambiguity largely unresolved and just focus on an efficient
administration of ambiguity. The Tomita parser \cite{tomita:book86} for example,
when running the sentence 
``{\it Labels can be assigned to a particular instruction step in a source program
to identify that step as an entry point for use in subsequent instructions.}''
on a relatively moderate sized grammar with 220 hand-coded rules
returns a parse forest representing 309 parsing alternatives.
For a more detailed grammar of 400 rules, the number of alternatives increases to 127,338.

Newer probabilistic approaches, often with only relatively restricted context sensitivity,
have achieved only limited success even when trained on very large corpora.
\citeA{magerman:acl95} and \citeA{collins:acl96} for example train on 40,000 sentences
from the Penn Treebank Wall Street Journal corpus and it is doubtful whether
the results can be further significantly increased by further enlarging the training
corpus, because (1) a parsing accuracy ceiling might have been reached due to model
limitations\footnote{For a more detailed discussion, see chapter~\ref{ch-related}, 
``Related Work''.}
and (2) a further significant enlargement of the manually annotated corpus is very 
expensive.

To cope with the complexity of unrestricted text, parse rules in any kind of formalism
will have to consider a complex context with many different morphological, syntactic
and semantic features.
This can present a significant problem, because even linguistically trained natural
language developers have great difficulties writing
explicit parse grammars covering a wide range of natural language.
On the other hand it is much easier for humans to decide how {\it specific} sentences
should be analyzed.

\section{Basic Parsing Paradigm}

We therefore propose an approach to parsing based on example-based learning with
a very strong emphasis on context.
After the segmentation of the sentence into words and punctuation marks, and a 
morphological analysis of the words, 
the parser transforms the resulting word sequence into an integrated phrase-structure 
and case-frame tree, powerful enough to be fed into a transfer and a generation 
module to complete the full process of machine translation.

The parser is trained on parse examples acquired under supervision. Since during
the training phase, the system is to be guided by a human supervisor, it is extremely 
important that the parsing process has a very transparent control structure that is
intuitive to a human.
As recent work in cognitive science, e.g.\ \cite{tanenhaus:acl96}, has confirmed again,
the human mind performs a continuous and deep interpretation of natural
language input.
A sentence is processed `left-to-right' in a single pass, 
integrating part-of-speech selection and syntactic and semantic analysis.

As the basic mechanism for parsing text we therefore choose a shift-reduce type parser 
\cite{marcus:book80}.
It breaks parsing into an ordered sequence of small and manageable parse actions such
as {\it shift} and {\it reduce}. With the parse sequence basically mimicking the order 
people process sentences, we don't only make the parsing process intuitive to the supervisor 
during the training phase, but, with this paradigm, we also can be confident that at any point
during the parse,
the current morphological, syntactic and semantic context information of a partially 
parsed sentence will be sufficient for the computer program to make good decisions
as to what parse action to perform next.

Like for humans, this approach also has the advantage of a single path, i.e.\ deterministic
parsing\footnote{For a brief discussion of the problem of garden path sentences, see 
section~\ref{sec-garden-path} later in this chapter.},
eliminating computation on `dead end' alternatives and resulting in a temporal
processing complexity that is {\bf linear} in the length of a sentence,
making parsing very fast.

Applying machine learning techniques, the system uses parse examples
acquired under supervision to generate a deterministic shift-reduce type parser
in the form of a decision structure. 
This decision structure's classification of a given parse state is the parse action to be
performed next. The parse state used by the decision structure is described by its context 
{\it features}.

Relieving the NL-developer from the hard if not impossible task of writing an explicit 
grammar, the focus on relevant features at the same time requires a relatively modest 
number of training examples when compared to more statistically oriented approaches like
for example \cite{magerman:acl95,collins:acl96}, where the parser is trained on 40,000
WSJ sentences. Our system is currently trained on only 256 WSJ sentences. The following
chapter discusses parsing experiments and the influence of the number of features in detail.

The approach we take provides a high degree of manageability of the `grammar' when
increasing its coverage. All that is necessary, is to add more examples. 
The advantage of examples is that they are much more modular than rules. All we assert
for an individual example is the next parse action that has to be taken for the specific
parse state of a specific sentence, regardless of how different that parse action might 
be for a similar parse state of another sentence. We therefore don't have to change
old examples when we try to increase coverage to include new sentences, however
much a decision structure based on the old examples might have misparsed the new sentences
by misclassifying new parse states. After new examples have been added, we automatically
compute a new, more refined decision structure that will also cover the new sentences.

On the other hand,
when increasing coverage in a rule-based system, it is typically not sufficient to add more
rules. Old rules will often have to be modified, e.g.\ by specializing their antecedents.
Given the enormous complexity of natural language, it can easily happen that sentences
that were parsed properly before a rule modification no longer work afterwards.
To make sure that the modified rules work, testing becomes necessary, and with that,
examples. So we see that extending rule sets is not only qualitatively more difficult
than extending example sets, but, at least in practise, examples are indispensable in
some form or other anyway.

\section{The Core Parsing Mechanism}
\label{sec-core-parsing-mechanism}

As just described, we choose a shift-reduce type parser
as our basic mechanism for parsing text into a shallow semantic representation.
Parsing is broken down into an ordered sequence of small and manageable parse actions
such as {\it shift} and {\it reduce}.

\begin{figure}[htbp]
\epsfxsize=10.0cm
\centerline{\epsfbox{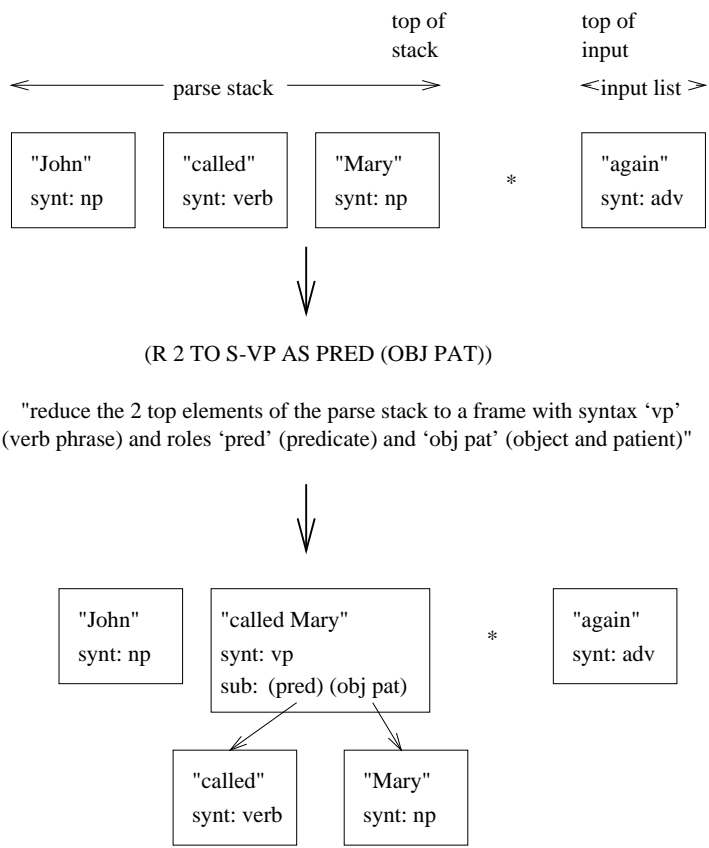}}
\caption{A typical parse action (simplified); boxes represent {\it parse entries}}
\label{fig-parsing_ex1}
\end{figure}

The central data structure for the parser contains a parse stack and an input list.
The parse stack and the input list contain trees of parse entries of words or phrases. 
Core slots of parse entries are surface and lexical form, syntactic and semantic category, 
subentries (of type parse entry) with syntactic and semantic roles, and
form restrictions such as number, person, and tense.
Other slots can include information like the numerical value of number words, flags
whether or not a (German) verb has a separable or inseparable prefix etc. 

Initially, the parse stack is empty and the input list contains the simple parse entries
produced by the morphological analyzer. 
After initialization, the {\it deterministic parser} applies a sequence of {\it parse actions} 
to the parse structure. 
The most frequent parse actions are 
   {\it shift}, which shifts a parse entry from the input list onto the parse stack or vice versa,
   and {\it reduce}, which combines one or several parse entries on the parse stack 
       into one new parse entry. The parse entries to be combined are typically, but not necessarily, 
       next to each other at the top of the stack. As shown in figure~\ref{fig-parsing_ex1},
       the action 

\begin{verbatim}
    (R 2 TO VP AS PRED (OBJ PAT))
\end{verbatim}

\noindent
for example reduces the two top parse entries of the stack into a new parse entry that is marked
as a verb phrase and contains the next-to-the-top parse entry as its predicate (or head)
and the top parse entry of the stack as its object and patient.
Other parse actions include `add-into' which adds parse entries arbitrarily
deep into an existing parse entry tree, `mark' that marks some slot of some parse entry with some 
value and operations to introduce empty categories (traces and `PRO', as 
in ``She$_{i}$ wanted PRO$_{i}$ to win.''). Parse actions can have numerous arguments,
making the {\it parse action language} very powerful. In particular, when shifting in a new word,
the mandatory argument of {\it shift} specifies the part-of-speech of the possibly ambiguous word
that is being shifted in, thereby effectively performing the so-called tagging as part of the parsing.

\section{The Parse Actions in Detail}

This section describes the various parse action in more detail. We present the various basic
types of parse action (shift, reduce, add-into, empty category instantiation, co-index, mark, expand
as well as the pseudo-action `done'). Many of these parse actions allow a sophisticated parameterization.
Often these parameters designate parse entries that participate in an operation.

In the English parsing training examples, we use seven basic types of parse actions (all except
`expand' which is only used for German) with a current total of 265 different individual parse actions.
These are quite unevenly distributed with the 10 most frequent parse actions accounting for 53\%
of all examples and 78 parse actions occurring only once, accounting for 0.66\% of all examples.

The various parse actions are not preselected in any way.
Instead, during training, as further explained in section~\ref{sec-training-the-parser}, 
each parse action example is assigned the parse action that the supervisor deems to be 
appropriate in the context of a specific, partially parsed text.
The range of parse actions therefore depends directly and exclusively on the specific set of 
training examples.

The first of the following subsections describes how parse entries that are referred to in features
can be identified. The description of the individual parse action types is then followed
by subsections on how these can be combined into multiple parse actions and how they implicitly trigger
a number of internal computations.

\subsection{Parse Entry Paths}
\begin{figure}[htbp]
\epsfxsize=16.4cm
\centerline{\epsfbox{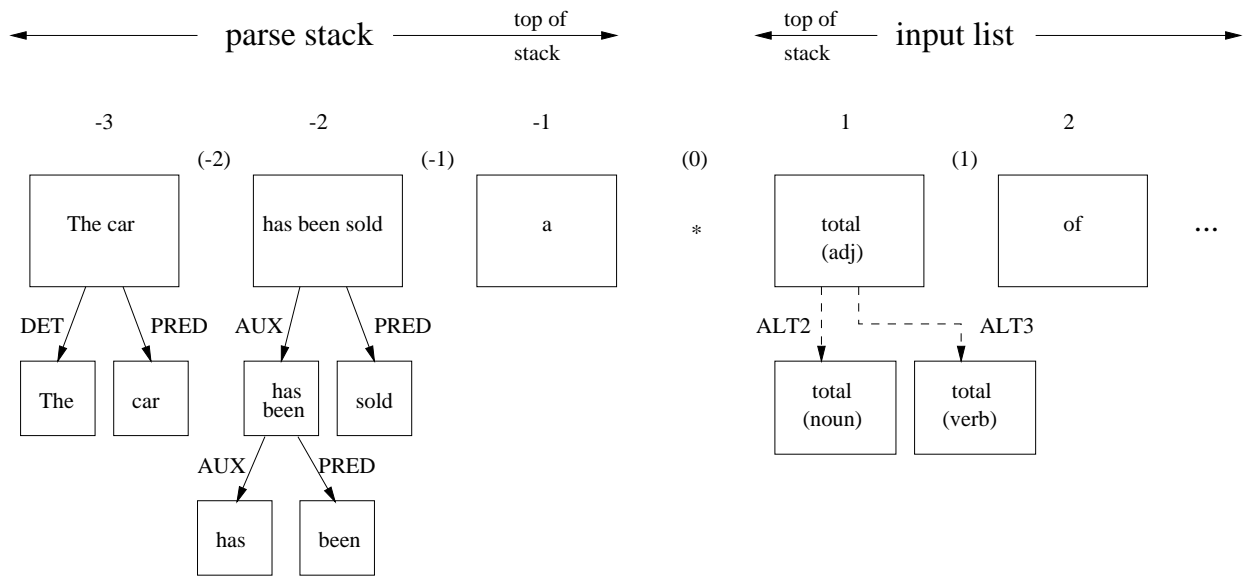}}
\caption{Parse stack and input list}
\label{fig-stacks}
\end{figure}

Many parse actions have parameters that refer to parse entries on the parse stack or the input list.
The participating parse entries are identified by {\it parse entry paths} with various degrees of complexity.
Parse entries that are elements of the input list are identified by positive integers. {\it n}
designates the {\it nth} element on the input list. Parse entries that are elements of the parse
stack are identified by negative integers such that {\it n} designates the $|n|th$ element of
the parse stack. The top row of numbers in figure~\ref{fig-stacks} illustrates this scheme.
The numbers in parentheses indicate positions between parse entries. Note that the active ends
of the parse stack and the input list\footnote{The input `list' is actually a stack too, since the
shift operation, as described in section~\ref{sec-garden-path}, can also shift parse entries back
onto the input `list'.} face each other. The active position (``0'') of a partially parsed sentence is
marked by an asterisk (*).

Sub entries are described by paths such as (DET OF -3), 
in the current example the parse entry for `{\it The}'; (AUX OF -2), the parse entry for
`{\it has been}'; and
(PRED OF AUX OF -2), the parse entry for `{\it been}'. {\it PRED*} designates the `ultimate' predicate
of a parse entry, i.e.\ the parse entry that is reached by repeatedly descending to sub-entry
PRED until this is no longer possible. In figure ~\ref{fig-stacks}, (AUX* OF -2) would designate the
parse entry for `{\it has}', but the asterisk-extension is only used for predicates (``PRED'').

Besides these concrete path descriptions, the system allows abstract paths that contain 
predefined elements
such as {\it NP-1, ACTIVE-FILLER-1} and {\it (SURF-NP -2 BEFORE -1)}. Descriptors of form
$<$restr$>$-$<$n$>$ designate the nth last parse entry on the parse stack for which $<$restr$>$ holds. NP-1
is the top (= rightmost) noun phrase parse entry on the parse stack, in the current figure
`{\it The car}'. ACTIVE-FILLER-1 is the top parse entry on the parse stack that has a slot {\it gap}
with the value `{\it active filler}'. (SURF-NP -2 BEFORE -1) is the penultimate surface noun phrase
before position -1; for the SURF-NP construction, all right-branching embedded phrases of the last
element before the position marked by the BEFORE argument count.

Path descriptions with such abstract elements can often better describe the essence of the path
that leads to the involved path entry. Consider for example these two sentence fragments, separated by a `+':
{\it I know the artist of this masterpiece,} + {\it which was painted 200 years ago}.
Relative phrases like in this example often refer to the last surface noun phrase, here {\it
this masterpiece}.

To express this relationship, no matter how deep this noun phrase is already embedded in a parse stack
parse entry, we use (SURF-NP -1 BEFORE -1), indicating the last surface noun phrase before 
position -1. We could have used (PRED-COMPL OF MOD OF OBJ OF -2), the predicate complement 
of the modifier {\it of this masterpiece} of the object {\it the artist of this masterpiece} of 
the second highest parse entry on the parse stack, but descriptions like that last concrete path
can vary enormously from example to example and would not lend themselves very well to generalization.
In order to {\it learn} parse actions well, it is important that essentially similar contexts should
call for the same action to be taken. Abstract paths such as (SURF-NP -1 BEFORE -1) allow
this parse action stability and therefore makes it easier for the machine learning component to
learn appropriate parse actions.

\subsection{Shift}
\label{sec-shift}

Parse entries can be shifted from the input-list to the parse-stack and vice versa.
The by far more common type of shift is from the input-list to the parse-stack (examples (1) and (2) below). 
Output products from morphological processing, i.e.\ parse entries in the input-list, can still be ambiguous, 
e.g.\ `{\it left}' can be interpreted as both a verb or an adjective. 
Since further processing on the parse stack requires syntactically disambiguated parse entries,
this shift requires an argument identifying the proper alternative. When shifting in an ambiguous parse
entry, the shift operation eliminates all alternatives that are not covered by the shift operation argument. 
Assigning a specific part-of-speech to a word is called {\it tagging}. Note that in our system, 
tagging is naturally integrated into the
parsing process, whereas for some other parsers, tagging is required as a separate preliminary step.

In the current system, only syntactic restrictor arguments are used, as in example (1), 
but semantic restrictors as in example (2) could be used to make a yet more specific choice,
such as for a word like {\it buck}.\\

\noindent
\begin{tabular}{llp{10.7cm}}
  (1) & {\bf (S S-VERB)} & shifts parse-entry of syntactic type {\it verb} from input-list to parse-stack \\
  (2) & {\bf (S I-EN-ANIMATE)} & shifts parse-entry of semantic type {\it animate} from input-list to parse-stack \\
  (3) & {\bf (S -2)} & shifts two parse entries from the parse-stack back onto the input-list \\
\end{tabular}\\

The other type of shift (example (3)) shifts parse entries back onto the input list. The absolute value
of the first argument indicates how many parse entries are to be shifted back. To better understand
the purpose of this reverse shifting, consider the case of a noun phrase followed by a preposition. 
The prepositional phrase that this preposition probably starts might or might not belong to the preceding 
noun phrase. Making an attachment decision already at this point is typically still premature, 
because the decision would have
to be based on the yet unprocessed components of the prepositional phrase.
So instead of making an early PP attachment decision, the system shifts in the components of the prepositional 
phrase, processes it, and
upon completion, decides whether or not an attachment to the preceding noun phrase is appropriate. 
If so, it performs the attachment
through a {\it reduce} operation. Otherwise, it shifts the prepositional phrase, that turned out not to
attach to the noun phrase after all, back onto the input-list so that the system can process the noun
phrase otherwise, e.g.\ by attaching it to the preceding verb.
Most shift-back actions involve a single parse entry, but some of them move two or even three parse entries
at a time.

\subsection{Reduce}

The {\it reduce} operation typically combines one or more parse entries into a new parse entry.
A typical parse action was already presented in figure~\ref{fig-parsing_ex1}. In the following
examples, the plus sign (+) denotes the boundary between two adjacent parse stack elements.\\

\noindent
\begin{tabular}{lp{15.1cm}}
  (1) & {\bf (R 2 TO S-PP AS PRED PRED-COMPL CLASS C-BY-AGENT)}\\
      & reduces the top two parse-stack parse entries to a
    prepositional phrase of class {\it C-BY-AGENT} with roles {\it pred} and {\it pred-compl} respectively.
    Example: {\it by} + {\it a shareholder}\\
  (2) & {\bf (R 3 AS MOD DUMMY PRED)}\\ & reduces the top three parse-stack parse entries to a parse-entry
    with roles {\it mod}, {\it dummy} and {\it pred} respectively.
    Example: {\it heart} + {\it `-'} + {\it attack}\\
    & For the parser, `dummy' is treated like any other role. We use it to mark components without any
    semantic meaning, mostly punctuation, that can later be discarded during transfer.\\
  (3) & {\bf (R (-3 -1) AS AUX PRED)}\\ & reduces the first and third parse entry of the parse-stack.
    Example: {\it have} + {\it clearly} + {\it put}\\
  (4) & {\bf (R (-3 -1) AS AUX PRED AT -2)}\\ & does the same, but places the result at position 2 of the
         parse stack.\\
  (5) & {\bf (R 2 AS SAME (OBJ QUANT))}\\ & merges the top parse entry into the second parse entry
         with roles {\it obj} and {\it quant}. Example: {\it will cost} + {\it \$15}\\
  (6) & {\bf (R (-2) AS SAME)}\\ & moves the second parse entry on the parse stack to the top.\\
  (7) & {\bf (R 2 TO LEXICAL)}\\ & reduces the two top parse-stack parse entries to a parse entry 
         that is listed in the lexicon. Example: {\it because} + {\it of}.\\
  (8) & {\bf (R 1 TO S-NP AS PRED)}\\ & upgrades the first parse entry on the parse stack to a noun phrase.\\
\end{tabular}\\

Often, the parse
entries to be reduced are next to each other at the top of the parse tree (examples (1) (2) (5) (7) (8)).
Disjoint constituents, e.g.\ {\it has} and {\it been} in ``{\it has always been}'', typically require a disjoint
reduction such as in examples (3) and (4). Constituents also don't necessarily have to be at the top of the
parse tree (example 6). Unless otherwise specified, as in example (4), the resulting parse entry is placed
at the top of the parse stack. Also, unless otherwise specified, as in example (1), the new parse entry inherits
properties like part-of-speech and semantic class from the component marked as predicate 
({\it pred} or {\it same}).
Sub parse entries can have one or more roles in the super parse entry. The second component in example (5) for example
has the roles {\it obj} and {\it quant}.
Reductions like the one in example (7), e.g.\ {\it New York} or {\it because of} reduce the components to
a parse entry with an entry in the lexicon. The entry in the lexicon then determines the part-of-speech and
semantic class of the new parse entry.

\begin{figure}[htbp]
\epsfxsize=13.4cm
\centerline{\epsfbox{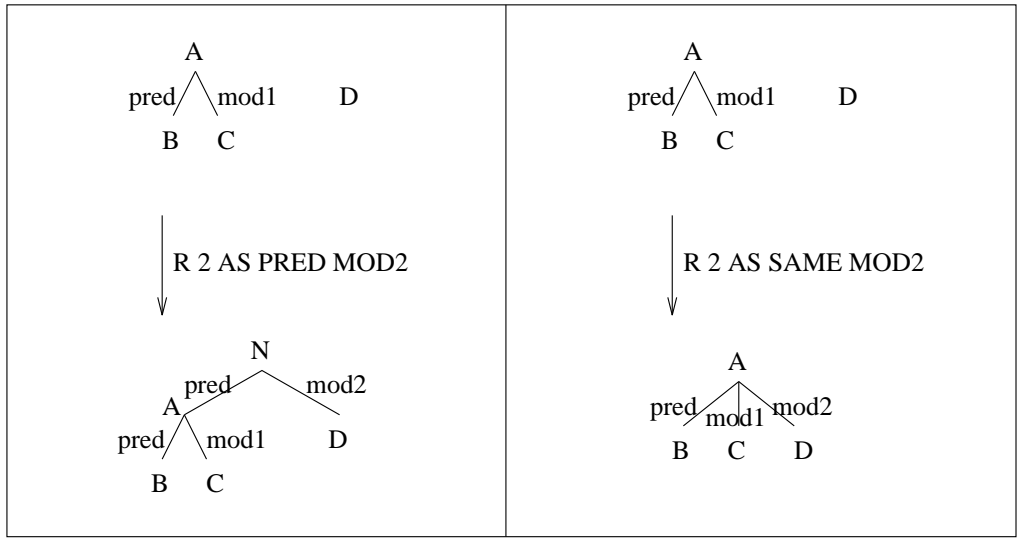}}
\caption{Difference between keywords 'pred' and 'same'}
\label{fig-pred_vs_same}
\end{figure}

When reducing elements,
a new level of hierarchy can be created (indicated by the keyword `PRED'),
or non-head elements are just
added as extra sub-parse-entries to an existing complex parse entry, keeping the hierarchy flat 
(indicated by the keyword `SAME' instead of `PRED'), as illustrated in figure~\ref{fig-pred_vs_same}.

\subsection{Add Into}

Just like the {\it ``reduce''} action, the {\it ``add into''} action is used to combine parse entries. 
It is used to add more sub-entries to parse entries that are no longer at the top level of the parse 
stack, but rather (already) embedded underneath. 

When reaching a comma while parsing sentences, it is often a good heuristic to use it as a signal 
to reduce all of what has been encountered so far before proceeding.
In the following example (1), the main clause up to the comma is parsed into a sentence, before anything to
the right of the comma is processed. When the sentence fragments {\it or} and {\it 89 cents a share} are
ready, they need to be added to the preceding disjunction element {\it a record \$9.9 million}, the object
of the third parse entry on the parse stack. The add-into action of example (1) accomplishes just that.
Like for {\it reduce} actions, the {\it as}-clause provides the roles of the newly added sub-entries in the
super entry.\\

\noindent
\begin{tabular}{lp{15.1cm}}
  (1) & {\bf (A (-2 -1) TO (SURF-NP -1 BEFORE -2) AS CONJ COORD))}\\
      & adds the top two parse-stack parse entries into the last surface noun phrase before position -2
        with roles {\it conj} and {\it coord} respectively.
    Example: {\it In fiscal 1986, it earned a record \$9.9 million,} + {\it or} + {\it 89 cents a share}\\
\end{tabular}\\
\begin{tabular}{lp{15.1cm}}
  (2) & {\bf (A -1 TO (PRED* OF -2) AS PARTICLE)}\\
      & adds the top parse-stack parse entry into the ultimate predicate of the second parse-stack entry
        as a particle.
    Example: {\it had already been taken} + {\it out}\\
\end{tabular}\\
\begin{tabular}{lp{15.1cm}}
  (3) & {\bf (A (-2 -1) TO (THAT-CLAUSE -1 BEFORE -2) AS CONJ COORD)}\\
      & adds the top two parse-stack parse entries into the last `that-clause' before position -2
        with roles {\it conj} and {\it coord} respectively.
    Example: {\it U.S. officials estimate that the move will result in \$85 million in additional U. S. 
         sales this year,} + {\it and} + {\it that such sales eventually could grow to \$300 million annually}\\
\end{tabular}\\
\begin{tabular}{lp{15.1cm}}
  (4) & {\bf (A -1 TO (INDEXED-NP -1 BEFORE -1) AS MOD)}\\
      & adds the top parse-stack parse entry into the last noun phrase to the before position -1 that
        has a matching index with a role of {\it mod}.
    Example: {\it The sales totaled $<\$100 million>_{1}$,} + 
       {\it [$<\$100 million>_{1}$] compared with \$90 million a year before}\\
\end{tabular}\\

Example (2) differs from the other examples in that the {\it PARTICLE} role triggers a special
treatment. The particle ({\it out}) is merged with its verb ({\it taken}) at the deepest level,
updating the verb concept from {\it I-EV-TAKE}\/ to {\it I-EV-TAKE-OUT}.

\subsection{Empty Category Instantiation}
\label{sec-empty-category}

Empty categories \cite{chomsky:gb88,lasnik:gb88} are phonologically unrealized noun phrases including 
NP-traces such as {\it t} in ``{\it $John_i$ was believed $t_i$ to be clever.}'', 
variables, such as the wh-trace {\it t} in ``{\it $What_i$ are you looking at $t_i$?}\/''
and PROs such as in ``{\it $He_{i}$ wanted $PRO_{i}$ to win.}''.
Empty categories are useful tools to describe sentences where noun phrases have a semantic role
in more than one clause (PRO) or where they are realized in a phrase other than the one where they have
a semantic role (traces). 

In order to have enough noun phrase representatives for noun phrases that play a role in more than
one phrase, empty category tokens are created from noun phrase parse entries when necessary. The
newly created parse entry is automatically co-indexed with the parse entry it is derived from.
With several parse entries representing a noun phrase with a single phonological realization,
these parse entries can now be assigned different roles in different phrases. E.g.\ in the
sentence ``{\it $He_{i}$ wanted $PRO_{i}$ to win.}'', `{\it He}' will be assigned the roles {\it
subject} and {\it experiencer}, whereas the corresponding `{\it PRO}'
will be assigned the role {\it agent}. In a sentence like ``{\it $Sales_i$
are expected $t_i$ to increase.}'', `{\it Sales}' will be assigned the roles {\it subject}
and {\it dummy}, whereas `{\it $t_i$}' will be the {\it theme} in the embedded clause.\\

\noindent
\begin{tabular}{lp{15.1cm}}
  (1) & {\bf (EMPTY-CAT FROM ACTIVE-FILLER-1 AT 0)}\\
      & creates an empty-category parse entry from the last entry marked as an `active filler', 
        e.g.\ {\it where} in the following example, and places it at the top of the stack.
    Example: {\it He asked} + {\it where} + {\it I} + {\it came} + {\it from}\\
  (2) & {\bf (EMPTY-CAT FROM NP-1 AT -1)}\\
      & creates an empty-category parse entry from the highest parse-stack parse entry that is a noun phrase
        and places it at position -1.
    Example: {\it Sales} + {\it are expected} + {\it to increase}\\
\end{tabular}\\

\newpage
\subsection{Co-Index}

\noindent
This action co-indexes the top parse entry on the parse-stack with the parse entry specified in the 
{\it WITH} clause. This is used particularly for co-indexing phonologically realized relative pronouns 
with their antecedents.
In the representation the co-indexing is realized by annotating the primary parse entry with 
{\it (INDEX $<$index$>$)} and by annotating all other entries that refer to it with {\it (REF $<$index$>$)},
where $<$index$>$ is a system generated integer.
When an empty category parse entry is created, this is performed automatically. Non-relative pronoun
resolution is currently done in a special phase at the end of parsing 
(see section~\ref{sec-anaphora-resolution}).\\

\noindent
\begin{tabular}{lp{15.1cm}}
  (1) & {\bf (CO-INDEX WITH -2)}\\
      & co-indexes the top parse entry of the parse-stack with the parse entry to its left.
    Example: {\it someone} + {\it who}\\
  (2) & {\bf (CO-INDEX WITH (SURF-NP -2 BEFORE -1))}\\
      & co-indexes the top parse entry of the parse-stack with the penultimate noun phrase on 
        the surface to the left of position -1. 
    Example: {\it The owner of the car,} + {\it who}, where {\it who} would be co-indexed with
        {\it the owner}.
\end{tabular}\\

\subsection{Mark}

\noindent
The parse action {\bf (M $<$path$>$ $<$slot$>$ $<$value$>$)} assigns slot $<$slot$>$ of the parse entry specified 
by $<$path$>$ the value $<$value$>$. This parse action is typically part of a complex parse action, i.e.
mark actions typically occur in a list of two basic parse actions starting with
a shift, reduce, or an empty category instantiation, followed by a mark action.\\

\noindent
\begin{tabular}{lp{15.1cm}}
  (1) & {\bf (M -1 GAP ACTIVE-FILLER)}\\
      & marks the top parse-stack parse entry as being an active gap. 
    Example: {\it What} * {\it do you want ...} This mark-up can later be exploited for proper
      empty category instantiation (see subsection~\ref{sec-empty-category}).\\
  (2) & {\bf (M -1 IN-PARENTHESES TRUE)}\\
      & marks the top parse-stack parse entry as being being in parentheses.\\
\end{tabular}\\

\subsection{Expand}
The parse action {\bf (EXPAND)} expands contractions of two or more words. An example in English
is `{\it won't}', a contraction of {\it will} and {\it not}. Examples from other languages are
{\it im = in + dem} in German, {\it aux = \`{a} + les} in French, {\it al = a + le} in Spanish and
{\it nel = in + il} in Italian, all of which are contractions of a preposition with a definite article.
In our system, the English verbal contractions are however already handled in the segmentation and 
morphology module, while the German contractions are expanded during parsing. The expand action 
does not take any arguments and always applies to the top parse-stack entry.

\subsection{Done}
The parameterless {\bf (DONE)} parse `action' signals that parsing is complete. Normally, this should
only occur when there is a single item left on the parse stack and the input list is empty. That one
parse entry is then returned as the result of the parse.

This parse action is also used in the case of serious problems during the parse, e.g.\ when strong 
symptoms of a potential endless loop are detected or when regularly proposed parse actions are overruled by the
sanity checker, which tests whether a parse action can actually be performed in a specific context
(more details in section ~\ref{sec-guards}).
The DONE parse action is the only one that can always be used as a last resort, because
it can be `executed' in any parse state. When reaching such 
a pathological case, the system combines all remaining items on the parse stack and the input 
list into a new `holding' parse entry.

\subsection{Multiple Parse Actions}
The parser decides what to do next by classifying the current partially parsed text. The classification
is technically a list of parse actions. In most cases (99.8\%), this parse action list contains only a single
parse action, but sometimes multiple parse actions are used:

\begin{enumerate}
\item ((S S-ADV) (M -1 GAP ACTIVE-FILLER))\\
      Example: I know * {\it where} he is.\\
      Example: I know * {\it where} he is from.
\item ((EMPTY-CAT FROM NP-1 AT 0) (M -1 GAP ACTIVE-FILLER))\\
      Example: I know the {\it restaurant} * you were talking about.
\item ((EMPTY-CAT FROM ACTIVE-FILLER-1 AT 0) (M ACTIVE-FILLER-1 GAP SATURAT\-ED-FILLER))\\
      Example: I know the $restaurant_{i}$ {\it $PRO_{i}$} you were talking about *.
\item ((R 2 TO S-ADJP AS COMP PRED) (M -1 GRADE SUPERLATIVE))\\   
      Example: the {\it most famous} * building
\end{enumerate}

Multiple parse actions are used when shifting in interrogative pronouns or adverbs, as in
example (1), in relative phrases without a phonologically realized relative pronoun, as in
example (2), when resolving wh-traces, as in example (3), or when reducing periphrastic
comparatives or superlatives, as in example (4).
As usual, the asterisk (*) marks the current parse position.

Even though the parse action language allows an arbitrary number of elementary parse actions
and does not have any restrictions on combinations, we only use up to two elementary parse
action, and this only in a few cases (0.2\%), all involving an `empty-category' or `mark' action.
Mark actions often set parse entry slots that are only used in a few special
circumstances. Theoretically the parser could of course learn to perform the {\it mark} action in
a separate step. But since the {\it mark} actions are conceptually closely tied to their preceding
action, it is simpler to learn the combination.

\subsection{Triggered Computations}

It is important that the supervisor does not get overwhelmed when providing
the correct parse action. The actions should be as simple as possible. This is partly
achieved by providing defaults, e.g.\ by letting a new parse entry inherit properties like its 
semantic class and its part of speech from its predicate sub-parse-entry.

In other cases, like compound verbs, the notion of inheritance is less useful.
Consider for example the need to determine the form of a compound verb (e.g.\ ``has'' $+$ ``given'').
The parser needs to know the tense, voice, aspect etc.\ of the compound.
One way to gain this information would be to let the supervisor provide this information as part
of the action of reducing an auxiliary and a main verb to a compound verb.
A more elegant and less burdensome way for the supervisor is to let the reduction of 
an auxiliary and a main verb trigger the computation of the form of the complex verb.
These computations don't depend on a lot of context and can therefore be performed using
straightforward algorithms that don't involve any learning.

As another example, the reduction of the subject noun phrase and a verb phrase,
e.g.\ ``{\it The fish} + {\it were alive.}\/'', triggers the computation of the form of a sentence
({\it third person plural}).

Currently, triggered computations are limited to determine the forms (e.g.\ tense or number) of 
new parse entries.  Word sense disambiguation within a specific part of speech is currently successfully
`delegated' to the transfer process and anaphora resolution is performed at the end of core parsing.
If, in a later extension, these tasks have to be moved up into core parsing, they could also
be performed as a triggered computation, a solution superior to explicit parse actions,
because the triggered computations would not only keep the parse action sequence simpler, 
but also keep parse action sequences impervious to lexicon additions that might turn a previously 
`unambiguous' word
like {\it pen} into an ambiguous one, necessitating a modification of all parse action sequences
with such a `newly ambiguous' word. Since such potential modifications of parse action sequences 
are highly undesirable, a triggered action is clearly preferable. The computations could be triggered
by the evaluation of a feature that is used in the parse action decision making process and depends 
on the specific sense of a word or its antecedent.

Such additional triggered computations could principally be so complex that they could benefit 
from machine learning. However, this wasn't found to be necessary for the currently implemented 
triggered computations.

\section{Anaphora Resolution}
\label{sec-anaphora-resolution}
While the computations described in the last section can easily be linked to triggering parse
actions, another computation, anaphora resolution, could be performed
at several points during the parse process. For convenience's sake, we choose to perform it
at the end of the parse action sequence.

Even in closely related languages such as English and German, which both differentiate pronouns
with respect to number (singular/plural), person (1-3) and gender (masculine/feminine/neu\-ter),
pronouns don't match one to one. Compared to English, German has a much stronger notion of an independent
grammatical gender, so that for example {\it der L\"{o}ffel (the spoon)} is masculine,
{\it die Gabel (the fork)} is feminine, and {\it das Messer (the knife)} is neuter.
Now consider the translation of the English pronoun {\it it}. The gender of its equivalent German 
pronoun depends on the grammatical gender of its German antecedent, which is the translation of
the antecedent of {\it it} in English. So, to pick up on the previous example, when translating the 
English sentence ``{\it The spoon/fork/knife is expensive, because it is made out of silver.}'', the
English pronoun {\it it} would have to be translated as
{\it er} (masc.), {\it sie} (fem.), or {\it es} (neuter), depending on what eating utensil it 
referred to.
As this example shows, anaphora resolution is necessary for proper translation.

To identify the antecedent of an anaphor, the system first finds the syntactically permissible
antecedent candidates, using for example number agreement and the linguistic relationship {\it c-command} 
(for details see for example \cite{riemsdijk:rw86}). It then eliminates some candidates by using a
few simple semantic restrictions,
e.g.\ that a pronoun that fills the role of an agent must have an antecedent that semantically can
be an agent.
Consider for example the pronoun in the sentence ``{\it The airline bought the plane because it had already
decided to do so earlier}''. Syntactically, both {\it The airline} and {\it the plane} qualify as an
antecedent for {\it it}, but {\it the plane} is ruled out, because it can not be an agent as required for subjects
of ``to decide'' (in the active voice).
Finally, among any remaining candidates, the syntactically closest is picked.

Relatively simple heuristics were sufficient to cover all anaphora cases in the 48 training sentences.
In a more advanced anaphora resolver, these heuristics have to be elaborated further. With enough
complexity, machine learning might again prove useful in deciding which antecedent to pick.
Previous work in this area includes \cite{aone:acl95}, which describes an anaphora resolution system
trained on examples from Japanese newspaper articles.

%A core issue here is that of optimal use of a set of contributing
%factors: these include, for instance, gender and number agreement,
%c-command constraints, semantic consistency, syntactic parallelism,
%semantic parallelism, salience, proximity and so forth.

\section{Parsing Safeguards}
\label{sec-guards}

The parser contains a few safeguards that suppress the attempt to execute unexecutable parse actions,
detect potential endless loops and handle incomplete parses.

Before a proposed parse action is actually executed, the {\it sanity checker} tests whether this
can actually be done. If it is impossible, e.g.\ in the case of a reduction of n elements on a parse
stack that actually contains less than n elements or a shift-in on an empty input list, the
sanity checker forces another action to be chosen.
The sanity checker also suppresses a shift-in that would immediately follow a shift-out.

If the decision structure can naturally provide an alternative acceptable parse action, 
that action is chosen, otherwise, a new word is shifted in, or, if the input list is empty, 
the parse action `done' is selected.

Another source of potential complications are endless loops. While the simple shift-in shift-out loop is already
suppressed by the sanity-checker, other loops are still possible, e.g.\ a loop with the iteration
{\it (M -1 GAP ACTIVE-FILLER) (EMPTY-CAT FROM NP-1 AT 0) (R 2 AS QUANT SAME)}. 

To detect a loop, the parser limits the number of allowed remaining parse actions to four
times the number of elements on the parse stack and the input list, about twice the amount of parse
actions typically expected. Since the number of remaining parse actions allowed is limited, all 
actual loops are detected. All suspected loops actually turned out to be loops, as inspections showed.
When a loop is detected, the parse is terminated by choosing the action `done'.
Fortunately, when the system is trained on at least a few dozen sentences, loops become quite rare.

If parsing is terminated before all words have been combined into a single parse entry, e.g.\ due to
a `done' parse action prescribed by 
the sanity checker, the endless loop detector or just an incorrect classification,
all remaining elements on the parse stack and input list are lumped into a holding parse entry
with uncommitted syntactic and semantic class ({\it S-SYNT-ELEM} and {\it I-EN-THING}\/) and uncommitted
roles ({\it CONC}\/) for the elements. This operation allows the parsed sentences to be further processed.
Obviously, we expect subsequent evaluations to yield substandard results, particularly in recall and
translation, but it is important that parsing quality can also be measured for pathological cases.

\section{Features}
\label{sec-features}

To make good parse decisions, a wide range of features at various degrees of abstraction have 
to be considered.
To express such a wide range of features, we define a {\it feature language}. Given a particular
parse state and a feature, the system can interpret the feature and compute its value for the 
given parse state, often using additional knowledge resources such as 
\begin{enumerate}
\item a general {\it knowledge base} (KB), which currently consists of a directed acyclic 
    graph of concepts, with currently 3608 {\it is-a}-relationship links,\\
    e.g. ``$CAR_{NOUN-CONCEPT}$ is-a $VEHICLE_{NOUN-CONCEPT}$''; for more details see 
section~\ref{sec-kb}
\item {\it subcategorization tables} that describe the syntactic and semantic role structure(s) for verbs
    and nouns with currently a total of close to 200 entries; 
    for more details see section~\ref{sec-subcat-tables}
\end{enumerate}

The following examples,
for easier understanding rendered in English and not in feature language syntax,
illustrate the expressiveness of the feature language:
\begin{itemize}
\item the general syntactic class of the top element of the stack (e.g.\ adjective, noun phrase), 
\item the specific finite tense of the second stack element (e.g.\ present tense, past tense),
\item whether or not some element could be a nominal degree adverb,
\item whether or not some phrase already contains a subject,
\item the semantic role of some noun phrase with respect to some verb phrase (e.g.\ agent, time; this 
      involves pattern matching with corresponding entries in the verb subcategorization table),
\item whether or not some noun and verb phrase agree.
\end{itemize}

Features can in principal refer to any element on the parse stack or input list, 
and any of their subelements, at any depth. Since all of the currently 205 features are supposed
to bear some linguistic relevance, none of them actually refer to anything too far removed from
the current focus of a parse state. A complete list of all 205 features can be found in
appendix section~\ref{sec-features-for-engl}.
The set of features is used for all parse examples for a specific language and
can easily be extended when the need arises. 

The current set of 205 features has been collected manually.
The feature collection is basically independent from the supervised parse action
acquisition. Before learning a decision structure for the first time, the supervisor
has to provide an initial set of features that can be considered obviously relevant.
During early development phases of our system, this set was increased whenever
parse examples had identical values for all features (so far) but nevertheless demanded
different parse actions.
Given a specific conflict pair of partially parsed sentences, which is signaled during
the machine learning process if it occurs, the supervisor would add
a new feature that relevantly discriminates between the two examples. Such an addition
requires fairly little supervisor effort.
Relevant context features refer to parse entries
that are shifted, reduced or otherwise directly participate in an operation,
as well as parse entries that describe context in the narrow sense.

As explained further in the next section, parse examples are generated from parse states and
parse actions.
Given the state of a partially parsed sentences, the system computes the values for all features
in the parsing feature list of a specific language. These values are then combined into a feature
vector, a list of values, which, together with the parse action,
%to be taken in the context of the partially parsed sentence as represented by the feature vector, 
form the core of a parse example.
If the set of features changes, the parse examples have to be recomputed, because the feature
vectors change. However, this can be done fully automatically since parse actions have been recorded
in log files.

The feature sets for parsing in English and German are by and large the same. However, a few 
features, such as ``{\it Is the second word on the parse stack `there'?}'', as it might be useful
to characterize expressions like ``there is''/``there are'', are language specific. We expect the 
feature set to grow, possibly to 300 features, when many more training examples from the Wall Street
Journal are added, and probably even more when expanding into new domains, but this does not 
appear to be very critical, because adding new features is easy and requires little time and because 
the subsequent parse action generation update is fully automatic.

The following subsections explain in detail how features can be defined.

\subsection{General Feature Structure}

The typical structure for a feature is 
\begin{verbatim}
   (<predicate> of <path> at <super-hierarchy-level>)
\end{verbatim}

E.g.\ the feature {\it (tense of pred of -2 at f-finite-tense)} describes
the tense of the sub-parse-entry with role {\it pred} of the second element of the
parse stack at hierarchy-level {\it f-finite-tense}. Possible values for this feature
are {\it f-pres-tense, f-past-tense, f-perf-tense, f-past-perf-tense, f-fut-tense} and 
{\it f-fut-perf-tense}.

Values are hierarchical. Consider a parse stack with a parse entry for ``to read'' at its top.
While the value of feature {\it (synt of -1 at s-synt-elem)}\footnote{{\it
s-synt-elem} denotes the top level of any syntactic category}
is {\it s-verb} in this example, the value for feature {\it (synt of -1 at s-verb)} would be
{\it s-tr-verb}, meaning that the parse entry stands for a transitive verb. The `super-hierarchy-level'
field in the above feature template is hence used to specify the hierarchy level of the value.
When evaluating a feature, the system will return the proper value directly below the 
`super-hierarchy-level' node, if there is any, and, otherwise, the special value {\it unavail}.

For parsing, the reference point for the paths is the current position. Many paths just
consist of a negative or positive integer, denoting a parse entry on the parse stack or the input list
respectively. From any parse entry, one can further navigate into sub-parse-entries using roles,
either syntactic or semantic. The feature {\it (class of pred-compl of co-theme of -1 at i-en-quantity)}
for example explores the type of quantity of the predicate complement of the co-theme of the top element
on the parse stack. There is no set limit for the length of the path, but reasonable linguistically
relevant features tend to have a limited path length, up to 3 in our current system.

Besides roles, the positional path elements {\it first, last, last-mod} and {\it last-coord} are
available. E.g.\ {\it first} accesses the first sub-parse-entry of a parse entry, and {\it last-mod}
accesses the last sub-entry with role {\it mod}. 

The path element {\it parent} accesses the parent entry. It is useful for describing paths that
have their reference points somewhere in a parse entry tree, as is the case for describing features
in the transfer module (see chapter~\ref{ch-transfer}).

{\it PRED*} designates the `ultimate' PRED of a parse entry, i.e.\ the parse entry that is 
reached by repeatedly descending to sub-entry PRED until this is no longer possible.
{\it npp-1} designates the last noun phrase or noun phrase within a prepositional phrase of
a sentence.

Elements on the input list can have several alternatives, typically when a word has several 
parts of speech. Unless otherwise specified, paths always select the first alternative.
Using path elements {\it alt2, alt3, alt-nom, alt-adv}, e.g.\ as in {\it (synt of alt2 of 1 at s-synt-elem)}
or {\it (classp of i-en-agent of alt-nom of 2 at m-boolean)}, select the second, third, nominal or
adverbial alternative.

Obviously, some paths don't lead to any parse entry. In that case, the system returns a value
of {\it unavail} for the feature.

\subsection{Syntactic and Semantic Category Features}

{\it ({\bf synt} of $<$path$>$ at $<$super-hierarchy-level$>$)} denotes the syntactic class of the parse entry
accessed through $<$path$>$ at $<$super-hierarchy-level$>$. {\it (synt of -1 at s-synt-elem)}, the general
part of speech of the top parse stack element, is the most often feature.

\noindent
{\it ({\bf class} of $<$path$>$ at $<$super-hierarchy-level$>$)} provides the same for the semantic class.

\noindent
{\it ({\bf syntp} of $<$syntactic class$>$ of $<$path$>$ at m-boolean)} is a boolean feature that holds iff
the parse entry at the end of $<$path$>$ is of the right $<$syntactic class$>$.\footnote{`m-boolean' is the 
direct super concept of `true' and `false'; the `m' prefix (`mathematical concept') is used to distinguish it 
clearly from the concept representing the adjective `boolean'.}

\noindent
{\it ({\bf classp} of $<$semantic class$>$ of $<$path$>$ at m-boolean)} is the corresponding 
semantic binary feature.

\subsection{Syntactic and Semantic Role Features}
The four features in this subsection draw on the background knowledge of the subcategorization
table, as described in section~\ref{sec-subcat-tables}.

{\it ({\bf semrole} of $<$position1$>$ of $<$position2$>$)} is the feature to describe the semantic role of the
parse entry at position 1 in a pattern best matched by the parse entry at position 2. 
If for example the two top parse stack entries are {\it (The student)} + {\it (read a book)},
the feature {\it (semrole of -2 of -1)} would have a value of {\it agent}.

\noindent
The corresponding binary feature type {\it ({\bf semrolep} of $<$position1$>$ of $<$position2$>$)} checks
if the parse entry at position 1 has {\it some} role in a pattern
best matched by the parse entry at position 2. If for example the
three top parse stack elements are {\it (sent) (a book) (to New York)}, the feature
{\it (semrolep of -1 of -3)} would be {\it true}, indicating that {\it (to New York)}
can have a meaningful role in a sentence with the predicate {\it to send}, which in
turn suggests that {\it (to New York)} should probably be attached to the verb and not
to the preceding noun phrase.

{\it (semrole of $<$syntactic role$>$ of $<$position2$>$)} describes the semantic role of the argument
with $<$syntactic role$>$
in the pattern best matched by the parse entry at $<$position2$>$. If for example the top element of
the parse stack is {\it (ate potatoes)}, the feature {\it (semrole of subj of -1)} would be
{\it AGENT}. {\it (semrolep of $<$syntactic role$>$ of $<$position2$>$)} checks whether $<$syntactic role$>$
is meaningful for the pattern best matched by position 2.

For verbal patterns, the parse entry at $<$position2$>$ can be a verb, 
a verb phrase or a sentence.  The verbal parse entry can be without any arguments, 
already have some attached, or even all attached.
When assigning values to such role features, the system first compares the verb 
structure with the patterns in the verb subcategorization table and identifies 
the pattern that fits best, using a scoring system in which
reward and penalty points are given for various types of matches
and mismatches of individual components:
\begin{itemize}
   \item -10,000 for each missing mandatory component
   \item -1,000 for each spurious primary (i.e.\ non-advp/pp) component
   \item +100 for each mandatory component covered before any spurious component
   \item +70 for each mandatory component covered after spurious advp/pp component
   \item +30 for each mandatory component covered after spurious primary component
   \item +10 for each optional component covered before any spurious component
   \item +7 for each optional component covered after spurious advp/pp component
   \item +3 for each optional component covered after spurious primary component
   \item -1 for each spurious app/pp component
\end{itemize}
The components of each pattern, whether mandatory or optional, 
have syntactic and semantic roles. A more detailed
description of subcategorization tables can be found in section~\ref{sec-subcat-tables}.

The same holds in principle for nominal patterns, but they are currently not used.
``{\bf syntrole}'' and ``{\bf syntrolep}'' are interpreted analogously.

\subsection{Form Features}

These features can be used to access the tense, aspect, person, number, case, 
gender, mode, voice and mood of a parse entry. The boolean features of examples 1-4 check whether 
a particular form value is compatible with the parse entry. Examples 5 and 6 can have vales 
{\it f-active} or {\it f-passive} and {\it f-masc}, {\it f-fem} or {\it f-neut} respectively.

\noindent
\begin{enumerate}
\item (f-finite-tense of -1 at m-boolean)
\item (f-part of 1 at m-boolean)
\item (f-pres-part of 1 at m-boolean)
\item (f-third-p of -1 at m-boolean)
\item (voice of -1 at f-voice)
\item (gender of -2 at f-gender)
\end{enumerate}

\subsection{Other Unary Boolean Features}
The following features check whether or not the parse entry referenced by $<$path$>$
is an abbreviated word, whether it is indexed (antecedent) or a referent (anaphor),
whether it is capitalized, or markedly capitalized (i.e.\ the capitalization is not
due to the placement of the word at the beginning of a sentence), or whether or not
the parse entry is lexical, i.e.\ represents a single word or another lexical unit 
such as {\it New York}.

\noindent
\begin{enumerate}
\item (is-abbreviation of $<$path$>$ at m-boolean)
\item (is-indexed of $<$path$>$ at m-boolean)
\item (is-ref of $<$path$>$ at m-boolean)
\item (capitalization of $<$path$>$ at m-boolean)
\item (marked-capitalization of $<$path$>$ at m-boolean)
\item (lexical of $<$path$>$ at m-boolean)
\end{enumerate}

\subsection{Binary Boolean Features}
The following features are special in that they involve two different parse entries.
They check whether the parse entry pair described by $<$path1$>$ and $<$path2$>$ are compatible
in terms of number, person and case as subject and verb phrase of a sentence (example 1),
whether they are compatible to form a compound verb as auxiliary and main verb (example 2),
and whether or not the concepts of the two parse entries are similar in that they
both represent agents, places, temporal intervals, or compatible types of quantities (example 3).

\noindent
\begin{enumerate}
\item (np-vp-match of $<$path1$>$ with $<$path2$>$ at m-boolean)
\item (compl-v-match of $<$path1$>$ with $<$path2$>$ at m-boolean)
\item (similar of $<$path1$>$ with $<$path2$>$ at m-boolean)
\end{enumerate}

\subsection{Other Features}
The first two example features show how optional parse entry slots such as {\it gap}
can be used. Notice the abstract path in example 2, which refers to the last parse stack
entry that has an active filler. Since the filler status can only be {\it active-filler},
{\it saturated-filler} or {\it unavail}, example 2 is used to check whether there exists
an active filler on the parse stack.
The third example feature checks, whether an integer is more likely to designate
the day of a month (1-31) or a year (e.g.\ 1997), which can be useful to distinguish between
{\it July 4} and {\it July 1776}.

\noindent
\begin{enumerate}
\item (gap of -2 at m-filler-status)
\item (gap of active-filler-1 at m-filler-status)
\item (day-or-year of -1 at i-enum-cardinal)
\end{enumerate}

\section{Training the Parser}
\label{sec-training-the-parser}

The decision structure that is used to parse sentences by deciding what parse action(s) to take
next is learnt from parse action examples.
The acquisition of these examples is a central task and is done in interactive training with
a supervisor.

For each training sentence, the system and the supervisor parse the sentence step by step,
starting with segmented and morphologically analyzed words on the input list and an empty stack,
e.g.\ ``{\it * The senators left .}''\footnote{As usual, the asterisk * denotes the current position,
between the parse stack and the input list.}
At the beginning, when there are no parse action examples and thus no decision structure,
the supervisor has to enter all parse actions by hand. In the given example, the first parse action
would be {\it (S S-ART)}, which shifts {\it The} from the input list to the parse stack, represented
by ``{\it (The) * senators left .}'' The representation of the elements on the parse stack uses 
brackets, because the parse entries can consist of complex parse entries that contain several words.

The complete parse action sequence for the given example is shown in 
table~\ref{fig-parse-action-sequence-senators}.
\begin{table}[htbp]
\begin{center}
\begin{tabular}{|l|l|} \hline
Parse state & parse action\\ \hline
* The senators left .     & (S S-ART)\\
(The) * senators left .   & (S S-NOUN)\\
(The) (senators) * left . & (R 1 TO S-NP AS PRED)\\
(The) (senators) * left . & (R 2 AS DET SAME)\\
(The senators) * left .   & (S S-ADJ)\\
(The senators) (left) * . & (R 1 TO S-VP AS PRED)\\
(The senators) (left) * . & (R 2 TO S-SNT AS (SUBJ THEME) SAME)\\
(The senators left) * .   & (S D-DELIMITER)\\
(The senators left) (.) * & (R 2 AS SAME DUMMY)\\
(The senators left.) *    & (DONE)\\ \hline
\end{tabular}
\caption{The complete parse action sequence for a very simple sentence. 
See page~\pageref{page-parse-tree-senators} for the resulting parse tree.}
\label{fig-parse-action-sequence-senators}
\end{center}
\end{table}
A detailed parse action sequence for a more interesting sentence is shown in appendix~\ref{app-parse-example}.

The system records the parse action sequence for a trained sentence in a {\it log file}.
Later, the system can automatically generate parse action examples from such log files.
This separation is very useful, because it allows examples to be regenerated automatically
if the set of features is changed.

After the system records the parse action, it interprets the command, executes it, and
displays the resulting parse state. During the training phase, the parse states are displayed in short
form, just as indicated in the table above. Internally of course, each element is a full fledged parse entry, or,
on the input list, a list of alternative parse entries, e.g.\ the following for {\it left}:

\newpage 
\begin{small}
\noindent
\begin{verbatim}
(OR  
 "left":
    synt:    S-ADJ
    class:   I-EADJ-LEFT
    forms:   (NIL)
    lex:     "left"
    props:   ((ADJ-TYPE S-NON-DEMONSTR-ADJ))  
 "left":
    synt:    S-VERB
    class:   I-EV-LEAVE
    forms:   (((TENSE F-PAST-PART)) ((TENSE F-PAST-TENSE)))
    lex:     "leave"
\end{verbatim}
\end{small}

The system repeats recording and interpreting parse actions 
until it reaches the parse action {\it done}. For the above example, at that point, the system contains
this final parse result on the parse stack:

\label{page-parse-tree-senators}
\begin{small}
\begin{verbatim}
"The senators left.":
    synt:    S-SNT
    class:   I-EV-LEAVE
    forms:   (((PERSON F-THIRD-P) (NUMBER F-PLURAL) 
               (CASE F-NOM) (TENSE F-PAST-TENSE)))
    lex:     "leave"
    subs:    
    (SUBJ THEME)  "The senators":
        synt:    S-NP
        class:   I-EN-SENATOR
        forms:   (((NUMBER F-PLURAL) (PERSON F-THIRD-P)))
        lex:     "senator"
        subs:    
        (DET)  "The":
            synt:    S-DEF-ART
            class:   I-EART-DEF-ART
            forms:   (NIL)
            lex:     "the"
            props:   ((CAPITALIZATION TRUE))
        (PRED)  "senators":
            synt:    S-COUNT-NOUN
            class:   I-EN-SENATOR
            forms:   (((NUMBER F-PLURAL) (PERSON F-THIRD-P)))
            lex:     "senator"
    (PRED)  "left":
        synt:    S-VERB
        class:   I-EV-LEAVE
        forms:   (((TENSE F-PAST-PART)) ((TENSE F-PAST-TENSE)))
        lex:     "leave"
    (DUMMY)  ".":
        synt:    D-PERIOD
        lex:     "."
\end{verbatim}
\end{small}

To generate parse action examples from parse action sequences recorded in a log file,
the system steps through the parse sequence of a sentence again. At each step, the system first
computes the values for all parse features, as described
in the previous section. The resulting feature vector plus the parse action provided by the supervisor form the
core of the parse action example. Other parse action example components include 
an identifier, e.g. `{\it (LOBBY 3 20 ``(The) * senators left .'')}', which links the example back to a
specific partially parsed sentence, e.g.\ sentence {\it 3} from corpus {\it LOBBY} step 20.

As described in the next section, parse action examples can be used to build a decision structure,
which can be used to predict the parse action for a given parse state. Even with only a few parse action
examples, e.g.\ from a single phrase, such decision structures immensely support further parse action
acquisition. For new training sentences, the decision structure proposes the parse action as a default, 
and for those that the system predicted right, the supervisor can just confirm the default by hitting
the return key, thereby no longer having to type it in. The learning curve of the decision structure
is initially so steep, that already for the second sentence, the supervisor can expect to enter more than half
of the parse actions by confirming the system proposed parse action default.

\begin{figure}[htbp]
\epsfxsize=16.4cm
\centerline{\epsfbox{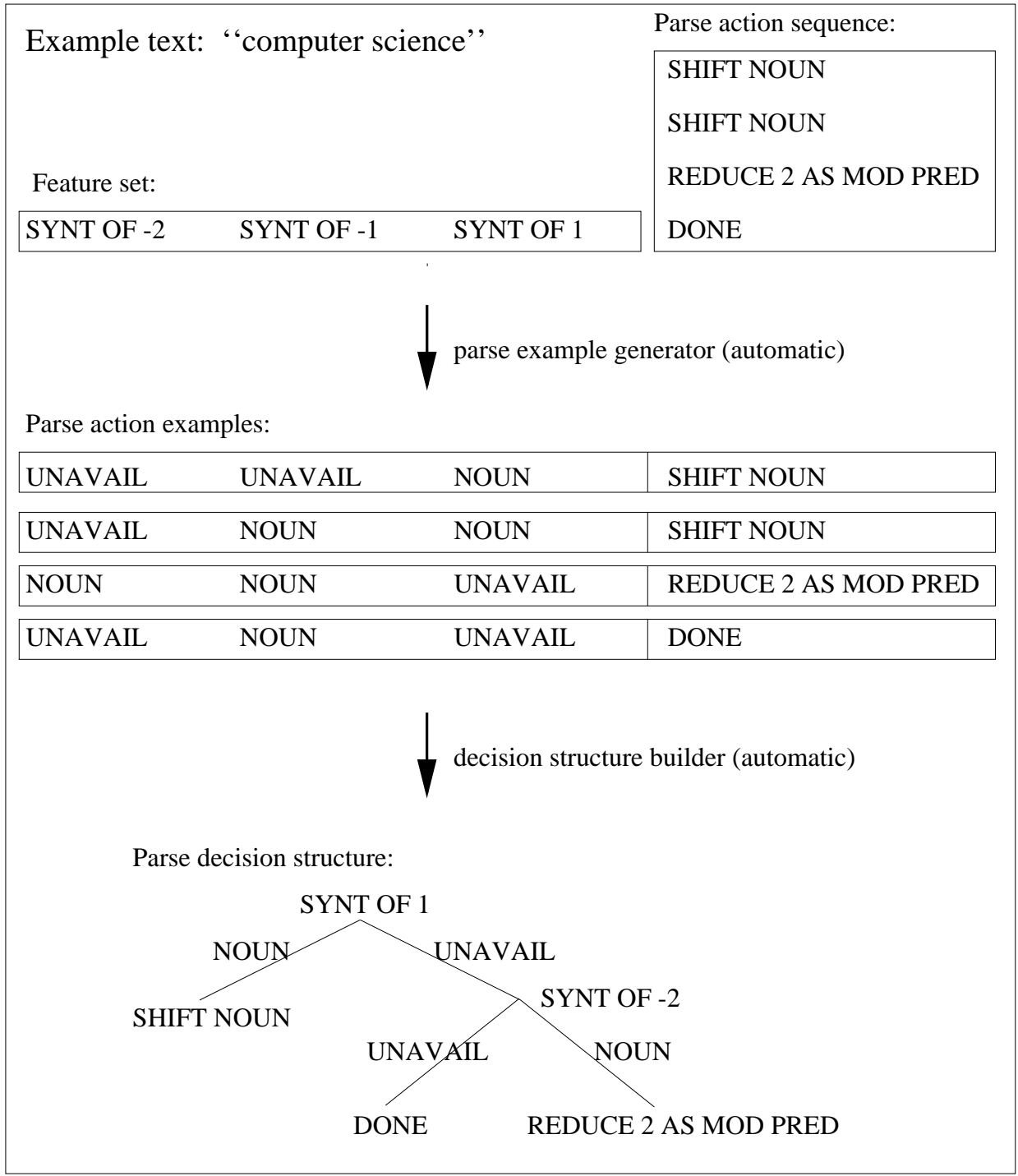}}
\caption{An example for generating parse action examples and learning a parse decision structure.}
\label{fig-parse_action_examples}
\end{figure}
The more parse action examples are acquired, the better the resulting decision structure will be,
so that the supervisor has to overrule the system with decreasing frequency. Soon, the actual work of
typing in parse action becomes almost negligible compared to the supervisor's effort to decide what the 
proper parse action of a sentence should be.

\section{Learning Decision Structures}

Traditional statistical techniques also use features, but often have to sharply limit
their number (for trigram approaches to three fairly simple features) to avoid
the loss of statistical significance.

Given that we use more than 200 features, it is obviously 
very critical to choose a decision structure and corresponding construction algorithm
that match the application well by exploiting domain knowledge to limit and/or bias
the selection of discriminating features or rules.

In parsing, only a very small number of features are crucial over a wide range 
of examples, while most features are critical in only a few examples, being used to 
`fine-tune' the decision structure for special cases. 

In order to overcome the antagonism between the importance of having a large number of 
features and the need to control the number of examples required for learning, particularly
when acquiring examples under supervision,
we choose a decision-tree based learning algorithm, which recursively selects the
most discriminating feature of the corresponding subset of training examples, 
eventually ignoring all locally irrelevant features, thereby tailoring the size of the 
final decision structure to the complexity of the training data.
Our decision structure is a hybrid that combines elements of decision trees,
decision lists and decision hierarchies.

\subsection{Review of Decision Trees and Lists}

Based on a training set of classified examples, algorithms like ID3 \cite{quinlan:mlj86}
and C4.5 \cite{quinlan:book93} build so-called {\it decision trees} that can classify
further (yet unclassified) cases. Consider the small training set with 14 examples
in table~\ref{fig-weather-training-set}. In each example, the weather is classified as
``{\it Play}'' or ``{\it Don't Play}''.

\begin{table}[htbp]
\begin{center}
\begin{tabular}{|cccc|c|} \hline
Outlook & Temp $(^{\circ}\hspace{-0.7mm}F)$ & Humidity (\%) & Windy? & Class \\ \hline
sunny         & 75 & 70 & true  & Play \\
sunny         & 80 & 90 & true  & Don't Play \\
sunny         & 85 & 85 & false & Don't Play \\
sunny         & 72 & 95 & false & Don't Play \\
sunny         & 69 & 70 & false & Play \\
overcast      & 69 & 90 & true  & Play \\
overcast      & 83 & 78 & false & Play \\
overcast      & 64 & 65 & true  & Play \\
overcast      & 81 & 75 & false & Play \\
rain          & 71 & 80 & true  & Don't Play \\
rain          & 65 & 70 & true  & Don't Play \\
rain          & 75 & 80 & false & Play \\
rain          & 68 & 80 & false & Play \\
rain          & 70 & 96 & false & Play \\ \hline
\end{tabular}
\caption{A small training set; from (Quinlan, 1993).}
\label{fig-weather-training-set}
\end{center}
\end{table}

\begin{figure}[htbp]
\epsfxsize=9.0cm
\centerline{\epsfbox{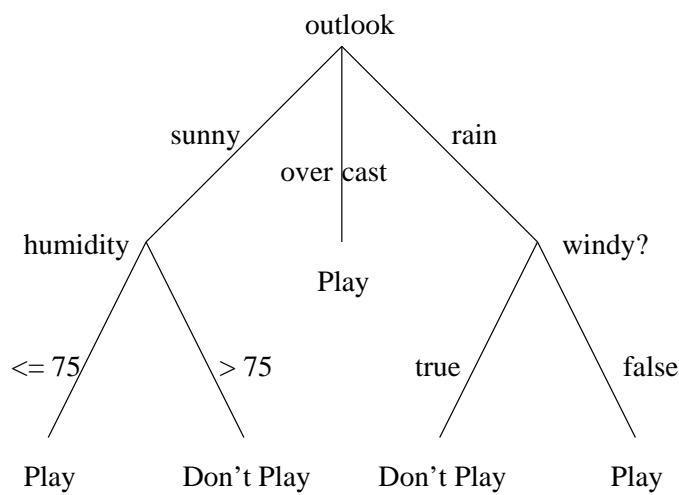}}
\caption{Decision tree built from training set shown in figure~\ref{fig-weather-training-set}}
\label{fig-weather-decision-tree}
\end{figure}

Given the examples of table~\ref{fig-weather-training-set}, the program C4.5 builds the
decision tree shown in figure~\ref{fig-weather-decision-tree}. This tree can not only be used
to reproduce the classes of the examples used to build it, but also to classify new examples,
e.g.\ (overcast/$75^{\circ}$F/40\% humidity/not windy) as ``{\it Play}''. The algorithm builds the
decision tree recursively top down. At each node, it selects the feature with the highest {\it gain ratio},
and repeats the decision tree building for each value (or value interval) of the selected
feature along with the corresponding example subset. The algorithm terminates when all remaining
training examples share the same class. 

\begin{table}[htbp]
\begin{center}
\begin{tabular}{|p{13.5cm}|} \hline
     \[ gain~ratio(X) = gain(X)/split~info(X) \]
     \[ split~info(X) = - \sum_{i=1}^{n} \frac{|T_{i}|}{|T|} * log_{2}\frac{|T_{i}|}{|T|} \]
     \[ gain(X) = info(T) - info_{X}(T) \]
     \[ info_{X}(T) = \sum_{i=1}^{n}\frac{|T_{i}|}{|T|}*info(T_{i}) \]
     \[ info(T) = - \sum_{j=1}^{k} \frac{freq(C_{j}, T)}{|T|} * log_{2}\frac{freq(C_{j}, T)}{|T|} bits\]  \\
     where $freq(C_{j}, T)$ is the frequency of the {\it j-th} class in example set {\it T}
     and the $T_{i}$'s are the different example subsets of {\it T} according to partitioning 
     test {\it X}. \\ \\ \hline
\end{tabular}
\end{center}
\caption{Definition of {\it gain ratio}. For a detailed motivation and explanation of the gain ratio, 
see (Quinlan, 1993), pp.\ 20 ff.}
\label{fig-gain-ratio-definition}
\end{table}

The gain ratio, as defined in table~\ref{fig-gain-ratio-definition} is a measure that describes 
how well a feature divides the example set into subsets 
that are to be as homogeneous as possible with respect to their classes.
We chose to use the {\it gain ratio} and not the plain {\it gain}, because the gain tends to bias
feature selection towards features with many values over those with only a few, and because
the feature set of our system has features with two to over 30 values, indeed a considerable
variation.

In our example, when computing the gain ratios for all four features (outlook, temperature, humidity, windy) 
at the top level, the highest gain ratio turns out to be for the `outlook' feature,
which partitions the example set into subsets for `sunny', `overcast', and `rain'.
All examples for `overcast' are classified as `Play', so the subtree for `overcast'
is a single node marked `Play'. The other two example subsets have to be split one more
time before all example subsets share the same class.

\begin{figure}[htbp]
\epsfxsize=8.0cm
\centerline{\epsfbox{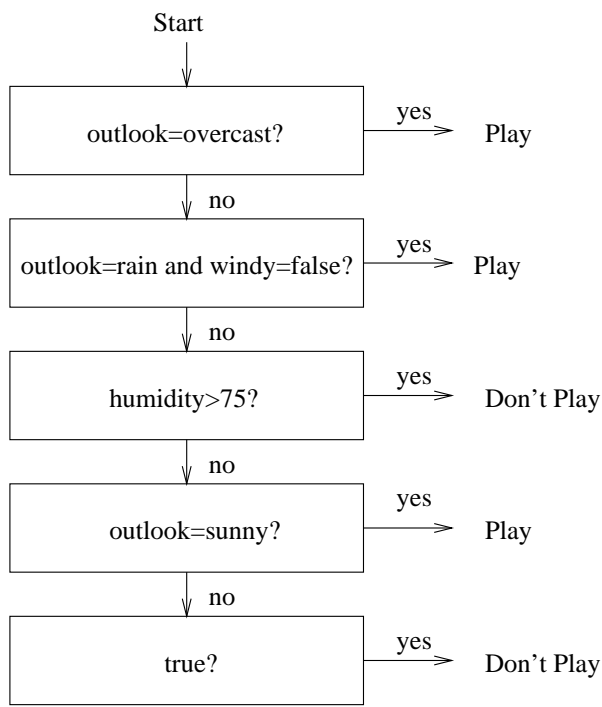}}
\caption{Decision list built from training set shown in figure~\ref{fig-weather-training-set}}
\label{fig-weather-decision-list}
\end{figure}

Alternatively, decision list algorithms \cite{rivest:ml87} generate a list of conjunctive rules,
where rules are tested in order and the first one that matches an instance is used to classify it.
%We use an implementation of the FOIL algorithm \cite{quinlan:mlj90}.
Figure~\ref{fig-weather-decision-list} shows the resulting decision list for the same weather 
training set.

\subsection{Decision Hierarchies}

Our first extension to the standard algorithms for decision trees and decision lists is the
concept of a decision hierarchy. Recall that while
parse actions might be complex for the action interpreter, they are atomic with respect to 
the decision structure learner; e.g.\ ``(R 2 TO VP AS PRED (OBJ PAT))'' would be such an atomic 
{\it classification}.
However, different parse actions are often related and apply in very similar contexts.
Consider for example the parse actions {\it (R 2 TO S-SNT AS (SUBJ AGENT) SAME)}
and {\it (R 2 TO S-SNT AS (SUBJ THEME) SAME)}. Both parse actions are used in a very similar
context: a noun phrase and a verb phrase are combined into a sentence, which then contains
all the components of the verb the {\bf same} way as in the verb phrase,
plus the noun phrase as a subject. The only difference is that in the first case, the subject
plays the semantic role of `agent', whereas in the second case it plays the semantic role 
of `theme'.

\begin{figure}[htbp]
\epsfxsize=10.0cm
\centerline{\epsfbox{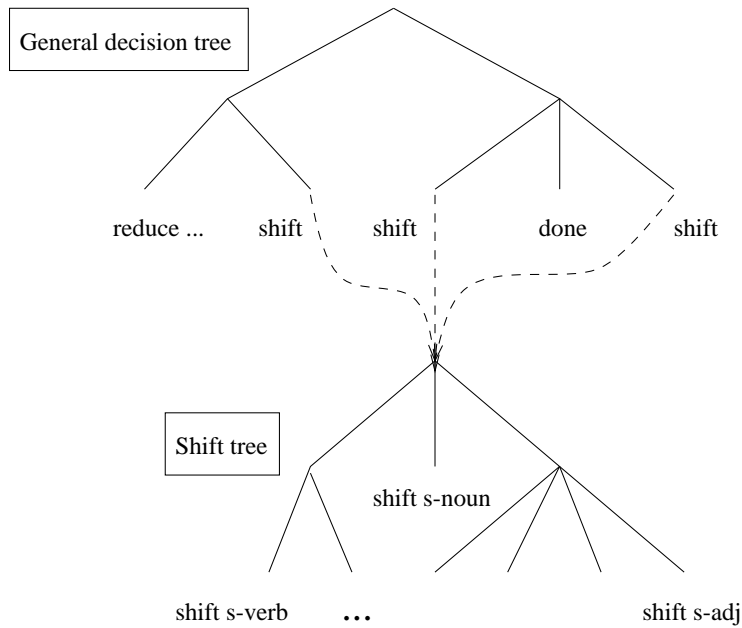}}
\caption{A two level decision hierarchy, where all shift parse actions have been grouped together.}
\label{fig-dec-hier}
\end{figure}

The knowledge of this type of similarity of parse actions can be exploited by defining sets of 
{\it similarity classes} for parse actions, e.g.\ the set of all parse actions that reduce a noun phrase
and a verb phrase to a sentence with the noun phrase as the subject, regardless of the subject's
semantic role.  Decision structures can then be constructed in two steps. During the first step,
a decision structure is built as usual except that all parse actions belonging to the same 
similarity class are treated as if the were the same.  In a second step, a special decision 
structure is built for the parse actions within each similarity class.

This allows the decision structure to split up the classification process into a coarse classification
and a fine classification step as illustrated in figure~\ref{fig-dec-hier}.
The advantage of this approach is that the system can `bundle' 
examples that are associated with different leaves of the coarse decision structure and thereby gain
a stronger example base for the fine classification. In principal, a decision structure can be split
up into an arbitrary number of levels of such decision hierarchies. 
The similarity classes are defined by predicates on parse
actions; all parse actions for which the predicate holds are defined to be members of the corresponding
similarity class. The predicates are defined manually. In our best-performing decision structure, 
we use two such decision hierarchy similarity classes: the one already mentioned in the preceding 
example and one which contains all parse actions that reduce a preposition and a noun phrase into a 
preposition phrase.

\subsection{Syntax of Decision Structure Definitions}

It is not only possible to define decision trees of decision trees, as illustrated in 
figure~\ref{fig-dec-hier}, but in fact any decision structure (e.g.\ trees or lists) can have any type
of sub-decision-structure, at any number of levels. In order to describe
such a complex multi-level hybrid structure, we introduce a formal syntax:

\begin{table}[htbp]
\begin{center}
\begin{tabular}{|lp{5cm}p{4cm}|} \hline
Entity & Definition & Comment \\ \hline
PRED: & $<$a parse action predicate$>$ & \\ 
DSTRUCT: & (PRED \{tree  $DSTRUCT^{*}$\}) & decision tree\\
         & (PRED dlist $DSTRUCT^{*}$) & decision list\\
         & (PRED slist $DSTRUCT^{+}$) & decision structure list\\ \hline
\end{tabular}
\caption{Syntax of decision structures.}
\label{fig-decision-structure-syntax}
\end{center}
\end{table}

Parse action predicates (``PRED'') are implemented functions that map parse actions to true or false.
Examples are 
{\it any-operation-p}, which holds for all parse actions;
{\it reduce-2-to-snt-operation-p}, which holds for all parse actions of the form {\it (R 2 TO S-SNT ...)}; and
{\it reduce-2-to-lexical-operation-p}, which only holds for the single parse action {\it (R 2 TO LEXICAL)}.

A decision structure (``DSTRUCT'') is a hybrid multi-layer structure composed of
decision trees, lists, and structure lists (as further explained in the following subsection).
Decision trees and lists can have any type of decision structure as a sub-component;
for decision structure lists, the specification of such sub-components is required.
The top level predicate (PRED) should be the all inclusive {\it any-operation-p}.
Table~\ref{fig-decision-structure-examples} gives examples of decision structure definitions.

\begin{table}[htbp]
\begin{center}
\begin{tabular}{|p{5.4cm}p{7.7cm}|} \hline
definition                         & description \\ \hline
(any-operation-p)                  & simple (non-hierarchical) decision tree \\
(any-operation-p tree)             & same as above \\
(any-operation-p dlist)            & simple decision list \\
(any-operation-p tree              & defines a hierarchical decision tree with \\
 \hspace*{1cm}(reduce-operation-p) & \hspace*{5mm} sub-decision-trees for reduce and shift \\
 \hspace*{1cm}(shift-operation-p)) & \hspace*{5mm} parse actions\\
(any-operation-p dlist             & defines a hierarchical decision list with \\
 \hspace*{1cm}(reduce-operation-p dlist) & \hspace*{5mm} a sub-decision-list for reduce parse actions \\
 \hspace*{1cm}(shift-operation-p)) & \hspace*{5mm} a sub-decision-tree for shift parse actions \\
(any-operation-p slist             & defines the decision structure list \\
 \hspace*{1cm}(done-operation-p)   & \hspace*{5mm} shown in figure~\ref{fig-dec-structure-list} \\
 \hspace*{1cm}(reduce-operation-p) & \\
 \hspace*{1cm}(shift-operation-p)) & \\
\hline
\end{tabular}
\caption{Examples of decision structures. Note: `any-operation-p', `shift-operation-p', 
`reduce-operation-p', and `done-operation-p' are predefined predicates for similarity classes, 
containing all parse actions, all shift parse actions, all reduce parse actions, and all
done parse actions respectively.}
\label{fig-decision-structure-examples}
\end{center}
\end{table}

\subsection{Decision Structure Lists}
\label{subsec-hybrid_dec_str_list}
The final extension to decision trees and lists we introduce is the {\it decision structure list}.
Recall that the major motivation for decision hierarchies was that we wanted to exploit our knowledge
that some of the currently 265 different parse actions are quite related to each other; using
decision hierarchies, we can split the decision structure learning into a coarse and a fine learning
part with the advantage that the examples within a similarity class, no matter where in a coarser
tree they might end up, are bundled back together, thereby providing a larger example base for
`fine-tuning'.

Another piece of linguistic background knowledge that we can exploit is the notion of
exceptionality vs.\ generality of parse actions. Natural language exhibits a complex pattern of 
regularities, sub-regularities, pockets of exceptions, and idiosyncratic exceptions.
Consider for example the noun phrase ``New York exchanges''.
We don't want to treat ``New'' like a normal adjective, and in particular, we don't want
to attach it as a standard modifier to the noun-compound ``York exchanges''. 
Instead, we want to take advantage of a special lexicon entry for ``New York'' and perform a 
special reduction (``R 2 TO LEXICAL'') that is principally lexically motivated.
Since parse actions often reflect this notion of degree of exceptionality, we can steer the
more exceptional types of parse actions towards an early treatment, so that they later don't
`confuse' the more normal parse actions like shift and reduce. We achieve this by defining
a list of similarity classes that we order such that more exceptional types of parse actions
come first and more standard ones later. Figure~\ref{fig-dec-structure-list} gives an example.
The system first separates the relatively special `done' operations from the rest of the flock,
and then proceeds with `reduce' operations before finally dealing with the fairly standard
`shift' operations. When building the decision structure, all `done' examples are discarded
when subsequent decision structure list sub-structures are built. 
We basically aim for a decrease of exceptional examples that are likely to `pollute' 
the patterns of more regular types of parse actions.

\subsubsection{Definition of Decision Structure Lists}

\begin{figure}[htbp]
\epsfxsize=16.4cm
\centerline{\epsfbox{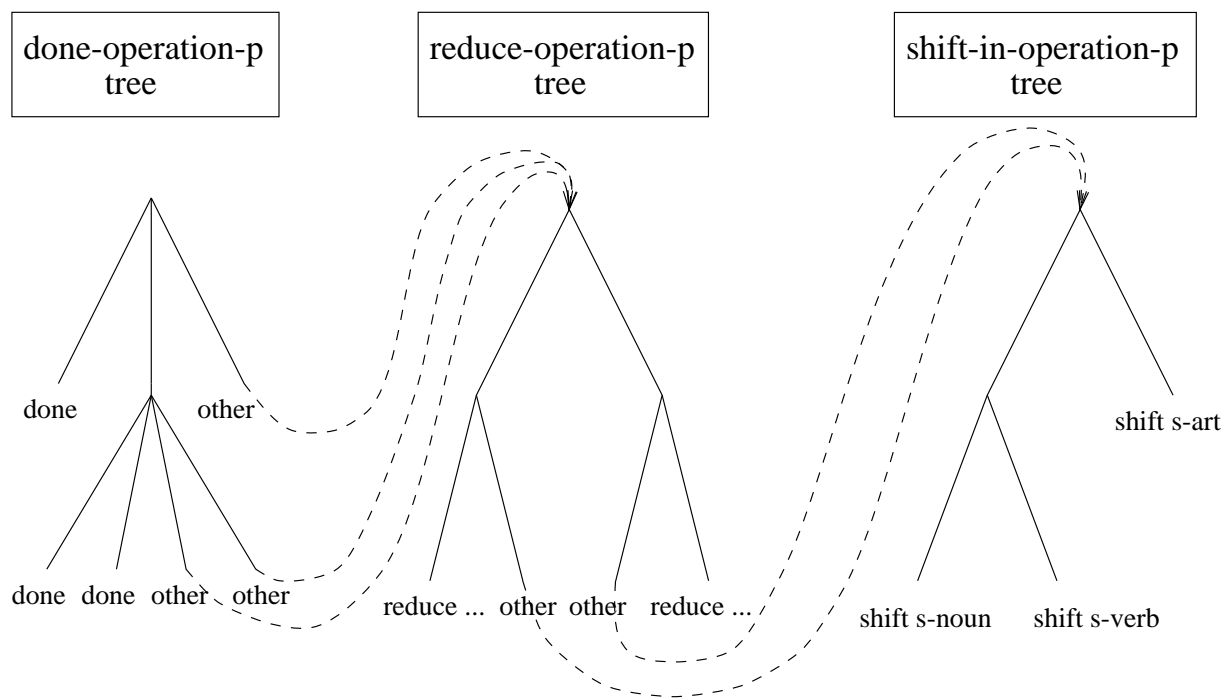}}
\caption{A decision structure list}
\label{fig-dec-structure-list}
\end{figure}

More formally, 
a {\it decision structure list} is a list of element decision structures each of which classifies 
examples by assigning normal classes or the special class {\it OTHER}. 
If an element decision structure assigns a normal
class, that class also becomes the class of the entire decision structure list; otherwise the
example is classified by the next element decision structure in the decision structure list. This is
repeated until the example is assigned a normal class. The last element decision structure should never
assign a class {\it other}, because it would be undefined. As mentioned before, 
figure~\ref{fig-dec-structure-list} gives an example of a decision structure list; 
it is generated from the definition
(any-operation-p slist (done-operation-p) (reduce-operation-p) (shift-in-operation-p)).

A decision structure list is defined by a tuple (PRED slist $DSTRUCT_{1}$, $DSTRUCT_{2}$, ... $DSTRUCT_{n}$),
where {\it n} must be at least one, but should be at least two to make any sense.
Each sub-structure $DSTRUCT_{i}$ is headed by a predicate, which we shall refer to as $PRED_{i}$.

For each $PRED_{i}$, the decision structure list algorithm
produces an element decision structure according to the type of $DSTRUCT_{i}$,
but with the modification that all parse actions for which $PRED_{i}$ does not hold are
classified as {\it OTHER}, a special class that is linked to the 
following element decision structure of the decision structure list.

\subsubsection{Differences between Decision Structure Lists and Decision Hierarchies}

The decision hierarchies and the decision structure lists share the concept of similarity
classes and both build a larger acyclic graph out of smaller decision structures. But while
decision hierarchies combine all classes {\it within} a similarity class, decision structure lists
combine all classes {\it outside} the similarity class. And while the multiple similarity classes 
within a decision hierarchy will cause their corresponding sub-decision-structures to be arranged 
side by side, the various elements of a decision structure list are arranged in sequence.

\subsubsection{Example of a Hybrid Decision Structure}

\begin{figure}[htbp]
\epsfxsize=16.4cm
\centerline{\epsfbox{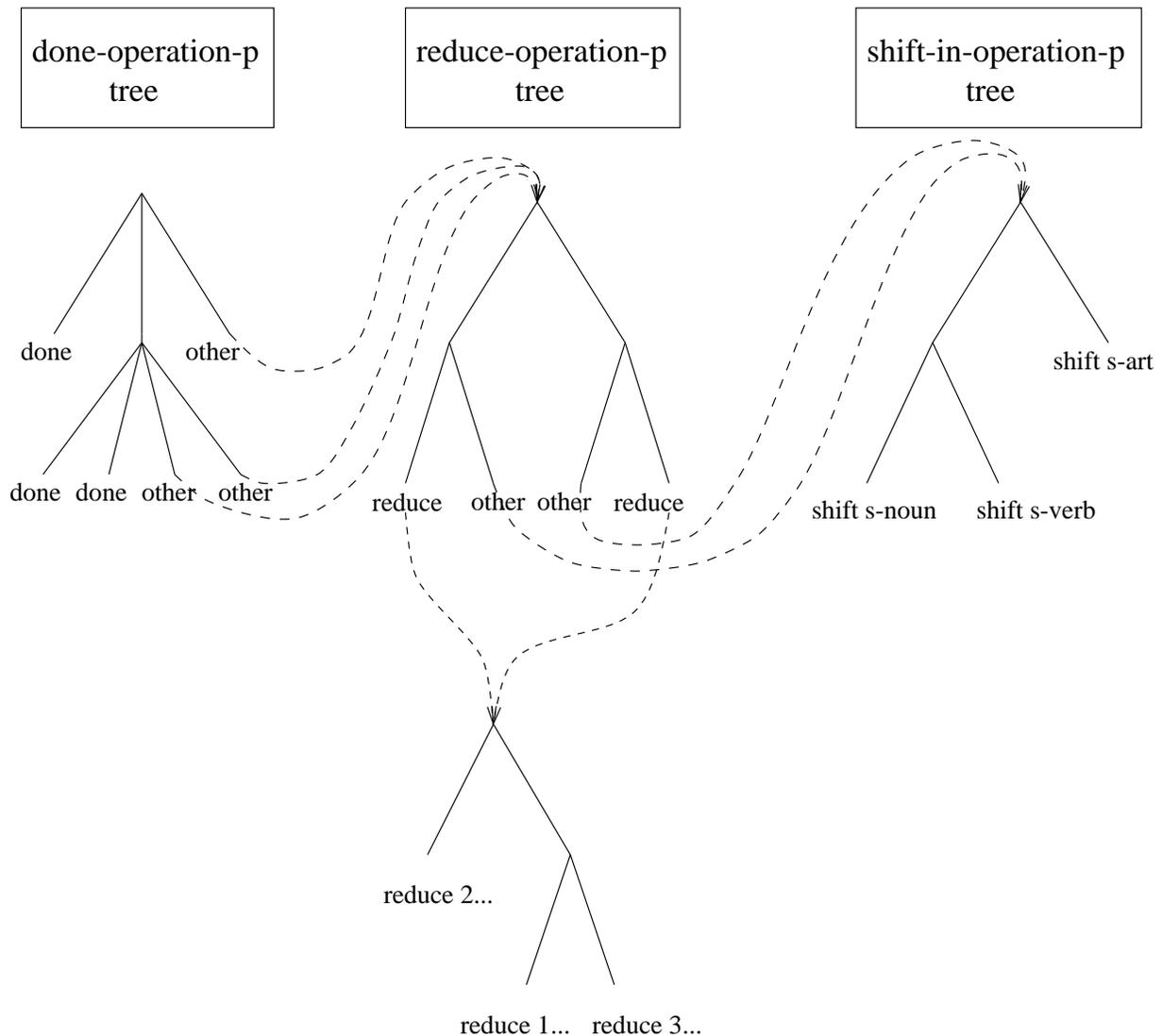}}
\caption{Example of a hybrid decision structure}
\label{fig-hier_dec_str_list}
\end{figure}

Figure~\ref{fig-hier_dec_str_list} shows a hybrid decision structure that is similar to the
one depicted in figure~\ref{fig-dec-structure-list}. It introduces an additional level
of decision hierarchy under the reduce-operation-p tree, reducing the number of different
classifications in that tree to two, `reduce' and `other'. Since there is only a single
`done' operation anyway, a similar sub-tree wouldn't have made sense for the done-operation-p
tree.

\subsubsection{The Hybrid Decision Structure Used For Parsing}

The decision structure list shown in table~\ref{fig-top-decision-structure-list}
defines the hybrid decision structure used in our parsing 
experiments. The structure has been constructed manually, following two principles:
\begin{enumerate}
  \item Group similar parse actions together, e.g.\ all `co-index' parse actions or
        all those of the form ``(r 2 to s-pp ...)''.
  \item Sort the resulting groups such that the more exceptional and specialized ones
        precede the more general groups.
\end{enumerate}
Since the sub-decision-structures are built sequentially, with all examples covered by
a previously listed predicate discarded, the more exceptional examples are no longer
around when the more general sub-decision-structures are built.

\begin{table}[htbp]
\begin{verbatim}
(any-operation-p slist
   (reduce-2-to-lexical-operation-p)
   (reduce-3-to-lexical-operation-p)
   (reduce-5-to-lexical-operation-p)
   (done-operation-p)
   (expand-operation-p)
   (co-index-operation-p)
   (add-operation-p)
   (empty-cat-operation-p)
   (mark-operation-p)
   (shift-out-operation-p)
   (reduce-4-operation-p)
   (reduce--3--1-operation-p)
   (reduce--4--1-operation-p)
   (reduce-non-contiguous-operation-p)
   (reduce-3-as-dummy-pred-operation-p)
   (reduce-3-as-comp-conj-pred-operation-p)
   (reduce-3-as-same-operation-p)
   (reduce-3-operation-p)
   (reduce-1-operation-p)
   (reduce-2-as-pred-or-same-dummy-operation-p)
   (reduce-2-to-pp-operation-p tree 
     (reduce-2-to-pp-operation-p))
   (reduce-2-to-snt-operation-p tree 
     (reduce-2-to-snt-operation-p))
   (reduce-2-operation-p)
   (shift-in-operation-p))
\end{verbatim}
\caption{Definition of the hybrid decision structure used in parsing experiments.}
\label{fig-top-decision-structure-list}
\end{table}

Based on the decision structure list definition in table~\ref{fig-top-decision-structure-list}, 
the system first builds a decision
tree (default) that classifies all examples as some {\it reduce-2-to-lexical-operation-p} parse action
or {\it OTHER}. Then, using all examples that have previously been classified as other, it builds the
next decision tree and so forth. The union of decision structure list parse action classes 
should always be the full set of parse examples, but the classes are allowed to overlap.

Note that the predicates partitioning the examples are not necessarily disjoint.
In particular, all examples covered by {\it reduce-3-as-comp-conj-pred-operation-p}
are also covered by {\it reduce-3-operation-p}, but this is no problem, because all
examples by the former are filtered out, so that the sub-decision-structure for
{\it reduce-3-operation-p} is trained only on all remaining examples. It is important
however that the disjunction of all sub-decision-structure predicates cover the
entire example space.

Notice further the double occurrence of {\it reduce-2-to-pp-operation-p}. The first
occurrence causes all {\bf non}-{\it reduce-2-to-pp-operation-p} parse actions to be
lumped together as {\it ``other''}, whereas the second occurrence groups all
{\it reduce-2-to-pp-operation-p} parse actions together. As a result, the corresponding
decision tree has exactly two classifications: a token representing 
{\it reduce-2-to-pp-operation-p}, and {\it ``other''}. A lower level decision tree
then separates the different parse actions within {\it reduce-2-to-pp-operation-p}.

After initially obtaining a significant improvement using a decision structure list
along the two principles outlined above,
further variation did not yield significant additional improvement. This seems to indicate
that the two basic principles above have merit, but that the results don't depend very
much on the detailed configuration of the decision structure list.

\subsection{Feature Selection Bias in Decision Trees}

In parsing, certain features are much more specialized than others. For example, consider the feature 
{\it (CLASSP OF I-EN-INTERR-PRONOUN OF -2 AT M-BOOLEAN)}, which checks whether or not the second
element on the parse stack is an interrogative pronoun, and {\it (SYNT OF -2 AT S-SYNT-ELEM)},
which checks the general part of speech of the same element. 
For better generalization when building decision trees and also
for a potentially faster learning time, we want to restrict or discourage the selection
of very specialized features before certain more general features
have already been used higher up in the decision structure.
For this we define a feature dependency graph, which for each feature specifies
which other features should have been used before when building a sub-decision-structure.

In our system, the use of a feature is discouraged, if the feature(s) describing the general part(s)
of speech of the one
or more parse stack or input list elements that are accessed in the feature have not been
used yet. 
The use of feature {\it (CLASSP OF I-EN-INTERR-PRONOUN OF -2 AT M-BOOLEAN)} for example would be 
discouraged until {\it (SYNT OF -2 AT S-SYNT-ELEM)} has been used.

Features checking the general part of speech are discouraged if there is still a feature for the
part of speech of some element closer to position 0 that has not been used yet. 
This means that the only features that are always without any such bias are {\it (SYNT OF -1 AT S-SYNT-ELEM)}
and {\it (SYNT OF 1 AT S-SYNT-ELEM)}.
Features that have the same value for all remaining examples have an information gain of 0 and
are considered to have been used, regardless of whether or not they actually have been.

Features are discouraged by dividing the ``raw'' {\it gain-ratio} by
\[1 + c_{1} * d^{c_{2}}\]
where {\it d} is the degree of dependency violation (e.g.\ 2 for 
{\it (SYNT OF -4 AT S-SYNT-ELEM)} if (SYNT OF -1 AT S-SYNT-ELEM), but neither 
(SYNT OF -2 AT S-SYNT-ELEM) nor (SYNT OF -3 AT S-SYNT-ELEM) had been used before. 
$c_{1}$ and $c_{2}$ are constants.

Experiments varying the strength of this bias towards more general features by choosing different
values for $c_{1}$ and $c_{2}$, from `no bias' 
($c_{1}$ = 0) over several intermediate values to `very strong bias'
(high values for both $c_{1}$ and $c_{2}$)
have shown that stronger generalization biases produce better decision structures. 
In fact, the extreme bias of not considering any feature with any dependency violation
leads to the best results. This allows the
decision tree builder to ignore many (yet too) specialized features before even computing a gain ratio,
thus considerably accelerating the decision tree building process. 

\subsection{Feature Selection Preferences}
\label{subsec-feature_selection_preferences}
The supervisor can optionally mark a parse example with one or more features indicating
that these are relevant for the specific example. Later on, this information can be used
to bias the feature selection towards those features which have been marked as relevant in the
example set that a remaining decision tree is built from. This bias is implemented by increasing
the ``raw'' gain ratio in a way similar to the one we described in the last subsection.

To a degree this option is a remnant of our previous approach of collecting,
from selected examples, features that were explicitly designated as relevant, 
instead of full feature vectors for all examples.
Out of a total of 11,822 parse action examples, 702 or 5.9\% contain such individual feature 
selection preference indications, with a total of 812 preference indications.\\

Simple decision trees, simple decision lists, hierarchical decision trees and 
hybrid decision structures containing hierarchical decision trees were all used
in our parsing experiments. The results are compared in 
section~\ref{sec-contribution-of-decision-structure-type}.

\section{Garden Paths}
\label{sec-garden-path}

Garden path sentences are sentences that initially mislead the reader in their syntactic analysis.
Classical examples \cite{marcus:book80} are:
\begin{enumerate}
\item The horse raced past the barn fell.
\item Cotton clothing is made of grows in Mississippi.
\end{enumerate}
Since local parsing preferences, e.g.\ the preference to interpret {\it raced} in example (1) as a finite verb or
to interpret {\it Cotton clothing} in example (2) as a compound noun, aren't recognized as incorrect
until much later in a left-to-right deterministic parse,
these garden path sentences present a difficulty, which is not surprising, because
the deterministic parser basically mimics a human reader in the way it processes sentences and
human readers experience these problems as well.

With regard to attachment decisions, our parser follows a relatively `conservative' policy and makes
the attachments relatively late (because the supervisor trained it that way). For example, it postpones 
reducing subject and verb until the entire verb phrase has
been processed. This allows it to circumnavigate the trap laid out by garden path sentence (1). In
other garden path sentences, such as (2), where our system will interpret {\it Cotton clothing} as a 
complex noun, a locally preferred choice later turns out to be wrong.

Good writers can avoid garden path sentences. ``The horse {\it which was} raced past the barn fell.''
and ``Cotton {\it that} clothing is made of grows in Mississippi.'' read much more easily.
Poor writing will never disappear, but garden path sentences are rare enough in practical
environments that quantitatively their nuisance is minor compared to other parse errors.

Nevertheless there are solutions to cope with garden path sentences in a deterministic parsing
paradigm. The system can learn to recognize {\it when} it has been misled, even if it is already too late
for the `proper' parse. Many garden path sentences are based on a few types of structural
ambiguity, caused for example by phonologically unrealized relative pronouns. When training a parser on
garden path traps, the parser could either be guided to directly repair the current partial path,
which would typically have to include some sort of `unreduce' operations, or it might somehow mark the 
trouble spot and use this mark when reaching the critical point of the sentence again after a parsing
restart. This method certainly somewhat breaks out of the normal deterministic paradigm, but that should be
acceptable, because human readers have to resort to this as well.

During the testing of parsing and translation of WSJ sentences, the parser in one case actually followed 
a wrong path: {\it The Federal Farm Credit Banks Funding Corp. plans to offer \$1.7 billion of 
bonds Thursday.} It is questionable whether or not this is a true garden path sentence, since the
local choice of whether {\it plans} is a noun or a verb might have a preference for a verbal
interpretation anyway; however, even though the system recognizes a part of speech ambiguity for
{\it plans}, it chooses the nominal interpretation, because none of the comparable training instances
that the parser was trained on
called for a verbal interpretation. (When training the system on 256 sentences, which includes 40
examples where a new word is shifted in that could be both a noun or a verb and that has a noun at position -1,
i.e.\ at the top of the parse stack, all those 40 examples select {\it noun} as the proper part
of speech.)
               %\chapter{Parsing} \label{ch-parsing}
\chapter{Parsing Experiments}          
\label{ch-p_exp}

This chapter presents results on training and testing a prototype implementation 
of our system with sentences from the Wall Street Journal, a prominent corpus of
`real' text, as collected on the ACL-CD.

\section{Corpus}
\label{ch-p_exp_corpus}

In order to limit the size of the required lexicon, we work on a reduced corpus
that includes all those sentences that are fully covered by the 3000 most
frequently occurring words (ignoring numbers etc.) in the entire corpus.
The lexically reduced corpus contains 105,356 sentences, a tenth of the full corpus.
3000 words is typically considered to be the size of the basic vocabulary of
a language. The lexical size is small enough to build a lexicon and corresponding 
KB entries with moderate tools in reasonable time, while still allowing a sufficiently
large number of sentences that still include a rich linguistic diversity.

For our training and testing we use the first 272 of the 105,356 sentences.
They vary in length from 4 to 45 words, averaging at 17.1 words and 43.5 parse
actions per sentence.
These 272 sentences are from a section of the Wall Street Journal dated March 23/24, 1987.
One of these sentence is ``{\it Canadian manufacturers' new orders fell to
\$20.80 billion (Canadian) in January, down 4\% from December's \$21.67 billion
on a seasonally adjusted basis, Statistics Canada, a federal agency, said.}''.
A complete listing of the 272 training and testing sentences can be found in
appendix~\ref{app-corpus}.

\section{Test Methodology and Evaluation Criteria}
\label{sec-parse-eval-methodology}

As a result of supervised acquisition, as described in section~\ref{sec-training-the-parser}, 
the correct parse action sequences for the 272 WSJ sentences have been recorded.
For the following parsing test series, the corpus of these 272 sentences
is divided into 17 blocks of 16 sentences each. The 17 blocks are then consecutively
used for testing. For each of the 17 sub-tests, a varying number of
sentences from the {\it other} blocks is used for training the parse decision structure,
so that within a sub-test, none of the training sentences are ever used as a test sentence.
The results of the 17 sub-tests of each series are then averaged. Such a test is also
referred to as a 17-fold cross-validation.

The following standard \cite{goodman:acl96} evaluation criteria are used:\\

\begin{tabbing}
\hspace*{1cm} $ {\bf Precision} = \frac{C_{system}}{N_{system}} $
%{\bf Precision:}
%\begin{tabular}{c}
%$C_{system}$ \\ \hline
%$N_{system}$ \\
%\end{tabular}\\

\hspace{2cm} \=

$ {\bf Labeled~precision} = \frac{L_{system}}{N_{system}} $ \\[1cm]
%{\bf Labeled precision:}
%\begin{tabular}{c}
%$L_{system}$ \\ \hline
%$N_{system}$ \\
%\end{tabular}\\

\hspace*{1cm} $ {\bf Recall} = \frac{C_{system}}{N_{logged}} $
%{\bf Recall:}
%\begin{tabular}{c}
%$C_{system}$ \\ \hline
%$N_{logged}$ \\
%\end{tabular}\\

\>

$ {\bf Labeled~recall} = \frac{L_{system}}{N_{logged}} $ \\
%{\bf Labeled recall:}
%\begin{tabular}{c}
%$L_{system}$ \\ \hline
%$N_{logged}$ \\
%\end{tabular}\\

\end{tabbing}

\noindent where\\
$N_{system}$  = number of constituents in system parse \\
$N_{logged}$  = number of constituents in logged parse \\
$C_{system}$  = number of correct constituents in system parse \\
$L_{system}$  = number of correct constituents with correct syntactic label in system parse \\

\noindent
{\bf Tagging accuracy:} percentage of words with correct part of speech assignment.\\
{\bf Crossing brackets:} number of constituents in system parse which violate constituent 
boundaries with a constituent in the logged parse.\\
{\bf Correct operations} measures the number of correct operations during a parse that is
continuously corrected based on the logged sequence.
A sentence has a correct {\it operating sequence}, {\bf OpSequence},
if the system fully predicts the logged parse action sequence,
and a correct {\it structure and labeling}, {\bf Struct\&Label}, if the structure
and syntactic labeling of the final system parse of a sentence is 100\% correct,
regardless of the operations leading to it.

\section{Accuracy on Seen Sentences}

The current set of 205 features was sufficient to always discriminate examples
with different parse actions. This guarantees an 100\% accuracy on sentences
already seen during training.
While that percentage is certainly less important than the accuracy figures for
unseen sentences, it nevertheless represents an upper ceiling, which for many
statistical systems lies significantly below 100\%.

\section{Testing on Unseen Sentences}

The first test series was done with a varying number of training sentences. Here,
as well as in all other test series, results are for training 
with all 205 features and using the hybrid decision structure list described in 
subsection~\ref{subsec-hybrid_dec_str_list}.

In some test cases, parsing could not be fully completed. This can happen, when the
decision structure prematurely selects {\it ``done''} as the next parse action.
This happens in particular when the system proposes a parse action that is actually
undefined in the specific current parse state and therefore overruled by the sanity
checker; if the decision structure can naturally provide an alternative legal parse 
action, e.g.\, in a decision list, by proceeding through the remainder of the list,
that action is chosen; otherwise, the system resorts to {\it ``done''} as
a last resort. In that case, all elements on the parse stack and input list are 
lumped together under a new parse entry node with the generic syntactic label 
{\it S-SYNT-ELEM} and the generic semantic label {\it I-EN-THING} and with the 
unspecified roles {\it CONC} for its components.
This way, there is at least formally a parse tree that can be evaluated or
passed on to transfer for full translation.

The other anomaly is the `endless' loop in which the system spins into a repetitive
parse action sequence, in which the parser for example keeps inserting empty categories,
or keeps reducing the same single parse entry over and over again. This occurs only
very rarely when the system has been trained on sufficient data, an adequate feature 
set and an appropriate type of decision structure. Nevertheless, the parser is equipped
with an `endless' loop detector, that basically measures parsing progress in terms
of the number of elements on the parse stack and input list, which gradually decreases,
though not necessarily monotonically. By allowing up to four
more parse steps for each remaining parse stack or input list element, the `endless'
loop detector was able to detect all loops fairly quickly and never terminated any
parsing sequence that was in fact still promising.
When an `endless' loop is detected, the parsing is stopped by selecting the parse
action {\it ``done''} and by then proceeding as described in the previous paragraph.

These pathological cases are fully included in the following test data, not surprisingly
with an overall negative impact. The number of test sentences caught in an `endless' 
loop is explicitly included in the following tables. Prematurely terminated sentences
obviously produce a lower recall score and have no chance at scoring as a fully correct
operation sequence or even produce a parse tree with correct structure and labeling.
However they can also lead to lower crossings, since the constituent structure is
often still incomplete.

\noindent
\begin{table}[htb]
\begin{center}
\begin{tabular}{|l|c|c|c|c|c|} \hline \hline
   Number of training sentences    & 16       & 32       & 64       & 128      & 256      \\ \hline
   Precision                       & 85.1\%   & 86.6\%   & 87.7\%   & 90.4\%   & 92.7\%   \\
   Recall                          & 82.8\%   & 85.3\%   & 87.7\%   & 89.9\%   & 92.8\%   \\
   Labeled precision              & 77.2\%   & 80.4\%   & 82.5\%   & 86.6\%   & 89.8\%   \\
   Labeled recall                 & 75.0\%   & 77.7\%   & 81.6\%   & 85.3\%   & 89.6\%   \\
   Tagging accuracy                & 96.6\%   & 96.5\%   & 97.1\%   & 97.5\%   & 98.4\%   \\
   Crossings per sentence          & 2.5      & 2.1      & 1.9      & 1.3      & 1.0      \\
   Sent.\ with 0 crossings         & 27.6\%   & 35.3\%   & 35.7\%   & 50.4\%   & 56.3\%   \\
   Sent.\ with up to 1 crossing    & 44.1\%   & 50.7\%   & 54.8\%   & 68.4\%   & 73.5\%   \\
   Sent.\ with up to 2 crossings   & 61.4\%   & 65.1\%   & 66.9\%   & 80.9\%   & 84.9\%   \\
   Sent.\ with up to 3 crossings   & 72.4\%   & 77.2\%   & 79.4\%   & 87.1\%   & 93.0\%   \\
   Sent.\ with up to 4 crossings   & 84.9\%   & 86.4\%   & 89.7\%   & 93.0\%   & 94.9\%   \\
   Correct operations              & 79.1\%   & 82.9\%   & 86.8\%   & 89.1\%   & 91.7\%   \\
   Sent.\ with correct OpSequence  &  1.8\%   &  4.4\%   &  5.9\%   & 10.7\%   & 16.5\%   \\
 Sent.\ with correct Struct\&Label &  5.5\%   &  8.8\%   & 10.3\%   & 18.8\%   & 26.8\%   \\
   Sentences with endless loop     &   13     &   6      &    0     &    1     &    1     \\ \hline
\end{tabular}
\end{center}
\caption{Evaluation results with varying number of training sentences; with 205 features}
\label{fig-pex_training_size}
\end{table}
\begin{figure}[htbp]
\epsfxsize=13.9cm
\centerline{\epsfbox{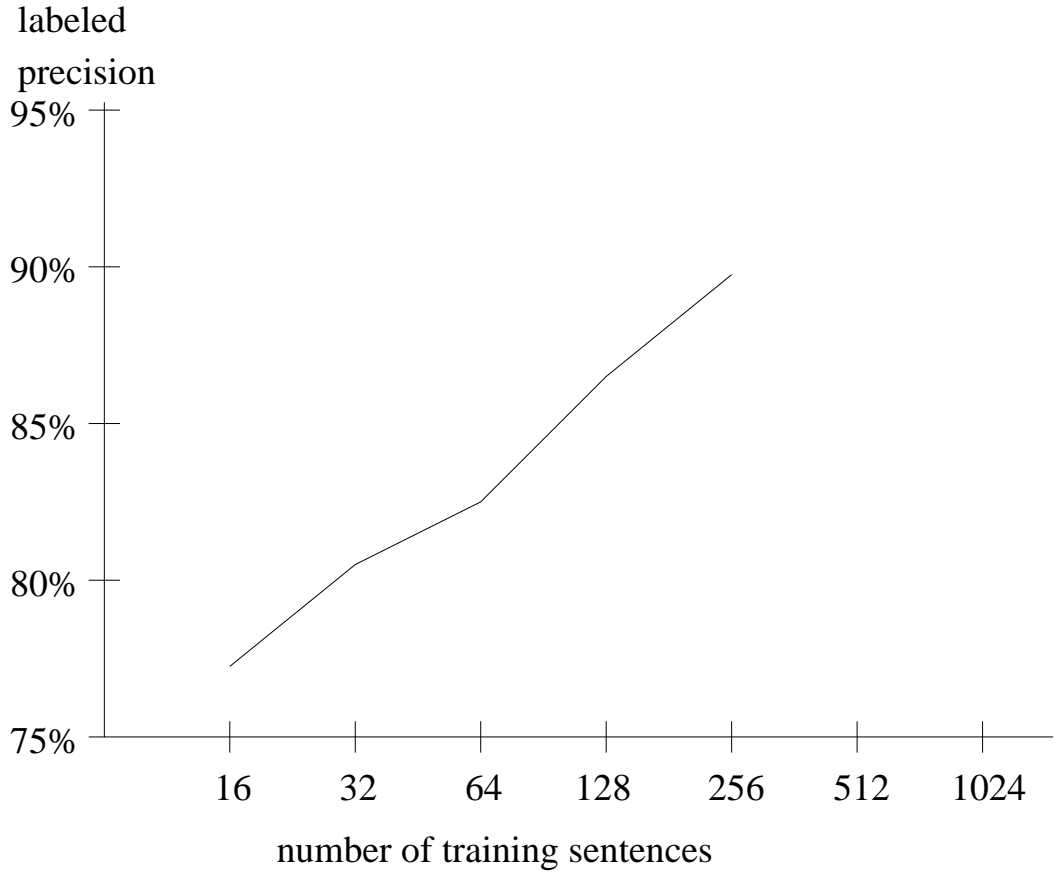}}
\caption{Learning curve for labeled precision (see table~\ref{fig-pex_training_size})}
\label{fig-lab_prec_learning_curve}
\end{figure}
\begin{table}[htbp]
\begin{center}
\begin{tabular}{|l|c|c|c|c|c|} \hline \hline
   Number of training sentences    & 16       & 32       & 64       & 128      & 256      \\ \hline
   Precision                       & 84.6\%   & 86.6\%   & 88.3\%   & 90.5\%   & 91.5\%   \\
   Recall                          & 82.3\%   & 85.2\%   & 88.1\%   & 90.3\%   & 91.6\%   \\
   Labeled precision              & 76.5\%   & 79.6\%   & 82.9\%   & 86.2\%   & 87.9\%   \\
   Labeled recall                 & 74.0\%   & 77.9\%   & 82.2\%   & 85.4\%   & 87.7\%   \\
   Tagging accuracy                & 96.5\%   & 96.7\%   & 97.0\%   & 97.9\%   & 98.2\%   \\
   Crossings per sentence          & 2.6      & 2.1      & 1.7      & 1.4      & 1.2      \\
   Sent.\ with 0 crossings         & 26.1\%   & 33.8\%   & 39.3\%   & 46.3\%   & 53.3\%   \\
   Sent.\ with up to 1 crossing    & 43.8\%   & 50.7\%   & 57.4\%   & 67.7\%   & 69.5\%   \\
   Sent.\ with up to 2 crossings   & 60.7\%   & 64.0\%   & 72.4\%   & 80.9\%   & 81.3\%   \\
   Sent.\ with up to 3 crossings   & 73.9\%   & 76.9\%   & 84.2\%   & 87.9\%   & 90.8\%   \\
   Sent.\ with up to 4 crossings   & 80.9\%   & 85.7\%   & 90.8\%   & 91.9\%   & 94.1\%   \\
   Correct operations              & 78.3\%   & 81.9\%   & 85.8\%   & 88.5\%   & 90.7\%   \\
   Sent.\ with correct OpSequence  &  2.2\%   &  1.5\%   &  5.1\%   & 10.3\%   & 14.3\%   \\
 Sent.\ with correct Struct\&Label &  3.7\%   &  6.6\%   & 12.1\%   & 17.3\%   & 22.1\%   \\
   Sentences with endless loop     &    7     &  15      &    2     &    0     &    1     \\ \hline
\end{tabular}
\caption{Evaluation results without using relevant feature information from individual examples}
\label{fig-pex_no_ind_rel_feature_info}
\end{center}
\end{table}
Table~\ref{fig-pex_training_size} clearly shows a continuous positive trend for all criteria.
The accuracy for an individual parse action is 91.7\% for 256 training sentences. 
Even though
the number appears to be good at first glance, the fact that there are an average of 43.5
parse actions per sentence means, that, assuming independence of parse action errors, the
probability of a completely correct operations sequence for an average length sentence is 
as low as $.917^{43.5}$
= 2.3\%. The much higher percentages of test sentences with correct operation sequence or 
at least correct structure and labeling already show that this independence assumption 
fortunately does not hold.
An individual analysis of parsing errors shows
that many mistakes are due to encountering constructions that just have not been
seen before at all, typically causing several erroneous parse decisions in a row.
This observation also supports our hope that with more training sentences,
the accuracy for unseen sentences will still rise significantly.

\subsection{Contribution of Individual Feature Selection Preferences}

Recall that the supervisor can optionally mark a parse example with one or more features 
indicating that these are relevant for the specific example, as already described in
subsection~\ref{subsec-feature_selection_preferences}.
Table~\ref{fig-pex_no_ind_rel_feature_info} shows the relatively modest contribution of
these individual feature selection preferences.
Without these individual feature preferences, the parse action accuracy drops by about 1\%,
regardless of training size.

\subsection{Contribution of the Subcategorization Table}

\noindent
\begin{table}[htbp]
\begin{center}
\begin{tabular}{|l|c|c|c|c|c|} \hline \hline
   Number of training sentences    & 16       & 32       & 64       & 128      & 256      \\ \hline
   Precision                       & 85.1\%   & 86.4\%   & 88.6\%   & 90.6\%   & 92.4\%   \\
   Recall                          & 83.1\%   & 85.0\%   & 88.2\%   & 90.5\%   & 92.4\%   \\
   Labeled precision              & 78.1\%   & 79.8\%   & 83.5\%   & 86.9\%   & 89.8\%   \\
   Labeled recall                 & 75.2\%   & 77.6\%   & 82.1\%   & 85.8\%   & 89.3\%   \\
   Tagging accuracy                & 96.4\%   & 96.5\%   & 97.1\%   & 97.9\%   & 98.3\%   \\
   Crossings per sentence          & 2.6      & 2.2      & 1.8      & 1.3      & 1.1      \\
   Sent.\ with 0 crossings         & 28.7\%   & 32.4\%   & 37.9\%   & 48.5\%   & 57.0\%   \\
   Sent.\ with up to 1 crossing    & 42.7\%   & 49.3\%   & 57.4\%   & 67.7\%   & 73.2\%   \\
   Sent.\ with up to 2 crossings   & 60.3\%   & 64.7\%   & 70.2\%   & 80.9\%   & 84.6\%   \\
   Sent.\ with up to 3 crossings   & 75.4\%   & 77.9\%   & 81.3\%   & 87.9\%   & 91.5\%   \\
   Sent.\ with up to 4 crossings   & 83.1\%   & 87.1\%   & 89.0\%   & 94.1\%   & 95.2\%   \\
   Correct operations              & 78.8\%   & 82.0\%   & 85.6\%   & 87.6\%   & 90.1\%   \\
   Sent.\ with correct OpSequence  &  1.5\%   &  3.3\%   &  4.0\%   &  7.7\%   & 10.3\%   \\
 Sent.\ with correct Struct\&Label &  5.5\%   &  7.4\%   & 10.7\%   & 20.2\%   & 27.9\%   \\
   Sentences with endless loop     &   13     &   8      &    0     &    1     &    2     \\ \hline
\end{tabular}
\end{center}
\caption{Evaluation results without subcategorization features}
\label{fig-pex_wo_subcat}
\end{table}
Table~\ref{fig-pex_wo_subcat} shows the contribution of the subcategorization table.
The 10 features that access the subcategorization module have been discarded in this test series.
While the figures for precision, recall, tagging accuracy and crossings hardly degrade at all,
we observe a significant deterioration in correct operations, however not in structure and label
correctness of the final parse tree. This means that so far the primary benefit from
the subcategorization table lies in the assignment of proper roles to phrase components,
which on one side is not a surprise, because the subcategorization table quite explicitly 
lists which semantic roles are to be assigned to various syntactic components. On the other side this
means that the contribution towards the resolution of structural ambiguity does not depend as
much on the subcategorization table as might have been expected.

\subsection{Contribution of Rich Context}

The following tables (\ref{fig-pex100}, \ref{fig-pex50}, \ref{fig-pex25}, \ref{fig-pex12}, 
\ref{fig-pex6}) show the impact of reducing the feature set to a set of {\it n} core features. 
We always chose those {\it n} features that appeared to be the most important {\it n} features,
relying on older smaller versions of feature lists and general experience acquired during the 
development process.
When 25 or fewer features are used, all of them are of a syntactic nature.
\noindent
\begin{table}[htbp]
\begin{center}
\begin{tabular}{|l|c|c|c|c|c|} \hline \hline
   Number of training sentences    & 16       & 32       & 64       & 128      & 256      \\ \hline
   Precision                       & 85.4\%   & 86.4\%   & 88.5\%   & 90.1\%   & 91.7\%   \\
   Recall                          & 83.1\%   & 85.6\%   & 88.5\%   & 89.7\%   & 91.7\%   \\
   Labeled precision              & 77.7\%   & 79.8\%   & 83.4\%   & 85.8\%   & 88.6\%   \\
   Labeled recall                 & 75.1\%   & 78.5\%   & 82.3\%   & 85.1\%   & 88.1\%   \\
   Tagging accuracy                & 96.5\%   & 96.5\%   & 96.9\%   & 97.4\%   & 98.2\%   \\
   Crossings per sentence          & 2.5      & 2.3      & 1.8      & 1.4      & 1.1      \\
   Sent.\ with 0 crossings         & 26.8\%   & 32.4\%   & 38.6\%   & 47.8\%   & 54.0\%   \\
   Sent.\ with up to 1 crossing    & 43.0\%   & 51.1\%   & 57.0\%   & 68.4\%   & 72.1\%   \\
   Sent.\ with up to 2 crossings   & 62.5\%   & 63.2\%   & 71.7\%   & 80.9\%   & 84.2\%   \\
   Sent.\ with up to 3 crossings   & 73.5\%   & 75.0\%   & 81.6\%   & 87.5\%   & 92.3\%   \\
   Sent.\ with up to 4 crossings   & 84.2\%   & 84.2\%   & 89.3\%   & 91.2\%   & 94.5\%   \\
   Correct operations              & 78.9\%   & 82.8\%   & 86.6\%   & 88.7\%   & 90.7\%   \\
   Sent.\ with correct OpSequence  &  1.8\%   &  4.8\%   &  5.1\%   &  8.1\%   & 13.6\%   \\
 Sent.\ with correct Struct\&Label &  5.9\%   &  8.5\%   &  9.6\%   & 15.4\%   & 23.5\%   \\
   Sentences with endless loop     &   11     &   1      &    2     &    2     &    2     \\ \hline
\end{tabular}
\end{center}
\caption{Evaluation results using only 100 features}
\label{fig-pex100}
\end{table}
\noindent
\begin{table}[htbp]
\begin{center}
\begin{tabular}{|l|c|c|c|c|c|} \hline \hline
   Number of training sentences    & 16       & 32       & 64       & 128      & 256      \\ \hline
   Precision                       & 84.6\%   & 85.7\%   & 87.9\%   & 89.6\%   & 90.8\%   \\
   Recall                          & 82.6\%   & 85.2\%   & 88.0\%   & 89.4\%   & 90.8\%   \\
   Labeled precision              & 76.9\%   & 79.0\%   & 82.5\%   & 84.7\%   & 87.2\%   \\
   Labeled recall                 & 75.1\%   & 78.0\%   & 82.2\%   & 84.2\%   & 86.9\%   \\
   Tagging accuracy                & 96.4\%   & 96.4\%   & 97.1\%   & 97.3\%   & 98.1\%   \\
   Crossings per sentence          & 2.7      & 2.3      & 1.9      & 1.5      & 1.3      \\
   Sent.\ with 0 crossings         & 25.4\%   & 30.9\%   & 36.0\%   & 46.0\%   & 50.4\%   \\
   Sent.\ with up to 1 crossing    & 42.3\%   & 48.2\%   & 56.6\%   & 62.1\%   & 70.6\%   \\
   Sent.\ with up to 2 crossings   & 59.6\%   & 65.8\%   & 69.5\%   & 78.7\%   & 80.5\%   \\
   Sent.\ with up to 3 crossings   & 72.8\%   & 76.5\%   & 80.9\%   & 84.9\%   & 88.6\%   \\
   Sent.\ with up to 4 crossings   & 82.7\%   & 83.5\%   & 88.6\%   & 91.5\%   & 93.8\%   \\
   Correct operations              & 78.5\%   & 81.7\%   & 85.3\%   & 87.0\%   & 88.9\%   \\
   Sent.\ with correct OpSequence  &  2.2\%   &  3.7\%   &  4.4\%   &  7.0\%   &  8.8\%   \\
 Sent.\ with correct Struct\&Label &  5.5\%   &  7.4\%   &  8.5\%   & 12.9\%   & 15.1\%   \\
   Sentences with endless loop     &    4     &   0      &    0     &    2     &    2     \\ \hline
\end{tabular}
\end{center}
\caption{Evaluation results using only 50 features}
\label{fig-pex50}
\end{table}
As table~\ref{fig-pex100} show, the reduction of features to 100 has no significant impact
on any criterion when training with under 100 sentences. 
When training on 256 sentences, precision and recall drop 1\% and compound test characteristics,
i.e.\ the percentage of sentences with a totally correct operation sequence, or at least the
correct final structure and labeling, decrease by about 3\%.

Cutting the number of features in half again, table~\ref{fig-pex50}, basically continues this
trend. Compound test characteristics start to deteriorate for medium numbers of training
sentences and, for training on 256 sentences, are only little more than half of when using all 
features.

When using 25 (or fewer features), table~\ref{fig-pex25}, all remaining features are syntactic.
Crossing brackets have increased significantly and now only little over 1\% of the sentences 
achieve a perfect operation sequence.
\noindent
\begin{table}[htbp]
\begin{center}
\begin{tabular}{|l|c|c|c|c|c|} \hline \hline
   Number of training sentences    & 16       & 32       & 64       & 128      & 256      \\ \hline
   Precision                       & 84.7\%   & 85.7\%   & 86.8\%   & 87.4\%   & 88.7\%   \\
   Recall                          & 82.3\%   & 84.3\%   & 86.4\%   & 87.4\%   & 88.7\%   \\
   Labeled precision              & 76.9\%   & 78.5\%   & 77.0\%   & 82.4\%   & 86.7\%   \\
   Labeled recall                 & 74.5\%   & 76.9\%   & 79.8\%   & 81.8\%   & 84.1\%   \\
   Tagging accuracy                & 96.3\%   & 96.4\%   & 97.1\%   & 97.5\%   & 97.9\%   \\
   Crossings per sentence          & 2.7      & 2.4      & 2.1      & 2.0      & 1.7      \\
   Sent.\ with 0 crossings         & 24.3\%   & 29.4\%   & 33.1\%   & 36.0\%   & 43.4\%   \\
   Sent.\ with up to 1 crossing    & 43.4\%   & 48.9\%   & 52.6\%   & 52.9\%   & 59.6\%   \\
   Sent.\ with up to 2 crossings   & 59.9\%   & 61.8\%   & 65.4\%   & 70.6\%   & 73.9\%   \\
   Sent.\ with up to 3 crossings   & 72.4\%   & 73.9\%   & 77.2\%   & 80.9\%   & 84.9\%   \\
   Sent.\ with up to 4 crossings   & 82.4\%   & 83.5\%   & 88.2\%   & 88.2\%   & 89.7\%   \\
   Correct operations              & 74.8\%   & 76.8\%   & 79.2\%   & 81.0\%   & 81.9\%   \\
   Sent.\ with correct OpSequence  &  0.0\%   &  1.1\%   &  1.1\%   &  1.1\%   &  1.5\%   \\
 Sent.\ with correct Struct\&Label &  3.7\%   &  5.5\%   &  6.3\%   &  7.0\%   &  9.2\%   \\
   Sentences with endless loop     &    5     &  12      &    8     &    3     &    2     \\ \hline
\end{tabular}
\end{center}
\caption{Evaluation results using only 25 features, all of them syntactic}
\label{fig-pex25}
\end{table}
\noindent
\begin{table}[htbp]
\begin{center}
\begin{tabular}{|l|c|c|c|c|c|} \hline \hline
   Number of training sentences    & 16       & 32       & 64       & 128      & 256      \\ \hline
   Precision                       & 84.0\%   & 82.5\%   & 85.4\%   & 85.2\%   & 86.4\%   \\
   Recall                          & 81.3\%   & 83.6\%   & 85.5\%   & 86.3\%   & 87.4\%   \\
   Labeled precision              & 66.6\%   & 17.9\%   & 36.2\%   & 26.0\%   & 27.6\%   \\
   Labeled recall                 & 72.7\%   & 75.2\%   & 78.3\%   & 79.9\%   & 82.0\%   \\
   Tagging accuracy                & 96.4\%   & 96.5\%   & 97.0\%   & 97.5\%   & 97.9\%   \\
   Crossings per sentence          & 2.7      & 2.5      & 2.1      & 2.0      & 1.7      \\
   Sent.\ with 0 crossings         & 25.7\%   & 26.8\%   & 32.7\%   & 33.8\%   & 43.0\%   \\
   Sent.\ with up to 1 crossing    & 43.0\%   & 44.1\%   & 52.9\%   & 52.9\%   & 58.8\%   \\
   Sent.\ with up to 2 crossings   & 59.9\%   & 58.8\%   & 64.3\%   & 68.4\%   & 74.6\%   \\
   Sent.\ with up to 3 crossings   & 69.5\%   & 71.0\%   & 77.2\%   & 80.5\%   & 84.6\%   \\
   Sent.\ with up to 4 crossings   & 79.4\%   & 82.7\%   & 86.8\%   & 87.5\%   & 89.3\%   \\
   Correct operations              & 74.2\%   & 76.3\%   & 78.5\%   & 80.1\%   & 80.9\%   \\
   Sent.\ with correct OpSequence  &  1.1\%   &  1.1\%   &  1.1\%   &  1.5\%   &  1.5\%   \\
 Sent.\ with correct Struct\&Label &  3.7\%   &  4.0\%   &  5.9\%   &  7.0\%   &  8.5\%   \\
   Sentences with endless loop     &   16     &   17     &   20     &   17     &   11     \\ \hline
\end{tabular}
\end{center}
\caption{Evaluation results using only 12 features, all of them syntactic}
\label{fig-pex12}
\end{table}
\noindent
\begin{table}[htbp]
\begin{center}
\begin{tabular}{|l|c|c|c|c|c|} \hline \hline
   Number of training sentences    & 16       & 32       & 64       & 128      & 256      \\ \hline
   Precision                       & 83.8\%   & 83.6\%   & 85.1\%   & 86.1\%   & 88.0\%   \\
   Recall                          & 81.5\%   & 83.5\%   & 85.1\%   & 85.9\%   & 87.3\%   \\
   Labeled precision              & 59.8\%   & 24.6\%   & 46.9\%   & 69.9\%   & 79.8\%   \\
   Labeled recall                 & 73.3\%   & 75.5\%   & 77.9\%   & 79.5\%   & 81.5\%   \\
   Tagging accuracy                & 96.6\%   & 96.7\%   & 96.8\%   & 97.2\%   & 97.6\%   \\
   Crossings per sentence          & 2.7      & 2.5      & 2.2      & 2.1      & 1.8      \\
   Sent.\ with 0 crossings         & 24.3\%   & 26.1\%   & 30.2\%   & 32.4\%   & 39.0\%   \\
   Sent.\ with up to 1 crossing    & 43.0\%   & 44.9\%   & 48.5\%   & 51.8\%   & 57.4\%   \\
   Sent.\ with up to 2 crossings   & 57.7\%   & 60.3\%   & 63.2\%   & 65.8\%   & 72.1\%   \\
   Sent.\ with up to 3 crossings   & 72.4\%   & 70.2\%   & 77.9\%   & 79.4\%   & 82.7\%   \\
   Sent.\ with up to 4 crossings   & 80.9\%   & 83.1\%   & 86.8\%   & 86.0\%   & 89.0\%   \\
   Correct operations              & 73.7\%   & 75.6\%   & 78.0\%   & 79.3\%   & 80.6\%   \\
   Sent.\ with correct OpSequence  &  1.1\%   &  1.1\%   &  1.1\%   &  1.1\%   &  2.6\%   \\
 Sent.\ with correct Struct\&Label &  4.0\%   &  4.0\%   &  4.8\%   &  5.9\%   &  8.8\%   \\
   Sentences with endless loop     &   19     &   14     &   15     &   14     &    8     \\ \hline
\end{tabular}
\end{center}
\caption{Evaluation results using only 6 features, all of them syntactic}
\label{fig-pex6}
\end{table}

With only 12 features, table~\ref{fig-pex12}, we notice a dramatic increase of sentences whose
parsing has to be prematurely stopped because loops have been detected. This problem also
causes a multitude of bad nodes, as particularly manifested in low labeled precisions; such
bad nodes are often `hordes' of incorrect empty category tokens etc.

Cutting the number of features one final time, table~\ref{fig-pex6}, produces relatively little
further degradation. Let us now compare the results when using 6 features to those using the full
complement of 205. For small training sizes, e.g.\ 6, we see only a very modest drop in precision
(from 85.1\% to 83.8\%) and even a slight increase of labeled recall (from 75.0\% to 77.4\%),
no loss of tagging accuracy (still 96.6\%), a modestly higher number of crossings (from 2.5 per
sentence to 2.7) and even for compound test characteristics a relatively moderate drop (e.g.\
from 5.5\% to 4.0\% for the percentage of sentences with perfect structure and labeling after 
parsing). For the full complement of 256 training sentences on the other hand, we observe a 
roughly 5\% drop in labeled and unlabeled precision and recall, almost twice as many crossings,
and a sharp decrease in compound test characteristics (from 16.7\% to 2.6\% totally correct
parse sequences and from 26.8\% to 8.8\% for parse sequences leading to the proper structure 
and labeling).

So, while the loss
of a few specialized features will not cause a major degradation, the relatively high number
of features used in our system finds a clear justification when evaluating compound
test characteristics.
When comparing the decrease of parsing accuracy for the various training sizes, we observe
that the accuracy loss is much more pronounced for higher number of training sentences.
As one might have expected, this means that larger training corpora can exploit our rich 
feature set better. When the training size is increased beyond 256, we can therefore
rightfully expect the advantage of a rich context to increase even further.

\subsection{Contribution of Decision Structure Type}
\label{sec-contribution-of-decision-structure-type}
\vspace{-1cm}
\noindent
\begin{table}[htbp]
\begin{center}
\begin{tabular}{|l|c|c|c|c|} \hline \hline
   Type of decision structure      & simple  & simple & hier.  & hybrid       \\
                                  & decision & decision & decision & decision \\
                                   & tree    & list   & list   & structure    \\ \hline
   Precision                       & 87.6\%  & 87.8\% & 91.0\% & 92.7\%       \\
   Recall                          & 89.7\%  & 89.9\% & 88.2\% & 92.8\%       \\
   Labeled precision              & 38.5\%  & 28.6\% & 87.4\% & 89.8\%       \\
   Labeled recall                 & 85.6\%  & 86.1\% & 84.7\% & 89.6\%       \\
   Tagging accuracy                & 97.9\%  & 97.9\% & 96.0\% & 98.4\%       \\
   Crossings per sentence          & 1.3     & 1.2    & 1.3    & 1.0          \\
   Sent.\ with 0 crossings         & 51.5\%  & 55.2\% & 52.9\% & 56.3\%       \\
   Sent.\ with up to 1 crossing    & 65.8\%  & 72.8\% & 71.0\% & 73.5\%       \\
   Sent.\ with up to 2 crossings   & 81.6\%  & 82.7\% & 82.7\% & 84.9\%       \\
   Sent.\ with up to 3 crossings   & 90.1\%  & 89.0\% & 89.0\% & 93.0\%       \\
   Sent.\ with up to 4 crossings   & 93.4\%  & 93.4\% & 93.4\% & 94.9\%       \\
   Correct operations              & 90.2\%  & 86.5\% & 90.3\% & 91.7\%       \\
   Sent.\ with correct OpSequence  & 13.6\%  & 12.9\% & 11.8\% & 16.5\%       \\
 Sent.\ with correct Struct\&Label & 21.7\%  & 22.4\% & 22.8\% & 26.8\%       \\
   Sentences with endless loop     & 32      & 26     & 23     &    1         \\ \hline
\end{tabular}
\end{center}
\caption{Evaluation results comparing different types of decision structures}
\label{fig-pex_dstr_types}
\end{table}

Table~\ref{fig-pex_dstr_types} compares four different machine learning variants: 
a plain decision tree,
a plain decision list,
a hierarchical decision list,
and finally a hybrid decision structure,
namely a decision list of hierarchical decision trees, as sketched in
figure~\ref{fig-hier_dec_str_list}. The results show that extensions to the basic
decision tree model can significantly improve learning results.
The hybrid decision structure is superior to the other three models with respect to each and
every criterion. While the three simpler decision structure have parsing loop problems
for around 10\% of the 272 sentences, the hybrid decision structure has only one such problem
sentence. For compound test characteristics, sentences with the perfect operation
sequence or at least the correct final structure and labeling, the decision
structure list scores 3\% and 4\% respectively higher than any other decision structure.

         %\chapter{Parsing Experiments} \label{ch-p_exp}
\chapter{Transfer}          
\label{ch-transfer}

The transfer module maps the final source language parse tree to a corresponding
target language tree. Before the actual transfer, the parse tree is normalized
and some language specific morphological and syntactic information is discarded.
The normalization includes the addition of back-pointers to parent nodes
and the backpropagation of forms. 
To understand what this backpropagation does, consider for example the sentence
{\it ``The deer are hungry.''}, where the number of the noun phrase and the person 
and number of the verb are locally ambiguous. After constraint unification
during parsing, form information such as number and person, unambiguous at the 
sentence level, is now propagated back to the individual components of the sentence, 
so that {\it the deer} is marked as plural and {\it are} as third person plural.

The other part of pre-transfer processing is the deletion of the inner structure 
of a compound verb, punctuation and other syntactic dummy components like for example
the empty or `prop' {\it it}\footnote{as e.g.\ in {\it ``It rained.'', 
``What time is it?''}} in English.
The inner structure of verbs, e.g. [[AUX: [AUX: had] [PRED: been]] [PRED: dissatisfied]],
is quite language specific and is therefore eliminated, leaving just the verb
concept (I-EV-DISSATISFY) along with form information (past perfect, passive, etc.).
Punctuation is also discarded. The main purpose of punctuation is to help structure
sentences, something that can and should be exploited during parsing, but once the
parse tree has been formed and the sentence has been assigned a structure, punctuation
no longer carries any additional information, and since its usage is also language 
specific, it is not transferred to the target parse tree.

So the main information transferred to a target language tree is the structure of the
sentence, its concepts and forms (such as tense), and roles of components in their
respective phrases.
The task of determining the correct word order, finding the specific surface words
for the various concepts and punctuation is left to the generation module.

Bilingual dictionaries are fully symmetric and can be used for translations in either direction.

\section {Simple Transfer}
The structure of the parse tree and the roles are basically preserved, leaving the
transfer of concepts as the core task. This concept transfer can happen for both
individual concepts as well as concept clusters, representing phrases such as
{\it ``to reach [an] agreement''} or {\it ``to comment on SOMETHING\_1''}.
This is done using a {\it bilingual dictionary}, which was already described in
section~\ref{sec-dictionary} in the chapter on background knowledge.
Recall that the {\it surface} dictionary, which has a very user friendly and intuitive
format similar to traditional (paper) dictionaries, is compiled into an 
{\it internal} dictionary, which links concepts (for individual words) or parse trees 
(for phrases) between languages.
In the simplest case, an individual source language concept will correspond to exactly 
one target language concept, in which case that concept is selected.

One could ask, why in such 
a case the words from the respective languages aren't just linked to the same KB concept
in the first place. The answer is that we don't want to certify that a specific word 
pair has a perfect conceptual match. While two words might appear to have a one-to-one
correspondence, there might be an obscure case or context where this does not hold
after all. We want to keep the information contained in the lexical entries micromodular
and be able to deal with any unforeseen mismatches by just adding another entry to the
bilingual dictionary.

\section {Ambiguous Transfer}
\label{sec-ambiguous-transfer}

The much more interesting and fairly frequently occurring case is
when words and phrases do not correspond one-to-one between languages.
Consider for example the two dictionary entries for ``know'':

\begin{center}
\begin{tabular}{|ll|} \hline
``know''                                & S-VERB  \\  ``kennen'' & \\ \hline
``know''                                & S-VERB  \\  ``wissen'' & \\ \hline
\end{tabular}\\
\end{center}

Depending on whether the object of ``know'' is 
a person, a place, etc., or a fact, ``know'' maps to ``kennen'' or ``wissen''
in German (and ``conocer''/``saber'' in Spanish; ``conna\^{\i}tre''/``savoir'' in French).
Lexical ambiguity can be divided into three levels: part of speech ambiguity,
and heterogeneous and homogeneous ambiguity within the same part of speech.
The transfer module assumes that the word has already been disambiguated with
respect to the part of speech. 
A disambiguation of heterogeneous (homonymous) meanings
(e.g.\ pen (for writing) vs.\ pen (to keep animals)) is currently not done during
parsing, because it hasn't found to be necessary for our WSJ sentences. When such
disambiguation should turn out to become necessary, e.g.\ to resolve structural 
ambiguity or case role assignment, it could be added, but so far in our research 
it has been sufficient to delay decisions on heterogeneous meanings until transfer.
The partitioning of fairly homogeneous (polysemous) senses, like ``know'', is very
language dependent though. It is hard, if not impossible, to do a proper partitioning
without having a particular target language in mind. This type of disambiguation 
is clearly more appropriate to be resolved in the transfer module. So, ambiguous
transfer currently has to resolve both heterogeneous and homogeneous ambiguity within 
the same part of speech. This however does not present any additional difficulty, 
because from the perspective of transfer, there is no substantive difference 
between heterogeneous and homogeneous ambiguity.

For choosing the appropriate target concept, we reuse the methods developed for parsing 
with only minor adaptations. Again, we provide examples and features, from which we compute
complete examples that include a full feature vector plus classification, and feed these
into our machine learning module. While there is one large example collection and one big
feature set for parsing, there are separate sets of examples and features for each concept
to be disambiguated. Consider the following three {\it transfer entries}:

\begin{verbatim}
(I-EV-KNOW
   ((SYNTP OF S-SUB-CLAUSE OF THEME OF PARENT AT M-BOOLEAN)
    (CLASSP OF I-EN-TANGIBLE-OBJECT OF THEME OF PARENT AT M-BOOLEAN))
   ((I-GV-KENNEN "I know the old man." (SHORT 5 1) 10)
    (I-GV-WISSEN "I know that the old man bought a car." 
           (SHORT 7 1) 20)))

(I-EP-BY
   ((SYNT OF PARENT OF PARENT AT S-SYNT-ELEM) ;;; S-CLAUSE, S-NP
    (CLASSP OF I-EN-AGENT OF PRED-COMPL OF PARENT AT M-BOOLEAN)
    (CLASSP OF C-AT-TIME OF PARENT AT M-BOOLEAN))
   ((I-GP-VON "It will be sold by a shareholder." nil 10)
    (I-GP-BIS "The transaction will be finished by May 10." nil 20)
    (I-GP-DURCH "Sales will be hurt by the losses." nil 30)
    (I-GP-GEN-CASE-MARKER 
           "The estimates by the bureau were published in May." 
           nil 40)))

(I-EN-FUTURE
   ((NUMBER AT F-NUMBER))
   ((I-GN-ZUKUNFT "future" nil 10)
    (I-GN-TERMINGESCHAEFT "He bought futures." nil 20)))
\end{verbatim}

Each transfer entry consists of the source concept, a list of features, and a set of examples.
The transfer examples consist of a target concept, an example text, an optional example reference
and an example identifier. The example text is often a sentence, but can basically be anything
that provides enough context for the features. The optional reference, e.g.\ (SHORT 5 1),
contains both a corpus reference, here sentence {\it 5} of corpus {\it SHORT}, and an occurrence identifier,
here {\it 1}.

The transfer examples are processed by first parsing the example text. This is done using the same
parser as for ordinary text. If a corpus reference is included and there is a log entry, i.e.\ a
supervisor-provided parse action sequence, that parse action sequence is used, because it has already
been certified as correct. Otherwise the parse is `free', but since the examples are typically
concise, the error potential is very small in the first place and only errors that would change
any of the transfer entry features would be problematic. If example texts must approach or
surpass the limits of the parser coverage, it is advised to store the text in some corpus and
provide a log entry for it. The occurrence identifier, which has a default of 1, selects which
of the potentially multiple occurrences of the source concept in the example text
should serve as a reference point.

After the parse tree has been computed for the example text and the reference point is identified,
the system computes values for all features listed in the transfer entry. The feature language
is the same as for parsing. Note again though that the reference point here is not the current 
parse state with parse stack and input list, but a specific parse entry. Therefore the access
paths of the transfer entry features (e.g.\ {\it THEME OF PARENT}) typically do not contain an
integer, as the parsing features (e.g.\ {\it THEME OF -1}) typically do.

Once all examples in a transfer entry have been expanded to include their full feature-value vector, 
standard decision tree learning is applied. The number of examples and features tends to be fairly
small per source concept, so compared to parsing, which uses a single huge decision structure,
we use many (one for each ambiguous concept) small decision trees for ambiguous transfer.

Since, for each source concept, there are relatively few target concepts (classifications), examples, 
and features compared to parsing, a hand coded approach would be more feasible than in parsing.
For several reasons we nevertheless believe that machine learning is better suited. Features would
have to be provided for both learning and hand-coding, for learning though only once, whereas
in hand-coding a feature might have to appear several times. While examples are explicitly
required for learning they are practically also required for hand-coding, namely for testing and
documentation purposes; the examples needed, typically short, unannotated sentences, can normally 
be provided fairly easily.
With learning, the decision structures are built automatically; coverage of the examples is guaranteed
unless there is an unresolvable example conflict, which the system signals automatically. ``Testing
the test sentences'' is in a sense done implicitly. Hand-coding on the other side would require
explicit rule construction and, in practise, repeated testing.

\section {Complex Transfer}
\begin{figure}[htb]
\par
\epsfxsize=12.0cm
\centerline{\epsfbox{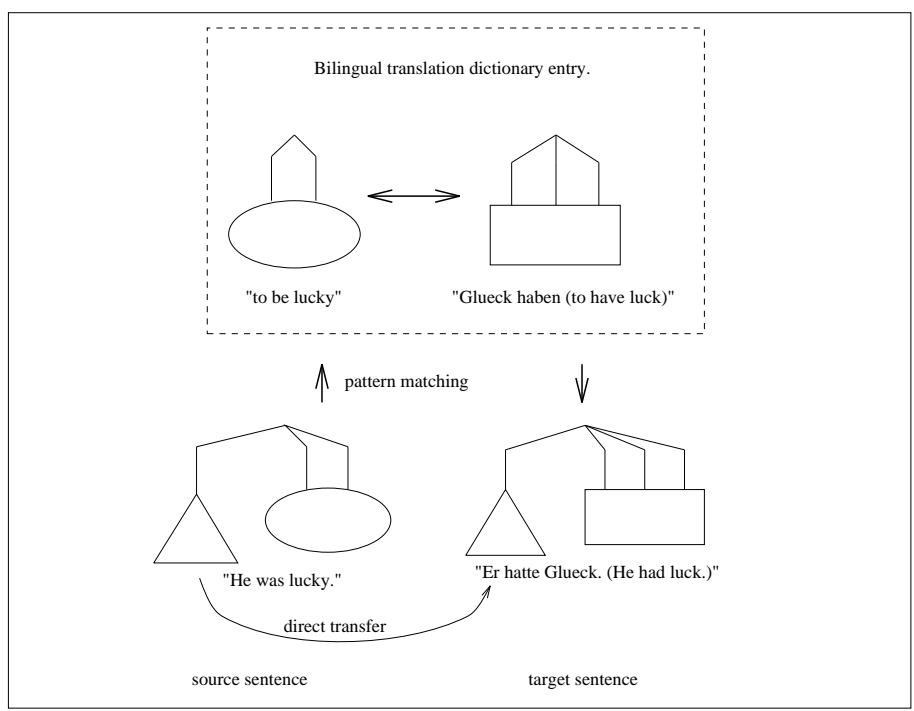}}
\caption{Complex transfer with pattern matching}
\label{csg_trans_match1}
\end{figure}

``Structural mismatches'' present another major issue. Often, the structures of the
source and target language parse trees don't match. To handle such cases, the bilingual
dictionary contains potentially complex phrases, represented as parse trees in the {\it internal}
dictionary. 
During transfer, a pattern matching module tries to match source sentences subtrees
to dictionary source trees and, if successful, maps the
source sentence tree to a target sentence tree using the dictionary target tree
linked to the successfully matched dictionary source tree as illustrated in 
figure~\ref{csg_trans_match1}.

The parse tree pair in the dashed box is an entry in the {\it internal} dictionary. To translate
the English sentence {\it ``He was lucky.''}, we first parse it to get a parse tree as depicted in the
lower left area of figure~\ref{csg_trans_match1}. The source sentence tree is transferred recursively.
At each level we check whether the current subtree matches a pattern in the dictionary. For a valid
match, all components of the dictionary source tree have to be covered, but the inverse does not
need to hold. In our example for example the source sentence parse tree contains an additional subject
component. Dictionary entries have been stripped of many of their forms, e.g.\ their infinitive 
tense, since such forms were meant to represent the word generically. As a consequence, the past 
tense form of {\it ``He was lucky.''} can be considered an additional component as well.

In case of a match, the target sentence tree is built based on the dictionary target tree. Additional
components from the source sentence tree (here {\it He} and {\it past-tense}) are then transferred 
separately and the resulting target components are finally properly attached to the target sentence 
tree.

\begin{center}
\begin{tabular}{|ll|} \hline
``to be lucky''              & S-VP    \\  ``Gl\"{u}ck haben'' & \\ \hline
``downtown PLACENAME\_1''    & S-ADV   \\  ``im Stadtzentrum von ORTSNAME\_1''\footnote{This 
phrase literally means {\it ``in\_the towncenter of PLACENAME\_1''}} & S-PP \\ \hline
``it takes SOMEBODY\_3 SOMETHING\_1     & S-CLAUSE \\
                                    TO\_DO\_SOMETHING\_2'' & \\
     ``JEMAND\_3 braucht ETWAS\_ACC\_1,                           & \\
                                        um ETWAS\_ZU\_MACHEN\_2`` & \\ \hline
``to reach SOME\_1 agreement'' & S-VP \\   ``zu EINER\_1 \"{U}bereinkunft kommen'' & \\ \hline
``interest rate'' & S-NOUN \\ ``Zinssatz'' & \\ \hline
\end{tabular}\\
\end{center}

\noindent The patterns can contain partially restricted variables, e.g.\ ``PLACENAME\_1'' in the 
selection above.\footnote{This phrase literally means {\it in\_the towncenter of PLACENAME\_1}}
Subtrees from the source sentence tree can match such variables if they fulfill the restrictions
that are the lexicon associates with these variables, typically reflected in the name of the variable.
In case of a match such subtrees are then transferred separately and the resulting target trees are
finally properly attached to the target sentence tree at the place marked by the target variable
with the same index.

The dummies of the form {\it SOME\_i} are special `collectors' which can be matched to a whole set 
of noun modifiers. This allows the tree for {\it ``to reach SOME\_1 agreement''} to match all of
the following:

\newpage
\begin{itemize}
   \item We reached an agreement.
   \item We didn't reach any agreement.
   \item No agreement was reached.
   \item We reached many new traffic agreements of great importance.
   \item Candidates Reached Agreement
\end{itemize}
A dictionary entry with {\it ``to reach an agreement''} would only match in the first example.

The search for possibly matching patterns is sped up through the use of the head concept as
an index and a quick pre-check, in which the cached set of all other obligatory leaf concepts 
in the dictionary source tree has to be covered by the source sentence tree.

If no complex match is found for a source sentence subtree, the various components of that subtree
are transferred separately in further recursion and put together using the same structure as the 
source sentence tree.

Even if there is no structural mismatch, as for example in the last entry of the last table,
where there is a complex noun consisting of two nouns in both English and German\footnote{{\it Zinssatz}
is a compound of {\it Zins} and {\it Satz}}, complex transfer can be used as a simple tool for
disambiguation, reducing the need for examples and features. In English, both {\it interest} and
{\it rate} are semantically ambiguous, as also reflected by different German translations, e.g.\
{\it Interesse, Zins(en), Anteil} for {\it interest} and {\it Rate, Satz, Kurs} for {\it rate}.
As a compound however, {\it interest rate} unambiguously maps to {\it Zinssatz}. Since complex
transfer precludes any individual and possibly ambiguous transfer at a lower level, no ambiguous
transfer examples and features are necessary for {\it interest} or {\it rate} 
when they occur in a joint compound.

\section{Added Concepts}

It can sometimes become necessary to add a concept to the target parse tree that is not based
on any specific concept or group of concepts in the source parse tree. Examples for such added 
concepts are articles, pronouns or adverbs. Japanese, which does not have articles, for example does
not mark its noun phrases as definite or indefinite. It also often omits pronouns that would be
obligatory in English. When performing parse tree transfer to a language that requires overt
components that don't have some overt corresponding
concept in the source parse tree, we have to break through the basic transfer paradigm and 
basically have to create ``something out of nothing''. Certain source parse tree patterns
have to trigger a context based decision making process that will add any appropriate concept
to the target parse tree. Even though this process depends strongly on the target language and
therefore might conceivably be included as an early step in generation, we deal with this issue
at the end of transfer, because the work that is actually necessary also strongly depends on the 
specific source language.

English and German both require overt pronouns and generally share the same notion of definiteness,
including the usage of definite and indefinite articles. The differences in article usage are
however strong enough that they manifested themselves several time in the WSJ translation development
corpus. The noun phrase in ``{\it Life can be difficult.}'' for example must be definite in German 
and French.

In our system, at the end of the core transfer, we traverse the parse tree and determine for each noun
phrase whether an additional definite article is necessary. While this is currently hard-coded,
we could possibly see machine learning as a useful tool when further refining this process.
%see function add-extra-articles-to-german-parse-entry-where-necessary in file order-german.lsp

The reverse process, deleting concepts, can be handled more easily by using the marker ``nil'' in
the bilingual dictionary entry.
According to the following example, the English word ``some'' can be translated by ``einige'' or
by nothing\footnote{As in ``Did you bring some records?'' $\rightarrow$ ``Hast Du Schallplatten
mitgebracht?''}.
Unlike all other dictionary entries, which can be used in both directions,
those dictionary entries that contain ``nil'' can only be used with ``nil'' as a target.\\

\begin{center}
\begin{tabular}{|ll|} \hline
``some''                                & S-ADJ   \\  ``einige'' & \\ \hline
``some''                                & S-ADJ   \\  nil & \\ \hline
\end{tabular}\\
\end{center}

In English to German transfer, only few minor concepts have to be added to cover the `nil to 
something' direction, but for other language pairs, in particular for less related languages, 
concepts might have to be added in more cases and also require more context.

              %\chapter{Transfer} \label{ch-transfer}
\chapter{Generation}
\label{ch-generation}

Following parsing and transfer, generation completes the process of machine translation
by ordering the components of phrases, adding appropriate punctuation, propagating 
morphologically relevant information, and finally generating the proper surface words
and phrases in the target language. 

In many natural language applications, generation can be quite formidable, because it 
involves tasks such as selecting the contents of a text and planning the discourse 
structure of the selected material. Such {\it strategic} generation, as it is often referred 
to, is however not necessary in machine translation, because the contents and overall 
discourse structure can be copied from the source text. This leaves the so
called {\it tactical} generation, consisting of the above mentioned 
subtasks.\footnote{When translating between very different languages, it might be
necessary to precede the generation steps described in this chapter by an additional 
{\it operational}
generation step to bridge general structural differences that can not be linked
to specific lexical items in the mismatching phrases and therefore can not be covered 
by the transfer module.}

Further reasons for the relative simplicity of tactical generation as we need it for our 
system are that
\begin{itemize}
  \item ordering basically depends on only fairly local syntactic properties of the phrase 
        components,
  \item the actual surface words basically depend on only fairly local morphological properties,
  \item a single good output is sufficient while the parser has to be able to cope with 
        all good and sometimes even not so good formulation alternatives of a sentence,
  \item difficult decision issues, like for example idiomatic expressions, have already been
        handled in transfer.
\end{itemize}
Since the steps necessary for generation are already linguistically well researched and
described in detail in grammar books, e.g.\ in \cite{lederer:dt69,engel:dt88} for German,
we don't use any learning for generation, but rather follow standard phrase constructions
from literature.

The input to generation is a transfer tree, an integrated phrase-structure and case-frame tree,
very similar to a parse tree, but typically not ordered, without punctuation,
with less morphological information and no surface forms. The generation module then manipulates the
transfer tree by reordering subtrees and propagating morphological information. It gradually
adds surface forms, the target character string associated with each subtree. The tree that is
finally formed after these manipulations then contains the translation result at the surface
form slot of the top node.
The following sections now describe the ordering and the morphological propagation and generation
in more detail.

\section{Ordering}

For each type of phrase of a given language to be ordered, the system uses a specific and
deterministic scheme. For German sentences or verbal clauses, the most complex case in that 
language, the system for example first determines whether the clause is finite, passive, 
a relative phrase, a main clause, and whether it needs a ``prop'' {\it es}\footnote{as in 
``{\it Es} regnete.'', the English equivalent of ``{\it It} rained.''} as a subject,
needs to be enclosed by commas or needs to have the conjugated part of the verb split off.

Based on this information, it arranges the sentence in a specific order:
\begin{enumerate}
\item coordinate conjunction
\item subordinate conjunctions
\item topic
\item the conjugated part of the verb (if the verb has to be split)
\item interrogative pronoun
\item pronominal or definite subject
\item pronominal direct object
\item pronominal indirect object
\item (indefinite) subject
\item definite indirect object
\item definite direct object
\item pronominal genitive complement
\item assessorial adjuncts
\item situational adjuncts
\item negational adjunct
\item other quantifiers
\item (indefinite) indirect object
\item (indefinite) direct object
\item (definite or indefinite) genitive complement
\item misc.\ other modifiers
\item other complements
\item adverbial complement
\item adjectival complement
\item nominal complement*
\item predicate
\item severed (``heavy'') relative clause
\item infinitival complement
\item particle phrase clauses*
\item other subclauses*
\item plus possibly enclosing commas\footnote{If a sentence
element fits more than one of these categories, it is placed at the first applicable position, 
unless an applicable category is marked by star (*), in which case it is placed at the last such 
position.}
\end{enumerate}
Every specific clause contains of course only some of these components; so the preceding list only
prescribes in which order the existing components are to be arranged with respect to each other.

The structure of the generated sentence
thus follows a syntactical pattern that can be characterized as `canonical', `safe' and `mainstream',
but does not necessarily reflect all the variance that is acceptable.
Similar schemes have be written for noun phrases, prepositional phrases, particle phrases, and
complex nouns.

\begin{figure}[htbp]
\epsfxsize=16.4cm
\centerline{\epsfbox{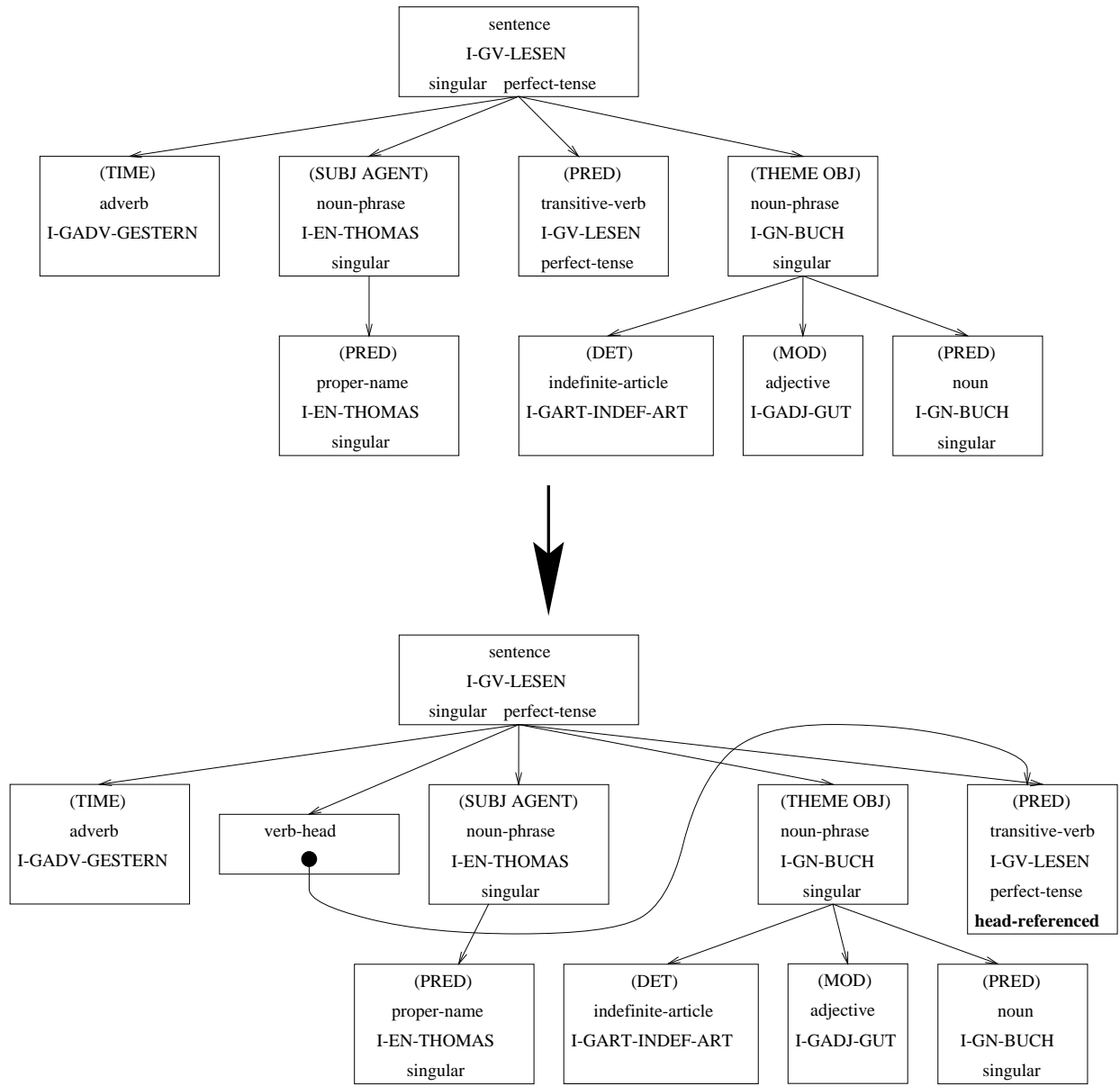}}
\caption{Ordering the target tree}
\label{fig-ordering}
\end{figure}

Figure~\ref{fig-ordering} illustrates this ordering process. The target tree output from the transfer
module still reflects the word order of the source sentence {\it ``Yesterday, Thomas read 
a good book.''}. As usual in German main clauses, the verb needs to split into the conjugated part
and the remainder. This is done by creating a special verb-head unit that points to the full verb and
symbolizes the conjugated part of the verb it points to; the full verb entry is marked as
{\it head-referenced} to indicate that it is being pointed to by such a special verb-head unit.

(Severed (``heavy'') relative clauses that have to separated from the noun phrases they belong to 
are treated the same way.)
%\clearpage
%\noindent

The components in the bottom tree of figure~\ref{fig-ordering} follow the order given in the 
above scheme: topic (time can qualify as such),
the conjugated part of the verb, definite subject, (indefinite) direct object, and predicate.
After the tree is ordered at the sentence level, the ordering recursively proceeds to the individual
sentence components.

\section{Morphological Propagation and Generation}

During morphological propagation, the carriers of gender, number, case, tense etc.\ (nouns,
prepositions, verbs etc.) propagate their forms to the constituents that they morphologically 
control.

\begin{figure}[htbp]
\epsfxsize=10cm
\centerline{\epsfbox{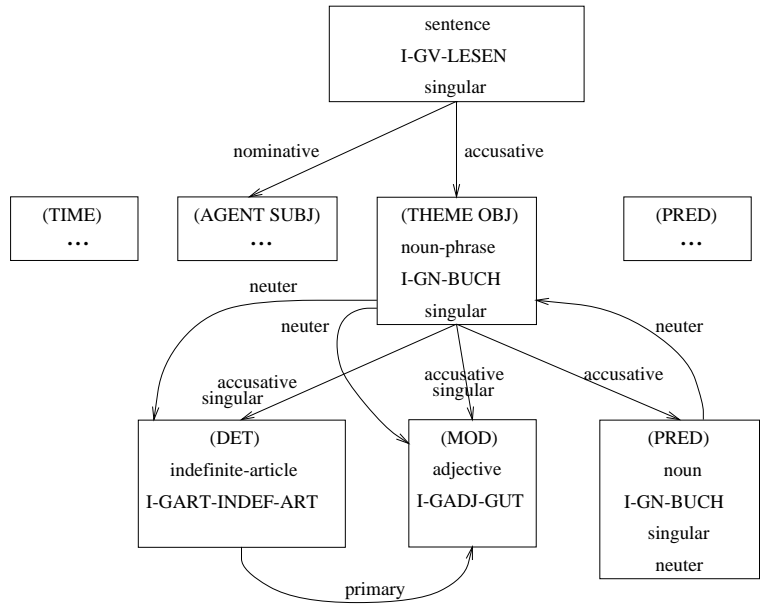}}
\caption{Morphological propagation}
\label{fig-morph_prop}
\end{figure}

In figure~\ref{fig-morph_prop}, we show how this is done for a portion of the graph from
figure~\ref{fig-ordering}. In order to form surface words, the tree nodes need to contain
complete form information. In our example from German, the adjective I-GADJ-GUT for example needs 
case, number, gender and a primary/secondary indicator. Case for noun phrases is assigned by
the verbs or prepositions that govern them, i.e.\ through the roles in the verbal clause they are 
a direct component of, or by the preposition with which they share a prepositional phrase.
In German, the subject is assigned the nominative case and the (direct)
object is assigned the accusative case. 
Many prepositions always assign the same specific case, e.g.\ ``{\it wegen}'' (``because of'')
should always assign genitive case, but some assign different cases, depending on the semantics 
of the prepositional phrase (e.g.\ {\it in} assigns
accusative for C-TO-LOCATION and dative for C-AT-LOCATION). The case of a noun phrase is passed
on to its components. Number is carried over from the source tree in the transfer process.
It is propagated to all parts of the noun phrase, including adjectives, because in German,
the number information is needed everywhere. All German nouns have a specific grammatical
gender, which is listed in the (monolingual) lexicon. The German word {\it Buch} is neuter,
and so this information gets passed up to the dominating noun phrase and then back down to
the other components of the noun phrase. Finally, based on the article I-GART-INDEF-ART, 
the system signals the adjective to carry the primary ending, because this specific article 
does not have a primary ending itself\footnote{While
definite articles always carry a primary ending, therefore causing the form {\it secondary} 
to be sent to any following adjectives, indefinite singular articles of certain cases and genders, 
incl.\ the one in our example, do not have such primary ending and therefore have the 
form {\it primary} sent to any following adjectives.}.

\begin{figure}[htbp]
\epsfxsize=16.4cm
\centerline{\epsfbox{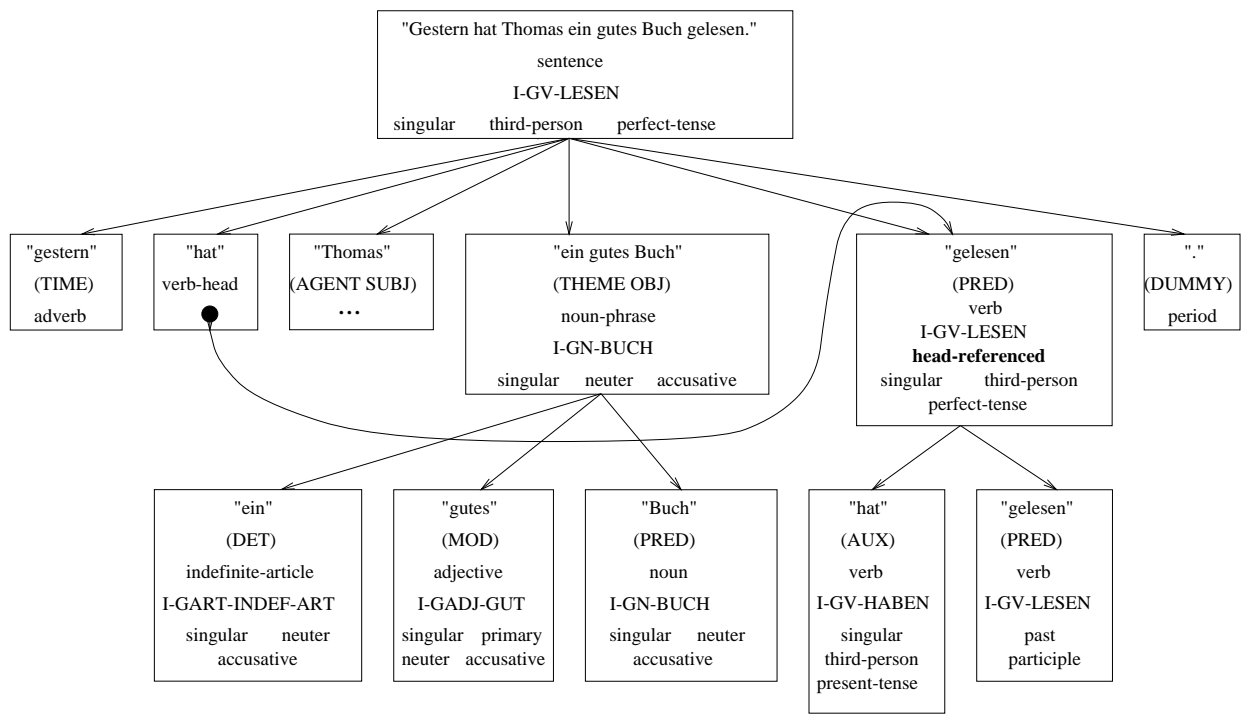}}
\caption{Generating the surface strings}
\label{fig-surface_words}
\end{figure}

Once all forms are propagated, the morphological module determines the surface forms of all words.
Starting with the leaves of the tree, surface forms are computed using the morphology module.
In a bottom up fashion, strings are then basically concatenated until the full target sentence is
stored at the surface form slot of the top node. The string in that very slot is the final
translation of the original source sentence.

As one can see in figure~\ref{fig-surface_words}, disjoint components like split verbs need
a little bit of a special treatment. The verb-head unit, pointing to a full verb complex,
and representing the conjugated part of the verb, takes the surface string of the conjugated
part of the verb, here ``{\it hat}'' of the German auxiliary verb ``{\it haben}''.
In the main verb entry, the flag {\it head-referenced} indicates that the conjugated part
of the verb is already been used by some verb-head unit, so that the surface string of
that part does not get included in the surface string of the main verb component.

So, in contrast to parsing and transfer, the generation module in our system does not require 
any learning.  Standard recursive techniques turned out to be sufficient, at least so far.
During further refinement, it might turn out that for relatively small subtasks in generation,
such as in the decision what case a specific preposition should assign to the noun phrase it 
governs, learning might be useful after all, but we expect that in the overall control of the
generation module, learning will remain inexpedient.

            %\chapter{Generation} \label{ch-generation}
\chapter{Translation Experiments}
\label{ch-t_exp}

\section{Test Methodology and Evaluation Criteria}

For training and testing the transfer and generation modules for translations from
English to German, we use sentences from the same Wall Street Journal corpus as for 
parsing.

The development of the transfer and generation modules including the selection of
example sentences and features for ambiguous translation was largely corpus driven
and based on the first 48 sentences (WSJ 0-47). The goal was to refine the transfer
and generation modules to the point that the system could produce both syntactically 
and lexically good translations for all 48 ``training'' sentences. It is important
to note that the expression ``training sentence'' here is not used in the strict
sense of machine learning, but rather as a sub-corpus that provides guidance as to
what linguistic phenomena need to be covered. Since the sub-corpus also already contains
a number of words with ambiguous translation targets, it also drove the selection of
succinct example sentences and corresponding features for the {\it transfer entries}
that were used to build decision trees for ambiguous concept transfer.

Section~\ref{sec-ambiguous-transfer} in the chapter on transfer already included
examples of transfer entries. A list of the various features used in English to 
German translation disambiguation decision structures can be found in 
section~\ref{sec-features-for-engl-german}.

When the system, based on correct parse trees for the 48 ``training sentences'', 
eventually produced good translations, we then tested the entire system on the 
following 32 sentences (WSJ 48-79).
The system input was an English sentence in the form of a simple character string, 
without any annotation. The output was a German sentence, again as a simple character
string.

In order to better be able to evaluate the contribution of the parser on one side and
the transfer and generation modules on the other side, we tested our system in two
versions. In one version, the parser was trained on 256 sentences excluding the
test sentences\footnote{(training sentences WSJ 0-47 \& 64-271 for test sentences 
WSJ 48-63 and training sentences WSJ 0-63 \& 80-271 for test sentences WSJ 64-79)},
thereby performing ``full'' test translations on sentences that had been used for
neither training the parser nor any other module. In the other version, the parser
was trained on the test sentences, meaning that the system by construction basically
started with a correct parse, subjecting only the transfer and generation modules to 
sentences that those modules had not been exposed to before. Comparing the results
of both versions helped us to better understand the contributions of the different
phases of translation.

\subsection{Comparative Translations}

To obtain some interesting comparisons, we used the same 32 test sentences on three
commercially available systems, Logos, SYSTRAN, and Globalink. The translation with
these systems were made over the Internet, all in October 1996. 

Globalink's Web site at http://www.globalink.com/ offers free translations for text
of limited length from English to French, Spanish, German and Italian and vice versa.
The source text is typed into a multiline text field form of the Web page and
the translation results are sent back by email. We had to split up the 32 sentences 
into three subsets because our text was longer than the Web interface accepted at any 
one time.

SYSTRAN's Web site at http://systranmt.com/ offers free translation of Web pages.
Given the fully qualified URL of a source page, the system allows translations
from English to French, Spanish, German, Italian, Portuguese and vice versa as well
as Russian to English. The translated Web page is displayed in a new browser window.
We created a Web page with the 32 test sentences and had them translated to German.

Logos' Web site at http://www.logos-ca.com/ did not provide free translation to the
general public, but the company was kind enough to provide a confidential Web site 
and let us use their system free of charge for our 32 test sentences.
Logos offers to translate from English or German to German, French, Spanish, Italian
or English.  As an additional input parameter, the
system asked for a domain choice, offering {\it General, Computers, Telecommunication}
and {\it Business} as alternatives. We chose the last option as the clearly most
appropriate for our Wall Street Journal sentences.

Two details about these three comparative translations:
\begin{itemize}
\item Since in testing
our system, we translated one sentence at a time, we ensured that the commercial
systems were at no disadvantage in this respect by separating input sentences by 
blank lines, if needed.
\item The results from SYSTRAN and Logos were returned so promptly that any human post
editing can be excluded just based on time. Based on the quality of the results
from Globalink, plus given that the translations were free of charge, human post editing 
is also most unlikely for that system.
\end{itemize}

Recall that the subcorpus we use is lexically limited to the 3000 words most 
frequently occurring in the Wall Street Journal. This allowed us to have a complete
monolingual lexicon for parsing and bilingual dictionary for transfer, so that in
the tests, our system was not confronted with unknown words. The commercial 
translation systems were designed for a wider range of domains and they also have significantly
larger lexicons and dictionaries. The only English words in the 32 test sentences
that apparently were unknown were {\it seasonally} (Logos) and {\it nationwide} 
(Logos and Globalink). Globalink also `translated' ``holiday schedule'' as
``holidayschedule'', ``currency exchange'' as ``currencyexchange'', 
``economic policy'' as ``economicpolicy'', and ``[over] the counter'' as
``[\"{u}ber] thecounter'', but since for example ``exchange'' was actually translated 
in another sentence, we conclude that there are entries for presumably at least
most of the individual words in the lexicon and that these translation failures
are primarily due to some shortcoming in the Globalink translation algorithm.
SYSTRAN, a veteran of machine translation systems, did not exhibit any lexical gaps.
So, with an average of one unknown word for all 32 test sentences, the commercial
systems were at no significant disadvantage with respect to unknown words.

As an additional comparison, the author of this dissertation, with a minor degree in
business and previous experience as a professional translator from English to German,
translated the 32 test sentences ``by hand''.

\subsection{Evaluators and Evaluation Criteria}

The translations were evaluated by ten bilingual graduate students at the University
of Texas at Austin. Seven of the volunteers were native speakers of German, two
were native speakers of English teaching German at the university and one was a
native speakers of English who has lived in Germany for many years, speaks German 
without an accent and has at least the German proficiency of the two teaching German.
Professional translations and proofreading of translations is normally performed by
native speakers of the target language, in our case German. This certainly justifies
the dominance of native German speakers in our group of evaluators.

Each of the 10 evaluators was given a list of the 32 original sentences (in English)
along with up to six German translations for each original sentence.
The translations always included those from the two versions of our system and
the three commercial systems. The human translations were given only to half of the
evaluators. This was done to control any possible influence of that translation on
the evaluation of the machine translations. When two or more translations were identical,
the translation was presented only once. The translations were listed in randomized
order, separately randomized for each WSJ sentence, and without identification.

For each of the translations, the evaluators were asked to assign two grades, one
for the grammatical correctness and the other for meaning preservation. 
Table~\ref{fig-grading_guidelines} and the following example
were given as a guideline to ensure a relative uniform standard.

\noindent
\begin{table}[htb]
\begin{tabular}{|c|p{15.1cm}|}
\multicolumn{2}{l}{{\bf Grammar (syntax and morphology)}} \\ \hline
{\it Grade} & {\it Usage} \\ \hline
1 & Correct grammar, including word order, word endings; the sentence reads fluently. \\ \hline
2 & Basically correct grammar, but not very fluent. \\ \hline
3 & Mostly correct grammar, but with significant shortcomings. \\ \hline
4 & The grammar is acceptable only in parts of the sentence. \\ \hline
5 & The grammar is generally so bad that the entire sentence becomes very hard to read. \\ \hline
6 & The grammar is so bad that the sentence becomes totally incomprehensible. \\ \hline
\multicolumn{2}{c}{} \\
\multicolumn{2}{c}{} \\
\multicolumn{2}{l}{{\bf Meaning (semantics)}} \\ \hline
{\it Grade} & {\it Usage} \\ \hline
1 & The meaning is fully preserved and can easily be understood. \\ \hline
2 & The meaning is mostly preserved and can be understood fairly well. \\ \hline
3 & The general idea of the sentence is preserved. \\ \hline
4 & Contains some useful information from the original sentence. \\ \hline
5 & A reader of the translated sentence can guess what the sentence is about,
     but the sentence provides hardly any useful information. \\ \hline
6 & The sentence is totally incomprehensible or totally misleading. \\ \hline
\end{tabular}
\caption{Grading guidelines for the syntax and semantics}
\label{fig-grading_guidelines}
\end{table}

\noindent (EXAMPLE)\\
{\bf Yesterday, I ate a red apple.}\\
\noindent \begin{tabular}{@{\hspace{0mm}}lll} (a) &
G\"{a}stern, ich haben essen Apfl-rot.
 & {\it Grammar:\rule{2mm}{.2mm}5\rule{2mm}{.2mm} Meaning:\rule{2mm}{.2mm}}2\rule{2mm}{.2mm} \\ (b) &
Meine roten \"{A}pfel haben viel gegessen.
 & {\it Grammar:\rule{2mm}{.2mm}1\rule{2mm}{.2mm} Meaning:\rule{2mm}{.2mm}}6\rule{2mm}{.2mm} \\ \end{tabular}\\

The scale is like the one used in the German education system: 1 = sehr gut {\it (excellent)};\\
2 = gut {\it (good)}; 3 = befriedigend {\it (satisfactory)}; 
4 = ausreichend {\it (passing)}; 5 = mangelhaft {\it (poor)};\\
6 = ungen\"{u}gend {\it (unsatisfactory)}.\\

The following is one of the 32 translation evaluation blocks of the questionnaire,
incidently the one with the funniest\footnote{Back translations: 
(a) He said that Pan Am currently has " over \$150 million " in the cash. 
{\it Note: `Million' should have been `Millionen' (plural).}
(b) He said Pan Am currently has over 150 million dollars in cash. 
(c) He said that frying-pan-in-the-morning currently has "more than 150 million dollars" in cash.
(d) He said that frying-pan is, has currently-" in surplus of \$150 million" in in cash.
{\it Note: `million' should have been `Millionen' (capitalized and plural).}}
machine translations:\\

\newpage
\noindent (WSJ 75)\\
{\bf He said Pan Am currently has "in excess of \$150 million" in cash.} \\
\noindent \begin{tabular}{@{\hspace{0mm}}lp{10.0cm}p{3.0cm}} (a) &
Er sagte, da{\ss} Pan Am z.Z.\ " \"{u}ber \$150 Million " im Bargeld hat.
 & {\it Grammar:\rule{8mm}{.2mm} Meaning:\rule{8mm}{.2mm}} \\ (b) &
Er sagte, Pan Am habe gegenw\"{a}rtig \"{u}ber 150 Mio.\ Dollar in Bargeld.
 & {\it Grammar:\rule{8mm}{.2mm} Meaning:\rule{8mm}{.2mm}} \\ (c) &
Er sagte, da{\ss} Pfannenvormittags zur Zeit "mehr als 150 Millionen Dollar" in Bargeld hat.
 & {\it Grammar:\rule{8mm}{.2mm} Meaning:\rule{8mm}{.2mm}} \\ (d) &
Er sagte, da{\ss} Pfanne ist, hat gegenw\"{a}rtig-" in \"{U}berschu{\ss} von \$150 million" in in bar.
 & {\it Grammar:\rule{8mm}{.2mm} Meaning:\rule{8mm}{.2mm}} \\ \end{tabular} \\

The questionnaires handed out for translation evaluation were hardcopies of the two Web pages
http://www.cs.utexas.edu/users/ulf/eval\_tegm.html, the version that always
includes human translations, and
http://www.cs.utexas.edu/users/ulf/ eval\_tego.html, the version that does not
include human translations, unless it happens to match one of the machine
translations.
Appendix~\ref{app-quest} contains the questionnaire including human translations
in dissertation format.

\section{Test Results}

\subsection{Overview}

Table~\ref{fig-translation_summary} summarizes the evaluation results of translating 32 sentences
from the Wall Street Journal from English to German.
\begin{table}[htbp]
\begin{center}
\begin{tabular}{|l|c|c|} \hline \hline
   System                          & Syntax  & Semantics \\ \hline
   Human translation               & 1.18     & 1.41     \\
   {\sc Contex} on correct parse         & 2.20     & 2.19     \\
   {\sc Contex} (full translation)       & 2.36     & 2.38     \\
   Logos                           & 2.57     & 3.24     \\
   SYSTRAN                         & 2.68     & 3.35     \\
   Globalink                       & 3.30     & 3.83     \\ \hline \hline
\end{tabular}\\
\caption{Summary of translation evaluation results (best possible = 1.00, worst possible = 6.00)}
\label{fig-translation_summary}
\end{center}
\end{table}

Our system performed significantly better than the commercial systems, but this has to
be interpreted with caution, since our system was trained and tested on sentences
from the same lexically limited corpus (but of course without overlap), whereas
the other systems were developed on and for texts from a larger variety of domains,
making lexical choices more difficult in particular. Additionally, the syntactic
style varies from to domain to domain, but we believe that this is less critical
for this translation quality evaluation,
because the Wall Street Journal already covers a wide range of syntactic constructs.

Note that the translation results using our parser are fairly close to those
starting with a correct parse.
This means that the errors made by the parser have had a relatively moderate impact
on translation quality. The transfer and generation modules were developed and trained
based on only 48 sentences, so we expect a significant translation quality improvement
by further development of those modules.

Recall that Logos required a translation domain as an extra input parameter. To get
an idea of the influence of this parameter, we repeated the translation with a domain
choice of `general' as opposed to `business'. In 32 sentences, only four words/expressions
were translated differently, one equally bad, one better for `business' and two better
for `general'. The difference is statistically insignificant so that for our test
sentences, the domain parameter apparently did not play a crucial role.

The impact of parse errors on the final translation quality depends of course greatly
on the specific language pair. Germanic and Romance languages like English, German,
French and Spanish are not only linked by springing from the same Western branch within
the Indo-European language family; due to intensive political, commercial and cultural
contacts of the native speakers of these languages throughout history, the languages
continuously kept influencing each other.

Many ambiguities, like for example ``with''-clauses, which in the instrumental case
(``eat pasta with a fork'') attach to the verb and which in the complementary case
(``eat pasta with sauce'') attach to the preceding noun phrase,
are lexically and structurally homomorphic among these languages,
so that an incorrect attachment would not be visible in the final translation,
whereas other languages, as for example Japanese,
make a clear lexical distinction in this case.\footnote{The complementary {\it with} translates 
as {\it ``to''}, whereas the instrumental {\it with} translates as {\it ``de''}.}
Therefore we would clearly expect a higher
impact of parse errors on translation quality for an English to Japanese translation.
On the other hand, translations between more closely related languages, e.g.\ between
Swedish, Norwegian and Danish, would suffer even less from parse errors.

Despite the relative proximity of English and German, the impact of parse errors in our
system on translations is certainly smaller than had been expected. This seems to indicate 
that our parser is already fairly robust where it counts.

\subsection{Variation Analysis}

In a more detailed analysis of the evaluation results, we computed standard
deviations for the various grades and grade differences as shown in
table~\ref{fig-translation_stats}.
\begin{table}[htbp]
\begin{center}
\begin{tabular}{|l|cc|cc|} \hline \hline
   System                          & \multicolumn{2}{c|}{Syntax} & \multicolumn{2}{c|}{Semantics} \\ 
                                   & average  & stand.dev. & average  & stand.dev. \\ \hline
   Human translation               & 1.18    & 0.29       & 1.41      & 0.31       \\
   \hspace*{1cm}{\it difference}   & 1.02    & 0.41       & 0.78      & 0.39       \\
   {\sc Contex} starting on correct parse  & 2.20  & 0.45       & 2.19      & 0.46       \\
   \hspace*{1cm}{\it difference}   & 0.16    & 0.08       & 0.19      & 0.09       \\
   {\sc Contex} (full translation)       & 2.36    & 0.43       & 2.38      & 0.47       \\
   \hspace*{1cm}{\it difference}   & 0.21    & 0.25       & 0.86      & 0.72       \\
   Logos                           & 2.57    & 0.60       & 3.24      & 0.82       \\
   \hspace*{1cm}{\it difference}   & 0.11    & 0.21       & 0.12      & 0.17       \\
   SYSTRAN                         & 2.68    & 0.52       & 3.35      & 0.86       \\
   \hspace*{1cm}{\it difference}   & 0.62    & 0.13       & 0.48      & 0.25       \\
   Globalink                       & 3.30    & 0.60       & 3.83      & 0.73       \\ \hline \hline
\end{tabular}\\
\end{center}
\caption{Standard deviation analysis of translation evaluation results}
\label{fig-translation_stats}
\end{table}

For machine translations, we find standard deviations in the range of 0.43 to 0.60 for syntax
and 0.46 to 0.86 for semantics. However, the standard deviations for {\it differences} between
adjacently ranking machine translations are much lower (0.08 to 0.25 for syntax and 0.09 to 0.72
for semantics). This means that while there are considerable differences in grading from one
evaluator to another, the relative grading differences between systems are relatively small
across different evaluators. So, while some evaluators graded more strictly than others,
they all came up with fairly similar relative ratings.

Note in particular that while the difference in both syntax and semantics between the two
versions of our system is quite small (0.16 and 0.19 respectively), the corresponding
standard deviations are even smaller (0.08 and 0.09) indicating that while the evaluators rank
the translations based on a correct parse only a little better than the full translations,
this difference is quite consistent; in fact all evaluators graded the {\sc Contex} translations
starting on a correct parse to be better for both syntax and semantics.

Note that on the other hand the small differences between Logos and Systran (0.11 and 0.12)
are accompanied by relatively big standard deviations (0.21 and 0.17); this is reflected 
by the fact that a minority of evaluators ranked SYSTRAN better than Logos.
\begin{table}[htbp]
\begin{center}
\begin{tabular}{|l|cc|cc|} \hline \hline
   \hfill  Evaluators     & \multicolumn{2}{c|}{American (3)} & \multicolumn{2}{c|}{German (7)} \\
   System                          & Syntax  & Semantics & Syntax  & Semantics \\ \hline
   Human translation               & 1.39     & 1.54     & 1.10     & 1.36     \\
   {\sc Contex} starting on correct parse  & 2.30   & 2.34     & 2.16     & 2.13     \\
   {\sc Contex} (full translation)       & 2.44     & 2.55     & 2.33     & 2.30     \\
   Logos                           & 2.83     & 3.32     & 2.45     & 3.20     \\
   SYSTRAN                         & 2.71     & 3.36     & 2.67     & 3.35     \\
   Globalink                       & 3.30     & 3.75     & 3.30     & 3.87     \\ \hline \hline
\end{tabular}\\
\caption{Evaluation results differentiated by nationality of evaluator}
\label{fig-translation_american_german}
\end{center}
\end{table}
By and large the results from native English and German speaking evaluators are similar.
However, the Americans tended to give worse grades. The most striking difference is for the
syntax of the manual translations, which, relative to the best possible grade (1.00), are almost 
four times as negative
for the native English speakers than for the countrymen of the translator. However, the data has 
to be interpreted very cautiously, because the sub-sample sizes are quite small.

Another somewhat unexpected result was the grade for meaning preservation for the human
control translation (1.41). A closer analysis revealed that more than a third of the difference 
from the optimal 1.00 resulted
from disagreement over a single sentence (WSJ 53). All except one evaluator interpreted the
sentence ``{\it The St.\ Louis-based bank holding company previously traded on the American Stock
Exchange.}'' as meaning that the bank holding company was the agent of the trading, and therefore
gave extremely bad grades to the human control translation which interpreted the sentence such that 
stocks of the bank holding company were traded on the American Stock Exchange. Monolingual domain
experts (incl.\ from the University of Texas Business School), who were 
given this sentence without context, however confirm the passive reading of {\it traded}.
The passive reading is further corroborated by the context of the WSJ article, which, of course, 
had not been available to the translation programs, the translator or the evaluators.

While such a misunderstanding has occurred in only one sentence to such an extent, and to a lesser
degree also in a second sentence, and the overall consequences appear to be fairly limited,
the evaluation of the control translations nevertheless demonstrates that
  translations can be quite hard to evaluate, even for bilingual speakers.
  Evaluators sometimes don't agree on substantial semantic issues, which results
        in evaluation differences.\footnote{However in the case just discussed, all machine
        translation programs selected the active reading for ``traded'' and therefore did not 
        benefit or suffer from the controversial evaluations relative to each other.}
  Optimally, translators and evaluators should have top competency not only in the
        languages involved, but also in the translation subject domain area.
  Such a strong triple competency requirement greatly reduces the number of well qualified 
  professional translators and reviewers, which, incidentally, is one of the main reasons 
  why translation support by computers is so important:
  it can free the relatively scarce expert translators from the routine aspects of translation
  and thereby make them more efficient.

\subsection{Correlation Analysis}

Given evaluations on both parsing and full translation, an interesting question is whether
the parsing evaluation criteria we used are a useful predictor for actual translation quality.
To investigate this issue, we computed a matrix of correlation coefficients between our parsing 
evaluation criteria and the syntactic and semantic grades of the final (full) translation 
and show the results in table~\ref{fig-corr_matrix}.

\noindent
\begin{table}[htb]
\begin{center}
\begin{tabular}{|l|r@{\hspace{1.5mm}}r@{\hspace{1.5mm}}r@{\hspace{1.5mm}}r@{\hspace{1.5mm}}r@{\hspace{1.5mm}}r@{\hspace{1.5mm}}r@{\hspace{1.5mm}}r@{\hspace{1.5mm}}r@{\hspace{1.5mm}}r|r@{\hspace{1.5mm}}r|} \hline
%      & \multicolumn{1}{c}{pr} & \multicolumn{1}{c}{rec} & \multicolumn{1}{c}{l\_pr} & \multicolumn{1}{c}{l\_rec} & \multicolumn{1}{c}{t\_acc} & \multicolumn{1}{c}{cross} & \multicolumn{1}{c}{ops} & \multicolumn{1}{c}{opseq} & \multicolumn{1}{c}{match} & \multicolumn{1}{c|}{str\_l} & \multicolumn{1}{c}{synt} & \multicolumn{1}{c|}{sem} \hline
       &  pr   &  rec  & l\_pr & l\_rec & t\_acc & cross & ops & opseq & match & str\_l & synt &  sem   \\ \hline
pr     &  1.00 &  0.91 &  0.84 &  0.88 &  0.48 & -0.87 &  0.42 &  0.49 &  0.94 &  0.69 & -0.63 & -0.63  \\
rec    &  0.91 &  1.00 &  0.86 &  0.95 &  0.39 & -0.84 &  0.53 &  0.57 &  0.97 &  0.80 & -0.64 & -0.66  \\
l\_pr  &  0.84 &  0.86 &  1.00 &  0.92 &  0.64 & -0.66 &  0.56 &  0.57 &  0.95 &  0.80 & -0.75 & -0.78  \\
l\_rec &  0.88 &  0.95 &  0.92 &  1.00 &  0.54 & -0.71 &  0.54 &  0.55 &  0.98 &  0.78 & -0.65 & -0.65  \\
t\_acc &  0.48 &  0.39 &  0.64 &  0.54 &  1.00 & -0.25 &  0.21 &  0.20 &  0.54 &  0.28 & -0.66 & -0.56  \\
cross  & -0.87 & -0.84 & -0.66 & -0.71 & -0.25 &  1.00 & -0.42 & -0.47 & -0.79 & -0.66 &  0.58 &  0.54  \\
ops    &  0.42 &  0.53 &  0.56 &  0.54 &  0.21 & -0.42 &  1.00 &  0.77 &  0.54 &  0.72 & -0.45 & -0.41  \\
opseq  &  0.49 &  0.57 &  0.57 &  0.55 &  0.20 & -0.47 &  0.77 &  1.00 &  0.57 &  0.71 & -0.39 & -0.36  \\
match  &  0.94 &  0.97 &  0.95 &  0.98 &  0.54 & -0.79 &  0.54 &  0.57 &  1.00 &  0.80 & -0.70 & -0.71  \\
str\_l &  0.69 &  0.80 &  0.80 &  0.78 &  0.28 & -0.66 &  0.72 &  0.71 &  0.80 &  1.00 & -0.62 & -0.62  \\ \hline
synt   & -0.63 & -0.64 & -0.75 & -0.65 & -0.66 &  0.58 & -0.45 & -0.39 & -0.70 & -0.62 &  1.00 &  0.67  \\
sem    & -0.63 & -0.66 & -0.78 & -0.65 & -0.56 &  0.54 & -0.41 & -0.36 & -0.71 & -0.62 &  0.67 &  1.00  \\ \hline
\end{tabular}
\end{center}
\caption{Correlation matrix between various parsing and translation metrics.
  (pr = precision, rec = recall, l\_pr = labeled precision, l\_rec = labeled recall, 
  t\_acc = tagging accuracy, cross = number of crossing constituents, ops = number of correct operations,
  opseq = whether or not the entire operation sequence is correct, match = mixed index including
  precision, recall, labeled precision and labeled recall, str\_l = whether or not the parse tree has
  the proper structure and labeling, synt = syntactic grade of the translated sentence, sem  =
  semantic grade of the translated sentence)}
\label{fig-corr_matrix}
\end{table}

Assuming an approximate linear correlation between two variables {\it x} and {\it y}, 
the {\it correlation coefficient c} indicates the strength of the correlation between the two variables. 
$|c|$ near 1.00 indicates a very strong correlation,
whereas $|c|$ near 0.00 indicates a weak or no correlation.
A {\it positive c} indicates that {\it y} tends to increase with an increasing {\it x} whereas
a {\it negative c} indicates that {\it y} tends to decrease with an increasing {\it x}.

The correlation coefficient of a variable with itself is always 1.00. Correlation coefficients
are symmetrical with respect to their two variables. An example for a high correlation is
the one
between {\it match} and {\it precision, recall, labeled precision} and {\it labeled recall}
(0.94, 0.97, 0.95, and 0.98 respectively),
which is no big surprise because {\it match} is an index composed of just those criteria.
An example for a low correlation is between tagging accuracy and the number of crossing
constituents, -0.25 (not because of the minus sign, but because of the low absolute value).

As it turns out, the best predictor for translation quality is labeled precision, with
correlation coefficients of -0.75 and -0.78 for syntax and semantics respectively. 
Recall that on the German grading scale, 1 is the best possible and 6 the worst possible
grade. Therefore a negative correlation coefficient here means that the better the labeled
precision, the better the expected syntactic and semantic quality of the translation.
A more detailed analysis has shown that labeled precision is also more strongly correlated 
to translation quality than any combination of labeled and unlabeled precision.
The various parsing metrics have about the same correlation with the syntactic and semantic
grades. Only tagging accuracy has a noticeably larger impact on syntax than on semantics
(-0.66 and -0.56 respectively).

\begin{figure}[htb]
\epsfxsize=13.4cm
\centerline{\epsfbox{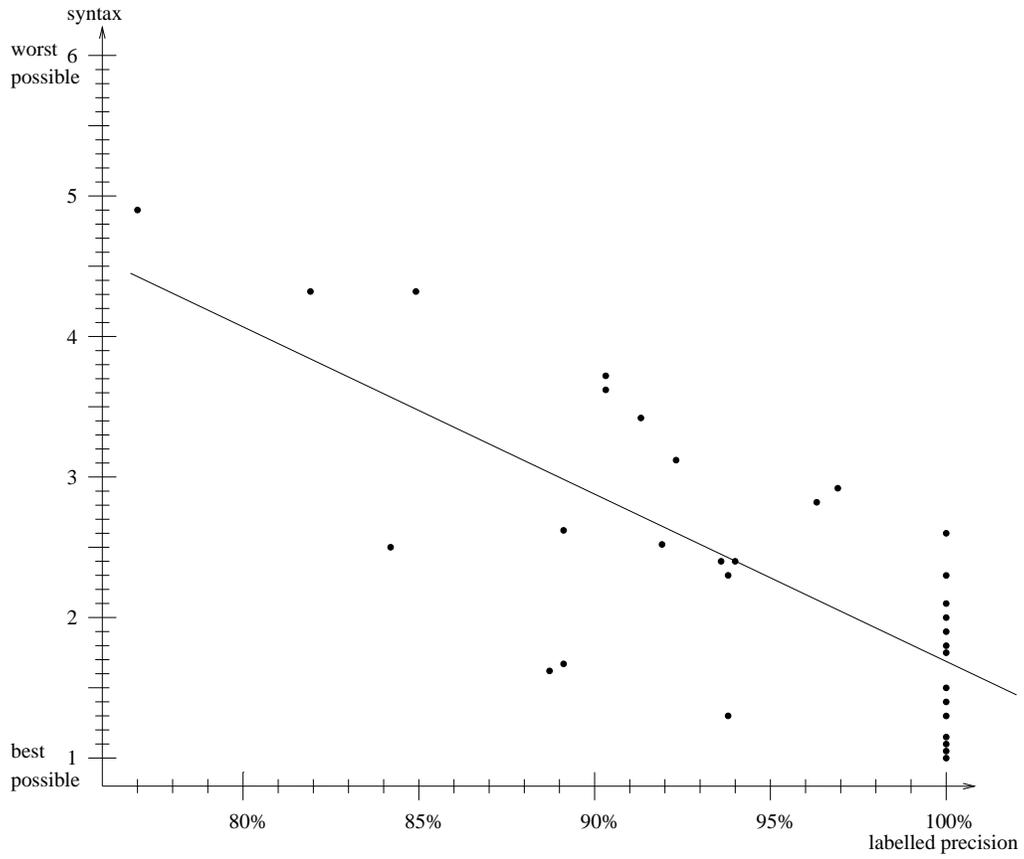}}
\caption{Correlation between labeled precision and syntax; coefficient = -0.75}
\label{fig-lab_pr_synt_corr}
\end{figure}
% y = 13.722433 - 12.039495 x

\begin{figure}[htb]
\epsfxsize=13.4cm
\centerline{\epsfbox{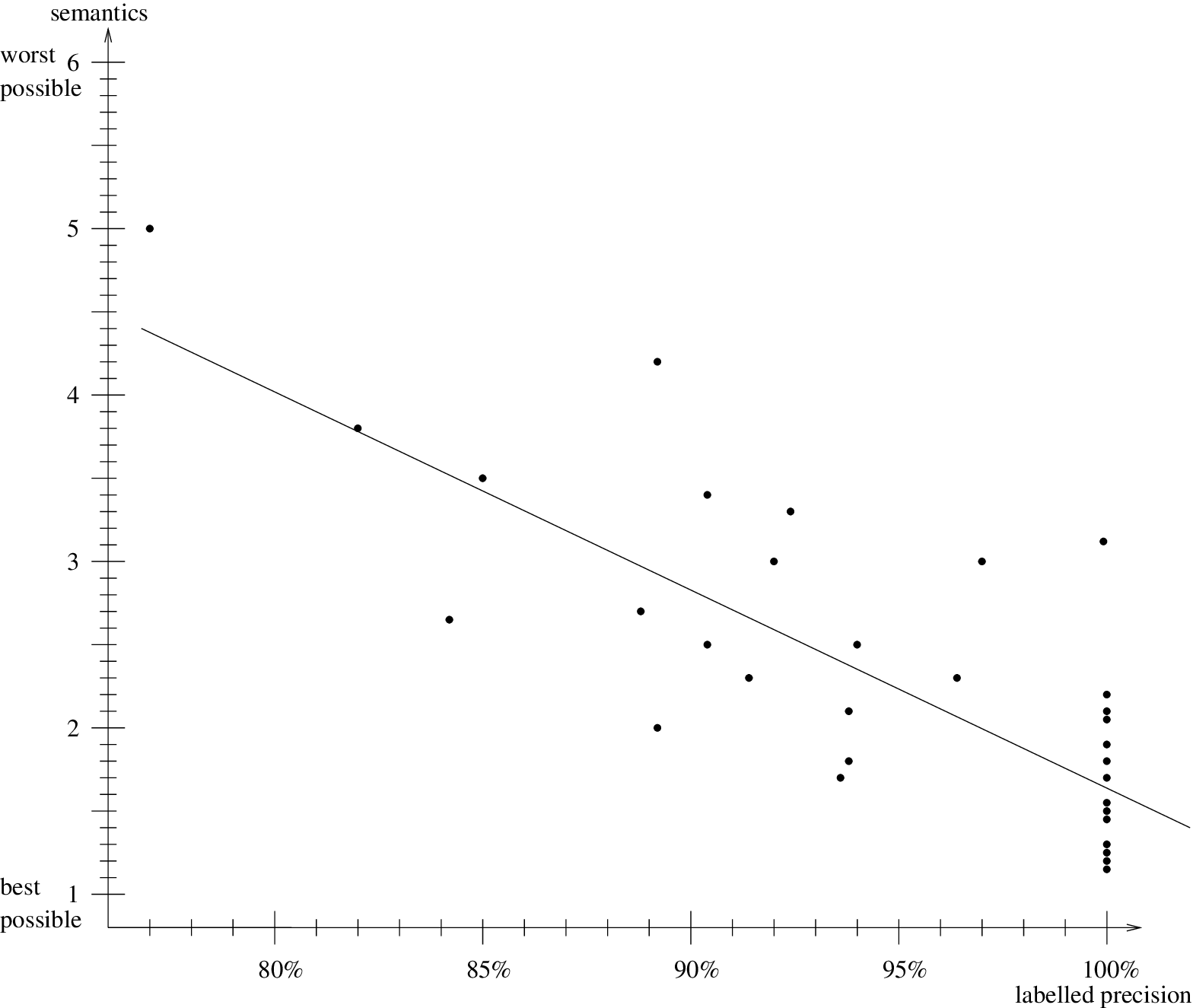}}
\caption{Correlation between labeled precision and semantics; coefficient = -0.78}
\label{fig-lab_pr_sem_corr}
\end{figure}
% y = 13.206429 - 11.468228 x

Figures \ref{fig-lab_pr_synt_corr} and \ref{fig-lab_pr_sem_corr} show the detailed distribution
of labeled precision and the syntactic and semantic grades.
The correlations are clearly visible, but by no means overwhelming, which concurs with our
previous observation that the impact of parse errors on translation quality in our system 
is only limited.
         %\chapter{Translation Experiments} \label{ch-t_exp}
\hyphenation{Mar-cin-kie-wicz}

\chapter{Related Work}
\label{ch-related}

In the introductory chapter, we already mentioned a number of papers representing the traditional
approach to parsing with hand-crafted rules, as well as a number of references to learning-oriented 
approaches of `sub-parsing' tasks. We now describe related empirical work on parsing.
 
\section{Simmons and Yu}
Our basic deterministic parsing and interactive training paradigm is based on \cite{simmons:compling92}.
Their ``context-dependent grammar'' (CDG) has the following characteristics:
\begin{itemize}
   \item It uses 10 features, the five syntactic categories to the left and the five to the right of the 
         current parse position.
   \item It has two parse actions, ``{\it shift}'' and ``{\it reduce $<$new syntactic category$>$}''.
   \item Words are pre-tagged, based on a {\it context-sensitive dictionary},
         which examines a context of plus and minus three words for part of speech assignment.
   \item Output is a binary syntactic phrase structure; a separate system produces case structures.
   \item For a given partially parsed state, the system proposes the next parse action based on the
         `closest' example, namely the one that maximizes
         \[\sum_{\mbox{f $\in$ feature set}}
                 \left\{ \begin{array}{ll}   %}
                                      weight(f) & \mbox{if examples have matching values for feature f} \\
                                      0         & \mbox{otherwise} \end{array} \right. \]
         where {\it weight(f)} is is the distance of syntactic element referred to by feature {\it f} 
         from the context window, i.e.\ 5 for the
         top element of the stack and the next input list element, and 1 less for each position that an 
         element is further removed from the `current position'.\footnote{Since the examples are stored
         in a hash table using the top two parse stack elements as a key, examples that don't match
         these two features are actually not considered at all.}
\end{itemize}

\noindent We have extended their work by
\begin{itemize}
   \item significantly increasing the expressiveness of the feature language, in particular by
         moving far beyond the 10 simple features that were limited to syntax only;
   \item significantly increasing the expressiveness of the parse action language by adding six
         new types of operations, adding (further) arguments to the shift and reduce operations,
         thereby, among other things, allowing discontinuous constituents and `empty categories';
   \item adding background knowledge, in particular a KB and a subcategorization table;
   \item building a much richer internal structure, including both phrase-structure and case-frame information;
   \item providing morphological pre-processing;
   \item introducing a sophisticated machine learning component.
\end{itemize}

\section{Parsers Learning From Treebank Examples}

Several researchers have now used the Penn Treebank \cite{marcus:compling93} to develop parsers.
The Penn Treebank is a corpus of over 4.5 million words of American English that has been 
annotated with part of speech. In addition, over half of it has been annotated for skeletal 
syntactic structure.
For a better comparison of the systems, we first describe some representational differences
between the Penn Treebank and our system, argue that parse action sequences carry more 
information, and finally discuss and compare specific treebank based parsers.

\subsection{Comparison between Penn Treebank and {\sc Contex} Parse Trees}

The Penn Treebank provides only a skeletal syntactic structure, in which, most prominently,
all internal structure of the noun phrase up through the head and including any single-word
post-head modifiers is left unannotated. Experiments at the University of Pennsylvania have
shown that this format increases the productivity (per word) of human annotators, a crucial
aspect when annotating such a large corpus.

The output of {\sc Contex} on the other hand provides a much more detailed
integrated phrase-structure and case-frame tree that contains both syntactic and semantic
classes and roles for the various constituents. Consider for example the representations
of the noun phrase ``{\it the October 1987 stock market crash}'' as shown in
table~\ref{fig-treebank-np-detail-comp}.

\begin{table}[htbp]
\noindent {\bf Penn Treebank}:
\begin{verbatim}
 (NP the October 1987 stock market crash)
\end{verbatim}

\noindent {\bf {\sc Contex}}:
\begin{verbatim}
 "the October 1987 stock market crash":
    synt:    S-NP
    class:   I-EN-CRASH
    forms:   (((NUMBER F-SING) (PERSON F-THIRD-P)))
    lex:     "crash"
    subs:    
    (DET)  "the":
        synt/class: S-DEF-ART/I-EART-DEF-ART
        forms: unrestricted 
    (TIME)  "October 1987":
        synt/class: S-NP/C-AT-TIME
        forms: (SING 3P)
        (PRED) "October 1987":
            (PRED)  "October"
            (MOD)  "1987"
    (PRED)  "stock market crash":
        synt/class: S-COUNT-NOUN/I-EN-CRASH
        forms: (SING 3P)  
        (MOD) "stock market":
            (MOD)  "stock"
            (PRED)  "market"
        (PRED)  "crash"
\end{verbatim}
\caption{Comparison of the syntactic structure of a complex NP 
         as provided by the Penn Treebank and {\sc Contex}.}
\label{fig-treebank-np-detail-comp}
\end{table}

In the {\sc Contex} representation, there is full syntactic and semantic class information plus
morphological, lexical and other applicable values at all levels. (However, in these examples,
only a reduced level of detail is actually displayed at lower levels to keep the outputs from 
quantitatively overwhelming the reader.)
Appendix section~\ref{sec-resulting-parse-tree} gives an example for a complete parse tree.

The second example (see tables~\ref{fig-treebank-snt-detail-penn} 
and~\ref{fig-treebank-snt-detail-contex}/\ref{fig-treebank-snt-detail-contex-continuation})
shows the different representations for a full sentence (``{\it She wanted
to avoid the morale-damaging public disclosure that a trial would bring.}'').
The asterisk (*) in the Penn Treebank representation stands for an ``understood'' subject,
and corresponds to the ``$<$REF$>$2'' in the {\sc Contex} representation.
Notice that the {\sc Contex} representation provides syntactic and semantic roles, 
semantic classes, co-indexing and miscellaneous morphological and lexical information,
but most important of all, once again, a more detailed structure.

\begin{table}[htbp]
\noindent {\bf Penn Treebank}:
\begin{verbatim}
 (S (NP she)
    (VP wanted
        (S (NP *)
           to
           (VP avoid
               (NP (NP the morale-damaging public disclosure)
                   (SBAR that
                         (S (NP a trial)
                            would
                            (VP bring))))))))
\end{verbatim}
\caption{Penn Treebank representation of the sentence ``{\it She wanted
to avoid the morale-damaging public disclosure that a trial would bring.}''.}
\label{fig-treebank-snt-detail-penn}
\end{table}

\begin{table}[htbp]
\noindent {\bf {\sc Contex}}:
\begin{verbatim}
"<She>2 wanted <REF>2 to avoid <the morale-damaging public disclosure 
 <that>1 a trial would bring>1.":
   synt:    S-SNT
   class:   I-EV-WANT
   forms:   (((GENDER F-FEM) (PERSON F-THIRD-P) (NUMBER F-SING)
              (CASE F-NOM) (TENSE F-PAST-TENSE)))
   lex:     "want"
   subs:    
   (SUBJ EXP)  <>2 "<She>2":
       synt/class: S-NP/I-EN-PERSONAL-PRONOUN
       forms: (SING 3P NOM FEM)  
       props:   ((INDEX 2) (ORIG-SURF "She") (INDEXED TRUE))
       (PRED)  "She"
   (PRED)  "wanted":
       synt/class: S-VERB/I-EV-WANT
       forms: (PAST or PAST-PART)
   (INF-COMPL
    THEME)  "<REF>2 to avoid <the morale-damaging public disclosure 
             <that>1 a trial would bring>1":
       synt/class: S-SNT/I-EV-AVOID
       forms: (TO-INF SING 3P NOM FEM)  
       (SUBJ AGENT)  <REF>2"<REF>2":
       (PRED) "to avoid":
           (AUX)  "to"
           (PRED)  "avoid"
       (OBJ
        THEME)  <>1"<the morale-damaging public disclosure <that>1 a 
                    trial would bring>1":
           (DET)  "the"
           (MOD) "morale-damaging":
               (MOD)  "morale"
               (DUMMY)  "-"
               (PRED)  "damaging"
           (MOD) "public":
               (PRED)  "public"
           (PRED)  "disclosure"
\end{verbatim}
{\it Continued in table~\ref{fig-treebank-snt-detail-contex-continuation}}
\caption{{\sc Contex} representation of the sentence ``{\it She wanted
to avoid the morale-damaging public disclosure that a trial would bring.}''.}
\label{fig-treebank-snt-detail-contex}
\end{table}
\begin{table}[htbp]
{\it Continued from table~\ref{fig-treebank-snt-detail-contex}}
\begin{verbatim}
           (PRED)  "disclosure"
           (MOD) "<that>1 a trial would bring":
               (OBJ THEME) "<that>1":
                   (PRED)  <>1 "<that>1"
               (SUBJ INSTR) "a trial":
                   (DET)  "a"
                   (PRED)  "trial"
               (PRED) "would bring":
                   (AUX)  "would"
                   (PRED)  "bring"
   (DUMMY)  ".":
       synt: D-PERIOD
\end{verbatim}
\caption{{\sc Contex} representation, continued from table~\ref{fig-treebank-snt-detail-contex}}
\label{fig-treebank-snt-detail-contex-continuation}
\end{table}

In neither system are the parse trees actually entered from scratch.
For the Penn Treebank sentences, a team of humans corrected automatically computed trees,
while for {\sc Contex}, a human corrected automatically computed parse actions\footnote{Except
for the very first sentence, for which a complete parse action sequence had to be entered
by hand; for all other sentences, a bootstrapping approach limited the manual entry
to corrections; see section~\ref{sec-training-the-parser} for more details.}
based on which the correct parse tree was continuously and automatically constructed.

\subsection{Declarative vs.\ Operational Parse Information}

This brings us to another essential difference with respect to parse information.
The Penn Treebank is actually a collection of trees, whereas our system is based on
parse action sequences. Since we can easily and automatically derive a parse tree
from a parse action sequence, the later obviously carries at least as much `linguistic'
information. We believe that it carries in fact more information and is much better
suited for training a parser, which has to break down the parsing of a sentence into
smaller pieces anyway. Consider the example in table~\ref{fig-treebank-pasta-snts} 
with the sentence pair (``I ate the pasta with the expensive cheese/fork.'') as 
represented in Penn Treebank format.

\begin{table}[htbp]
\begin{verbatim}
 (S (NP I)
    (VP ate
        (NP the pasta
            (PP with
                (NP the expensive cheese)))))
 (S (NP I)
    (VP ate
        (NP the pasta)
        (PP with
            (NP the expensive fork))))
\end{verbatim}
\caption{Penn Treebank representation of ``I ate the pasta with the expensive cheese/fork.''}
\label{fig-treebank-pasta-snts}
\end{table}

Despite the superficially similar surface structure of the sentences, the prepositional 
phrases have very different functions and accordingly attach differently:
the complementational PP is an adjunct to the preceding noun phrase,
whereas the instrumental PP is an adjunct to the verb.

Deriving the proper parse action sequences from a parse tree can now be quite tricky.
If the parse action sequence just follows the bottom-up, left-to-right order of the
parse tree, we can encounter problems as illustrated in the following example\footnote{As
usual, the asterisk (*) indicates the current position, the pluses (+) separate the current
partially parsed components.}:
\begin{enumerate}
 \item I + ate + the pasta * with + the + expensive + cheese + .
 \item I + ate + the pasta * with + the + expensive + fork + .
\end{enumerate}

Following the parse tree order, the parser should,
in the first example, shift in the preposition, because the prepositional phrase
is part of the pasta noun phrase, whereas in the second case, the pasta noun phrase should
now be combined with the verb. Assuming that the differentiating word ({\it cheese/fork})
is still out of the context window, the contexts of the two examples are identical and the
parser must process both cases equally. This does not only pose a problem for a deterministic
parser, but also for a statistical model, which would check both alternatives, noting the
probabilities associated with each choice. But since the rest of the parse is unambiguous,
there won't be any more probabilities below 1, resulting in total parse probabilities equal
to the probabilities at the critical point, and thus attaching the PP the same way in both
cases.

In our system on the other hand, we first parse the PP in both cases and then make
an attachment decision when the parse is advanced to ``{\it I} + {\it ate} + {\it the pasta}
+ {\it with the expensive cheese/fork} * .'' Then the parser decides whether or not to
attach the PP to the preceding noun phrase. If not,
the PP is shifted back out onto the input list, the object is combined with the verb,
the PP is shifted back in and combined with the verb phrase. This type of strategy is not
included in a parse tree, but is automatically part of a parse action sequence, just by
virtue of the supervisor guiding the parser that way during training.

\subsection{SPATTER}

\citeA{magerman:thesis94,magerman:acl95} uses a statistical decision tree model, 
training his system SPATTER with parse action sequences for 40,000 Wall Street Journal
sentences derived from the Penn Treebank.
Questioning the traditional n-grams, Magerman already advocates a heavier reliance on
contextual information. Going beyond Magerman's still relatively rigid set of a little over
36 features, we propose a yet richer, basically unlimited feature language set.
Our parse action
sequences are too complex to be derived from a treebank like Penn's. Not only do our
parse trees contain semantic annotations, roles and more syntactic detail, we also rely
on the more informative parse action sequence. While this necessitates the involvement
of a parsing supervisor for training, we are able to perform deterministic parsing
and get already very good test results for only 256 training sentences.

\subsection{IBM Language Modeling Group}

Magerman's work was significantly influenced by the IBM Language Modeling Group.
\citeA{black:acl92} show how, using a treebank, the various parse rules of a hand-coded
broad-coverage context-free phrase-structure grammar can automatically be assigned 
probabilities in order to better identify the parse tree that is most likely correct.
Their ``history-based grammar'' (HBG) system \cite{black:acl93} enhances
the probabilistic context-free grammar approach by providing more detailed
(incl.\ semantic) linguistic information to resolve ambiguity. 

Finally, the Candide system \cite{brown:compling90,berger:arpa94} presents a 
statistical approach to machine translation. A target sentence is built word by word
from left to right.
The system keeps a hypothesis set of ranked partial target sentences. 
A partial target sentence (hypothesis) is extended by adding one word, and then re-ranked
by estimating the probability that the new partial target sentence corresponds to the
source sentence and that the new partial target sentence would occur like that in the
target language.
Recent additions to the system include (1) pre-processing of the source text, including
part-of-speech tagging, morphological analysis, special treatment for numbers and names, 
and some local word reordering and normalizations, (2) a reverse post-processing
of the target text, and (3) the introduction of context sensitivity to translation
probabilities.

There is however no structural analysis of the source sentence. It seems that as a
consequence, Candide works relatively well with short sentences\footnote{In
\cite{berger:arpa94}, the authors report that their ``in-house evaluation methodology
consists of fully-automatic translation of 100 sentences of 15 words or less''.},
but it is not clear
how much it can be improved to produce good translations for longer sentences without
incorporating at least a shallow structural analysis. Reported
results from an {\sc ARPA} evaluation show that Candide's fluency of the target
language was better than Systran's, that its adequacy was remarkable (.67 on a 0 to 1
scale compared to .743 for Systran) and that Candide could at least serve as a time-saving 
tool to produce translations that can be manually post-edited faster than a manual 
translation from scratch would take.

\subsection{Bigram Lexical Dependencies}

Another treebank based statistical parser, the bigram lexical dependencies (BLD) 
system \cite{collins:acl96}, 
is based on probabilities between head-words in the parse tree, enhanced by additional
distance features that check for order and adjacency of dependency candidates as well
as intervening verbs, and intervening or surrounding commas.
Trained on the same 40,000 sentences as SPATTER,
it relies on a much more limited type of context than our system
and needs little background knowledge.

\subsection{Comparison of Results}

\begin{table}[htbp]
\begin{center}
\begin{tabular}{|l|c|c|c|c|c|} \hline \hline
   System                          & Spatter  & Spatter  & BLD      & \hspace*{-1mm}{\sc Contex}\hspace*{-1mm}   & \hspace*{-1mm}{\sc Contex}\hspace*{-1mm}   \\ \hline \hline
   Training Sentences              & 40,000   & 40,000   & 40,000   & {\bf 64} & {\bf 256} \\
   Background Knowledge            & little   & little   & little   & much     & much     \\
   Test Sentence Length Range    & {\bf 4-25} & {\bf 4-40} & $\leq$ 40 & 4-45  & 4-45     \\
   Test Sentence Length Average    & 16.8     & 22.3 & $\approx$ 22 & 17.1     & 17.1     \\ \hline
   Labeled precision              & 88.1\%   & 84.5\%   & 86.3\%   & 82.5\%   & 89.8\%   \\
   Labeled    recall              & 87.6\%   & 84.0\%   & 85.8\%   & 81.6\%   & 89.6\%   \\
   Crossings per sentence          & 0.63     & 1.33     & 1.14     & 1.87     & 1.02     \\
   Sent.\ with 0 crossings         & 69.8\%   & 55.4\%   & 59.9\%   & 35.7\%   & 56.3\%   \\
   Sent.\ with $\leq$ 2 crossings  & 92.1\%   & 80.2\%   & 83.6\%   & 66.9\%   & 84.9\%   \\ \hline \hline
\end{tabular}\\[2mm]
\end{center}
\caption{Comparing {\sc Contex} with Magerman's SPATTER and Collins' BLD, all trained and
         tested on sentences from the Wall Street Journal.}
\label{fig-spatter-bld-contex-comparison}
\end{table}

The results in table~\ref{fig-spatter-bld-contex-comparison} have to be interpreted cautiously 
since they are not based on the exact same sentences and detail of bracketing.
Due to lexical restrictions, our average sentence length (17.1) is below the one
used in SPATTER and BLD for a similar comparable sentence length range.
While the Penn Treebank leaves many phrases such as ``the New York Stock Exchange'' 
without internal structure, our system performs a complete bracketing, 
thereby increasing the risk of crossing brackets.

The labeled precision and recall for {\sc Contex} (trained on 256 sentences) surpass the rates 
for Spatter and BLD, even when compared to Spatter limited on sentences up to 25 words. This is
partly due to the finer granularity of our parse trees, because precision and recall rates tend
to be higher within the core\footnote{By {\it core} noun phrases, we here mean those noun
phrases that the Penn Treebank does not sub-structure any further.} noun phrases.
With respect to the three crossing criteria, {\sc Contex} fares better when compared with longer
sentences, and worse when compared with the shorter sentences. While the average sentence
length for the shorter sentences (16.8) is only slightly shorter then our average of 17.1,
the finer granularity that we use is now a disadvantage, because additional crossings are
possible within core noun phrases.

However, the results are very encouraging, in particular since we can still expect significant 
improvement by increasing the number of our training sentences.

          %\chapter{Related Work} \label{ch-related}
\chapter{Future Work}
\label{ch-future}

The ways to extend our system are numerous and promising. 
Future work could further scale up the system, deal with
incomplete knowledge, move beyond `English to German', and improve the run time of the system.

\section{Scaling Up Further}
The analysis of the evaluation results revealed that the grades for translations that started
with a correct parse are only very moderately better than those for a fully automated translation.
As a consequence, the largest potential for improvement in translation quality lies in the stages 
beyond parsing.
An investigation of the type of translation errors that occur when the system starts on a 
correct parse shows
that it is clearly the generation module that is most responsible.
Mistakes included wrong case assignment in prepositional phrases governed by prepositions
that can assign both the dative and accusative case, and incorrect word order within a participle phrase
and its place in the superior phrase. There are many other types of errors and it seems 
that the generation exhibits the need for modest fine-tuning in a multitude of places, 
rather than the resolution of a few principle problems.
An example of a `small', yet quite damaging shortcoming was the translation of mixed fractions,
as often used in stock market notations. The parser correctly analyzed the four tokens `3', `1', `/', and
`4' as the number representing $3 + \frac{1}{4}$, which it stored as the value of the number.
In generation, this mixed fraction was however printed as `13/4', the default format of
the programming language used in the implementation (Lisp). Even though $3 \frac{1}{4} = \frac{13}{4}$,
most evaluators counted the unified fraction as a very serious semantic error.
One extra line of code can solve this problem, but it becomes clear that generation has
to consider a host of smaller issues. Fortunately, the resolutions of the various problems
seem to be relatively independent. Overall, the shortcomings of the generation module are actually
quite understandable, given that it was developed based on only 48 sentences.

Considering that the parser was trained on 256 sentences, it already exhibits an
amazing degree of robustness. The learning `curve' of parsing quality, based on the number of 
training sentences, as shown in table~\ref{fig-pex_training_size}, strongly suggests that
more training examples will most likely still significantly improve parsing quality. This
estimate is corroborated by the observation that many parse errors are due to obviously
novel syntactic constructions that have not occurred in the training set.

\subsection{New Domains}
\label{sec-new-domains}
\begin{table}[htbp]
\begin{center}
\begin{tabular}{|p{13cm}|} \hline
\\
Parse tree for ``{\it On February 27 I I in school.}'':
\begin{verbatim}
 "On February 27 I I in school.":
    synt:    S-SYNT-ELEM
    class:   I-EN-THING
    subs:
    (CONC)  "On February 27 I I":
        synt/class/roles: S-PP/C-AT-TIME/(CONC)
        forms: unrestricted
        (PRED)  "On"
        (PRED-COMPL) "February 27":
            (PRED) "February 27":
                (MOD)  "February"
                (PRED)  "27"
        (PRED-COMPL) "I I":
            (MOD) "I":
                (PRED)  "I"
            (PRED)  "I"
    (CONC)  "in school.":
        synt/class/roles: S-PP/C-IN-PP/(CONC)
        forms: unrestricted
        (PRED)  "in"
        (PRED-COMPL) "school":
            (PRED)  "school"
        (DUMMY)  "."
\end{verbatim}\\ \\
Subsequent translation to German, based on above parse tree:
\begin{verbatim}
 "Am 27. Februar ich ich in der Schule."
\end{verbatim}\\ \hline
\end{tabular}
\end{center}

\caption{Parse tree and subsequent translation of an ill-formed sentence.}
The sentence is a `noisy' variation of ``On February 27 I was in school.'',
both based on a German sentence pair from \cite{wermter:jair97}.
As usual, the parse tree is printed with decreasing detail at lower levels.\\
The parse tree is somewhat pathological
(see section~\ref{sec-guards} for an explanation of {\it CONC} etc.),
but adequately preserves the ``healthy'' parts of the input sentence.
The German translation (which is based on the parse tree) precisely reflects the English
original. It is encouraging to find this robustness even though {\sc Contex} was never
trained on any ill-formed sentences.
\label{fig-ill-formed-sentence}
\end{table}

Another important avenue of future work is the extension of the current Wall Street Journal 
domain to others,
in particular to text based on spoken language, which often includes quite peculiar 
or even technically ill-formed constructions.

We have used our parser to process a few sentences that other authors have used to 
make a point about their systems and found our system to be quite competitive. Consider for
example the parse tree and translation of the ill-formed sentence shown in
table~\ref{fig-ill-formed-sentence}.
Even though such anecdotal evidence already suggests a degree of robustness in parsing
when moving to sentences that are ill-formed or from other domains, adding additional examples 
from other domains might be quite beneficial, in particular when text other than 
well-formed complete sentences is used.
Along with more examples, we might also add more features, particularly when moving to a 
new domain. We expect the number of `specialized' features to be relatively limited though.
In our current feature set, the most business oriented features are
{\it (classp of i-en-currency-unit of -1 at m-boolean)}\footnote{``Is the top element on the stack 
a currency unit?''} and
{\it (classp of i-en-monetarily-quantifiable-abstract of} -n {\it at m-boolean)}\footnote{``Is
the {\it nth} element on the stack a monetarily quantifiable abstract?''} for  n = 1, 2, 3.
It might be beneficial to have features indicating the domain, which might improve especially
the quality of choice of alternative transfer concepts. The domain can probably be identified
automatically by matching the vocabulary of the text to be translated with a typical domain
vocabulary profile. Eventually, subcategorization tables might be added for nouns or even
prepositions.

\subsection{Lexical Expansion}
\label{sec-lexical-expansion}
Even though the amount of work to increase the coverage of lexicon and KB is quite small
per word, the task to increase the currently slightly over 3000 entries of the English
lexicon by one order of magnitude for an advanced vocabulary or two orders of magnitude
to match Webster's Encyclopedic Unabridged Dictionary would have good use for further 
automation.

\citeA{riloff:ai96} describes research in dictionary construction automation. The techniques
described for information extraction applications can be adapted for our system.
A key idea of successful information mining is to look for patterns that are typical for
certain word classes and then have a supervisor confirm or reject likely candidates.

The bilingual dictionary is a prime candidate for automated acquisition. Bilingual
corpora like the Canadian Hansards (parliamentary proceedings) or articles from the
{\it Scientific American} along with their German translations in {\it Spektrum der
Wissenschaft} can be exploited to identify co-occurrences of sentences
\cite{gale:acl91}, words and expressions. \citeA{kay:compling93} even have been able
to extract a word alignment table containing German/English word pairs based solely on
the internal evidence of the bilingual corpus. As for the monolingual lexicon,
alignment programs could propose likely candidates to a supervisor for approval or
rejection. 

The Champollion program \cite{smadja:compling96} for example,
using aligned text from the Canadian Hansards and given a multi-word English collocation,
can identify the equivalent collocation in French with a precision of up to 78\%.
Champollion is limited to a statistical analysis of sets of words. The use of parsers and
moderate background knowledge can certainly further improve the accuracy.
\citeA{wu:mt95} use similar statistical techniques to automatically extract an English-Chinese
dictionary with 6429 English entries and report precision rates from 86\% to 96\%.

Finally, online dictionaries and other resources like lists of proper names can be
used for lexical expansion. Even if the knowledge encoded in these resources is not
deep enough, it is probably better than having no entry at all for a word or expression.
The following section describes how such incomplete knowledge might be handled.

\subsection{Considering Context Beyond the Sentence Boundary}
An extension of the feature-encoded context beyond the sentence boundary is simpler
than it might seem. The parse structure would just need additional slots for previous
and following sentences and the feature language could easily be extended to access
these slots. This might be useful for anaphora resolution, which sometimes depends on
previous sentences, and for a better identification of the general type of text, e.g.\
business, science, sports, weather. Such topic features could then be used to make 
better decisions for ambiguous transfer. The relative ease of such an extension shows
how useful it is that {\sc Contex} very naturally integrates features of very 
different types and that it can automatically recompute the parse action examples and 
the decision structure to incorporate new features.

\section{Incomplete Knowledge}
We have used a lexically limited corpus for our research. While the restriction to 3000
words still allowed sentences with a very diverse cross-section of text, representing
the enormous complexity of natural language, the problem of incomplete knowledge, like lexically
unknown words, and corresponding missing entries in the subcategorization tables and the knowledge
base, has been eliminated. For truly free text, any assumption of complete knowledge will
always be unrealistic. We believe however that our paradigm of learning from examples lends 
itself particularly well for handling incomplete knowledge.

\subsection{Unknown Words}
Webster's Encyclopedic Unabridged Dictionary of the English Language \cite{webster:dict94}
contains over 250,000 lexical entries. Nevertheless, many technical terms like ``{\it to
disambiguate}'' as well as most proper names are not included.
Even if unknown words are unavoidable in free text, we nevertheless want to proceed with
parsing, so that, for example, a sentence could still be translated, even if some individual
words have to be left in the source language.
To do proper parsing, we need to assign a part of speech to the unknown word. Since this is
just another classification problem, machine learning seems to be the natural solution again.
The context of the current parse state before shifting\footnote{Recall that the shift-in operation
includes an argument designating the part of speech; a shift-in operation therefore performs
the so-called tagging.} in the unknown word can provide
features, just like for `normal' parsing itself. Additional features that access word
endings or try to decompose a word into suffixes and known words will probably be useful.
The ending ``{\it -ated\/}'' as in ``{\it disambiguated\/}'' for example is a typical verbal ending
(past tense or past participle), and ``{\it unrealistic}'' could be identified as an adjective
composed of the prefix ``{\it un-}'' and the lexically known adjective ``{\it realistic}''. The
syntactic context will certainly help as well to identify the proper part of speech of the
unknown word. Endings might also be used to identify forms, such as tense for verbs and number 
for nouns. \citeA{weischedel:compling93} used features like these in a probabilistic model and
were able to achieve a tagging accuracy rate of 85\% for unknown words.

Examples for unknown word classification can be obtained by temporarily disabling some entries
in the lexicon and thus rendering previously known words in already acquired parse action
examples `unknown'. Disabling entries that were added to the lexicon more recently is probably
better than disabling lexicon entries at random, because the more recent lexicon additions
typically present a better cross-section of words that have not been covered yet.
While the ten most frequently occurring words of the full WSJ corpus (the, of, to, a, in, and, 
that, for, is, said) contain mostly words from closed word classes, i.e.\ word classes with
a limited, often fairly small number of members, like for example articles, prepositions and
conjunctions, later additions will at some point exclusively belong to open word classes 
like nouns (in particular proper names), verbs and adjectives. The 10 least frequent words
of the 3000 most frequent WSJ words for example are {\it sluggish, repeatedly, Lewis, Caesars,
quotas, Dominion, Lake, F, volatile} and {\it Wells}. Currently, the system just assumes that
any unknown word is a noun. To obtain these additional examples, we let our parser run
over the logged training sentences with part of the lexicon disabled and collect only
those examples where some feature(s) in the current feature set accesses the unknown word; all 
other examples have already been collected during the regular automatic full parse example
generation process (as described in section~\ref{sec-training-the-parser}).

\subsection{Incomplete Subcategorization Table}
Subcategorization entries can be missing for some verbs that are in the lexicon and
presumably just about all verbs that are not in the lexicon. In analogy to lexically unknown words,
we can temporarily disable some subcategorization table entries and gain examples that
lack any match in the subcategorization table. It is however important to distinguish
this case from a situation where a match has been found, but no role is available
for a specific potential phrase component.
Using techniques described in \cite{manning:acl93}, we could at least partially
automate the acquisition of subcategorization entries from corpora.

\subsection{Incomplete Knowledge Base}
When unknown words get assigned a part of speech, they also need to be assigned a 
semantic concept from the knowledge base. Unless the context can provide some more
specific semantic restrictions, this will have to be one of the very
generic concepts, e.g.\ `I-EADJ-ADJECTIVE' (some adjective), `I-EV-PROCESS-STATE'
(some process or state, for verbs), `I-EN-THING' (some thing, for nouns).
The entry for the unknown word should also be marked as semantically un(der)specified,
to distinguish cases of lexically covered and uncovered words. Consider for example
a test in which we check whether or not an entity can be an agent. While a negative
result for the lexically covered word `table' should, using the closed world assumption,
be treated at face value, an equally negative test result for an unknown word should probably
yield a special value like `{\it don't know}'. A feature value `{\it don't know}' can
then be used like any other value in the decision structure building process.

\section{Moving Beyond `English to German'}
In terms of functionality, our system could be expanded in several ways. Besides adding
the capability to also translate from German to English, other languages could be added
and our system might be used for other tasks such as a grammar checking.

\subsection{German to English}
For a translation in the opposite direction, much of the current system could be reused.

The only essential component that is totally missing for translation from German to
English is the English generation module. It would not be very hard and time consuming
to add such a module, because the basic program structure
would be exactly the same. Starting with the duplicate of the German generation module,
we would have to make a number of modifications, particularly with respect to the order
of components in sentences and noun phrases; in the morphological propagation part, most
of the changes would be deletions, since English is morphologically much simpler than
German.

The transfer component is bidirectional and basically does not require any extensions.
The only exception is the set of the ambiguous transfer examples, because transfer ambiguity
is not symmetric. The current English to German ambiguous transfer example file with 
only 62 lines is short enough to justify our estimate that the corresponding German to English
example file could be developed in a relatively short period of time.

A German parser already exists. It was trained to parse the German entries of the bilingual
dictionary. Since the phrases in dictionary entries typically exhibit significantly less 
ambiguity and variation, fewer training examples were needed. In fact, the log file for the
German examples is about 5\% of the size of the log file containing the parse action sequences
for the 272 Wall Street Journal sentences. For parsing quality comparable to the current
English parser, the number of training examples would have to be increased to a similar
level as for English. This would most likely be the most labor intensive aspect of an 
extension to translate from German to English.

Furthermore the German lexicon currently has fewer entries than the English lexicon
(1039 as opposed to 3015) and there aren't any subcategorization tables for German yet.
The morphological processing is fully bidirectional and thus does not need any further work.

\subsection{Adding More Languages}
For each additional language, a new morphological processor, a lexicon, a parse example 
training set, and a generation module would be required. Our dictionaries are {\bf bi}lingual,
but we don't necessarily have to develop a full matrix of bilingual dictionaries from scratch.
Obviously, we need at least a set of dictionaries that link the languages that we want
to translate to and from. For English/German/French/Spanish, we might for example have an
German/English, an English/French, and a French/Spanish dictionary. When translating,
we can then parse the source text using the source language parser, then perform a {\it series}
of tree transfers, based on the dictionary `graph', and then compute the final output using
the target language generation module. This is far better than making full translations to
intermediate languages, because we save both generation and parsing for each intermediate 
language, steps that can introduce a host of errors.

A more efficient approach though might be the `compilation' of a dictionary for languages A and B
and one for languages B and C into a dictionary for languages A and C. For each pair of matching entry
pairs $A_{i}$/$B_{j}$ and $B_{j}$/$C_{k}$, one could create an entry pair $A_{i}$/$C_{k}$ in the
compiled dictionary. For example, based on the entries 
``EINEN\_1 Kompromi{\ss} erreichen''/``to reach SOME\_1 compromise'' and 
``to reach SOME\_1 compromise''/``aboutir {\`{a}} UN\_1 compromis'', we would add
``EINEN\_1 Kompromi{\ss} erreichen''/``aboutir {\`{a}} UN\_1 compromis'' to the German/French
dictionary.
This can lead to overgeneration for ambiguous entries. Given the dictionary pairs 
``wissen''/``to know'', ``kennen''/``to know'', ``to know''/``savoir'', and 
``to know''/``conna\^{\i}tre'', we would get four word pairs for German/French, even
though only ``wissen''/``savoir'' and ``kennen''/``conna\^{\i}tre'' are actually valid.
These spurious dictionary entries can be deleted manually, or, based on ambiguous transfer
decision structures from English to German and English to French, automatically.

Finally, the corresponding ambiguous transfer examples should be compiled into direct
transfer examples, using the ambiguous transfer decision structures, and combining the
sets of relevant features. To minimize these ambiguity mismatches, it is apparently
wise to choose the original dictionary language pairs giving preference to related 
languages. E.g.\, given that Danish and Norwegian are very close to each other,
Danish/Norwegian and Norwegian/Japanese are better than Danish/Japanese
and Japanese/Norwegian when compiling a dictionary for the remaining language pair.

\subsection{Grammar Checking}
For any type of automatic grammar checker, it is hard to distinguish between the 
insufficiencies of the grammar checker and the text to be checked. For our system,
the robustness of our parser, a benefit when it comes to machine translation, can present
an additional disadvantage when it comes to grammar checking. In a sentence like
``A dogs eats a bone.'' for example, our parser would probably not even check the number 
agreement within the subject or between subject and verb, because it does not help the 
system in making parse decisions. Both the general and the specific problem can be
handled by actively training the system to find grammatical errors by also providing 
grammatically incorrect examples along with the error classification.

Two variants on how to do this come to mind:
\begin{enumerate}
  \item The sentence is first parsed as usual and then the `error examples' classify nodes
        of the parse tree. This resembles the decision making in transfer, which also
        operates on parse tree nodes.
  \item `Error examples' operate on parse states representing partially parsed sentences. 
        This resembles the decision making in parsing, which also operates on parse states.
\end{enumerate}
It seems that the first alternative is conceptually somewhat simpler, because it might
be hard to link an error to a specific parse state, but the second 
alternative could possibly interact better with the actual parsing, by not only signaling
a grammar problem, but also passing this information on to the `running' parser, which
might in some `obvious' cases then automatically correct the error and thereby avoid
follow-up problems.

\section{Speed-Ups}

Due to the deterministic nature of the parser, which keeps its time complexity {\bf linear}
with respect to the length of the sentence, the system is already very fast. 
On the hardware we use (a 70 Mhz SparcStation 5 with 32 MB of memory),
our current system needs about 2.4 sec in total CPU time (2.9 sec real time) to translate 
an average sentence with 17 words from English to German 
(incl.\ segmentation, morphological analysis, parsing, transfer and generation). 
To train our parser on 11,822 examples, acquired under supervision from 272 sentences
from the Wall Street Journal, the system needs 20 minutes in total CPU time (4.3 hours of real 
time).

We estimate\footnote{This estimate is based on previous experience of the author of this 
dissertation from work in the Knowledge Based Natural Language (KBNL) group at
the Microelectronics and Computer Technology Corporation (MCC) in Austin, Texas,
where the author reimplemented major parts of a Lisp parser in C++, achieving a speed-up of
one to two orders of magnitude, and speeding up the JUMAN Japanese language segmenter
\cite{matsumoto:tr94} by a factor of 60, mostly by caching information that was used repeatedly.}
that the system can be sped up by a factor of 10 to 100 by measures including the following steps:
\begin{itemize}
  \item reimplementation of the Lisp system in C++
  \item caching of information that is used repeatedly
  \item a more compressed format of examples
  \item adding a new parse mode that would allow parse actions to be destructive, i.e.\ don't force
        the parser to preserve all intermediate parse states; during supervised example acquisition
        it can be useful to `backstep' to a previous parse state, but in actual applications 
        of parsing such `backstepping' is not needed anyway.
\end{itemize}
These speed-ups would result in virtually instantaneous translations of sentences and short texts.
           %\chapter{Future Work} \label{ch-future}
\chapter{Conclusions}
\label{ch-conclusions}
 
Guided by the goal to advance the technology of robust and efficient parsers
and machine translation systems for free text,
we try to bridge the gap between the typically hard-to-scale hand-crafted
approach and the typically large-scale but context-poor probabilistic approach.\\

\noindent Using 
\begin{itemize}
  \item a rich and unified context with 205 features,
  \item a complex parse action language that allows integrated 
        part of speech tagging and syntactic and semantic processing,
  \item a sophisticated decision structure that generalizes traditional decision trees and lists,
  \item a balanced use of micromodular background knowledge and machine learning,
  \item a modest number of interactively acquired examples from the Wall Street Journal,
\end{itemize}
\noindent our system {\sc Contex}
\begin{itemize}
  \item computes parse trees and translations fast, because it uses a deterministic single-pass parser,
  \item shows good robustness when encountering novel constructions,
  \item produces good parsing results comparable to those of the leading probabilistic methods, and
  \item delivers competitive results for machine translations.
\end{itemize}

Finally, given that so far we trained our parser on examples from only 256 sentences,
and developed our generation module based on only 48 sentences,
we can still expect to improve our results very significantly, because
the learning curve hasn't flattened out yet and
adding substantially more examples is still very feasible. Besides scaling up further,
extensions in other directions, such as more languages and grammar checking,
are numerous and very promising.\\

\newpage
\begin{tabbing}
\indent \= {\it Question}: \= So when will the general purpose machine translation system\\
        \>                 \> with 100\% accuracy be ready? \\
        \> {\it Short answer:} Never.\\
\end{tabbing}

\noindent As many scientists have already pointed out, natural language processing in
general and machine learning in particular are ``AI-complete'',
i.e.\ NLP and MT can not be fully solved until artificial intelligence with all its other
subdisciplines like knowledge representation and problem solving have been solved as well ---
until the computer has become omniscient. We recommend that you don't hold your breath for that.

The good news is that the relation between knowledge and accuracy is not linear. We believe
that it is safe to assume that with less than ``1\% knowledge'' we can achieve more than 
``99\% accuracy'' and that even within that fraction of one percent, a minute sub-fraction will
yield a ``90\% accuracy''. 
We believe that this strongly unbalanced nature of knowledge suggests a hybrid approach to
machine translation. Since the ``core knowledge'' is relatively small yet critical for accuracy,
it seems reasonable to acquire it under very careful human supervision or even provide it
manually. The more we move away from the ``core knowledge'', the lower the rate of return will be.
With increasingly larger quantities of additional knowledge necessary to reduce the error rate 
by a specific factor, it becomes more and more acceptable to allow rough approximations of
knowledge. Probabilistic methods that can operate on large corpora of texts with little or no
human intervention at all become more and more attractive. Even if the knowledge they build
is not as ``perfect'' as if built under careful human supervision, it will still be useful, given
that very large scale manual knowledge support becomes prohibitive and assuming
that the final accuracy is still better than without such machine generated knowledge.
The natural combination of more manually acquired ``core knowledge'' and more automatically
acquired vast ``outer knowledge'' manifests itself in a tendency of significant MT systems to
move towards more hybrid solutions as the systems are developed. 
While for example the originally extremely probabilistic Candide system 
\cite{brown:compling90,berger:arpa94} has been augmented by a number of manual heuristics,
the originally fairly manual-knowledge-based {\sc Pangloss} Mark I/II/III system \cite{pangloss:amta94}
has incorporated less dogmatic example-based and transfer-based approaches.

Following the same trend, our system {\sc Contex}, when scaled up further, will have to include
ever more automatically acquired knowledge, but
we believe that {\sc Contex} already incorporates many characteristics of the successful 
hybrid system of the future: its machine learning can provide critical robustness, 
``real corpus''-based development can provide the necessary focus on critical knowledge, 
and a unified context of diverse features easily integrates knowledge from different sources.

           %\chapter{Conclusions} \label{ch-conclusions}
\appendix
\chapter{WSJ Corpus}
\label{app-corpus}

The WSJ corpus used in this work is a subset of the sentences from Wall Street Journal
articles from 1987, as provided on the ACL data-disc.
The WSJ corpus consists of the 272 first sentences that are fully covered by the 3000 most
frequently occurring words of the entire corpus.
These sentences are also available on the WWW at http://www.cs.utexas.edu/users/ulf/diss/wsj\_corpus272.html .\\

\noindent
{\bf (WSJ 0)} Industry sources put the value of the proposed acquisition at more than \$100 million.\\
{\bf (WSJ 1)} A seat on the Chicago Board Options Exchange was sold for \$340,000, up \$5,000 from the previous sale March 11.\\
{\bf (WSJ 2)} Of the total, 15 million shares will be sold by the company and the rest by a shareholder.\\
{\bf (WSJ 3)} The securities are convertible at a rate of \$27.61 of debentures for each common share.\\
{\bf (WSJ 4)} The utility said it expects an additional \$109 million (Canadian) in 1987 revenue from the rate increase.\\
{\bf (WSJ 5)} It was cause to wonder.\\
{\bf (WSJ 6)} The gold mines rose 12 points to 362.4.\\
{\bf (WSJ 7)} The airline said that it would report a loss for the first quarter but that it would be "substantially less" than the \$118.4 million deficit it posted in the first quarter of 1986.\\
{\bf (WSJ 8)} In 1986, while general inflation was only 2\%, hospital expenses grew 8\%.\\
{\bf (WSJ 9)} Banks will be offered a slightly higher interest-rate margin if they accept the notes than if they take a cash payment.\\
{\bf (WSJ 10)} "It's a real high," said one.\\
{\bf (WSJ 11)} "You can't get out of a falling market if you're in an index fund."\\
{\bf (WSJ 12)} We're part of our own problem.\\
{\bf (WSJ 13)} The average increase is about 1.8\%, Toyota said.\\
{\bf (WSJ 14)} This gave Japanese manufacturers a huge cost advantage over U.S.\ companies.\\
{\bf (WSJ 15)} He said that this was the first he had heard about any such deal.\\
{\bf (WSJ 16)} No one is particularly happy about the situation.\\
{\bf (WSJ 17)} South Korea posted a surplus on its current account of \$419 million in February, in contrast to a deficit of \$112 million a year earlier, the government said.\\
{\bf (WSJ 18)} Seats currently are quoted at \$330,000 bid, \$345,000 asked.\\
{\bf (WSJ 19)} Sales are expected to increase about 10\% to \$40 million from \$36.6 million in 1986.\\
{\bf (WSJ 20)} In over-the-counter trading Friday, On-Line common closed at \$22.25, down 50 cents.\\
{\bf (WSJ 21)} The issue is one of Germany's biggest.\\
{\bf (WSJ 22)} He said no.\\
{\bf (WSJ 23)} The stock exchange index rose 2.39 to 1860.70.\\
{\bf (WSJ 24)} For all of 1986, Pan Am reported a loss of \$462.8 million, compared with 1985 net income of \$51.8 million, or 45 cents a share.\\
{\bf (WSJ 25)} The for-profit hospital, in contrast, has a powerful social voice in the form of equity markets.\\
{\bf (WSJ 26)} Major U.S.\ accounting firms are split on how to value the notes.\\
{\bf (WSJ 27)} "It's like you're king of the hill."\\
{\bf (WSJ 28)} That makes index funds equal to 15\% of institutional equity holdings, or about 7\% of the entire stock market.\\
{\bf (WSJ 29)} American industry can't just keep running to government for relief.\\
{\bf (WSJ 30)} Toyota said the prices of 16 1987 models weren't increased.\\
{\bf (WSJ 31)} But by the end of January, Japanese compensation costs had risen to 79\% of the U.S.\ level, according to estimates by the Bureau of Labor Statistics.\\
{\bf (WSJ 32)} The statement reported total assets of \$40 million, net worth of \$39 million and annual income of \$908,000.\\
{\bf (WSJ 33)} There's nothing he can say.\\
{\bf (WSJ 34)} Industrial production in Italy declined 3.4\% in January from a year earlier, the government said.\\
{\bf (WSJ 35)} The record price of \$342,000 for a full membership on the exchange was set Feb.\ 27.\\
{\bf (WSJ 36)} The company said it doesn't expect the ruling to have a major impact on earnings because it had already set aside about \$14 million in reserves to cover the judgment and reached an agreement for a bank loan to pay the balance.\\
{\bf (WSJ 37)} The previous terms weren't ever disclosed.\\
{\bf (WSJ 38)} That would have valued the holding at between \$1.49 billion and \$1.71 billion.\\
{\bf (WSJ 39)} Others in Washington don't much care where new revenues come from.\\
{\bf (WSJ 40)} Volume was about 1.5 billion shares, up from 1.4 billion Thursday.\\
{\bf (WSJ 41)} Revenue declined 13\% to \$3.04 billion from \$3.47 billion.\\
{\bf (WSJ 42)} They represent only 17\% of U.S.\ hospitals, and some of them are run on a for-profit basis.\\
{\bf (WSJ 43)} But he declined to comment on the issue of the notes.\\
{\bf (WSJ 44)} The S\&P-500 stock index finished up 4.09 at 298.17 and the New York Stock Exchange composite index gained 2.09 to 169.37.\\
{\bf (WSJ 45)} By 10 a.m., he was done.\\
{\bf (WSJ 46)} It's a chance to get some improvement here.\\
{\bf (WSJ 47)} "What would it take to get people to pick this up?"\\
{\bf (WSJ 48)} Largely because of the falling dollar, West German labor costs rose to 120\% of those for U.S.\ production workers from 75\% in 1985.\\
{\bf (WSJ 49)} In high school, he was a member of the speech team.\\
{\bf (WSJ 50)} But the immediate concern is the short-term credits.\\
{\bf (WSJ 51)} Canadian manufacturers' new orders fell to \$20.80 billion (Canadian) in January, down 4\% from December's \$21.67 billion on a seasonally adjusted basis, Statistics Canada, a federal agency, said.\\
{\bf (WSJ 52)} The Federal Farm Credit Banks Funding Corp.\ plans to offer \$1.7 billion of bonds Thursday.\\
{\bf (WSJ 53)} The St.\ Louis-based bank holding company previously traded on the American Stock Exchange.\\
{\bf (WSJ 54)} The transaction is expected to be completed by May 1.\\
{\bf (WSJ 55)} A successor for him hasn't been named.\\
{\bf (WSJ 56)} The president's news conference was a much-needed step in that direction.\\
{\bf (WSJ 57)} The Tokyo exchange was closed Saturday as part of its regular holiday schedule.\\
{\bf (WSJ 58)} Pan Am said its full-year results were hurt by foreign currency exchange losses of \$46.8 million, primarily related to Japanese yen debt, compared with \$11.1 million in 1985.\\
{\bf (WSJ 59)} The American hospital sector spent \$181 billion in 1986.\\
{\bf (WSJ 60)} But the company inquiry, which began a few months ago, changed everything.\\
{\bf (WSJ 61)} Texas Instruments rose 3 1/4 to 175 1/2.\\
{\bf (WSJ 62)} If futures fell more than 0.20 point below the stocks, he would buy futures and sell stocks instead.\\
{\bf (WSJ 63)} "I'm willing to work for a foreign company," says the father of two.\\
{\bf (WSJ 64)} It estimated that sales totaled \$217 million, compared with the year-earlier \$257 million.\\
{\bf (WSJ 65)} With the falling dollar, U.S.\ manufacturers are in a pretty good position to compete on world markets.\\
{\bf (WSJ 66)} You go on to Bank B and get another loan, using some of Bank B's proceeds to pay back some of your debt to Bank A.\\
{\bf (WSJ 67)} The critical point is to know when Brazil will produce an economic policy to generate foreign exchange to pay its interest.\\
{\bf (WSJ 68)} Manufacturers' shipments followed the same trend, falling 1.5\% in January to \$21.08 billion, after a 2.8\% increase the previous month.\\
{\bf (WSJ 69)} The offerings will be made through the corporation and a nationwide group of securities dealers and dealer banks.\\
{\bf (WSJ 70)} The real estate services company formerly traded over the counter.\\
{\bf (WSJ 71)} The shares are convertible at a rate of \$9.50 of preferred for each common share.\\
{\bf (WSJ 72)} He continues as president of the company's corporate division.\\
{\bf (WSJ 73)} In tests it has worked for many heart-attack victims.\\
{\bf (WSJ 74)} The percentage change is since year-end.\\
{\bf (WSJ 75)} He said Pan Am currently has "in excess of \$150 million" in cash.\\
{\bf (WSJ 76)} These revenues could then have been used for many purposes, such as funding those people without medical insurance.\\
{\bf (WSJ 77)} The case continues in U.S.\ Bankruptcy Court in St.\ Louis.\\
{\bf (WSJ 78)} American Express fell 1 1/2 to 77 1/4 on more than 2.1 million shares.\\
{\bf (WSJ 79)} His typical trade involved \$30 million.\\
{\bf (WSJ 80)} Other voters reflect the divisions between their two senators.\\
{\bf (WSJ 81)} However, the indicated fourth-quarter net loss is about \$10 million.\\
{\bf (WSJ 82)} In fact, unit labor costs have risen less in the U.S.\ than in other industrial countries.\\
{\bf (WSJ 83)} First Interstate Bank of Denver -- \$5 million.\\
{\bf (WSJ 84)} "He knows when to act and when to do nothing."\\
{\bf (WSJ 85)} The company said it isn't aware of any takeover interest.\\
{\bf (WSJ 86)} Earnings included a \$615,000 write-down of an investment.\\
{\bf (WSJ 87)} The \$1.4375 Series A preferred stock is convertible at the rate of 1.0593 common shares for each preferred share.\\
{\bf (WSJ 88)} In fact, no investigation was completed, but one is under way.\\
{\bf (WSJ 89)} His approach never seems to fail.\\
{\bf (WSJ 90)} It should be approved.\\
{\bf (WSJ 91)} The card will cost \$15 a year, in addition to the \$45 that American Express charges for its regular card.\\
{\bf (WSJ 92)} At the end of 1986, long-term debt totaled \$830 million.\\
{\bf (WSJ 93)} Construction is to begin immediately, with completion expected in late 1989.\\
{\bf (WSJ 94)} Whatever the case, the three executives left within two days of one another.\\
{\bf (WSJ 95)} The company filed an offering of 15.7 million shares.\\
{\bf (WSJ 96)} You have to look for the hand at the other end.\\
{\bf (WSJ 97)} But other voters think the issue is more complex.\\
{\bf (WSJ 98)} The credit of about \$44 million made net income \$38.2 million, or \$5.26 a share.\\
{\bf (WSJ 99)} A look at Japan shows why.\\
{\bf (WSJ 100)} That, too, was just fine with Mr.\ Clark, because Hutton also had an office in the building.\\
{\bf (WSJ 101)} "They've clearly put the company up for sale."\\
{\bf (WSJ 102)} Terms weren't disclosed, but the industry sources said the price was about \$2.5 million.\\
{\bf (WSJ 103)} The company expects the sale to be completed later this month, but said it still requires a definitive agreement.\\
{\bf (WSJ 104)} It has 5,225,000 units outstanding.\\
{\bf (WSJ 105)} In fiscal 1986, it earned a record \$9.9 million, or 89 cents a share, on record revenue of \$138.5 million.\\
{\bf (WSJ 106)} Financial regulators closed several thrifts and banks over the weekend, including a Texas savings and loan association with \$1.4 billion in assets.\\
{\bf (WSJ 107)} That was two years ago.\\
{\bf (WSJ 108)} But it will extend 13.5\% credit.\\
{\bf (WSJ 109)} Pan Am's unions, meanwhile, are trying to find another merger partner for the company.\\
{\bf (WSJ 110)} "We needed someone who would be in it for the long run."\\
{\bf (WSJ 111)} Has there been any improvement since then?\\
{\bf (WSJ 112)} Pan Am was unchanged at 4 3/8 in active trading.\\
{\bf (WSJ 113)} "It isn't an emergency situation where people have to trade at all costs," he says.\\
{\bf (WSJ 114)} "It could help steel, but it would hurt coal because they have to sell out of the country."\\
{\bf (WSJ 115)} The company currently has 10 million shares outstanding.\\
{\bf (WSJ 116)} But when the yen wages are converted to dollars to show the impact of exchange rates, compensation rose 47\% to the equivalent of \$9.50 in 1986 from \$6.45 in 1985.\\
{\bf (WSJ 117)} "Banks are highly competitive in trying to place good loans," he says.\\
{\bf (WSJ 118)} He declined to elaborate.\\
{\bf (WSJ 119)} The plant will be used by Texas Instruments' Defense Systems and Electronics Group to produce electronic equipment.\\
{\bf (WSJ 120)} Canada's seasonally adjusted consumer price index rose 0.3\% in February, Statistics Canada, a federal agency, said.\\
{\bf (WSJ 121)} It said then that it would make a final decision by late March.\\
{\bf (WSJ 122)} There is quite a difference.\\
{\bf (WSJ 123)} It had assets of \$115 million.\\
{\bf (WSJ 124)} Yet last year Congress approved \$900 million in grant military aid, the highest level since the end of the Vietnam War.\\
{\bf (WSJ 125)} American Express says only a limited number of existing customers will be offered the new card.\\
{\bf (WSJ 126)} They have made their own cost-cutting proposals, but the two sides haven't reached any agreements.\\
{\bf (WSJ 127)} In December, the company sold six radio stations for \$65.5 million.\\
{\bf (WSJ 128)} Right on target, for 1979 or 1981.\\
{\bf (WSJ 129)} A bankruptcy judge approved the company's \$100 million financing agreement, an important step in its bankruptcy reorganization.\\
{\bf (WSJ 130)} Growth in such investing was slow in the 1970s.\\
{\bf (WSJ 131)} The only thing all voters seem to agree on is that something must be done.\\
{\bf (WSJ 132)} It didn't give a reason for the sales.\\
{\bf (WSJ 133)} Although the dollar began to fall two years ago, U.S.\ trade performance is just beginning to turn around.\\
{\bf (WSJ 134)} The banking system, he says, "is built on trust."\\
{\bf (WSJ 135)} He suggested that the two executives' interests might not reflect shareholders' interests.\\
{\bf (WSJ 136)} But the company wouldn't elaborate.\\
{\bf (WSJ 137)} The February rise followed increases of 0.2\% in January and 0.4\% in December.\\
{\bf (WSJ 138)} It has 5.8 million units outstanding.\\
{\bf (WSJ 139)} For the fourth quarter, the insurance-brokerage company posted a loss of \$26.3 million, compared with a loss of \$87.4 million in the year-earlier quarter.\\
{\bf (WSJ 140)} First American will assume deposits of about \$37 million in 7,200 accounts.\\
{\bf (WSJ 141)} Of the administration's proposed \$20 billion foreign-affairs budget for 1988, \$15 billion would go to foreign aid.\\
{\bf (WSJ 142)} Banks also believe that the American Express estimates are too modest, and some fear a plastic rate war.\\
{\bf (WSJ 143)} We all had to take out second mortgages to help finance our investments in Stanley.\\
{\bf (WSJ 144)} Per-share earnings were adjusted to reflect a 2-for-1 stock split paid in January.\\
{\bf (WSJ 145)} "Reform" is not "revolution".\\
{\bf (WSJ 146)} The next few days may determine whether the small, closely held airline has finally run out of fuel.\\
{\bf (WSJ 147)} That's when they really decided to let it grow.\\
{\bf (WSJ 148)} "There's no guarantee for anything in global competition."\\
{\bf (WSJ 149)} Allegheny International makes consumer and industrial products.\\
{\bf (WSJ 150)} About \$184 million of debt is affected.\\
{\bf (WSJ 151)} And a person intent on bank fraud "is going to get it done."\\
{\bf (WSJ 152)} They did not give us any response to our offer.\\
{\bf (WSJ 153)} Terms of the contract haven't been disclosed.\\
{\bf (WSJ 154)} First Federal, which has about \$270 million in assets, currently has 566,100 common shares outstanding.\\
{\bf (WSJ 155)} Details about the proceeds weren't immediately available.\\
{\bf (WSJ 156)} Revenue rose 11\% to \$99.3 million from \$89.4 million.\\
{\bf (WSJ 157)} New City had assets of about \$21 million.\\
{\bf (WSJ 158)} Of that \$15 billion, 38\% is for development aid, up from the administration's request of 33\% last year.\\
{\bf (WSJ 159)} Card rates haven't fallen nearly as sharply as other interest rates since 1980.\\
{\bf (WSJ 160)} This came to a head during an early planning session when a production manager asked what the company policy was on a particular matter.\\
{\bf (WSJ 161)} Industry officials said any combination probably would be the first step toward a merger of all five New York exchanges.\\
{\bf (WSJ 162)} He had led her to believe it was worth more than \$1 million.\\
{\bf (WSJ 163)} Air Atlanta officials said the company may be able to get through the week with cash on hand.\\
{\bf (WSJ 164)} It's expensive.\\
{\bf (WSJ 165)} But it is likely that its economy won't expand this year at even the modest 2\% rate forecast only a few months ago -- and that it will actually decline in the current quarter.\\
{\bf (WSJ 166)} "This means that, to a significant degree, we're using up the inventory found prior to 1970 at an increasing rate," he said.\\
{\bf (WSJ 167)} S\&P cited "expectations of continued profit pressures" for the company.\\
{\bf (WSJ 168)} Mr.\ Lewis couldn't be reached for comment.\\
{\bf (WSJ 169)} The truck maker said the group agreed not to buy Clark voting stock for 10 years.\\
{\bf (WSJ 170)} Merrill Lynch also has an option to extend the agreement through February 1993.\\
{\bf (WSJ 171)} "We now have a relatively small presence there."\\
{\bf (WSJ 172)} As it happened, Far Eastern interest was limited.\\
{\bf (WSJ 173)} Latest results included \$24.2 million in losses from discontinued operations.\\
{\bf (WSJ 174)} Hardly any of the participants in the tax debate, however, are taking account of the positive revenue effects of lower tax rates.\\
{\bf (WSJ 175)} The total cost of those proposed issues to shareholders will be about the equivalent of \$897.3 million.\\
{\bf (WSJ 176)} In a Securities and Exchange Commission filing, the group including the New York investment company said it holds 555,057 shares, including 59,800 purchased Jan.\ 23 through Feb.\ 6 for \$14 to \$14.37 each.\\
{\bf (WSJ 177)} "I don't know," I said.\\
{\bf (WSJ 178)} Moreover, any merger would take time; industry executives suggested it would take more than two years to work out.\\
{\bf (WSJ 179)} Then the rally will continue, he said.\\
{\bf (WSJ 180)} Despite the airline's problems, company officials say they believe Air Atlanta will get the capital needed to put it on its feet.\\
{\bf (WSJ 181)} "It doesn't take much to realize that we have a very big problem."\\
{\bf (WSJ 182)} Germany's inflation-adjusted growth of 2.4\% last year was its worst showing since the economy climbed out of recession in 1983.\\
{\bf (WSJ 183)} The coming period is a very critical economic period.\\
{\bf (WSJ 184)} Data General couldn't be reached for comment.\\
{\bf (WSJ 185)} Mr.\ Smith, who is a former partner of Bear, Stearns \& Co., didn't return phone calls.\\
{\bf (WSJ 186)} In composite trading on the Big Board, Clark common shares closed at \$25, down \$1.875.\\
{\bf (WSJ 187)} The suit seeks class-action status on behalf of other company shareholders.\\
{\bf (WSJ 188)} Ohio Power Co., a unit of American Electric Power Co., said it will redeem \$34.4 million of its first mortgage bonds June 1.\\
{\bf (WSJ 189)} The securities are convertible at a rate of \$44.25 of debentures for each common share.\\
{\bf (WSJ 190)} Revenue rose 8.4\% to \$390.9 million from \$360.5 million in 1985.\\
{\bf (WSJ 191)} Recent events have served to focus the debate.\\
{\bf (WSJ 192)} London shares advanced to a record close Friday in moderate trading.\\
{\bf (WSJ 193)} The company's shares rose 12.5 cents a share, to \$14.125, in American Stock Exchange composite trading Friday.\\
{\bf (WSJ 194)} "It's our company now; we have to create the policy."\\
{\bf (WSJ 195)} But Sen.\ Dole announced Friday that he decided against offering the amendment.\\
{\bf (WSJ 196)} In Chicago, the June contract on the Standard \& Poor's 500-stock index soared to a record high of 299.80, up 3.80 for the day.\\
{\bf (WSJ 197)} One reason the company is negotiating with potential investors is to provide money to lease additional aircraft.\\
{\bf (WSJ 198)} It's a price we had to charge to stay in business.\\
{\bf (WSJ 199)} That happened, to some extent.\\
{\bf (WSJ 200)} It's an improvement over the first proposal, all right, but that's not saying very much at all.\\
{\bf (WSJ 201)} The issue next goes to the cabinet-level Economic Policy Council, which has scheduled a meeting for mid-week.\\
{\bf (WSJ 202)} The company previously declared a 3-for-2 split in November 1986.\\
{\bf (WSJ 203)} In late New York trading Friday, the pound stood at \$1.6042, up from \$1.6003 Thursday, but eased to 2.9333 marks, from 2.9338.\\
{\bf (WSJ 204)} The transaction is subject to regulatory approval.\\
{\bf (WSJ 205)} Ohio Edison owns about 30\% of the unit.\\
{\bf (WSJ 206)} In composite trading on the New York Stock Exchange Friday, the company's common shares closed at \$35.125, down 25 cents.\\
{\bf (WSJ 207)} The deficit in January was \$2.17 billion.\\
{\bf (WSJ 208)} No doubt taxes are easier to reform in times of high economic growth.\\
{\bf (WSJ 209)} The Financial Times industrial-share index rose 17.3 to 1598.9.\\
{\bf (WSJ 210)} Pan Am Corp.\ reported a \$197.5 million fourth-quarter loss, worse than it had predicted, and said it expects to post a deficit in the first quarter.\\
{\bf (WSJ 211)} Soon, they were no longer talking about what they had to have, but about what they no longer needed.\\
{\bf (WSJ 212)} President Reagan suggested Friday that he would accept the earlier Senate version.\\
{\bf (WSJ 213)} Stock prices started to rise as trading got under way Friday.\\
{\bf (WSJ 214)} "What looks to the public like we were acting late was actually a situation in which we thought we were acting fast," the executive said.\\
{\bf (WSJ 215)} The government has to take a lead role.\\
{\bf (WSJ 216)} Export volume grew less than 1\% last year, compared with a 6\% increase in 1985.\\
{\bf (WSJ 217)} Currently, there's still some air around Wright's museum.\\
{\bf (WSJ 218)} Current practices require U.S.\ officials to draw up a list of potential targets and allow affected parties here to comment on their likely impact.\\
{\bf (WSJ 219)} Attorneys for Mr.\ Boesky declined to comment.\\
{\bf (WSJ 220)} Gold was quoted at \$406 an ounce in early trading Monday in Hong Kong.\\
{\bf (WSJ 221)} The financial services company said about 100 employees of American Health will lose their jobs after the sale is completed in August.\\
{\bf (WSJ 222)} Ohio Edison said the investors will pay it \$509 million for the stake, then lease it back for 29 years at a rate estimated at 8.5\% to 9\% of the cash payment.\\
{\bf (WSJ 223)} Both the government and the opposition have refused to compromise on election reform.\\
{\bf (WSJ 224)} The Reagan administration's most recent estimate for the fiscal 1987 deficit is \$143.91 billion.\\
{\bf (WSJ 225)} The next move is up to the opposition Liberal-National coalition, which has still not announced its tax policy.\\
{\bf (WSJ 226)} They expect retail issues to benefit from the budget's personal-tax cuts and the drop in British banks' lending rates, a trader said.\\
{\bf (WSJ 227)} In the 1985 fourth quarter, Pan Am had net income of \$241.4 million, or \$1.79 a share, which included a \$341 million gain from the Pacific division sale.\\
{\bf (WSJ 228)} The bank holding company is expected to make the announcement today.\\
{\bf (WSJ 229)} "I am in full support of reasonable funding levels for these programs similar to the legislation passed by the Senate," he said in a statement put out by the White House.\\
{\bf (WSJ 230)} Airline, drug, oil, technology, and some brokerage firm stocks took off.\\
{\bf (WSJ 231)} He predicted that the break-even level will be reached in 1987's second half.\\
{\bf (WSJ 232)} We have a lot of poor patients, and I would like to see them on the drug.\\
{\bf (WSJ 233)} Capital spending by German firms this year is expected to run only slightly ahead of 1986 levels, after gains of 9\% in 1985 and 5\% last year.\\
{\bf (WSJ 234)} The addition doesn't, of course, need to be this high.\\
{\bf (WSJ 235)} U.S.\ officials estimate that the move will result in \$85 million in additional U.S.\ sales this year, and that such sales eventually could grow to \$300 million annually.\\
{\bf (WSJ 236)} Competitors have recently supported the company's efforts to get regulation removed.\\
{\bf (WSJ 237)} Every one-cent rise in the gasoline tax would raise \$900 million to \$1 billion a year.\\
{\bf (WSJ 238)} Substantially all the delayed shipments should be made in the second quarter, the company said.\\
{\bf (WSJ 239)} Ohio Edison said the transaction will allow the equity investors to take advantage of federal tax benefits.\\
{\bf (WSJ 240)} The company had 32.9 million shares outstanding at Dec.\ 31.\\
{\bf (WSJ 241)} The deficit for all of fiscal 1986 was \$220.7 billion.\\
{\bf (WSJ 242)} We view it in one way.\\
{\bf (WSJ 243)} And the benefits of lower mortgage costs, resulting from falling interest rates, a drop in money supply and stable inflation, will be "larger than the gains" from tax cuts, the trader said.\\
{\bf (WSJ 244)} Without that gain, Pan Am would have reported a 1985 fourth-quarter loss of about \$100 million.\\
{\bf (WSJ 245)} Approval of the agreement between the Dallas-based aerospace, energy and steel concern and its 22-member bank group had been delayed since January.\\
{\bf (WSJ 246)} It also raises the accounting question of how banks will value such notes.\\
{\bf (WSJ 247)} Morgan Stanley \& Co.\ reported shortly after 3 p.m.\ that it had blue-chip stock buy orders totaling \$1.1 billion.\\
{\bf (WSJ 248)} Now managing more than \$60 billion for clients, mainly big pension funds, the firm has become the biggest single investor in the stock market.\\
{\bf (WSJ 249)} We expect that they will.\\
{\bf (WSJ 250)} The bureau said the number of housing units in the country will reach 100 million by the end of March.\\
{\bf (WSJ 251)} We do this kind of work every day.\\
{\bf (WSJ 252)} It also called on Japan to allow U.S.\ producers a larger share of Japan's own market.\\
{\bf (WSJ 253)} Brazil's short-term financing from foreign banks could soon be cut by as much as \$3 billion, according to bankers in the U.S.\ and Brazil.\\
{\bf (WSJ 254)} The tax boost would be part of a \$36 billion deficit-reduction package the House Democratic leadership seeks for the coming fiscal year.\\
{\bf (WSJ 255)} Under the plan, the holder of each common share will receive on March 31 the right to buy one additional share for \$3.\\
{\bf (WSJ 256)} The following issues recently were filed with the Securities and Exchange Commission:\\
{\bf (WSJ 257)} Holders are to vote at the April 23 annual meeting.\\
{\bf (WSJ 258)} The government paid \$13.7 billion in interest on the federal debt in February, up from \$13.49 billion in January.\\
{\bf (WSJ 259)} We're sure we can reach a compromise.\\
{\bf (WSJ 260)} Foreign demand, Wall Street's performance Thursday and buying ahead of the new trading quarter, which starts today, helped as well, traders said.\\
{\bf (WSJ 261)} In the 1986 fourth quarter, revenue declined 12\%, to \$797.3 million from \$906.7 million in the year-earlier quarter.\\
{\bf (WSJ 262)} No one at the unit could be reached for comment.\\
{\bf (WSJ 263)} Talks on those issues will resume later this week.\\
{\bf (WSJ 264)} It's not going to be up that much.\\
{\bf (WSJ 265)} In contrast, conventional managers usually hold some cash reserves, which have hurt their performance recently.\\
{\bf (WSJ 266)} Why didn't you set it at \$100,000?\\
{\bf (WSJ 267)} "There is a lot of housing out there," Mr.\ Young said.\\
{\bf (WSJ 268)} Let's keep it the way Wright built it.\\
{\bf (WSJ 269)} It raised more money from British institutions and took out its first major bank loans.\\
{\bf (WSJ 270)} About \$10 billion of trade credits and \$5 billion of money-market deposits fall due March 31.\\
{\bf (WSJ 271)} An additional \$18 billion would come from spending cuts.\\
                %\chapter{Corpus} \label{app-corpus}
\chapter{Features}
\label{app-features}

For an explanation of the structure of the following 205 features, please
refer to section~\ref{sec-features}.

\section{Features Used for Parsing English}
\label{sec-features-for-engl}
\begin{flushleft}
\mbox{[1]} {\bf (synt of -5 at s-synt-elem)}\\
\mbox{[2]} {\bf (synt of -4 at s-synt-elem)}\\
\mbox{[3]} {\bf (synt of -3 at s-synt-elem)}\\
\mbox{[4]} {\bf (synt of -2 at s-synt-elem)}\\
\mbox{[5]} {\bf (synt of -1 at s-synt-elem)}\\
\mbox{[6]} {\bf (synt of 1 at s-synt-elem)}\\
\mbox{[7]} {\bf (synt of 2 at s-synt-elem)}\\
\mbox{[8]} {\bf (synt of 3 at s-synt-elem)}\\
\mbox{[9]} {\bf (similar of -3 with -1 at m-boolean)}\\
\mbox{[10]} {\bf (class of -2 at c-app)}\\
\mbox{[11]} {\bf (class of -1 at c-app)}\\
\mbox{[12]} {\bf (class of 1 at c-app)}\\
\mbox{[13]} {\bf (class of -2 at c-clause)}\\
\mbox{[14]} {\bf (class of -1 at c-clause)}\\
\mbox{[15]} {\bf (synt of -3 at d-other-delimiter)}\\
\mbox{[16]} {\bf (synt of -2 at d-other-delimiter)}\\
\mbox{[17]} {\bf (synt of last of -2 at d-dividing-delimiter)}\\
\mbox{[18]} {\bf (synt of det of -1 at s-synt-elem)}\\
\mbox{[19]} {\bf (synt of time of -2 at s-synt-elem)}\\
\mbox{[20]} {\bf (synt of -1 at d-dividing-delimiter)}\\
\mbox{[21]} {\bf (synt of -1 at d-other-delimiter)}\\
\mbox{[22]} {\bf (synt of 1 at d-dividing-delimiter)}\\
\mbox{[23]} {\bf (synt of 1 at d-other-delimiter)}\\
\mbox{[24]} {\bf (f-finite-tense of -1 at m-boolean)}\\
\mbox{[25]} {\bf (f-non-finite-tense of -1 at m-boolean)}\\
\mbox{[26]} {\bf (f-part of -1 at m-boolean)}\\
\mbox{[27]} {\bf (f-part of 1 at m-boolean)}\\
\mbox{[28]} {\bf (f-pres-part of -1 at m-boolean)}\\
\mbox{[29]} {\bf (f-pres-part of 1 at m-boolean)}\\
\mbox{[30]} {\bf (f-inf of -1 at m-boolean)}\\
\mbox{[31]} {\bf (f-to-inf of -1 at m-boolean)}\\
\mbox{[32]} {\bf (f-passive of -1 at m-boolean)}\\
\mbox{[33]} {\bf (class of -2 at i-ec-conjunction)}\\
\mbox{[34]} {\bf (class of 1 at i-en-agent)}\\
\mbox{[35]} {\bf (class of subj of -1 at i-en-interrogative)}\\
\mbox{[36]} {\bf (class of -2 at i-en-quantity)}\\
\mbox{[37]} {\bf (class of -1 at i-en-quantity)}\\
\mbox{[38]} {\bf (class of pred-compl of co-theme of -1 at
                  i-en-quantity)}\\
\mbox{[39]} {\bf (classp of
                  i-ec-conjunction-introducing-correlative-apposition
                  of -2 at m-boolean)}\\
\mbox{[40]} {\bf (classp of i-eart-indef-art of -2 at m-boolean)}\\
\mbox{[41]} {\bf (classp of i-en-prename-title of -1 at m-boolean)}\\
\mbox{[42]} {\bf (classp of i-en-prename-title of -2 at m-boolean)}\\
\mbox{[43]} {\bf (classp of i-eadv-verbal-degree-adverb of -2 at
                  m-boolean)}\\
\mbox{[44]} {\bf (classp of i-eadv-adverbial-degree-adverb of -2 at
                  m-boolean)}\\
\mbox{[45]} {\bf (classp of i-eadv-adjectival-degree-adverb of -2 at
                  m-boolean)}\\
\mbox{[46]} {\bf (classp of i-eadv-nominal-degree-adverb of -2 at
                  m-boolean)}\\
\mbox{[47]} {\bf (classp of i-eadv-verbal-degree-adverb of -1 at
                  m-boolean)}\\
\mbox{[48]} {\bf (classp of i-eadv-adverbial-degree-adverb of -1 at
                  m-boolean)}\\
\mbox{[49]} {\bf (classp of i-eadv-adjectival-degree-adverb of -1 at
                  m-boolean)}\\
\mbox{[50]} {\bf (classp of i-eadv-adjectival-degree-adverb of alt-adv
                  of 1 at m-boolean)}\\
\mbox{[51]} {\bf (classp of i-eadv-nominal-degree-adverb of -1 at
                  m-boolean)}\\
\mbox{[52]} {\bf (classp of i-en-temporal-unit of alt-nom of 2 at
                  m-boolean)}\\
\mbox{[53]} {\bf (classp of i-en-temporal-concept of -1 at m-boolean)}\\
\mbox{[54]} {\bf (classp of i-en-temporal-interval of -1 at
                  m-boolean)}\\
\mbox{[55]} {\bf (classp of i-en-temporal-interval of -2 at
                  m-boolean)}\\
\mbox{[56]} {\bf (classp of i-en-process of -1 at m-boolean)}\\
\mbox{[57]} {\bf (class of pred-compl of -1 at i-en-process)}\\
\mbox{[58]} {\bf (classp of i-e-process of -2 at m-boolean)}\\
\mbox{[59]} {\bf (classp of i-en-reason of -1 at m-boolean)}\\
\mbox{[60]} {\bf (class of -1 at i-en-unit)}\\
\mbox{[61]} {\bf (day-or-year of -1 at i-enum-cardinal)}\\
\mbox{[62]} {\bf (day-or-year of 1 at i-enum-cardinal)}\\
\mbox{[63]} {\bf (classp of i-en-month-of-the-year of -1 at
                  m-boolean)}\\
\mbox{[64]} {\bf (class of -4 at i-ep-preposition)}\\
\mbox{[65]} {\bf (class of -3 at i-ep-preposition)}\\
\mbox{[66]} {\bf (class of -2 at i-ep-preposition)}\\
\mbox{[67]} {\bf (class of 2 at i-ep-preposition)}\\
\mbox{[68]} {\bf (class of 1 at i-ev-process-state)}\\
\mbox{[69]} {\bf (is-abbreviation of -1 at m-boolean)}\\
\mbox{[70]} {\bf (is-indexed of -1 at m-boolean)}\\
\mbox{[71]} {\bf (capitalization of pred* of -2 at m-boolean)}\\
\mbox{[72]} {\bf (capitalization of pred* of -1 at m-boolean)}\\
\mbox{[73]} {\bf (capitalization of pred* of 1 at m-boolean)}\\
\mbox{[74]} {\bf (marked-capitalization of pred* of -1 at m-boolean)}\\
\mbox{[75]} {\bf (is-abbreviation of -2 at m-boolean)}\\
\mbox{[76]} {\bf (is-ref of -1 at m-boolean)}\\
\mbox{[77]} {\bf (is-ref of subj of -1 at m-boolean)}\\
\mbox{[78]} {\bf (is-ref of subj of inf-compl of -1 at m-boolean)}\\
\mbox{[79]} {\bf (is-ref of -2 at m-boolean)}\\
\mbox{[80]} {\bf (np-vp-match of -2 with -1 at m-boolean)}\\
\mbox{[81]} {\bf (np-vp-match of -1 with 1 at m-boolean)}\\
\mbox{[82]} {\bf (compl-v-match of -2 with -1 at m-boolean)}\\
\mbox{[83]} {\bf (compl-v-match of -1 with 1 at m-boolean)}\\
\mbox{[84]} {\bf (marked-capitalization of 1 at m-boolean)}\\
\mbox{[85]} {\bf (syntp of unavail of lexical of 5 at m-boolean)}\\
\mbox{[86]} {\bf (syntp of unavail of lexical of 3 at m-boolean)}\\
\mbox{[87]} {\bf (syntp of unavail of lexical of 2 at m-boolean)}\\
\mbox{[88]} {\bf (syntp of unavail of lexical of ((pred* of -2) -1) at
                  m-boolean)}\\
\mbox{[89]} {\bf (syntp of unavail of lexical of (-1 1) at m-boolean)}\\
\mbox{[90]} {\bf (syntp of unavail of lexical of (-1 1 2) at
                  m-boolean)}\\
\mbox{[91]} {\bf (syntp of unavail of lexical of (-2 -1 1) at
                  m-boolean)}\\
\mbox{[92]} {\bf (semrole of -1 of -2)}\\
\mbox{[93]} {\bf (semrole of -1 of -4)}\\
\mbox{[94]} {\bf (semrole of 1 of -1)}\\
\mbox{[95]} {\bf (semrole of -1 of -3)}\\
\mbox{[96]} {\bf (semrolep of -1 of -3)}\\
\mbox{[97]} {\bf (semrole of -2 of -3)}\\
\mbox{[98]} {\bf (semrolep of -2 of -3)}\\
\mbox{[99]} {\bf (semrolep of obj of -1)}\\
\mbox{[100]} {\bf (semrole of subj of -1)}\\
\mbox{[101]} {\bf (syntrole of -2 of -1)}\\
\mbox{[102]} {\bf (synt of -1 at s-clause)}\\
\mbox{[103]} {\bf (synt of -2 at s-clause)}\\
\mbox{[104]} {\bf (classp of i-en-interr-pronoun-whatever of -2 at
                   m-boolean)}\\
\mbox{[105]} {\bf (classp of i-eprt-adv-particle of -1 at m-boolean)}\\
\mbox{[106]} {\bf (classp of i-en-letter-character of -1 at
                   m-boolean)}\\
\mbox{[107]} {\bf (classp of i-en-day-of-the-week of -1 at m-boolean)}\\
\mbox{[108]} {\bf (classp of i-en-day-of-the-week of alt-nom of 1 at
                   m-boolean)}\\
\mbox{[109]} {\bf (classp of i-eadj-able of -1 at m-boolean)}\\
\mbox{[110]} {\bf (classp of i-eadj-able of 1 at m-boolean)}\\
\mbox{[111]} {\bf (synt of -1 at s-particle)}\\
\mbox{[112]} {\bf (synt of 1 at s-particle)}\\
\mbox{[113]} {\bf (syntp of unavail of vp-1 at m-boolean)}\\
\mbox{[114]} {\bf (syntp of unavail of npp-1 of -2 at m-boolean)}\\
\mbox{[115]} {\bf (syntp of unavail of npp-1 of -3 at m-boolean)}\\
\mbox{[116]} {\bf (syntp of unavail of conj of -1 at m-boolean)}\\
\mbox{[117]} {\bf (syntp of unavail of pred of -1 at m-boolean)}\\
\mbox{[118]} {\bf (syntp of unavail of pred of -2 at m-boolean)}\\
\mbox{[119]} {\bf (syntp of unavail of from-quant of -2 at m-boolean)}\\
\mbox{[120]} {\bf (syntp of unavail of pred-compl of -1 at m-boolean)}\\
\mbox{[121]} {\bf (syntp of unavail of time of pred-compl of -1 at
                   m-boolean)}\\
\mbox{[122]} {\bf (syntp of d-dividing-delimiter of last of -1 at
                   m-boolean)}\\
\mbox{[123]} {\bf (syntp of unavail of obj of -1 at m-boolean)}\\
\mbox{[124]} {\bf (syntp of unavail of inf-compl of -1 at m-boolean)}\\
\mbox{[125]} {\bf (synt of alt2 of 1 at s-synt-elem)}\\
\mbox{[126]} {\bf (synt of alt3 of 1 at s-synt-elem)}\\
\mbox{[127]} {\bf (synt of -3 at s-verb)}\\
\mbox{[128]} {\bf (syntp of s-aux of -1 at m-boolean)}\\
\mbox{[129]} {\bf (classp of i-en-point of -1 at m-boolean)}\\
\mbox{[130]} {\bf (classp of c-at-time of -1 at m-boolean)}\\
\mbox{[131]} {\bf (classp of c-at-time of -2 at m-boolean)}\\
\mbox{[132]} {\bf (classp of i-eadv-there of -2 at m-boolean)}\\
\mbox{[133]} {\bf (classp of i-ev-be of -1 at m-boolean)}\\
\mbox{[134]} {\bf (classp of i-ep-to of -1 at m-boolean)}\\
\mbox{[135]} {\bf (classp of i-ep-to of 1 at m-boolean)}\\
\mbox{[136]} {\bf (classp of i-en-agent of -5 at m-boolean)}\\
\mbox{[137]} {\bf (classp of i-en-place of -5 at m-boolean)}\\
\mbox{[138]} {\bf (classp of i-en-quantity of -5 at m-boolean)}\\
\mbox{[139]} {\bf (classp of i-en-quantity of -4 at m-boolean)}\\
\mbox{[140]} {\bf (classp of i-en-agent of -3 at m-boolean)}\\
\mbox{[141]} {\bf (classp of i-en-place of -3 at m-boolean)}\\
\mbox{[142]} {\bf (classp of i-en-quantity of -3 at m-boolean)}\\
\mbox{[143]} {\bf (classp of i-en-tangible-object of -3 at m-boolean)}\\
\mbox{[144]} {\bf (classp of i-en-pronoun of -3 at m-boolean)}\\
\mbox{[145]} {\bf (classp of i-en-pronoun of -2 at m-boolean)}\\
\mbox{[146]} {\bf (classp of i-en-interr-pronoun of -2 at m-boolean)}\\
\mbox{[147]} {\bf (classp of i-en-agent of -2 at m-boolean)}\\
\mbox{[148]} {\bf (classp of i-en-proper-place of -2 at m-boolean)}\\
\mbox{[149]} {\bf (classp of i-en-place of -2 at m-boolean)}\\
\mbox{[150]} {\bf (classp of i-en-quantity of -2 at m-boolean)}\\
\mbox{[151]} {\bf (classp of i-en-temporal-entity of -2 at m-boolean)}\\
\mbox{[152]} {\bf (classp of i-en-currency-unit of -1 at m-boolean)}\\
\mbox{[153]} {\bf (classp of i-en-agent of -1 at m-boolean)}\\
\mbox{[154]} {\bf (classp of i-en-place of -1 at m-boolean)}\\
\mbox{[155]} {\bf (classp of i-en-percent of -1 at m-boolean)}\\
\mbox{[156]} {\bf (classp of i-en-percentage of -1 at m-boolean)}\\
\mbox{[157]} {\bf (classp of i-en-pronoun of -1 at m-boolean)}\\
\mbox{[158]} {\bf (classp of i-en-personal-pronoun of -1 at
                   m-boolean)}\\
\mbox{[159]} {\bf (classp of i-en-personal-pronoun of -2 at
                   m-boolean)}\\
\mbox{[160]} {\bf (classp of i-en-monetarily-quantifiable-abstract of
                   -1 at m-boolean)}\\
\mbox{[161]} {\bf (classp of i-en-monetarily-quantifiable-abstract of
                   -2 at m-boolean)}\\
\mbox{[162]} {\bf (classp of i-eadv-quantifier of -2 at m-boolean)}\\
\mbox{[163]} {\bf (classp of i-en-mod-abstract of -3 at m-boolean)}\\
\mbox{[164]} {\bf (classp of i-en-monetarily-quantifiable-abstract of
                   -3 at m-boolean)}\\
\mbox{[165]} {\bf (classp of i-en-quantifying-abstract of -3 at
                   m-boolean)}\\
\mbox{[166]} {\bf (syntp of s-indef-pron of pred* of -3 at m-boolean)}\\
\mbox{[167]} {\bf (syntp of d-delimiter of last of -3 at m-boolean)}\\
\mbox{[168]} {\bf (synt of -2 at s-conj)}\\
\mbox{[169]} {\bf (classp of i-eadv-temporal-quantifier of -1 at
                   m-boolean)}\\
\mbox{[170]} {\bf (classp of i-en-tangible-object of -1 at m-boolean)}\\
\mbox{[171]} {\bf (classp of i-en-quantity of -1 at m-boolean)}\\
\mbox{[172]} {\bf (classp of i-en-temporal-entity of -1 at m-boolean)}\\
\mbox{[173]} {\bf (classp of i-en-unit of -1 at m-boolean)}\\
\mbox{[174]} {\bf (classp of i-en-agent of subj of -1 at m-boolean)}\\
\mbox{[175]} {\bf (classp of i-en-place of subj of -1 at m-boolean)}\\
\mbox{[176]} {\bf (classp of i-en-tangible-object of alt-nom of 1 at
                   m-boolean)}\\
\mbox{[177]} {\bf (classp of i-en-agent of alt-nom of 1 at m-boolean)}\\
\mbox{[178]} {\bf (classp of i-en-place of alt-nom of 1 at m-boolean)}\\
\mbox{[179]} {\bf (classp of i-en-temporal-entity of 1 at m-boolean)}\\
\mbox{[180]} {\bf (classp of i-en-unit of alt-nom of 1 at m-boolean)}\\
\mbox{[181]} {\bf (classp of i-en-agent of alt-nom of 2 at m-boolean)}\\
\mbox{[182]} {\bf (classp of i-en-place of alt-nom of 2 at m-boolean)}\\
\mbox{[183]} {\bf (classp of i-en-tangible-object of alt-nom of 2 at
                   m-boolean)}\\
\mbox{[184]} {\bf (classp of i-en-tangible-object of -2 at m-boolean)}\\
\mbox{[185]} {\bf (classp of i-en-temporal-entity of 2 at m-boolean)}\\
\mbox{[186]} {\bf (classp of i-en-unit of alt-nom of 2 at m-boolean)}\\
\mbox{[187]} {\bf (classp of i-en-interrogative of active-filler-1 at
                   m-boolean)}\\
\mbox{[188]} {\bf (classp of i-en-interrogative of -2 at m-boolean)}\\
\mbox{[189]} {\bf (classp of i-en-interrogative of -1 at m-boolean)}\\
\mbox{[190]} {\bf (classp of i-en-interr-pronoun of -1 at m-boolean)}\\
\mbox{[191]} {\bf (classp of i-en-interrogative of 1 at m-boolean)}\\
\mbox{[192]} {\bf (classp of i-eadv-at-location of -1 at m-boolean)}\\
\mbox{[193]} {\bf (classp of i-eadv-at-location of alt-adv of 1 at
                   m-boolean)}\\
\mbox{[194]} {\bf (classp of i-eadv-to-location of alt-adv of 1 at
                   m-boolean)}\\
\mbox{[195]} {\bf (gap of active-filler-1 at m-filler-status)}\\
\mbox{[196]} {\bf (gap of -2 at m-filler-status)}\\
\mbox{[197]} {\bf (synt of -2 at s-app)}\\
\mbox{[198]} {\bf (synt of -1 at s-app)}\\
\mbox{[199]} {\bf (synt of 1 at s-app)}\\
\mbox{[200]} {\bf (synt of -3 at s-numeral)}\\
\mbox{[201]} {\bf (synt of -2 at s-numeral)}\\
\mbox{[202]} {\bf (synt of -1 at s-numeral)}\\
\mbox{[203]} {\bf (synt of 1 at s-numeral)}\\
\mbox{[204]} {\bf (synt of 1 at s-conj)}\\
\mbox{[205]} {\bf (classp of i-en-demonstr-pronoun of -2 at
                   m-boolean)}
\end{flushleft}

\section{Features Used in English to German Translation Disambiguation Decision Structures}
\label{sec-features-for-engl-german}

The following is a list of features used for English to German translation disambiguation.
Recall that the transfer examples and features are organized in {\it transfer entries}, 
each of which contains the example phrases and features for a specific source concept.
For each individual transfer entry, only a small subset of the following features
is used.

\begin{flushleft}
\mbox{[1]} {\bf (class of parent of parent at c-app)}\\
\mbox{[2]} {\bf (classp of c-at-time of parent at m-boolean)}\\
\mbox{[3]} {\bf (classp of c-to-quant of parent at m-boolean)}\\
\mbox{[4]} {\bf (classp of i-eadv-neg-adverb of quantifier of parent at m-boolean)}\\
\mbox{[5]} {\bf (classp of i-en-agent of pred-compl of parent at m-boolean)}\\
\mbox{[6]} {\bf (classp of i-en-agent of theme of parent at m-boolean)}\\
\mbox{[7]} {\bf (classp of i-en-exchange-boerse of pred-compl of parent at m-boolean)}\\
\mbox{[8]} {\bf (classp of i-en-proper-place of mod of parent at m-boolean)}\\
\mbox{[9]} {\bf (classp of i-en-proper-place of pred-compl of parent at m-boolean)}\\
\mbox{[10]} {\bf (classp of i-en-tangible-object of theme of parent at m-boolean)}\\
\mbox{[11]} {\bf (number at f-number)}\\
\mbox{[12]} {\bf (synt of agent of parent at s-synt-elem)}\\
\mbox{[13]} {\bf (synt of parent of parent at s-synt-elem)}\\
\mbox{[14]} {\bf (syntp of s-np of theme of parent at m-boolean)}\\
\mbox{[15]} {\bf (syntp of s-sub-clause of theme of parent at m-boolean)}\\
\mbox{[16]} {\bf (syntp of unavail of det of pred-compl of parent at m-boolean)}\\
\mbox{[17]} {\bf (syntp of unavail of obj of parent at m-boolean)}\\
\mbox{[18]} {\bf (voice of parent at f-voice)}
\end{flushleft}
         %\chapter{Features} \label{app-features}
\chapter{Parse Example}
\label{app-parse-example}

The following sections show the sentence
``{\it I like the house Mr. Miller built.}''
after segmentation and morphological analysis, the complete sequence of partial parse
states and parse actions, and finally the resulting parse tree.

\section{After Segmentation and Morphological Processing}

{\footnotesize
\begin{verbatim}
 "I":
    synt:    S-PRON
    class:   I-EN-PERSONAL-PRONOUN
    forms:   (((NUMBER F-SING) (PERSON F-FIRST-P) (CASE F-NOM)))
    lex:     "PRON"
    props:   ((CAPITALIZATION TRUE))
(OR  
 "like":
    synt:    S-PREP
    class:   I-EP-LIKE
    forms:   (NIL)
    lex:     "like"  
 "like":
    synt:    S-VERB
    class:   I-EV-LIKE
    forms:   (((PERSON F-SECOND-P) (NUMBER F-SING)
               (TENSE F-PRES-TENSE))
              ((PERSON F-FIRST-P) (NUMBER F-SING) (TENSE F-PRES-TENSE))
              ((NUMBER F-PLURAL) (TENSE F-PRES-TENSE))
              ((TENSE F-PRES-INF)))
    lex:     "like")
 "the":
    synt:    S-DEF-ART
    class:   I-EART-DEF-ART
    forms:   (NIL)
    lex:     "the"
 "new":
    synt:    S-ADJ
    class:   I-EADJ-NEW
    forms:   (NIL)
    lex:     "new"
    props:   ((ADJ-TYPE S-NON-DEMONSTR-ADJ))
 "house":
    synt:    S-COUNT-NOUN
    class:   I-EN-HOUSE
    forms:   (((NUMBER F-SING) (PERSON F-THIRD-P)))
    lex:     "house"
 "Mr":
    synt:    S-COUNT-NOUN
    class:   I-EN-MISTER-TITLE
    forms:   (((NUMBER F-SING) (PERSON F-THIRD-P)))
    lex:     "Mr"
    props:   ((CAPITALIZATION TRUE) (IS-ABBREVIATION TRUE))
 ".":
    synt:    D-PERIOD
    lex:     "."
 "Miller":
    synt:    S-PROPER-NAME
    class:   I-EN-MILLER
    forms:   (((NUMBER F-SING) (PERSON F-THIRD-P)))
    lex:     "Miller"
    props:   ((CAPITALIZATION TRUE))
 "Miller":
    synt:    S-PROPER-NAME
    class:   I-EN-MILLER
    forms:   (((NUMBER F-SING) (PERSON F-THIRD-P)))
    lex:     "Miller"
    props:   ((CAPITALIZATION TRUE))
 "built":
    synt:    S-VERB
    class:   I-EV-BUILD
    forms:   (((TENSE F-PAST-PART)) ((TENSE F-PAST-TENSE)))
    lex:     "build"
 ".":
    synt:    D-PERIOD
    lex:     "."
\end{verbatim}
}

\section{Parse State and Actions Step by Step}

{\footnotesize
\begin{verbatim}
* I like the new house Mr . Miller built .  
   (S S-PRON)
(I) * like the new house Mr . Miller built .  
   (R 1 TO S-NP AS PRED) 
(I) * like the new house Mr . Miller built .  
   (S S-VERB)
(I) (like) * the new house Mr . Miller built .  
   (R 1 TO S-VP AS PRED) 
(I) (like) * the new house Mr . Miller built .  
   (S S-ART) 
(I) (like) (the) * new house Mr . Miller built .  
   (S S-ADJ) 
(I) (like) (the) (new) * house Mr . Miller built .  
   (R 1 TO S-ADJP AS PRED) 
(I) (like) (the) (new) * house Mr . Miller built .  
   (S S-NOUN) 
(I) (like) (the) (new) (house) * Mr . Miller built .  
   (R 1 TO S-NP AS PRED)
(I) (like) (the) (new) (house) * Mr . Miller built .  
   (R 2 AS MOD SAME
(I) (like) (the) (new house) * Mr . Miller built .  
   (R 2 AS DET SAME)
(I) (like) (the new house) * Mr . Miller built .  
   (EMPTY-CAT FROM NP-1 AT 0) (M -1 GAP ACTIVE-FILLER)
(I) (like) (<the new house>1) (<REF>1) * Mr . Miller built .  
   (S S-NOUN)
(I) (like) (<the new house>1) (<REF>1) (Mr) * . Miller built .  
   (S D-DELIMITER)
(I) (like) (<the new house>1) (<REF>1) (Mr) (.) * Miller built .  
   (R 2 AS PRED DUMMY)
(I) (like) (<the new house>1) (<REF>1) (Mr.) * Miller built .  
   (S S-NOUN)
(I) (like) (<the new house>1) (<REF>1) (Mr.) (Miller) * built .  
   (R 2 AS MOD PRED)
(I) (like) (<the new house>1) (<REF>1) (Mr. Miller) * built .  
   (R 1 TO S-NP AS PRED)
(I) (like) (<the new house>1) (<REF>1) (Mr. Miller) * built .  
   (S S-VERB)
(I) (like) (<the new house>1) (<REF>1) (Mr. Miller) (built) * .  
   (R 1 TO S-VP AS PRED)
(I) (like) (<the new house>1) (<REF>1) (Mr. Miller) (built) * .  
   (R 2 TO S-SNT AS (SUBJ AGENT) SAME)
(I) (like) (<the new house>1) (<REF>1) (Mr. Miller built) * .  
   (R 2 TO S-REL-CLAUSE AS (OBJ THEME) SAME)
(I) (like) (<the new house>1) (<REF>1 Mr. Miller built) * .  
   (R 2 AS SAME MOD)
(I) (like) (<the new house <REF>1 Mr. Miller built>1) * .  
   (R 2 AS SAME (OBJ THEME))
(I) (like <the new house <REF>1 Mr. Miller built>1) * .  
   (R 2 TO S-SNT AS (SUBJ EXP) SAME)
(I like <the new house <REF>1 Mr. Miller built>1) * .  
   (S D-DELIMITER)
(I like <the new house <REF>1 Mr. Miller built>1) (.) * 
   (R 2 AS SAME DUMMY)
(I like <the new house <REF>1 Mr. Miller built>1.) * 
   (DONE)
\end{verbatim}
}

\section{Resulting Parse Tree}
\label{sec-resulting-parse-tree}

{\footnotesize
\begin{verbatim}
"I like <the new house <REF>1 Mr. Miller built>1.":
    synt:    S-SNT
    class:   I-EV-LIKE
    forms:   (((CASE F-NOM) (TENSE F-PRES-TENSE) (PERSON F-FIRST-P)
               (NUMBER F-SING)))
    lex:     "like"
    props:   ((CACHED-BEST-PATTERN 25
               (((SUBJ EXP (I-EN-AGENT)) (OBJ THEME (I-EN-THING)))
                (( <>1 "<the new house <REF>1 Mr. Miller built>1" 1)
                 ( "I" 0)))))
    subs:    
    (SUBJ EXP)  "I":
        synt:    S-NP
        class:   I-EN-PERSONAL-PRONOUN
        forms:   (((NUMBER F-SING) (PERSON F-FIRST-P) (CASE F-NOM)))
        lex:     "PRON"
        subs:    
        (PRED)  "I":
            synt:    S-PRON
            class:   I-EN-PERSONAL-PRONOUN
            forms:   (((NUMBER F-SING) (PERSON F-FIRST-P) (CASE F-NOM)))
            lex:     "PRON"
            props:   ((CAPITALIZATION TRUE))
    (PRED)  "like":
        synt:    S-VERB
        class:   I-EV-LIKE
        forms:   (((PERSON F-SECOND-P) (NUMBER F-SING)
                   (TENSE F-PRES-TENSE))
                  ((PERSON F-FIRST-P) (NUMBER F-SING)
                   (TENSE F-PRES-TENSE))
                  ((NUMBER F-PLURAL) (TENSE F-PRES-TENSE))
                  ((TENSE F-PRES-INF)))
        lex:     "like"
    (OBJ THEME)  "<the new house <REF>1 Mr. Miller built>1":
        synt:    S-NP
        class:   I-EN-HOUSE
        forms:   (((NUMBER F-SING) (PERSON F-THIRD-P)))
        lex:     "house"
        props:   ((INDEX 1) (ORIG-SURF "the new house") (INDEXED TRUE))
        subs:    
        (DET)  "the":
            synt:    S-DEF-ART
            class:   I-EART-DEF-ART
            forms:   (NIL)
            lex:     "the"
        (MOD)  "new":
            synt:    S-ADJP
            class:   I-EADJ-NEW
            forms:   (NIL)
            lex:     "new"
            subs:    
            (PRED)  "new":
                synt:    S-ADJ
                class:   I-EADJ-NEW
                forms:   (NIL)
                lex:     "new"
                props:   ((ADJ-TYPE S-NON-DEMONSTR-ADJ))
        (PRED)  "house":
            synt:    S-COUNT-NOUN
            class:   I-EN-HOUSE
            forms:   (((NUMBER F-SING) (PERSON F-THIRD-P)))
            lex:     "house"
        (MOD)  "<REF>1 Mr. Miller built":
            synt:    S-REL-CLAUSE
            class:   I-EV-BUILD
            forms:   (((PERSON F-THIRD-P) (NUMBER F-SING) (CASE F-NOM)
                       (TENSE F-PAST-TENSE)))
            lex:     "build"
            subs:    
            (OBJ THEME)  "<REF>1":
                synt:    S-NP
                class:   I-EN-HOUSE
                forms:   (((NUMBER F-SING) (PERSON F-THIRD-P)))
                lex:     "house"
                props:   ((GAP ACTIVE-FILLER) (INDEXED TRUE) (REF 1)
                          (ORIG-SURF "the new house"))
                subs:    
                (DET)  "the":
                    synt:    S-DEF-ART
                    class:   I-EART-DEF-ART
                    forms:   (NIL)
                    lex:     "the"
                (MOD)  "new":
                    synt:    S-ADJP
                    class:   I-EADJ-NEW
                    forms:   (NIL)
                    lex:     "new"
                    subs:    
                    (PRED)  "new":
                        synt:    S-ADJ
                        class:   I-EADJ-NEW
                        forms:   (NIL)
                        lex:     "new"
                        props:   ((ADJ-TYPE S-NON-DEMONSTR-ADJ))
                (PRED)  "house":
                    synt:    S-COUNT-NOUN
                    class:   I-EN-HOUSE
                    forms:   (((NUMBER F-SING) (PERSON F-THIRD-P)))
                    lex:     "house"
            (SUBJ AGENT)  "Mr. Miller":
                synt:    S-NP
                class:   I-EN-MILLER
                forms:   (((NUMBER F-SING) (PERSON F-THIRD-P)))
                lex:     "Miller"
                subs:    
                (PRED)  "Mr. Miller":
                    synt:    S-PROPER-NAME
                    class:   I-EN-MILLER
                    forms:   (((NUMBER F-SING) (PERSON F-THIRD-P)))
                    lex:     "Miller"
                    subs:    
                    (MOD)  "Mr.":
                        synt:    S-COUNT-NOUN
                        class:   I-EN-MISTER-TITLE
                        forms:   (((NUMBER F-SING) (PERSON F-THIRD-P)))
                        lex:     "Mr"
                        subs:    
                        (PRED)  "Mr":
                            synt:    S-COUNT-NOUN
                            class:   I-EN-MISTER-TITLE
                            forms:   (((NUMBER F-SING)
                                       (PERSON F-THIRD-P)))
                            lex:     "Mr"
                            props:   ((CAPITALIZATION TRUE)
                                      (IS-ABBREVIATION TRUE))
                        (DUMMY)  ".":
                            synt:    D-PERIOD
                            lex:     "."
                    (PRED)  "Miller":
                        synt:    S-PROPER-NAME
                        class:   I-EN-MILLER
                        forms:   (((NUMBER F-SING) (PERSON F-THIRD-P)))
                        lex:     "Miller"
                        props:   ((CAPITALIZATION TRUE))
            (PRED)  "built":
                synt:    S-VERB
                class:   I-EV-BUILD
                forms:   (((TENSE F-PAST-PART)) ((TENSE F-PAST-TENSE)))
                lex:     "build"
    (DUMMY)  ".":
        synt:    D-PERIOD
        lex:     "."
\end{verbatim}
}

             %\chapter{Parse Example} \label{app-p_example}
\chapter{Translation Evaluation Questionnaire and Key}
\label{app-quest}

\hyphenation{
ja-pa-ni-schen
an-fing
}

\section{Translation Evaluation Questionnaire}

The questionnaires handed out for translation evaluation were hardcopies of the two Web pages
{\tt http://www.cs.utexas.edu/users/ulf/diss/eval\_tegm.html}, the version that always
includes human translations, and
{\tt http://www.cs.utexas.edu/users/ulf/diss/eval\_tego.html}, the version that does not
include human translations, unless it happens to match any of the machine translations.

The questionnaire starting on the next page is the dissertation format version
of\\
{\tt http://www.cs.utexas.edu/users/ulf/diss/eval\_tegm.html} .

\newpage \noindent
{\bf {\large Translation Evaluation of Wall Street Journal sentences 48-79}}\\
October 1996\\
http://www.cs.utexas.edu/users/ulf/diss/eval\_tegm.html\\

\noindent \hrulefill\\

\noindent {\bf Hello!} This evaluation will help the research of a dissertation in the area of natural 
language processing. The following includes the results of computer programs that 
translated sentences from the Wall Street Journal from English to German.

\noindent Please evaluate the following translations by {\bf assigning grades} for both grammatical
correctness and meaning preservation using the following tables as a guideline. 
Note that the scale is like the one used in the German education system (1 = sehr gut; 
2 = gut; 3 = befriedigend; 4 = ausreichend; 5 = mangelhaft; 6 = ungen\"{u}gend). \\

\noindent {\bf Grammar (syntax and morphology)}\\

\noindent \begin{tabular}{|c|p{12cm}|} \hline
{\it Grade} & {\it Usage} \\ \hline
1 & Correct grammar, including word order, word endings; the sentence reads fluently. \\ \hline
2 & Basically correct grammar, but not very fluent. \\ \hline
3 & Mostly correct grammar, but with significant shortcomings. \\ \hline
4 & The grammar is acceptable only in parts of the sentence. \\ \hline
5 & The grammar is generally so bad that the entire sentence becomes very hard to read. \\ \hline
6 & The grammar is so bad that the sentence becomes totally incomprehensible. \\ \hline
\end{tabular}\\[3mm]

\noindent {\bf Meaning (semantics)}\\

\noindent \begin{tabular}{|c|p{12cm}|} \hline
{\it Grade} & {\it Usage} \\ \hline
1 & The meaning is fully preserved and can easily be understood. \\ \hline
2 & The meaning is mostly preserved and can be understood fairly well. \\ \hline
3 & The general idea of the sentence is preserved. \\ \hline
4 & Contains some useful information from the original sentence. \\ \hline
5 & A reader of the translated sentence can guess what the sentence is about,
     but the sentence provides hardly any useful information. \\ \hline
6 & The sentence is totally incomprehensible or totally misleading. \\ \hline
\end{tabular}\\[7mm]

\noindent (EXAMPLE)\\
{\bf Yesterday, I ate a red apple.}\\
\noindent \begin{tabular}{@{\hspace{0mm}}lll} (a) & 
G\"{a}stern, ich haben essen Apfl-rot.
 & {\it Grammar:\rule{2mm}{.2mm}5\rule{2mm}{.2mm} Meaning:\rule{2mm}{.2mm}}2\rule{2mm}{.2mm} \\ (b) &
Meine roten \"{A}pfel haben viel gegessen.
 & {\it Grammar:\rule{2mm}{.2mm}1\rule{2mm}{.2mm} Meaning:\rule{2mm}{.2mm}}6\rule{2mm}{.2mm} \\ \end{tabular}\\

\noindent \hrulefill\\

\noindent The evaluation of the translations will take you {\bf about 45-60 minutes}, or about 1-3 minutes for
each of the 32 English sentences and their German translations.
If you have any questions, please don't hesitate to contact Ulf at 
\begin{itemize}
  \item +1 (512) 320-0650 (home; voice \& fax)
  \item +1 (512) 471-9777 (office)
  \item ulf@cs.utexas.edu
\end{itemize}
Please return the evaluations to:
\begin{itemize}
  \item {\it Campus mail:} Ulf Hermjakob - Dept.\ of Computer Sciences - Mail code C0500
  \item {\it Postal address:} Ulf Hermjakob - 600 W26th St \#A308 - Austin, TX 78705
  \item {\it Office location:} Taylor Hall 150B (with 24 hour accessible mailbox)
  \item or call or email Ulf for pick-up
\end{itemize}

\noindent \hrulefill\\

\newpage
\noindent (WSJ 48)\\
{\bf Largely because of the falling dollar, West German labor costs rose to 120\% of those for U.S.\ production workers from 75\% in 1985.} \\
\noindent \begin{tabular}{@{\hspace{0mm}}lp{12.6cm}p{3.0cm}} (a) & 
Im gro{\ss}en und ganzen wegen des fallenden Dollars stiegen Westliche deutsche Arbeitskosten 1985 zu 120\% von denen f\"{u}r die Produktionsarbeiter der USA von 75\%.
 & {\it Grammar:\rule{8mm}{.2mm} Meaning:\rule{8mm}{.2mm}} \\ (b) & 
Gro{\ss} wegen des fallenden Dollars, stiegen westdeutsche Lohnkosten zu 120\% von denen f\"{u}r US-Produktionsarbeiter von 75\% 1985.
 & {\it Grammar:\rule{8mm}{.2mm} Meaning:\rule{8mm}{.2mm}} \\ (c) & 
Zum gr\"{o}{\ss}ten Teil wegen des fallenden Dollars stiegen westdeutsche Arbeitskosten auf 120\% von denen f\"{u}r US Produktionsarbeiter von 75\% 1985.
 & {\it Grammar:\rule{8mm}{.2mm} Meaning:\rule{8mm}{.2mm}} \\ (d) & 
Zum gr\"{o}{\ss}ten Teil wegen des fallend Dollar, Westdeutscher wirtschaftliche Preise Rosen to120\% von jenen f\"{u}r AMERIKANISCHE Produktion-Arbeiter von 75\% in 1985.
 & {\it Grammar:\rule{8mm}{.2mm} Meaning:\rule{8mm}{.2mm}} \\ (e) & 
Zum gr\"{o}{\ss}ten Teil wegen des fallenden Dollars stiegen westdeutsche Lohnkosten auf 120\% von denen f\"{u}r US-Industriearbeiter von 75\% im Jahre 1985.
 & {\it Grammar:\rule{8mm}{.2mm} Meaning:\rule{8mm}{.2mm}} \\ (f) & 
Zum gr\"{o}{\ss}ten Teil wegen des fallenden Dollars stiegen westdeutsche Arbeitskosten f\"{u}r US Produktionsarbeiter auf 120\% von denen von 75\% 1985.
 & {\it Grammar:\rule{8mm}{.2mm} Meaning:\rule{8mm}{.2mm}} \\ \end{tabular} \\

\noindent (WSJ 49)\\
{\bf In high school, he was a member of the speech team.} \\
\noindent \begin{tabular}{@{\hspace{0mm}}lp{12.6cm}p{3.0cm}} (a) & 
Im Gymnasium war er ein Mitglied der Redemannschaft.
 & {\it Grammar:\rule{8mm}{.2mm} Meaning:\rule{8mm}{.2mm}} \\ (b) & 
In Gymnasium war er ein Mitglied der Rede-Mannschaft.
 & {\it Grammar:\rule{8mm}{.2mm} Meaning:\rule{8mm}{.2mm}} \\ (c) & 
Auf dem Gymnasium war er ein Mitglied der Redemannschaft.
 & {\it Grammar:\rule{8mm}{.2mm} Meaning:\rule{8mm}{.2mm}} \\ (d) & 
In der hohen Schule war er ein Mitglied der Redemannschaft.
 & {\it Grammar:\rule{8mm}{.2mm} Meaning:\rule{8mm}{.2mm}} \\ (e) & 
In High School war er ein Mitglied des Redeteams.
 & {\it Grammar:\rule{8mm}{.2mm} Meaning:\rule{8mm}{.2mm}} \\ \end{tabular} \\

\noindent (WSJ 50)\\
{\bf But the immediate concern is the short-term credits.} \\
\noindent \begin{tabular}{@{\hspace{0mm}}lp{12.6cm}p{3.0cm}} (a) & 
Aber das unmittelbare Anliegen ist die kurzfristigen Kredite.
 & {\it Grammar:\rule{8mm}{.2mm} Meaning:\rule{8mm}{.2mm}} \\ (b) & 
Aber das unmittelbare Anliegen sind die kurzfristigen Kredite.
 & {\it Grammar:\rule{8mm}{.2mm} Meaning:\rule{8mm}{.2mm}} \\ (c) & 
Aber die unmittelbare Angelegenheit ist die kurzfristigen Kredite.
 & {\it Grammar:\rule{8mm}{.2mm} Meaning:\rule{8mm}{.2mm}} \\ (d) & 
Aber das sofortige Interesse ist die kurzfristigen Gutschriften.
 & {\it Grammar:\rule{8mm}{.2mm} Meaning:\rule{8mm}{.2mm}} \\ (e) & 
Aber die unmittelbare Sorge ist die kurzfristigen Kredite.
 & {\it Grammar:\rule{8mm}{.2mm} Meaning:\rule{8mm}{.2mm}} \\ \end{tabular} \\

\newpage
\noindent (WSJ 51)\\
{\bf Canadian manufacturers' new orders fell to \$20.80 billion (Canadian) in January, down 4\% from December's \$21.67 billion on a seasonally adjusted basis, Statistics Canada, a federal agency, said.} \\
\noindent \begin{tabular}{@{\hspace{0mm}}lp{12.6cm}p{3.0cm}} (a) & 
Neue Auftr\"{a}ge der kanadischen Hersteller fielen auf \$20,80 Milliarde (Kanadier) im Januar, unten 4\% von Dezembers \$21,67 Milliarde auf einer jahreszeitlichen Grundlage, Statistiken Kanada, ein Bundesamt, gesagt.
 & {\it Grammar:\rule{8mm}{.2mm} Meaning:\rule{8mm}{.2mm}} \\ (b) & 
Kanadischer Hersteller neue Auftr\"{a}ge seien im Januar auf 20.8 Mrd.\ (kanadische) Dollar gefallen, ein Minus von 4\% von Dezembers 21.67 Mrd.\ Dollar auf einer saisonal bereinige Basis, sagte Statistik-Kanada eine Bundesbeh\"{o}rde.
 & {\it Grammar:\rule{8mm}{.2mm} Meaning:\rule{8mm}{.2mm}} \\ (c) & 
Neuen Auftr\"{a}ge kanadischer Hersteller f\"{a}llten zu 20,80 Milliarden Dollar (Kanadier) in Januar, herunter 4\% vom \$.\ Dezember 21,67 Milliarde auf einer von seasonally eingestellten Grundlage, statistische Angaben Kanada, eine Bundesagentur sagte.
 & {\it Grammar:\rule{8mm}{.2mm} Meaning:\rule{8mm}{.2mm}} \\ (d) & 
Die neuen Disziplinen kanadischer Hersteller fielen zu \$20.80 milliard (kanadisch) in Januar, besiegen Sie 4\% von Dezembers \$21.67 milliard auf einer saisongem\"{a}{\ss} eingestellt Basis, Statistiken, die Kanada, eine Bundes Agentur, sagte.
 & {\it Grammar:\rule{8mm}{.2mm} Meaning:\rule{8mm}{.2mm}} \\ (e) & 
Neue Auftr\"{a}ge kanadischer Hersteller seien im Januar auf 20.8 Mrd.\ (kanadische) Dollar gefallen, ein Minus von 4\% im Vergleich zu 21.67 Mrd.\ Dollar im Dezember auf einer saisonal bereinigten Basis, sagte Statistik-Kanada, eine Bundesbeh\"{o}rde.
 & {\it Grammar:\rule{8mm}{.2mm} Meaning:\rule{8mm}{.2mm}} \\ (f) & 
Kanadischer Hersteller neue Auftr\"{a}ge seien im Januar auf 20.8 Mrd.\ Dollar (kanadisch) gefallen, ein Minus von 4\% aus Dezember 21.67 Mrd.\ Dollar auf einer saisonal bereinige Basis, sagte Statistik-Kanada eine Bundesbeh\"{o}rde.
 & {\it Grammar:\rule{8mm}{.2mm} Meaning:\rule{8mm}{.2mm}} \\ \end{tabular} \\

\noindent (WSJ 52)\\
{\bf The Federal Farm Credit Banks Funding Corp.\ plans to offer \$1.7 billion of bonds Thursday.} \\
\noindent \begin{tabular}{@{\hspace{0mm}}lp{12.6cm}p{3.0cm}} (a) & 
Federal Farm Credit Banks Funding Corp.\ plant, \$1,7 Milliarde von Bindungen Donnerstag anzubieten.
 & {\it Grammar:\rule{8mm}{.2mm} Meaning:\rule{8mm}{.2mm}} \\ (b) & 
Die Bundes Bauernhof-Kredit-Banken, die finanzieren, AG plant, \$1.7 milliard von Banden Donnerstag anzubieten.
 & {\it Grammar:\rule{8mm}{.2mm} Meaning:\rule{8mm}{.2mm}} \\ (c) & 
Die Federal Farm Credit Banks Funding Corp.\ plant, Donnerstag 1.7 Mrd.\ Dollar in Obligationen anzubieten.
 & {\it Grammar:\rule{8mm}{.2mm} Meaning:\rule{8mm}{.2mm}} \\ (d) & 
Die Bundesstaatlichen Bauernhofkreditbankfinanzierungs Corp.\ Pl\"{a}ne zu Angebot 1,7 Milliarden Dollar \"{U}bereinkommen-Donnerstags.
 & {\it Grammar:\rule{8mm}{.2mm} Meaning:\rule{8mm}{.2mm}} \\ (e) & 
Die Bundesbauernhofkreditbankenfinanzgesellschaftpl\"{a}ne, Donnerstag 1.7 Mrd.\ Dollar in Obligationen anzubieten.
 & {\it Grammar:\rule{8mm}{.2mm} Meaning:\rule{8mm}{.2mm}} \\ (f) & 
Die Bundesbauernhofkreditbankenfinanzgesellschaft plant, Donnerstag 1.7 Mrd.\ Dollar in Obligationen anzubieten.
 & {\it Grammar:\rule{8mm}{.2mm} Meaning:\rule{8mm}{.2mm}} \\ \end{tabular} \\

\newpage
\noindent (WSJ 53)\\
{\bf The St.\ Louis-based bank holding company previously traded on the American Stock Exchange.} \\
\noindent \begin{tabular}{@{\hspace{0mm}}lp{12.6cm}p{3.0cm}} (a) & 
Die Str.\ Louis-gest\"{u}tzte Bankholdinggesellschaft n\"{u}tzte vorher die amerikanische B\"{o}rse aus.
 & {\it Grammar:\rule{8mm}{.2mm} Meaning:\rule{8mm}{.2mm}} \\ (b) & 
Die der Sankt Louis basierte Bankholdinggesellschaft handelte fr\"{u}her an die amerikanische B\"{o}rse.
 & {\it Grammar:\rule{8mm}{.2mm} Meaning:\rule{8mm}{.2mm}} \\ (c) & 
Die in St.\ Louis basierte Bankholdinggesellschaft wurde zuvor an der amerikanischen B\"{o}rse gehandelt.
 & {\it Grammar:\rule{8mm}{.2mm} Meaning:\rule{8mm}{.2mm}} \\ 
(d) & Die St.\ Louis basierte Bankholdinggesellschaft handelte fr\"{u}her an die amerikanische B\"{o}rse.
 & {\it Grammar:\rule{8mm}{.2mm} Meaning:\rule{8mm}{.2mm}} \\ 
\end{tabular} \\ \noindent \begin{tabular}{@{\hspace{0mm}}lp{12.6cm}p{3.0cm}} 
(e) & Die Str.\ Louis-louis-based Bank, die Firma tauschte h\"{a}lt vorher, auf der amerikanischen B\"{o}rse.
 & {\it Grammar:\rule{8mm}{.2mm} Meaning:\rule{8mm}{.2mm}} \\ (f) & 
Die Str.\ Louis-based, die Bank, die Gesellschaft vorher h\"{a}lt, auf der amerikanischen B\"{o}rse tauschte.
 & {\it Grammar:\rule{8mm}{.2mm} Meaning:\rule{8mm}{.2mm}} \\ \end{tabular}  \\

\noindent (WSJ 54)\\
{\bf The transaction is expected to be completed by May 1.} \\
\noindent \begin{tabular}{@{\hspace{0mm}}lp{12.6cm}p{3.0cm}} (a) & 
Die Verhandlung wird erwartet, durch den 1.\ Mai vervollst\"{a}ndigt zu werden.
 & {\it Grammar:\rule{8mm}{.2mm} Meaning:\rule{8mm}{.2mm}} \\ (b) & 
Die Verhandlung wird erwartet, bis Mai 1 durchgef\"{u}hrt zu werden.
 & {\it Grammar:\rule{8mm}{.2mm} Meaning:\rule{8mm}{.2mm}} \\ (c) & 
Es wird erwartet, da{\ss} die Transaktion bis zum erstem Mai abgeschlossen wird.
 & {\it Grammar:\rule{8mm}{.2mm} Meaning:\rule{8mm}{.2mm}} \\ (d) & 
Es wird bis zum erstem Mai erwartet, da{\ss} die Transaktion abgeschlossen ist.
 & {\it Grammar:\rule{8mm}{.2mm} Meaning:\rule{8mm}{.2mm}} \\ (e) & 
Die Transaktion soll im 1.\ Mai beendet werden.
 & {\it Grammar:\rule{8mm}{.2mm} Meaning:\rule{8mm}{.2mm}} \\ (f) & 
Es wird erwartet, da{\ss} die Transaktion bis zum 1.\ Mai abgeschlossen wird.
 & {\it Grammar:\rule{8mm}{.2mm} Meaning:\rule{8mm}{.2mm}} \\ \end{tabular}  \\

\noindent (WSJ 55)\\
{\bf A successor for him hasn't been named.} \\
\noindent \begin{tabular}{@{\hspace{0mm}}lp{12.6cm}p{3.0cm}} (a) & 
Ein Nachfolger f\"{u}r ihn ist nicht genannt worden.
 & {\it Grammar:\rule{8mm}{.2mm} Meaning:\rule{8mm}{.2mm}} \\ \end{tabular} \\

\noindent (WSJ 56)\\
{\bf The president's news conference was a much-needed step in that direction.} \\
\noindent \begin{tabular}{@{\hspace{0mm}}lp{12.6cm}p{3.0cm}} (a) & 
Die Nachrichtenkonferenz des Pr\"{a}sidenten war ein sehr notwendig Schritt in jener Richtung.
 & {\it Grammar:\rule{8mm}{.2mm} Meaning:\rule{8mm}{.2mm}} \\ (b) & 
Die Nachricht-Konferenz des Pr\"{a}sidenten war ein viel-gebraucht Schritt in jenem direction.
 & {\it Grammar:\rule{8mm}{.2mm} Meaning:\rule{8mm}{.2mm}} \\ (c) & 
Des Pr\"{a}sidenten Nachrichtenkonferenz war ein dringend notwendiger Schritt in dieser Richtung.
 & {\it Grammar:\rule{8mm}{.2mm} Meaning:\rule{8mm}{.2mm}} \\ 
\end{tabular} \\ \noindent \begin{tabular}{@{\hspace{0mm}}lp{12.6cm}p{3.0cm}}
(d) & Die Nachrichtenkonferenz des Pr\"{a}sidenten war ein dringend notwendiger Schritt in diese Richtung.
 & {\it Grammar:\rule{8mm}{.2mm} Meaning:\rule{8mm}{.2mm}} \\ (e) & 
Die Nachrichtenkonferenz des Pr\"{a}sidenten war ein dringend ben\"{o}tigter Schritt in dieser Richtung.
 & {\it Grammar:\rule{8mm}{.2mm} Meaning:\rule{8mm}{.2mm}} \\ \end{tabular} \\

\noindent (WSJ 57)\\
{\bf The Tokyo exchange was closed Saturday as part of its regular holiday schedule.} \\
\noindent \begin{tabular}{@{\hspace{0mm}}lp{12.6cm}p{3.0cm}} (a) & 
Die Tokyo-B\"{o}rse wurde Sonnabend als Teil seines normalen Feiertagzeitplans geschlossen.
 & {\it Grammar:\rule{8mm}{.2mm} Meaning:\rule{8mm}{.2mm}} \\ (b) & 
Der Tokyoaustausch war geschlossener Samstag als Teil seines regelm\"{a}{\ss}igen Feiertagzeitplanes.
 & {\it Grammar:\rule{8mm}{.2mm} Meaning:\rule{8mm}{.2mm}} \\ (c) & 
Der Tokio Austausch wurde Samstag als Teil seines regelm\"{a}{\ss}igen Feiertagplanes geschlossen.
 & {\it Grammar:\rule{8mm}{.2mm} Meaning:\rule{8mm}{.2mm}} \\ (d) & 
The, Tokyo Tausch wurde Samstag als Teil seines regul\"{a}ren holidayschedule geschlossen.
 & {\it Grammar:\rule{8mm}{.2mm} Meaning:\rule{8mm}{.2mm}} \\ (e) & 
Die Tokyoter B\"{o}rse wurde Sonnabend als Teil ihres normalen Feiertagszeitplans geschlossen.
 & {\it Grammar:\rule{8mm}{.2mm} Meaning:\rule{8mm}{.2mm}} \\ \end{tabular} \\

\noindent (WSJ 58)
{\bf Pan Am said its full-year results were hurt by foreign currency exchange losses of \$46.8 million, primarily related to Japanese yen debt, compared with \$11.1 million in 1985.} \\
\noindent \begin{tabular}{@{\hspace{0mm}}lp{12.6cm}p{3.0cm}} (a) & 
Pan Am sagte, seine Ganzjahresresultate seien durch 46.8 Mio.\ Dollar, verglichen mit 11.1 Mio.\ Dollar 1985, in ausl\"{a}ndischen W\"{a}hrungswechselverlusten in erster Linie zur japanischer Yenschuld bezogen geschadet worden.
 & {\it Grammar:\rule{8mm}{.2mm} Meaning:\rule{8mm}{.2mm}} \\ (b) & 
Pan Am, die seine Volljahr-Resultate besagt ist, wurden durch die Austauschverluste der ausl\"{a}ndischen W\"{a}hrung von \$46,8 Million verletzt, haupts\"{a}chlich bezogen auf der japanischen Yenschuld, verglichen mit \$11,1 Million 1985.
 & {\it Grammar:\rule{8mm}{.2mm} Meaning:\rule{8mm}{.2mm}} \\ (c) & 
Pfanne wird gesagt, da{\ss} seine voll-Jahr-Ergebnisse durch fremde currencyexchange-Verluste von \$46.8 million verletzt wurden, haupts\"{a}chlich erz\"{a}hlt zu Japanisch Yen-Schuld, die mit \$11.1 million in 1985 verglichen wurde.
 & {\it Grammar:\rule{8mm}{.2mm} Meaning:\rule{8mm}{.2mm}} \\ (d) & 
Pan Am sagte, seine vollen Jahrresultate seien durch 46.8 Mio.\ Dollar, verglichen 1985 mit 11.1 Mio.\ Dollar, in ausl\"{a}ndischen W\"{a}hrungswechselverlusten in erster Linie zur japanischer Yenschuld bezogen geschadet worden.
 & {\it Grammar:\rule{8mm}{.2mm} Meaning:\rule{8mm}{.2mm}} \\ (e) & 
Pfanne wird gesagt, da{\ss} seine voll-Jahr-Ergebnisse von Devisen-Austausch-Verlusten von 46,8 Millionen Dollar weh getan wurde, vorwiegend mit japanischer Yenschuld zusammenhing, verglichen 1985 mit 11,1 Millionen Dollar.
 & {\it Grammar:\rule{8mm}{.2mm} Meaning:\rule{8mm}{.2mm}} \\ (f) & 
Pan Am sagte, seine Ganzjahresresultate seien durch ausl\"{a}ndische W\"{a}hrungswechselverluste in H\"{o}he von 46.8 Mio.\ Dollar beeintr\"{a}chtigt worden, insbesondere durch Schulden in japanischen Yen, verglichen mit 11.1 Mio.\ Dollar im Jahre 1985.
 & {\it Grammar:\rule{8mm}{.2mm} Meaning:\rule{8mm}{.2mm}} \\ \end{tabular} \\

\newpage
\noindent (WSJ 59)\\
{\bf The American hospital sector spent \$181 billion in 1986.} \\
\noindent \begin{tabular}{@{\hspace{0mm}}lp{12.6cm}p{3.0cm}} (a) & 
Der amerikanische Krankenhaussektor gab 1986 181 Mrd.\ Dollar aus.
 & {\it Grammar:\rule{8mm}{.2mm} Meaning:\rule{8mm}{.2mm}} \\ (b) & 
Der amerikanische Krankenhaussektor gab 181 Milliarden Dollar 1986 aus.
 & {\it Grammar:\rule{8mm}{.2mm} Meaning:\rule{8mm}{.2mm}} \\ (c) & 
Der amerikanische Krankenhaussektor wendete \$181 Milliarde 1986 auf.
 & {\it Grammar:\rule{8mm}{.2mm} Meaning:\rule{8mm}{.2mm}} \\ (d) & 
Der amerikanische Krankenhaus-Sektor gab \$181 milliard in 1986 aus.
 & {\it Grammar:\rule{8mm}{.2mm} Meaning:\rule{8mm}{.2mm}} \\ \end{tabular}  \\

\noindent (WSJ 60)\\
{\bf But the company inquiry, which began a few months ago, changed everything.} \\
\noindent \begin{tabular}{@{\hspace{0mm}}lp{12.6cm}p{3.0cm}} (a) & 
Aber die Firmenanfrage, die vor einigem Monaten anfing, \"{a}nderte alles.
 & {\it Grammar:\rule{8mm}{.2mm} Meaning:\rule{8mm}{.2mm}} \\ (b) & 
Aber die Gesellschaftsanfrage, es einem wenigen Monat vor begann, \"{a}nderte alles.
 & {\it Grammar:\rule{8mm}{.2mm} Meaning:\rule{8mm}{.2mm}} \\ (c) & 
Aber die Firmenuntersuchung, die vor wenigen Monaten begann, \"{a}nderte alles.
 & {\it Grammar:\rule{8mm}{.2mm} Meaning:\rule{8mm}{.2mm}} \\ (d) & 
Aber die Gesellschaftsanfrage, die es einem wenigen Monat vor begann, \"{a}nderte alles.
 & {\it Grammar:\rule{8mm}{.2mm} Meaning:\rule{8mm}{.2mm}} \\ (e) & 
Aber die Firmaanfrage, die vor einigen Monaten anfing, \"{a}nderte alles.
 & {\it Grammar:\rule{8mm}{.2mm} Meaning:\rule{8mm}{.2mm}} \\ (f) & 
Aber die Gesellschaft-Anfrage, die vor einigen Monaten anfing, ver\"{a}nderte alles.
 & {\it Grammar:\rule{8mm}{.2mm} Meaning:\rule{8mm}{.2mm}} \\ \end{tabular} \\

\noindent (WSJ 61)\\
{\bf Texas Instruments rose 3 1/4 to 175 1/2.} \\
\noindent \begin{tabular}{@{\hspace{0mm}}lp{12.6cm}p{3.0cm}} (a) & 
Texas Instruments stieg um 13/4 auf 351/2.
 & {\it Grammar:\rule{8mm}{.2mm} Meaning:\rule{8mm}{.2mm}} \\ (b) & 
Texas Instrumente standen 3 1/4 bis 175 1/2 auf.
 & {\it Grammar:\rule{8mm}{.2mm} Meaning:\rule{8mm}{.2mm}} \\ (c) & 
Texas Instruments stieg 3 1/4 bis 175 1/2.
 & {\it Grammar:\rule{8mm}{.2mm} Meaning:\rule{8mm}{.2mm}} \\ (d) & 
Texas Instruments stiegen um 13/4 auf 351/2.
 & {\it Grammar:\rule{8mm}{.2mm} Meaning:\rule{8mm}{.2mm}} \\ (e) & 
Texas Instruments stieg um 3 1/4 auf 175 1/2.
 & {\it Grammar:\rule{8mm}{.2mm} Meaning:\rule{8mm}{.2mm}} \\ (f) & 
Texas Instrumente, die rosarot sind, 3 1/4 bis 175 1/2.
 & {\it Grammar:\rule{8mm}{.2mm} Meaning:\rule{8mm}{.2mm}} \\ \end{tabular} \\

\newpage
\noindent (WSJ 62)\\
{\bf If futures fell more than 0.20 point below the stocks, he would buy futures and sell stocks instead.} \\
\noindent \begin{tabular}{@{\hspace{0mm}}lp{12.6cm}p{3.0cm}} (a) & 
Wenn Zukunft mehr als 0.20 Punkt unter den Aktien f\"{a}llten, w\"{u}rdete er kaufen Sie Zukunft und verkaufen Sie Aktien stattdessen.
 & {\it Grammar:\rule{8mm}{.2mm} Meaning:\rule{8mm}{.2mm}} \\ (b) & 
Wenn Terminpapiere mehr als 0.2 Punkte unter die Aktien fielen, w\"{u}rde er Terminpapiere kaufen, und Aktien stattdessen verkaufen.
 & {\it Grammar:\rule{8mm}{.2mm} Meaning:\rule{8mm}{.2mm}} \\ (c) & 
Wenn Zukunft mehr als 0,20 Punkt unter die Vorr\"{a}te fiel, w\"{u}rde er Zukunft kaufen und Vorr\"{a}te anstatt verkaufen.
 & {\it Grammar:\rule{8mm}{.2mm} Meaning:\rule{8mm}{.2mm}} \\ 
(d) & 
Wenn Termingesch\"{a}fte mehr als 0.2 Punkt unter die Aktien fielen, w\"{u}rde er Termingesch\"{a}fte kaufen, und Aktien stattdessen verkaufen.
 & {\it Grammar:\rule{8mm}{.2mm} Meaning:\rule{8mm}{.2mm}} \\ 
\end{tabular} \\ \noindent \begin{tabular}{@{\hspace{0mm}}lp{12.6cm}p{3.0cm}}
(e) & 
Wenn Termingesch\"{a}fte mehr als 0.2 Punkt unter den Aktien fielen, w\"{u}rde er Termingesch\"{a}fte kaufen, und Aktien stattdessen verkaufen.
 & {\it Grammar:\rule{8mm}{.2mm} Meaning:\rule{8mm}{.2mm}} \\ (f) & 
Wenn Zukunft mehr f\"{a}llt als 0,20 unter den Anteilen richten, w\"{u}rde er Zukunft kaufen und Anteile stattdessen verkaufen.
 & {\it Grammar:\rule{8mm}{.2mm} Meaning:\rule{8mm}{.2mm}} \\ \end{tabular} \\

\noindent (WSJ 63)\\
{\bf "I'm willing to work for a foreign company," says the father of two.} \\
\noindent \begin{tabular}{@{\hspace{0mm}}lp{12.6cm}p{3.0cm}} (a) & 
"Ich bin bereit, f\"{u}r eine fremde Firma zu arbeiten," sage den Vater von zwei.
 & {\it Grammar:\rule{8mm}{.2mm} Meaning:\rule{8mm}{.2mm}} \\ (b) & 
Ich bin bereit zu arbeiten f\"{u}r eine ausl\"{a}ndische Gesellschaft sagt der Vater von 2.
 & {\it Grammar:\rule{8mm}{.2mm} Meaning:\rule{8mm}{.2mm}} \\ (c) & 
Ich sei bereit, f\"{u}r eine ausl\"{a}ndische Gesellschaft zu arbeiten, sagt der Vater vom 2.
 & {\it Grammar:\rule{8mm}{.2mm} Meaning:\rule{8mm}{.2mm}} \\ (d) & 
"Ich, der wollen arbeiten f\"{u}r eine fremde Gesellschaft," sagt den Vater von zwei.
 & {\it Grammar:\rule{8mm}{.2mm} Meaning:\rule{8mm}{.2mm}} \\ (e) & 
" ich bin bereit, f\"{u}r eine Auslandsgesellschaft zu arbeiten, " sagt den Vater von zwei.
 & {\it Grammar:\rule{8mm}{.2mm} Meaning:\rule{8mm}{.2mm}} \\ (f) & 
"Ich bin bereit, f\"{u}r eine ausl\"{a}ndische Gesellschaft zu arbeiten", sagt der Vater von zweien.
 & {\it Grammar:\rule{8mm}{.2mm} Meaning:\rule{8mm}{.2mm}} \\ \end{tabular} \\

\noindent (WSJ 64)\\
{\bf It estimated that sales totaled \$217 million, compared with the year-earlier \$257 million.} \\
\noindent \begin{tabular}{@{\hspace{0mm}}lp{12.6cm}p{3.0cm}} (a) & 
Es sch\"{a}tzte ab, da{\ss} Verk\"{a}ufe zusammen 217 Millionen Dollar z\"{a}hlten, verglichen 257 Million mit den jahresfr\"{u}heren \$.
 & {\it Grammar:\rule{8mm}{.2mm} Meaning:\rule{8mm}{.2mm}} \\ (b) & 
Es sch\"{a}tzte, da{\ss} Verk\"{a}ufe \$217 million zusammenz\"{a}hlten, verglich mit das Jahr-fr\"{u}her \$257 million.
 & {\it Grammar:\rule{8mm}{.2mm} Meaning:\rule{8mm}{.2mm}} \\
\end{tabular} \\ \noindent \begin{tabular}{@{\hspace{0mm}}lp{12.6cm}p{3.0cm}}
(c) &
Es sch\"{a}tzte, da{\ss} Verk\"{a}ufe 217 Mio.\ Dollar, verglichen mit dem 257 Mio.\ Dollar im Jahr zuvor, betrugen.
 & {\it Grammar:\rule{8mm}{.2mm} Meaning:\rule{8mm}{.2mm}} \\ (d) & 
Sie sch\"{a}tzte, da{\ss} Verk\"{a}ufe \$217 Million zusammenz\"{a}hlten, mit dem Jahr-fr\"{u}h \$257 Million verglichen.
 & {\it Grammar:\rule{8mm}{.2mm} Meaning:\rule{8mm}{.2mm}} \\ (e) & 
Es sch\"{a}tzte, da{\ss} Verk\"{a}ufe insgesamt 217 Mio.\ Dollar betrugen, verglichen mit 257 Mio.\ Dollar im Jahr zuvor.
 & {\it Grammar:\rule{8mm}{.2mm} Meaning:\rule{8mm}{.2mm}} \\ \end{tabular} \\

\newpage
\noindent (WSJ 65)\\
{\bf With the falling dollar, U.S.\ manufacturers are in a pretty good position to compete on world markets.} \\
\noindent \begin{tabular}{@{\hspace{0mm}}lp{12.6cm}p{3.0cm}} (a) & 
Mit dem fallenden Dollar sind US Hersteller in einer ziemlich guten Position, um auf Weltm\"{a}rkte zu konkurrieren.
 & {\it Grammar:\rule{8mm}{.2mm} Meaning:\rule{8mm}{.2mm}} \\ (b) & 
Mit dem fallenden Dollar sind die Hersteller der USA in der ziemlich guten Lage, auf Weltm\"{a}rkten zu konkurrieren.
 & {\it Grammar:\rule{8mm}{.2mm} Meaning:\rule{8mm}{.2mm}} \\ (c) & 
Mit dem fallend Dollar sind AMERIKANISCHE Hersteller in einer ganz guten Position, auf Welt-M\"{a}rkten zu konkurrieren.
 & {\it Grammar:\rule{8mm}{.2mm} Meaning:\rule{8mm}{.2mm}} \\ (d) & 
Mit dem fallenden Dollar sind-- US-Hersteller in einer h\"{u}bschen guten Position zum Konkurrieren in den Weltm\"{a}rkten.
 & {\it Grammar:\rule{8mm}{.2mm} Meaning:\rule{8mm}{.2mm}} \\ (e) & 
Mit dem fallenden Dollar sind US-Hersteller in einer ziemlich guten Position, auf Weltm\"{a}rkten zu konkurrieren.
 & {\it Grammar:\rule{8mm}{.2mm} Meaning:\rule{8mm}{.2mm}} \\ (f) & 
Mit dem fallenden Dollar sind US Hersteller in einer ziemlich guten Position, auf Weltm\"{a}rkte zu konkurrieren.
 & {\it Grammar:\rule{8mm}{.2mm} Meaning:\rule{8mm}{.2mm}} \\ \end{tabular} \\

\noindent (WSJ 66)\\
{\bf You go on to Bank B and get another loan, using some of Bank B's proceeds to pay back some of your debt to Bank A.} \\
\noindent \begin{tabular}{@{\hspace{0mm}}lp{12.6cm}p{3.0cm}} (a) & 
Sie gehen zu Bank B \"{u}ber und bekommen ein anderes Darlehen, was einige von Bank B Erl\"{o}sen benutzt, um etwas von Ihrer Schuld zu Bank A zur\"{u}ckzuzahlen.
 & {\it Grammar:\rule{8mm}{.2mm} Meaning:\rule{8mm}{.2mm}} \\ (b) & 
Sie gehen weiter, B zu \"{u}berh\"{o}hen und noch einen Kredit zu bekommen, das Benutzen von einigen von Bank B's geht weiter, einige Ihrer Schuld zur\"{u}ckzuzahlen, um A.\ zu \"{u}berh\"{o}hen
 & {\it Grammar:\rule{8mm}{.2mm} Meaning:\rule{8mm}{.2mm}} \\ (c) & 
**abend** Sie fortfahren on zu haben b und erhalten ein Darlehen, mit einig von Ertrag der Bank b zu zahlen zur\"{u}ck etwas von Ihr Schuld zu Bank a.
 & {\it Grammar:\rule{8mm}{.2mm} Meaning:\rule{8mm}{.2mm}} \\ (d) & 
Man f\"{a}hrt auf Bank-b fort, und bekommt ein weiter Kredit einige von Bank-bs Ertr\"{a}ge, einige von deiner Schuld zur  zur\"{u}ckzuzahlen, benutzend.
 & {\it Grammar:\rule{8mm}{.2mm} Meaning:\rule{8mm}{.2mm}} \\ (e) & 
Man f\"{a}hrt zur Bank b fort, und bekommt ein weiter Kredit, um einige von deiner Schuld zur Bank a zur\"{u}ckzuzahlen, einige von Bank-bs Ertr\"{a}ge benutzend.
 & {\it Grammar:\rule{8mm}{.2mm} Meaning:\rule{8mm}{.2mm}} \\ (f) & 
Man wendet sich dann an Bank B und bekommt dort einen weiteren Kredit, den man dann teilweise dazu benutzt, einen Teil der Schulden bei Bank A zur\"{u}ckzuzahlen.
 & {\it Grammar:\rule{8mm}{.2mm} Meaning:\rule{8mm}{.2mm}} \\ \end{tabular} \\

\newpage
\noindent (WSJ 67)\\
{\bf The critical point is to know when Brazil will produce an economic policy to generate foreign exchange to pay its interest.} \\
\noindent \begin{tabular}{@{\hspace{0mm}}lp{12.6cm}p{3.0cm}} (a) & 
Der kritische Punkt soll wissen, wann Brasilien eine Wirtschaftspolitik produziert, um Devisenkurs zu erzeugen, um sein Interesse zu zahlen.
 & {\it Grammar:\rule{8mm}{.2mm} Meaning:\rule{8mm}{.2mm}} \\ (b) & 
Der kritische Punkt ist, um zu kennen, wann Brasilien eine Wirtschaftspolitik produzieren wird, um ausl\"{a}ndischen Wechsel zu erzeugen, um seinen Zins zu zahlen.
 & {\it Grammar:\rule{8mm}{.2mm} Meaning:\rule{8mm}{.2mm}} \\ (c) & 
Der kritische Punkt sollte wissen, wenn Brasilien einen economicpolicy produzieren wird, um fremden Tausch zu erzeugen, um sein Interesse zu zahlen.
 & {\it Grammar:\rule{8mm}{.2mm} Meaning:\rule{8mm}{.2mm}} \\ (d) & 
Der kritische Punkt ist, zu wissen, wann Brazil, um, um seinen Zins zu zahlen, ausl\"{a}ndischen Wechsel zu erzeugen, eine Wirtschaftspolitik produzieren wird.
 & {\it Grammar:\rule{8mm}{.2mm} Meaning:\rule{8mm}{.2mm}} \\ (e) & 
Der kritische Punkt ist zu wissen, wann Brasilien eine Wirtschaftspolitik aufstellen wird, um Au{\ss}enhandel zu erzeugen, um seine Zinsen zu zahlen.
 & {\it Grammar:\rule{8mm}{.2mm} Meaning:\rule{8mm}{.2mm}} \\ (f) & 
Der kritische Punkt ist, zu wissen, wann Brasilien eine wirtschaftliche Politik produziert, um Devisen zu generieren, um seine Zinsen zu bezahlen.
 & {\it Grammar:\rule{8mm}{.2mm} Meaning:\rule{8mm}{.2mm}} \\ \end{tabular} \\

\noindent (WSJ 68)\\
{\bf Manufacturers' shipments followed the same trend, falling 1.5\% in January to \$21.08 billion, after a 2.8\% increase the previous month.} \\
\noindent \begin{tabular}{@{\hspace{0mm}}lp{12.6cm}p{3.0cm}} (a) & 
Die Sendungen von Herstellern folgten der gleichen Tendenz, fallenden 1,5\% im Januar zu 21,08 Milliarden Dollar, nachdem ein 2,8\% den vorhergehenden Monat vergr\"{o}{\ss}ert.
 & {\it Grammar:\rule{8mm}{.2mm} Meaning:\rule{8mm}{.2mm}} \\ 
\end{tabular} \\ \noindent \begin{tabular}{@{\hspace{0mm}}lp{12.6cm}p{3.0cm}}
(b) & Herstellersendungen folgten dem gleichen Trend und fielen im Januar um 1.5\% auf 21.08 Mrd.\ Dollar nach einer 2.8\% Zunahme im vorherigen Monat.
 & {\it Grammar:\rule{8mm}{.2mm} Meaning:\rule{8mm}{.2mm}} \\ 
\end{tabular} \\ \noindent \begin{tabular}{@{\hspace{0mm}}lp{12.6cm}p{3.0cm}}
(c) & Die Sendungen Hersteller folgten dem gleichen Trend, beim Fallen, 1.5\% in Januar zu \$21.08 milliard, nach einem 2.8\% Zunahme der vorausgehende Monat.
 & {\it Grammar:\rule{8mm}{.2mm} Meaning:\rule{8mm}{.2mm}} \\ 
\end{tabular} \\ \noindent \begin{tabular}{@{\hspace{0mm}}lp{12.6cm}p{3.0cm}}
(d) & Hersteller Sendungen folgten den gleichen Trend in Januar auf 21.08 Mrd.\ Dollar 1.5\% fallend nach einer 2.8\% Zunahme dem vorherig Monat.
 & {\it Grammar:\rule{8mm}{.2mm} Meaning:\rule{8mm}{.2mm}} \\ 
\end{tabular} \\ \noindent \begin{tabular}{@{\hspace{0mm}}lp{12.6cm}p{3.0cm}}
(e) & Hersteller Sendungen im Januar 1.5\% auf 21.08 Mrd.\ Dollar fallend folgten den gleichen Trend nach einer 2.8\% Zunahme der vorherige Monat.
 & {\it Grammar:\rule{8mm}{.2mm} Meaning:\rule{8mm}{.2mm}} \\ (f) & 
Versand der Hersteller folgte der gleichen Tendenz, fallenden 1,5\% im Januar bis \$21,08 Milliarde, nachdem eine Zunahme 2,8\% der vorhergehende Monat.
 & {\it Grammar:\rule{8mm}{.2mm} Meaning:\rule{8mm}{.2mm}} \\ \end{tabular} \\

\newpage
\noindent (WSJ 69)\\
{\bf The offerings will be made through the corporation and a nationwide group of securities dealers and dealer banks.} \\
\noindent \begin{tabular}{@{\hspace{0mm}}lp{12.6cm}p{3.0cm}} (a) & 
Die Angebote werden durch die Gesellschaft und eine landesweit Gruppe von Wertpapierh\"{a}ndlern und H\"{a}ndler Banken gemacht werden.
 & {\it Grammar:\rule{8mm}{.2mm} Meaning:\rule{8mm}{.2mm}} \\ 
\end{tabular} \\ \noindent \begin{tabular}{@{\hspace{0mm}}lp{12.6cm}p{3.0cm}}
(b) & Die Gaben werden durch die Firma gemacht werden, und anationwide gruppieren von Sicherheiten-H\"{a}ndlern und H\"{a}ndler-Banken.
 & {\it Grammar:\rule{8mm}{.2mm} Meaning:\rule{8mm}{.2mm}} \\ 
\end{tabular} \\ \noindent \begin{tabular}{@{\hspace{0mm}}lp{12.6cm}p{3.0cm}}
(c) & Zu den Angeboten wird durch das Unternehmen und eine nationwide Gruppe von Wertpapierh\"{a}ndlern und H\"{a}ndlerbanken gemacht.
 & {\it Grammar:\rule{8mm}{.2mm} Meaning:\rule{8mm}{.2mm}} \\ (d) & 
Die Angebote werden durch die Gesellschaft und eine landesweite Gruppe von Wertpapierh\"{a}ndlern und H\"{a}ndlerbanken gemacht werden.
 & {\it Grammar:\rule{8mm}{.2mm} Meaning:\rule{8mm}{.2mm}} \\ (e) & 
Die Opfer werden durch die Corporation und eine allgemein Gruppe Wertpapierh\"{a}ndler und H\"{a}ndlerbanken gebildet.
 & {\it Grammar:\rule{8mm}{.2mm} Meaning:\rule{8mm}{.2mm}} \\ \end{tabular} \\

\noindent (WSJ 70)\\
{\bf The real estate services company formerly traded over the counter.} \\
\noindent \begin{tabular}{@{\hspace{0mm}}lp{12.6cm}p{3.0cm}} (a) & 
Die wirkliche Gut-Dienste-Gesellschaft tauschte \"{u}ber thecounter ehemals.
 & {\it Grammar:\rule{8mm}{.2mm} Meaning:\rule{8mm}{.2mm}} \\ (b) & 
Die Immobilienservicegesellschaft handelte fr\"{u}her im Freiverkehr.
 & {\it Grammar:\rule{8mm}{.2mm} Meaning:\rule{8mm}{.2mm}} \\
\end{tabular} \\ \noindent \begin{tabular}{@{\hspace{0mm}}lp{12.6cm}p{3.0cm}}
(c) & Die Immobilienservicegesellschaft wurde fr\"{u}her im Freiverkehr gehandelt.
 & {\it Grammar:\rule{8mm}{.2mm} Meaning:\rule{8mm}{.2mm}} \\ 
\end{tabular} \\ \noindent \begin{tabular}{@{\hspace{0mm}}lp{12.6cm}p{3.0cm}}
(d) & Die Immobilien warten Firma, die fr\"{u}her \"{u}ber dem Z\"{a}hler getauscht war.
 & {\it Grammar:\rule{8mm}{.2mm} Meaning:\rule{8mm}{.2mm}} \\ (e) & 
Die Immobilienservice-Firma tauschte fr\"{u}her \"{u}ber dem Kostenz\"{a}hler.
 & {\it Grammar:\rule{8mm}{.2mm} Meaning:\rule{8mm}{.2mm}} \\ \end{tabular} \\

\noindent (WSJ 71)\\
{\bf The shares are convertible at a rate of \$9.50 of preferred for each common share.} \\
\noindent \begin{tabular}{@{\hspace{0mm}}lp{12.6cm}p{3.0cm}} (a) & 
Die Anteile sind mit einer Rate von \$9,50 von bevorzugt f\"{u}r jeden allgemeinen Anteil umwandelbar.
 & {\it Grammar:\rule{8mm}{.2mm} Meaning:\rule{8mm}{.2mm}} \\ (b) & 
Die Anteile sind umwandelbar bei einer Rate von \$9.50 von, zog vor f\"{u}r jeden gew\"{o}hnlichen Anteil.
 & {\it Grammar:\rule{8mm}{.2mm} Meaning:\rule{8mm}{.2mm}} \\ (c) & 
Die Aktien sind zu einem Kurs von 9.5 Dollar in bevorzugter pro Stammaktie konvertierbar.
 & {\it Grammar:\rule{8mm}{.2mm} Meaning:\rule{8mm}{.2mm}} \\ (d) & 
Die Aktien sind zu einem Kurs von 9.50 Dollar in Vorz\"{u}gen pro Stammaktie konvertierbar.
 & {\it Grammar:\rule{8mm}{.2mm} Meaning:\rule{8mm}{.2mm}} \\ (e) & 
Die Anteile sind bei einer Geschwindigkeit von 9,50 Dollar f\"{u}r jeden weitverbreiteten Anteil konvertibel von vorgezogen.
 & {\it Grammar:\rule{8mm}{.2mm} Meaning:\rule{8mm}{.2mm}} \\ (f) & 
Die Aktien sind konvertierbar zu einem Kurs von 9.5 Dollar von bevorzugt pro Stammaktie.
 & {\it Grammar:\rule{8mm}{.2mm} Meaning:\rule{8mm}{.2mm}} \\ \end{tabular} \\

\newpage
\noindent (WSJ 72)\\
{\bf He continues as president of the company's corporate division.} \\
\noindent \begin{tabular}{@{\hspace{0mm}}lp{12.6cm}p{3.0cm}} (a) & 
Er setzt als Pr\"{a}sident von der korporativen Teilung der Gesellschaft fort.
 & {\it Grammar:\rule{8mm}{.2mm} Meaning:\rule{8mm}{.2mm}} \\ (b) & 
Er macht als Pr\"{a}sident der Gesellschaft Firmenabteilung weiter.
 & {\it Grammar:\rule{8mm}{.2mm} Meaning:\rule{8mm}{.2mm}} \\ (c) & 
Er macht als Pr\"{a}sident der k\"{o}rperschaftlichen Division der Firma weiter.
 & {\it Grammar:\rule{8mm}{.2mm} Meaning:\rule{8mm}{.2mm}} \\ (d) & 
Er macht als Pr\"{a}sident der Firmenabteilung des Unternehmens weiter.
 & {\it Grammar:\rule{8mm}{.2mm} Meaning:\rule{8mm}{.2mm}} \\ (e) & 
Er f\"{a}hrt als Pr\"{a}sident der korporativen Abteilung der Firma fort.
 & {\it Grammar:\rule{8mm}{.2mm} Meaning:\rule{8mm}{.2mm}} \\ \end{tabular} \\

\noindent (WSJ 73)\\
{\bf In tests it has worked for many heart-attack victims.} \\
\noindent \begin{tabular}{@{\hspace{0mm}}lp{12.6cm}p{3.0cm}} (a) & 
In den Tests, die es f\"{u}r viele funktioniert hat, Opfer Herz-angreifen.
 & {\it Grammar:\rule{8mm}{.2mm} Meaning:\rule{8mm}{.2mm}} \\ (b) & 
In Pr\"{u}fungen hat es f\"{u}r viele Herz-Angriffs-Opfer gearbeitet.
 & {\it Grammar:\rule{8mm}{.2mm} Meaning:\rule{8mm}{.2mm}} \\ (c) & 
In Tests hat es f\"{u}r viele Herzinfarktopfer funktioniert.
 & {\it Grammar:\rule{8mm}{.2mm} Meaning:\rule{8mm}{.2mm}} \\ (d) & 
In Pr\"{u}fungen hat es f\"{u}r viele Herz-Angriff-Opfer gearbeitet.
 & {\it Grammar:\rule{8mm}{.2mm} Meaning:\rule{8mm}{.2mm}} \\ (e) & 
In Tests hat es f\"{u}r viele Herzinfarktopfer gearbeitet.
 & {\it Grammar:\rule{8mm}{.2mm} Meaning:\rule{8mm}{.2mm}} \\ \end{tabular} \\

\noindent (WSJ 74)\\
{\bf The percentage change is since year-end.} \\
\noindent \begin{tabular}{@{\hspace{0mm}}lp{12.6cm}p{3.0cm}} (a) & 
Die Prozentsatz\"{A}nderung ist seit Jahresende.
 & {\it Grammar:\rule{8mm}{.2mm} Meaning:\rule{8mm}{.2mm}} \\ (b) & 
Die Prozentsatz\"{a}nderung ist seit Jahresende.
 & {\it Grammar:\rule{8mm}{.2mm} Meaning:\rule{8mm}{.2mm}} \\ (c) & 
Die Prozent \"{A}nderung ist seit Jahr-Spitze.
 & {\it Grammar:\rule{8mm}{.2mm} Meaning:\rule{8mm}{.2mm}} \\ (d) & 
Die Prozentsatz\"{a}nderung ist seit year-end.
 & {\it Grammar:\rule{8mm}{.2mm} Meaning:\rule{8mm}{.2mm}} \\ (e) & 
Die Prozentsatz\"{a}nderung ist seit Jahres-Ende.
 & {\it Grammar:\rule{8mm}{.2mm} Meaning:\rule{8mm}{.2mm}} \\ \end{tabular} \\

\newpage
\noindent (WSJ 75)\\
{\bf He said Pan Am currently has "in excess of \$150 million" in cash.} \\
\noindent \begin{tabular}{@{\hspace{0mm}}lp{12.6cm}p{3.0cm}} (a) & 
Er sagte, da{\ss} Pan Am z.Z.\ " \"{u}ber \$150 Million " im Bargeld hat.
 & {\it Grammar:\rule{8mm}{.2mm} Meaning:\rule{8mm}{.2mm}} \\ (b) & 
Er sagte, Pan Am habe gegenw\"{a}rtig \"{u}ber 150 Mio.\ Dollar in Bargeld.
 & {\it Grammar:\rule{8mm}{.2mm} Meaning:\rule{8mm}{.2mm}} \\ (c) & 
Er sagte, da{\ss} Pfannenvormittags zur Zeit "mehr als 150 Millionen Dollar" in Bargeld hat.
 & {\it Grammar:\rule{8mm}{.2mm} Meaning:\rule{8mm}{.2mm}} \\ (d) & 
Er sagte, da{\ss} Pfanne ist, hat gegenw\"{a}rtig-" in \"{U}berschu{\ss} von \$150 million" in in bar.
 & {\it Grammar:\rule{8mm}{.2mm} Meaning:\rule{8mm}{.2mm}} \\ \end{tabular} \\

\noindent (WSJ 76)\\
{\bf These revenues could then have been used for many purposes, such as funding those people without medical insurance.} \\
\noindent \begin{tabular}{@{\hspace{0mm}}lp{12.6cm}p{3.0cm}} (a) & 
Diese Einkommen h\"{a}tten dann f\"{u}r viele Zwecke benutzt werden k\"{o}nnen, wie die Finanzierung jener Menschen ohne medizinische Versicherung.
 & {\it Grammar:\rule{8mm}{.2mm} Meaning:\rule{8mm}{.2mm}} \\ (b) & 
Diese Einkommen konnten f\"{u}r viele Zwecke, wie Finanzierung jener Leute ohne medizinische Versicherung dann benutzt worden sein.
 & {\it Grammar:\rule{8mm}{.2mm} Meaning:\rule{8mm}{.2mm}} \\ (c) & 
Diese Einnahmen h\"{a}tten dann f\"{u}r viele Zwecke benutzt werden k\"{o}nnen, wie z.B.\ der Bezuschussung von Leuten ohne Krankenversicherung.
 & {\it Grammar:\rule{8mm}{.2mm} Meaning:\rule{8mm}{.2mm}} \\ 
\end{tabular} \\ \noindent \begin{tabular}{@{\hspace{0mm}}lp{12.6cm}p{3.0cm}}
(d) & Diese Einnahmen gekonnt werden dann f\"{u}r viele Zwecke benutzt werden wie Finanzen diese Leute ohne die Krankenversicherung.
 & {\it Grammar:\rule{8mm}{.2mm} Meaning:\rule{8mm}{.2mm}} \\ 
\end{tabular} \\ \noindent \begin{tabular}{@{\hspace{0mm}}lp{12.6cm}p{3.0cm}}
(e) & Diese Einnahmen gekonnt werden dann f\"{u}r viele Zwecke wie finanzierend diese Leute ohne Krankenversicherung benutzt werden.
 & {\it Grammar:\rule{8mm}{.2mm} Meaning:\rule{8mm}{.2mm}} \\ (f) & 
Diese Einnahmen k\"{o}nnten f\"{u}r viel purposes,such als das Finanzieren jener Leute ohne medizinische Versicherung dann benutzt worden sein.
 & {\it Grammar:\rule{8mm}{.2mm} Meaning:\rule{8mm}{.2mm}} \\ \end{tabular} \\

\noindent (WSJ 77)\\
{\bf The case continues in U.S.\ Bankruptcy Court in St.\ Louis.} \\
\noindent \begin{tabular}{@{\hspace{0mm}}lp{12.6cm}p{3.0cm}} (a) & 
Der Fall macht im US Konkursgericht in St.\ Louis weiter.
 & {\it Grammar:\rule{8mm}{.2mm} Meaning:\rule{8mm}{.2mm}} \\ (b) & 
Der Fall macht in USA weiter.\ Konkurs-Hof in Str.\ Louis.
 & {\it Grammar:\rule{8mm}{.2mm} Meaning:\rule{8mm}{.2mm}} \\ (c) & 
Der Fall macht in das US Konkursgericht in St.\ Louis weiter.
 & {\it Grammar:\rule{8mm}{.2mm} Meaning:\rule{8mm}{.2mm}} \\ (d) & 
Der Fall wird am Konkursgericht in St.\ Louis fortgesetzt.
 & {\it Grammar:\rule{8mm}{.2mm} Meaning:\rule{8mm}{.2mm}} \\ (e) & 
Der Fall setzt in AMERIKANISCHEM Bankrott-Gericht in Str.\ Louis fort.
 & {\it Grammar:\rule{8mm}{.2mm} Meaning:\rule{8mm}{.2mm}} \\ (f) & 
Der Fall f\"{a}hrt in US fort.\ Konkursgericht in St.\ Louis.
 & {\it Grammar:\rule{8mm}{.2mm} Meaning:\rule{8mm}{.2mm}} \\ \end{tabular} \\

\newpage
\noindent (WSJ 78)\\
{\bf American Express fell 1 1/2 to 77 1/4 on more than 2.1 million shares.} \\
\noindent \begin{tabular}{@{\hspace{0mm}}lp{12.6cm}p{3.0cm}} (a) & 
Amerikanischer Expre{\ss} f\"{a}llt 1 1/2 bis 77 1/4 auf mehr als 2,1 Millionen Anteilen.
 & {\it Grammar:\rule{8mm}{.2mm} Meaning:\rule{8mm}{.2mm}} \\ (b) & 
Amerikanischer Bestimmter kahle Berg 1 1/2 bis 77 1/4 auf mehr als 2.1 millionshares.
 & {\it Grammar:\rule{8mm}{.2mm} Meaning:\rule{8mm}{.2mm}} \\ (c) & 
American Express fiel 3/2 auf 309/4 auf mehr als 2.1 Mio.\ Aktien.
 & {\it Grammar:\rule{8mm}{.2mm} Meaning:\rule{8mm}{.2mm}} \\ (d) & 
Der ausdr\"{u}ckliche Amerikaner fiel 1 1/2 bis 77 1/4 auf mehr als 2,1 Million Anteilen.
 & {\it Grammar:\rule{8mm}{.2mm} Meaning:\rule{8mm}{.2mm}} \\ (e) & 
American Express fiel 3/2 auf mehr als 2.1 Mio.\ Aktien auf 309/4.
 & {\it Grammar:\rule{8mm}{.2mm} Meaning:\rule{8mm}{.2mm}} \\ (f) & 
American Express fiel um 1 1/2 auf 77 1/4 bei mehr als 2.1 Mio.\ Aktien.
 & {\it Grammar:\rule{8mm}{.2mm} Meaning:\rule{8mm}{.2mm}} \\ \end{tabular} \\

\noindent (WSJ 79)\\
{\bf His typical trade involved \$30 million.} \\
\noindent \begin{tabular}{@{\hspace{0mm}}lp{12.6cm}p{3.0cm}} (a) & 
Sein typischer Handel umfa{\ss}te 30 Mio.\ Dollar.
 & {\it Grammar:\rule{8mm}{.2mm} Meaning:\rule{8mm}{.2mm}} \\ (b) & 
Sein typischer Handel bezog \$30 Million mit ein.
 & {\it Grammar:\rule{8mm}{.2mm} Meaning:\rule{8mm}{.2mm}} \\ (c) & 
Sein typischer Beruf brachte \$30 million mit sich.
 & {\it Grammar:\rule{8mm}{.2mm} Meaning:\rule{8mm}{.2mm}} \\ (d) & 
Sein typischer Handel betraf 30 Millionen Dollar.
 & {\it Grammar:\rule{8mm}{.2mm} Meaning:\rule{8mm}{.2mm}} \\ \end{tabular} \\

\noindent \hrulefill\\

\noindent {\it Thank you!}

\newpage
\section{Translation Evaluation Key}

The two numbers in parentheses associated with each translation are the averages 
of the syntactic and semantic grades that the evaluators have assigned to that
specific translation.

\noindent (WSJ 48) \\
\hspace*{1.2cm}(a): Logos (2.1, 3.3) \\
\hspace*{1.2cm}(b): {\sc Systran} (2.8, 2.9) \\
\hspace*{1.2cm}(c): {\sc Contex} on correct parse (2.0, 2.1) \\
\hspace*{1.2cm}(d): Globalink (4.6, 4.4) \\
\hspace*{1.2cm}(e): human translation (1.5, 1.2) \\
\hspace*{1.2cm}(f): {\sc Contex} (full translation) (1.7, 4.2) \\

\noindent (WSJ 49) \\
\hspace*{1.2cm}(a): {\sc Contex} (full translation), {\sc Contex} on correct parse (1.0, 1.5) \\
\hspace*{1.2cm}(b): Globalink (2.1, 1.4) \\
\hspace*{1.2cm}(c): human translation (1.0, 1.0) \\
\hspace*{1.2cm}(d): {\sc Systran} (1.1, 2.8) \\
\hspace*{1.2cm}(e): Logos (2.1, 1.9) \\

\noindent (WSJ 50) \\
\hspace*{1.2cm}(a): {\sc Contex} (full translation), {\sc Contex} on correct parse (2.3, 1.9) \\
\hspace*{1.2cm}(b): human translation (1.0, 1.5) \\
\hspace*{1.2cm}(c): Logos (2.4, 2.9) \\
\hspace*{1.2cm}(d): {\sc Systran} (2.3, 4.5) \\
\hspace*{1.2cm}(e): Globalink (2.3, 2.2) \\

\noindent (WSJ 51) \\
\hspace*{1.2cm}(a): {\sc Systran} (3.7, 3.1) \\
\hspace*{1.2cm}(b): {\sc Contex} on correct parse (3.4, 2.3) \\
\hspace*{1.2cm}(c): Logos (4.1, 3.5) \\
\hspace*{1.2cm}(d): Globalink (4.5, 4.9) \\
\hspace*{1.2cm}(e): human translation (1.5, 1.2) \\
\hspace*{1.2cm}(f): {\sc Contex} (full translation) (3.4, 2.3) \\

\noindent (WSJ 52) \\
\hspace*{1.2cm}(a): {\sc Systran} (2.3, 3.4) \\
\hspace*{1.2cm}(b): Globalink (4.0, 5.3) \\
\hspace*{1.2cm}(c): human translation (1.7, 1.5) \\
\hspace*{1.2cm}(d): Logos (5.3, 5.5) \\
\hspace*{1.2cm}(e): {\sc Contex} (full translation) (4.3, 3.5) \\
\hspace*{1.2cm}(f): {\sc Contex} on correct parse (1.7, 2.3) \\

\newpage
\noindent (WSJ 53) \\
\hspace*{1.2cm}(a): Logos (1.8, 5.3) \\
\hspace*{1.2cm}(b): {\sc Contex} (full translation) (3.1, 3.3) \\
\hspace*{1.2cm}(c): human translation (1.2, 4.5) \\
\hspace*{1.2cm}(d): {\sc Contex} on correct parse (2.5, 2.3) \\
\hspace*{1.2cm}(e): {\sc Systran} (5.0, 5.2) \\
\hspace*{1.2cm}(f): Globalink (4.8, 5.0) \\

\noindent (WSJ 54) \\
\hspace*{1.2cm}(a): Globalink (2.3, 4.4) \\
\hspace*{1.2cm}(b): {\sc Systran} (2.5, 3.1) \\
\hspace*{1.2cm}(c): {\sc Contex} on correct parse (1.5, 1.2) \\
\hspace*{1.2cm}(d): {\sc Contex} (full translation) (1.6, 2.7) \\
\hspace*{1.2cm}(e): Logos (2.2, 2.2) \\
\hspace*{1.2cm}(f): human translation (1.0, 1.0) \\

\noindent (WSJ 55) \\
\hspace*{1.2cm}(a): {\sc Contex} (full translation), {\sc Contex} on correct parse, Logos,\\
\hspace*{1.9cm} {\sc Systran}, Globalink, human translation (1.4, 1.5) \\

\noindent (WSJ 56) \\
\hspace*{1.2cm}(a): Logos (2.2, 1.5) \\
\hspace*{1.2cm}(b): Globalink (3.8, 2.9) \\
\hspace*{1.2cm}(c): {\sc Contex} (full translation), {\sc Contex} on correct parse (2.0, 1.8) \\
\hspace*{1.2cm}(d): human translation (1.0, 1.2) \\
\hspace*{1.2cm}(e): {\sc Systran} (1.3, 1.7) \\

\noindent (WSJ 57) \\
\hspace*{1.2cm}(a): {\sc Contex} (full translation), {\sc Contex} on correct parse (1.8, 1.7) \\
\hspace*{1.2cm}(b): {\sc Systran} (2.4, 4.1) \\
\hspace*{1.2cm}(c): Logos (1.4, 2.9) \\
\hspace*{1.2cm}(d): Globalink (3.7, 3.6) \\
\hspace*{1.2cm}(e): human translation (1.2, 1.0) \\

\noindent (WSJ 58) \\
\hspace*{1.2cm}(a): {\sc Contex} on correct parse (2.9, 2.9) \\
\hspace*{1.2cm}(b): {\sc Systran} (3.6, 3.7) \\
\hspace*{1.2cm}(c): Globalink (4.0, 5.3) \\
\hspace*{1.2cm}(d): {\sc Contex} (full translation) (2.9, 3.0) \\
\hspace*{1.2cm}(e): Logos (3.8, 4.9) \\
\hspace*{1.2cm}(f): human translation (1.6, 1.7) \\

\newpage
\noindent (WSJ 59) \\
\hspace*{1.2cm}(a): {\sc Contex} (full translation), {\sc Contex} on correct parse, human\\
\hspace*{1.9cm} translation (1.1, 1.3) \\
\hspace*{1.2cm}(b): Logos (1.9, 1.4) \\
\hspace*{1.2cm}(c): {\sc Systran} (2.0, 2.1) \\
\hspace*{1.2cm}(d): Globalink (3.0, 1.8) \\

\noindent (WSJ 60) \\
\hspace*{1.2cm}(a): Logos (2.1, 1.9) \\
\hspace*{1.2cm}(b): {\sc Contex} on correct parse (4.1, 3.5) \\
\hspace*{1.2cm}(c): human translation (1.0, 1.2) \\
\hspace*{1.2cm}(d): {\sc Contex} (full translation) (3.7, 3.4) \\
\hspace*{1.2cm}(e): {\sc Systran} (1.5, 2.0) \\
\hspace*{1.2cm}(f): Globalink (1.6, 2.5) \\

\noindent (WSJ 61) \\
\hspace*{1.2cm}(a): {\sc Contex} on correct parse (1.1, 3.1) \\
\hspace*{1.2cm}(b): Globalink (2.8, 4.8) \\
\hspace*{1.2cm}(c): {\sc Systran} (2.3, 2.4) \\
\hspace*{1.2cm}(d): {\sc Contex} (full translation) (1.5, 3.1) \\
\hspace*{1.2cm}(e): human translation (1.0, 1.0) \\
\hspace*{1.2cm}(f): Logos (3.9, 5.8) \\

\noindent (WSJ 62) \\
\hspace*{1.2cm}(a): Globalink (4.5, 4.7) \\
\hspace*{1.2cm}(b): human translation (1.5, 1.5) \\
\hspace*{1.2cm}(c): {\sc Systran} (2.9, 4.7) \\
\hspace*{1.2cm}(d): {\sc Contex} on correct parse (2.0, 1.7) \\
\hspace*{1.2cm}(e): {\sc Contex} (full translation) (2.4, 1.7) \\
\hspace*{1.2cm}(f): Logos (3.6, 4.8) \\

\noindent (WSJ 63) \\
\hspace*{1.2cm}(a): Logos (2.3, 2.5) \\
\hspace*{1.2cm}(b): {\sc Contex} (full translation) (2.6, 2.0) \\
\hspace*{1.2cm}(c): {\sc Contex} on correct parse (2.5, 3.2) \\
\hspace*{1.2cm}(d): Globalink (4.0, 3.9) \\
\hspace*{1.2cm}(e): {\sc Systran} (2.6, 2.2) \\
\hspace*{1.2cm}(f): human translation (1.0, 1.2) \\

\noindent (WSJ 64) \\
\hspace*{1.2cm}(a): Logos (3.3, 3.5) \\
\hspace*{1.2cm}(b): Globalink (3.5, 2.7) \\
\hspace*{1.2cm}(c): {\sc Contex} (full translation), {\sc Contex} on correct parse (2.6, 2.1) \\
\hspace*{1.2cm}(d): {\sc Systran} (3.6, 3.3) \\
\hspace*{1.2cm}(e): human translation (1.0, 1.0) \\

\noindent (WSJ 65) \\
\hspace*{1.2cm}(a): {\sc Contex} (full translation) (2.0, 1.2) \\
\hspace*{1.2cm}(b): Logos (1.5, 2.1) \\
\hspace*{1.2cm}(c): Globalink (2.8, 1.8) \\
\hspace*{1.2cm}(d): {\sc Systran} (2.9, 2.4) \\
\hspace*{1.2cm}(e): human translation (1.2, 1.0) \\
\hspace*{1.2cm}(f): {\sc Contex} on correct parse (1.8, 1.7) \\

\noindent (WSJ 66) \\
\hspace*{1.2cm}(a): Logos (2.8, 2.5) \\
\hspace*{1.2cm}(b): Globalink (3.7, 5.3) \\
\hspace*{1.2cm}(c): {\sc Systran} (5.3, 4.8) \\
\hspace*{1.2cm}(d): {\sc Contex} (full translation) (4.9, 5.0) \\
\hspace*{1.2cm}(e): {\sc Contex} on correct parse (3.8, 3.2) \\
\hspace*{1.2cm}(f): human translation (1.0, 1.2) \\

\noindent (WSJ 67) \\
\hspace*{1.2cm}(a): {\sc Systran} (2.5, 4.3) \\
\hspace*{1.2cm}(b): {\sc Contex} (full translation) (2.5, 3.0) \\
\hspace*{1.2cm}(c): Globalink (3.2, 4.3) \\
\hspace*{1.2cm}(d): {\sc Contex} on correct parse (3.3, 3.6) \\
\hspace*{1.2cm}(e): human translation (1.2, 1.5) \\
\hspace*{1.2cm}(f): Logos (2.1, 2.8) \\

\noindent (WSJ 68) \\
\hspace*{1.2cm}(a): Logos (3.4, 3.0) \\
\hspace*{1.2cm}(b): human translation (1.3, 1.0) \\
\hspace*{1.2cm}(c): Globalink (3.8, 3.0) \\
\hspace*{1.2cm}(d): {\sc Contex} (full translation) (3.6, 2.5) \\
\hspace*{1.2cm}(e): {\sc Contex} on correct parse (3.6, 3.0) \\
\hspace*{1.2cm}(f): {\sc Systran} (3.6, 3.0) \\

\noindent (WSJ 69) \\
\hspace*{1.2cm}(a): {\sc Contex} (full translation), {\sc Contex} on correct parse (2.3, 1.8) \\
\hspace*{1.2cm}(b): Globalink (3.8, 4.5) \\
\hspace*{1.2cm}(c): Logos (3.6, 4.5) \\
\hspace*{1.2cm}(d): human translation (1.0, 1.0) \\
\hspace*{1.2cm}(e): {\sc Systran} (3.1, 5.2) \\

\newpage
\noindent (WSJ 70) \\
\hspace*{1.2cm}(a): Globalink (3.4, 4.5) \\
\hspace*{1.2cm}(b): {\sc Contex} (full translation), {\sc Contex} on correct parse (1.3, 2.1) \\
\hspace*{1.2cm}(c): human translation (1.0, 3.7) \\
\hspace*{1.2cm}(d): Logos (3.9, 5.6) \\
\hspace*{1.2cm}(e): {\sc Systran} (1.8, 4.9) \\

\noindent (WSJ 71) \\
\hspace*{1.2cm}(a): {\sc Systran} (2.8, 3.2) \\
\hspace*{1.2cm}(b): Globalink (3.7, 4.3) \\
\hspace*{1.2cm}(c): {\sc Contex} on correct parse (2.3, 2.3) \\
\hspace*{1.2cm}(d): human translation (1.0, 1.0) \\
\hspace*{1.2cm}(e): Logos (2.8, 4.6) \\
\hspace*{1.2cm}(f): {\sc Contex} (full translation) (2.5, 2.7) \\

\noindent (WSJ 72) \\
\hspace*{1.2cm}(a): Globalink (1.9, 4.1) \\
\hspace*{1.2cm}(b): {\sc Contex} (full translation), {\sc Contex} on correct parse (1.9, 2.2) \\
\hspace*{1.2cm}(c): Logos (1.4, 2.7) \\
\hspace*{1.2cm}(d): human translation (1.0, 1.7) \\
\hspace*{1.2cm}(e): {\sc Systran} (1.3, 3.2) \\

\noindent (WSJ 73) \\
\hspace*{1.2cm}(a): {\sc Systran} (4.3, 5.0) \\
\hspace*{1.2cm}(b): Logos (2.1, 3.1) \\
\hspace*{1.2cm}(c): {\sc Contex} on correct parse, human translation (1.1, 1.1) \\
\hspace*{1.2cm}(d): Globalink (1.8, 2.9) \\
\hspace*{1.2cm}(e): {\sc Contex} (full translation) (1.1, 2.1) \\

\noindent (WSJ 74) \\
\hspace*{1.2cm}(a): {\sc Contex} (full translation), {\sc Contex} on correct parse (2.1, 1.2) \\
\hspace*{1.2cm}(b): human translation (1.0, 1.0) \\
\hspace*{1.2cm}(c): Globalink (2.8, 4.2) \\
\hspace*{1.2cm}(d): {\sc Systran} (2.4, 2.9) \\
\hspace*{1.2cm}(e): Logos (1.7, 1.2) \\

\noindent (WSJ 75) \\
\hspace*{1.2cm}(a): {\sc Systran} (2.0, 1.6) \\
\hspace*{1.2cm}(b): {\sc Contex} (full translation), {\sc Contex} on correct parse, human\\
\hspace*{1.9cm}     translation (1.3, 1.2) \\
\hspace*{1.2cm}(c): Logos (1.8, 5.2) \\
\hspace*{1.2cm}(d): Globalink (4.4, 5.2) \\

\newpage
\noindent (WSJ 76) \\
\hspace*{1.2cm}(a): Logos (1.3, 1.5) \\
\hspace*{1.2cm}(b): {\sc Systran} (2.8, 2.7) \\
\hspace*{1.2cm}(c): human translation (1.2, 1.6) \\
\hspace*{1.2cm}(d): {\sc Contex} (full translation) (4.3, 3.8) \\
\hspace*{1.2cm}(e): {\sc Contex} on correct parse (4.5, 3.7) \\
\hspace*{1.2cm}(f): Globalink (3.9, 3.6) \\

\noindent (WSJ 77) \\
\hspace*{1.2cm}(a): {\sc Contex} on correct parse (1.5, 2.1) \\
\hspace*{1.2cm}(b): Logos (3.8, 4.1) \\
\hspace*{1.2cm}(c): {\sc Contex} (full translation) (2.8, 2.3) \\
\hspace*{1.2cm}(d): human translation (1.0, 1.0) \\
\hspace*{1.2cm}(e): Globalink (3.3, 3.2) \\
\hspace*{1.2cm}(f): {\sc Systran} (3.6, 3.5) \\

\noindent (WSJ 78) \\
\hspace*{1.2cm}(a): Logos (2.7, 2.8) \\
\hspace*{1.2cm}(b): Globalink (5.0, 5.7) \\
\hspace*{1.2cm}(c): {\sc Contex} (full translation) (2.4, 2.5) \\
\hspace*{1.2cm}(d): {\sc Systran} (2.7, 5.1) \\
\hspace*{1.2cm}(e): {\sc Contex} on correct parse (2.6, 2.9) \\
\hspace*{1.2cm}(f): human translation (1.2, 1.2) \\

\noindent (WSJ 79) \\
\hspace*{1.2cm}(a): {\sc Contex} (full translation), {\sc Contex} on correct parse, human\\
\hspace*{1.9cm}     translation (1.1, 1.5) \\
\hspace*{1.2cm}(b): {\sc Systran} (1.3, 2.7) \\
\hspace*{1.2cm}(c): Globalink (1.4, 4.7) \\
\hspace*{1.2cm}(d): Logos (1.3, 2.5) \\

                 %\chapter{Translation Evaluation Questionnaire and Key} \label{app-quest}
\chapter{Abbreviations}          
 
\begin{tabular}{ll}
  A & add (operation)\\
  ACL & Association for Computational Linguistics \\
  ADJ & adjective\\
  ADJP & adjective phrase\\
  ADV & adverb\\
  ADVP & adverbial phrase\\
  AI & artificial intelligence\\
  AMTA & Association for Machine Translation in the Americas \\
  APP & adverbial or prepositional phrase\\
  ARPA & Advanced Research Projects Agency (a United States Department of \\
       & Defense agency for advanced technology research) \\
  ART & article\\
  AUX & auxiliary (verb)\\
  BEN & beneficiary\\
  CAT & category\\
  COMPL & complement\\
  CONC & concatenate, concatenated element\\
  CONJ & conjunction\\
  D-... & delimiter\\
  DEF & definite\\
  ...-E... & English\\
  ELEM & element\\
  F-... & form\\
  FEM & feminine\\
  ...-G... & German\\
  I-... & internal concept\\
  INDEF & indefinite\\
  INF & infinitive/infinitival\\
  INFL & inflection\\
  INTR & intransitive\\
  IOBJ & indirect object\\
\end{tabular} \newpage \begin{tabular}{ll}
  IRR & irregular\\
  KB & knowledge base\\
  KBMT & knowledge based machine translation\\
  LEX & lexicon/lexical\\
  M-... & mathematical (concept) \\
  MASC & masculine\\
  MT & machine translation\\
  N & noun\\
  nc & count noun\\
  NEUT & neuter\\
  NL & natural language\\
  NLP & natural language processing\\
  NP & noun phrase\\
  OBJ & object\\
  P & person\\
  PART & participle\\
  pat & patient (a semantic role)\\
  pn & proper name\\
  PP & prepositional phrase\\
  PRED & predicate\\
  PREP & preposition\\
  PRES & present (tense)\\
  R & reduce (operation)\\
  R-... & role\\
  QUANT & quantity\\
  S & shift (operation)\\
  S-... & syntactic\\
  SEM & semantic\\
  SEMROLE & semantic role\\
  SING & singular\\
  SUBJ & subject\\
  SYNT & syntactic\\
  SYNTROLE & syntactic role\\
  TR & transitive\\
  UNAVAIL & unavailable\\
  URL & universal resource locator (World Wide Web address)\\
  V & verb\\
  VP & verb phrase\\
  WSJ & Wall Street Journal\\
\end{tabular}
         %\chapter{Abbreviations} \label{app-abbreviations}

%\begin{thebibliography}{..}          %% Start your bibliography here; you can
%\bibitem{...} ...                    %% also use the \bibliography command
%\end{thebibliography}                %% to generate your bibliography.
\bibliographystyle{theapa}
\bibliography{lunar}

\begin{thesisauthorvita}             %% Write your vita here; it can be anything in LaTeX2e par-mode.
\hyphenation{Stu-dien-stif-tung}
Ulf Hermjakob was born in Meppen, Germany, on November 26, 1962.
After a few years in Basel, Switzerland, and Bad Oldesloe, Germany, his family moved to
B\"{u}nde, Germany, where he graduated from the Freiherr-vom-Stein Gymnasium in 1981. After spending
an academic year at the University of Georgia at Athens on a Georgia Rotary Student Program scholarship, 
he studied Computer Science with minors in business and (German) civil law at the University of Karlsruhe,
Germany, where he received his Vordiplom in 1984 and his Hauptdiplom in 1988. After briefly working
for Siemens in Paris, France, he entered the Computer Science Ph.D.\ program of the
University of Texas at Austin on a one-year scholarship from the ``Studienstiftung des deutschen
Volkes'' (German National Scholarship Foundation). After two years as Teaching Assistant for
Computer Graphics, he joined the Knowledge Based Natural Language group at the Microelectronics
and Computer Technology Corporation (MCC) in Austin, Texas, where he worked from 1991-1995.\\[1.5cm]
\begin{tabbing}
Email: \= ulf@cs.utexas.edu \\
URL:   \> http://www.cs.utexas.edu/users/ulf/ \\[-2.85cm]
\end{tabbing}
\end{thesisauthorvita}               %%

\end{document}